\documentclass[twoside,11pt]{article}
\usepackage[preprint]{jmlr2e}

\usepackage{amsmath,amssymb,amsfonts,mathtools}
\newtheorem{assumption}{Assumption}
\usepackage{booktabs}
\usepackage{enumitem}
\usepackage{xcolor}
\usepackage[most]{tcolorbox}
\usepackage{pgf}
\usepackage{tikz}
\usetikzlibrary{arrows.meta,positioning,calc,shapes.geometric}
\usetikzlibrary{decorations.pathmorphing}
\usetikzlibrary{decorations.pathreplacing}
\usetikzlibrary{fit,backgrounds}

\usepackage{subcaption}
\usepackage{algorithm}
\usepackage{algorithmic}
\newtheorem{defn}{Definition}

\newcommand{\dom}[1]{\operatorname{dom}[#1]}
\newcommand{\scmdo}{\operatorname{do}}

\newcommand{\scmsimple}{\mathcal{M}}
\DeclareMathOperator{\sign}{sign}
\DeclareMathOperator{\prox}{prox}

\newcommand{\scmbase}{\mathcal{M}^{\ell}}
\newcommand{\scmabst}{\mathcal{M}^{h}}

\newcommand{\intervsetbase}{\mathcal{I}^{\ell}}
\newcommand{\intervsetabst}{\mathcal{I}^{h}}

\newcommand{\structfuncset}{\mathcal{F}}
\newcommand{\scmdag}{\mathcal{G}}
\DeclareMathOperator{\proj}{proj}
\newcommand{\enbase}{\mathcal{X}^{\ell}}
\newcommand{\enabst}{\mathcal{X}^{h}}

\newcommand{\en}{\mathcal{X}}
\newcommand{\ex}{\mathcal{U}}

\newcommand{\prob}{\mathbb{P}}

\newcommand{\lintau}{\operatorname{T}}

\newcommand{\cD}{{\cal D}}

\newcommand{\cI}{{\cal I}}

\newcommand{\cM}{{\cal M}}

\newcommand{\cS}{{\cal S}}

\newenvironment{namedproof}[1]
{\par\noindent\textbf{Proof of #1.}\quad}
{\hfill\ensuremath{\blacksquare}\par}

\DeclareMathOperator*{\argmin}{arg\,min}
\DeclareMathOperator{\diag}{diag}
\DeclareMathOperator{\Tr}{Tr}

\newcommand{\Qcal}{\mathcal{Q}}

\newcommand{\dglyph}[3][magenta]{%
  \draw[#1!40,fill=#1!8,line width=0.4pt]  (#2,#3) circle (2.4mm);
  \draw[#1!65,fill=#1!18,line width=0.4pt] (#2,#3) circle (1.6mm);
  \fill[#1] (#2,#3) circle (0.9mm);
}
\newcommand{\hammer}[2]{%
  \begin{scope}[shift={(#1,#2)}]
    \fill[magenta] (0,-0.15) circle (0.8mm);
    \draw[magenta, line width=0.8pt] (0,-0.15) circle (0.8mm);
    
    \begin{scope}[shift={(0.31, -0.06)}, rotate=60, scale=0.4]
      \fill[black!90] (-0.07, 0.12) rectangle (0.07, 0.55);
      \fill[black!90] (-0.2, 0.5) rectangle (0.2, 0.8);
      \fill[black!90] (-0.28, 0.45) rectangle (-0.2, 0.85);
      \fill[black!90] (0.2, 0.8) -- (0.35, 0.8) -- (0.35, 0.6) -- (0.2, 0.5) -- cycle;
    \end{scope}
  \end{scope}
}
\newcommand{\inlinehammer}{%
  \tikz[baseline=-0.5ex, scale=0.45]{
    \begin{scope}[rotate=60]
      \fill[black!90] (-0.07, 0.12) rectangle (0.07, 0.55);
      \fill[black!90] (-0.2, 0.5) rectangle (0.2, 0.8);
      \fill[black!90] (-0.28, 0.45) rectangle (-0.2, 0.85);
      \fill[black!90] (0.2, 0.8) -- (0.35, 0.8) -- (0.35, 0.6) -- (0.2, 0.5) -- cycle;
    \end{scope}
  }%
}

\newcommand{\shiftlaw}[7]{%
  \foreach \rr in {1.0,0.62,0.32}
    {\draw[#3!65!black, fill=#3!20, line width=0.3pt, rotate around={#4:(#1,#2)}]
       (#1,#2) ellipse ({\rr*#5} and {\rr*0.24});}
  \draw[#3!70!black, dashed, line width=0.5pt, rotate around={#4:({#1+#6},{#2+#7})}]
       ({#1+#6},{#2+#7}) ellipse ({#5} and {0.24});
  \draw[#3!70!black, dashed, line width=0.45pt, rotate around={#4:({#1+#6},{#2+#7})}]
       ({#1+#6},{#2+#7}) ellipse ({0.62*#5} and {0.15});
  \draw[#3!75!black, -{Latex[length=3pt]}, line width=0.4pt]
       (#1,#2) -- ({#1+#6*0.7},{#2+#7*0.7});
}
\newcommand{\blob}[5]{%
  \foreach \rr in {1.0,0.68,0.42,0.22}
    {\draw[#3!65!black, fill=#3!18, line width=0.35pt, rotate around={#4:(#1,#2)}]
       (#1,#2) ellipse ({\rr*#5} and {\rr*0.26});}
  \fill[#3!80!black, rotate around={#4:(#1,#2)}] (#1,#2) ellipse ({0.11*#5} and {0.033});
}

\newcommand{\glyphcentre}{\tikz[baseline=-0.5ex]{\node[circle,draw=black!55,fill=gray!35,minimum size=2.2mm,inner sep=0pt]{};}}
\newcommand{\glyphwc}{\tikz[baseline=-0.5ex]{\node[diamond,draw=black!55,fill=gray!55,minimum size=2.6mm,inner sep=0pt]{};}}
\newcommand{\glyphtrue}{\tikz[baseline=-0.5ex]{\node[star,star points=5,draw=black!55,fill=gray!55,minimum size=2.6mm,inner sep=0pt]{};}}

\usepackage{lastpage}

\jmlrheading{}{2026}{}{}{}{}{Yorgos Felekis, Paris Giampouras, Fabio Massimo Zennaro and Theodoros Damoulas}

\ShortHeadings{Generalised Transportability via Causal Abstractions}{Felekis, Giampouras, Zennaro, Damoulas}
\firstpageno{1}

\begin{document}

\title{Generalised Transportability via Causal Abstractions}

\author{\name Yorgos Felekis \email yorgos.felekis@warwick.ac.uk \\
       \addr Department of Computer Science\\
       University of Warwick\\
       Coventry, CV4 7AL, UK
       \AND
       \name Paris Giampouras \email paris.giampouras@warwick.ac.uk \\
       \addr Department of Computer Science\\
       University of Warwick\\
       Coventry, CV4 7AL, UK
       \AND
       \name Fabio Massimo Zennaro \email fabio.zennaro@uib.no \\
       \addr Department of Informatics\\
        University of Bergen\\
       Bergen, N-5020, NO
       \AND
       \name Theodoros Damoulas \email t.damoulas@warwick.ac.uk\\
       \addr Departments of Computer Science \& Statistics\\
       University of Warwick\\
       Coventry, CV4 7AL, UK}

\editor{My editor}

\maketitle
\begin{abstract}
Transporting a causal conclusion from a source study population to a target one is a fundamental problem in causal inference. The theory of transportability provides a criterion for when this is possible: given experimental data from the source, observational data from the target, and a diagram showing where the two populations differ, it determines whether a target query is identifiable and does so completely; i.e., if the query can be transported, the criterion finds the exact formula. However, it works one query at a time and answers only whether transport is possible, without providing the query value. It is also silent in two practically important regimes: when the query is not transportable, even given target data, and when no target data exist at all. To tackle both, we take a model-level perspective on the transportability problem grounded in Causal Abstraction theory. Source and target share variables, graph, and interventions, differing only at a known set of mechanisms, which makes transportability a special case of same-level abstraction. Thus, instead of asking whether one query transports, we ask whether a single map aligns the source and target across their interventional behaviour. We characterise when such a map exists in both the Markovian and semi-Markovian settings; when it does, every target query transports at once. Our main contribution lies in the approximate case. When no exact map exists, the best approximate one still yields \emph{certified query intervals}, recasting abstraction error as a quantitative notion of \emph{approximate transportability}. We formulate model-level transport as distributionally robust optimisation over mechanism and environment perturbations of the unseen target and derive certificates that cover both challenging regimes: bounds for non-transportable queries and guarantees under target-agnostic settings. We evaluate the framework on synthetic Markovian and semi-Markovian benchmarks and a real ecological dataset, and we show that the certified intervals bracket the true interventional query. 
\end{abstract}


\section{Introduction}
\label{sec:introduction}
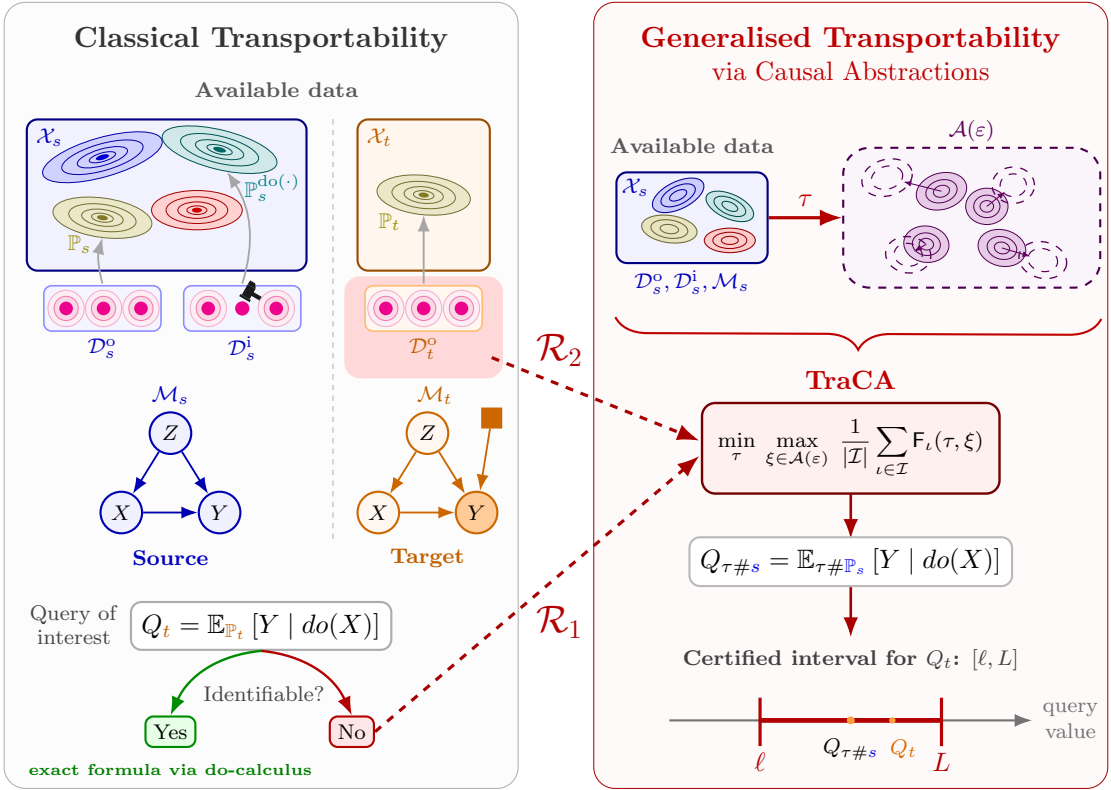
\begin{figure}[t]
\centering
\begin{tikzpicture}[
    >=Latex, font=\sffamily, line width=0.6pt,
    nd/.style={circle, draw=blue!70!black, fill=blue!6, thick, minimum size=5.5mm, inner sep=0pt, font=\scriptsize},
    ndorg/.style={circle, draw=orange!80!black, fill=orange!6, thick, minimum size=5.5mm, inner sep=0pt, font=\scriptsize},
    shifted/.style={circle, draw=orange!80!black, fill=orange!40, thick, minimum size=5.5mm, inner sep=0pt, font=\scriptsize},
    sel/.style={rectangle, draw=orange!80!black, fill=orange!80!black, thick, minimum width=2.5mm, 
    minimum height=2.5mm, inner sep=1pt, font=\tiny},
    redbox/.style={rounded corners=3pt, draw=red!70!black, fill=red!10, thick, inner sep=3pt, font=\scriptsize, align=center},
    grnbox/.style={rounded corners=3pt, draw=green!55!black, fill=green!10, thick, inner sep=3pt, font=\scriptsize, align=center},
    qbox/.style={rounded corners=4pt, draw=gray!65, fill=white, thick, inner sep=4pt, font=\small},
    srcbox/.style={rounded corners=3pt, draw=blue!45, fill=blue!4, line width=0.6pt},
    srclaw/.style={rounded corners=4pt, draw=blue!50!black, fill=blue!5, line width=0.8pt},
    tgtbox/.style={rounded corners=3pt, draw=orange!55, fill=orange!6, line width=0.6pt},
    tgtlaw/.style={rounded corners=4pt, draw=orange!60!black, fill=orange!7, line width=0.8pt},
    edge/.style={->, draw=blue!70!black, line width=0.7pt},
    edgeorg/.style={->, draw=orange!80!black, line width=0.7pt},
    flow/.style={->, draw=gray!55!black, line width=0.8pt},
    grnflow/.style={->, draw=green!55!black, line width=0.9pt},
    feed/.style={->, draw=black!35, line width=0.7pt},
    lbl/.style={font=\scriptsize, text=gray!70!black},
    slbl/.style={font=\scriptsize, text=gray!55!black},
    ptitle/.style={font=\normalsize\bfseries},
]

\begin{scope}[shift={(0,0)}]
\node[draw=gray!65, rounded corners=8pt, line width=0pt, fill=gray!2,
      minimum width=6.8cm, minimum height=10.4cm] (Lpanel) at (3.4, 0) {};
\node[ptitle, text=black!80] at (3.4, 4.7) {Classical Transportability};

\node[font=\scriptsize\bfseries, text=black!65] at (3.6, 4.05) {Available data};
\draw[gray!35, line width=0.7pt, dash pattern=on 2pt off 2pt] (4.35, -1.9) -- (4.35, 3.7);

\node[srclaw, minimum width=3.7cm, minimum height=2.0cm] (Xs) at (2.15, 2.65) {};
\node[slbl, text=blue!75!black, anchor=north west] at ($(Xs.north west)+(0.0,-0.00)$) {$\mathcal{X}_s$};
\blob{1.3}{3.15}{blue}{16}{0.82}
\blob{2.85}{3.25}{teal}{-14}{0.78}
\blob{2.55}{2.45}{red}{0}{0.6}
\blob{1.3}{2.35}{olive}{-8}{0.66}

\node[srcbox, minimum width=1.55cm, minimum height=0.6cm] (sobs) at (1.3, 1.15) {};
\dglyph{0.82}{1.15}\dglyph{1.315}{1.15}\dglyph{1.81}{1.15}
\node[slbl, text=blue!75!black] at (1.3, 0.62) {$\mathcal D_s^{\operatorname{o}}$};
\draw[feed] (1.3, 1.48) to[bend left=15] (1.3, 2.085);

\node[srcbox, minimum width=1.55cm, minimum height=0.6cm] (si1) at (3.15, 1.15) {};
\dglyph{2.7}{1.15}\hammer{3.15}{1.3}\dglyph{3.6}{1.15}
\node[slbl, text=blue!75!black] at (3.15, 0.62) {$\mathcal D_s^{\operatorname{i}}$};
\draw[feed] (3.15, 1.48) to[bend right=25] (2.95, 2.96);

\node[tgtlaw, minimum width=1.75cm, minimum height=2.0cm] (Xt) at (5.55, 2.65) {};
\node[slbl, text=orange!75!black, anchor=north west] at ($(Xt.north west)+(0.00,-0.00)$) {$\mathcal{X}_t$};
\blob{5.55}{2.65}{olive}{-6}{0.62}

\node[font=\scriptsize\bfseries, text=blue!75!black] at (2.2, -2.15) {Source};
\node[font=\scriptsize\bfseries, text=orange!75!black] at (5.6, -2.15) {Target};

\node[rounded corners=6pt, fill=red!15, minimum width=2.1cm, minimum height=1.35cm] at (5.55, 0.9) {};

\node[tgtbox, minimum width=1.55cm, minimum height=0.6cm] (tobs) at (5.55, 1.15) {};
\dglyph{5.05}{1.15}\dglyph{5.55}{1.15}\dglyph{6.05}{1.15}
\node[slbl, text=orange!75!black] at (5.55, 0.62) {$\mathcal D_t^{\operatorname{o}}
$};
\draw[feed] (5.55, 1.48) -- (5.55, 2.385);
\node[font=\small\bfseries, text=red!70!black] at (7.35, 0.6) {\Large $\mathcal{R}_2$};

\node[slbl, text=blue!75!black] at (2.2, -0.0) {$\mathcal{M}_s$};
\node[nd] (sZ) at (2.2, -0.5) {$Z$};
\node[nd] (sX) at (1.55, -1.55) {$X$};
\node[nd] (sY) at (2.85, -1.55) {$Y$};
\draw[edge] (sZ) -- (sX); \draw[edge] (sZ) -- (sY); \draw[edge] (sX) -- (sY);

\node[slbl, text=orange!75!black] at (5.7, -0.0) {$\mathcal{M}_t$};
\node[ndorg] (tZ) at (5.6, -0.5) {$Z$};
\node[ndorg] (tX) at (4.95, -1.55) {$X$};
\node[shifted] (tY) at (6.25, -1.55) {$Y$};
\node[sel] (sig) at (6.45, -0.3) {};
\draw[edgeorg] (tZ) -- (tX); \draw[edgeorg] (tZ) -- (tY); \draw[edgeorg] (tX) -- (tY);
\draw[->, orange!80!black, line width=0.7pt] (sig) -- (tY);

\node[qbox] (query) at (3.4, -3.05) 
{$Q_{\textcolor{orange!75!black}{t}}
=
\mathbb{E}_{\textcolor{orange!75!black}{\mathbb{P}_t}}
\left[Y \mid do(X)\right]$};

\node[slbl, anchor=west] at (0.2, -2.9) {Query of};
\node[slbl, anchor=west] at (0.3, -3.2) {interest};

\node[slbl, anchor=west] at (2.5, -3.95) {Identifiable?};
\node[grnbox] (yes) at (2.2, -4.45) {Yes};
\node[redbox] (no)  at (4.6, -4.45) {No};
\draw[grnflow] (query.south) to[bend right=30] (yes.north);
\draw[grnflow, draw=red!70!black] (query.south) to[bend left=30] (no.north);
\node[font=\tiny\bfseries, text=green!50!black] at (2.2, -4.95) {exact formula via do-calculus};
\node[font=\small\bfseries, text=red!70!black] at (7.35, -3.0) {\Large $\mathcal{R}_1$};

\node[font=\scriptsize\bfseries, text=olive] at (1, 2) {$\mathbb{P}_s$};

\node[font=\scriptsize, text=teal] at (3.55, 2.75) {$\mathbb{P}_s^{\operatorname{do}(\cdot)}$};

\node[font=\scriptsize\bfseries, text=olive] at (5.1, 2.3) {$\mathbb{P}_t$};
\end{scope}

\begin{scope}[shift={(7.8,0)}]
\node[draw=red!65!black, rounded corners=8pt, line width=0pt, fill=red!2,
      minimum width=6.8cm, minimum height=10.4cm] (Rpanel) at (3.4, 0) {};
\node[ptitle, text=red!75!black] at (3.4, 4.7) {Generalised Transportability};
\node[ptitle, text=red!75!black, font=\small] at (3.4, 4.3) {via Causal Abstractions};

\node[srclaw, thick, minimum width=2.0cm, minimum height=1.2cm] (sfam) at (1.3, 2.35) {};
\node[slbl, text=blue!75!black] at (0.55, 2.75) {$\mathcal{X}_s$};
\foreach \r/\c in {0.45/olive,0.30/olive,0.15/olive}
  {\draw[\c!60!black, fill=\c!20, line width=0.4pt, rotate around={-8:(1.0, 2.2)}] (1.0, 2.2) ellipse ({\r*0.82cm} and {\r*0.4cm});}
\foreach \r/\c in {0.45/teal,0.30/teal,0.15/teal}
  {\draw[\c!60!black, fill=\c!20, line width=0.4pt, rotate around={-24:(1.8, 2.5)}] (1.8, 2.5) ellipse ({\r*0.75cm} and {\r*0.36cm});}
\foreach \r/\c in {0.45/blue,0.30/blue,0.15/blue}
  {\draw[\c!60!black, fill=\c!20, line width=0.4pt, rotate around={26:(1.0, 2.2)}] (1.3, 2.55) ellipse ({\r*0.82cm} and {\r*0.4cm});}
\foreach \r/\c in {0.45/red,0.30/red,0.15/red}
  {\draw[\c!60!black, fill=\c!20, line width=0.4pt, rotate around={0:(1.8, 2.5)}] (1.8, 2.05) ellipse ({\r*0.75cm} and {\r*0.36cm});}

\node[rounded corners=10pt, draw=violet!60!black, dashed, fill=violet!4, line width=0.8pt, minimum width=3.4cm, minimum height=1.85cm] (ambset) at (5.0, 2.35) {};
\node[slbl, text=violet!70!black] at (5.0, 3.5) {$\mathcal{A}(\varepsilon)$};
\shiftlaw{4.6}{2.7}{violet}{16}{0.34}{-0.72}{0.18}
\shiftlaw{5.4}{1.95}{violet}{-14}{0.32}{0.54}{-0.16}
\shiftlaw{5.2}{2.5}{violet}{0}{0.28}{0.35}{0.28}
\shiftlaw{4.4}{2.0}{violet}{-8}{0.3}{-0.24}{-0.14}

\draw[->, draw=red!70!black, line width=1.2pt] (sfam.east) to
    node[above, font=\small\bfseries, text=red!75!black] {$\tau$} (ambset.west);

\draw[decorate, decoration={brace, amplitude=12pt, mirror, raise=0pt},
      line width=0.8pt, red!75!black]
      ($(sfam.south west)+(-0.0,-0.70)$) -- ($(ambset.south east)+(0.0,-0.40)$)
      coordinate[midway] (bracemid);

\node[font=\small\bfseries, text=red!75!black] at (3.4, 0.2) {TraCA};

\node[font=\scriptsize\bfseries, text=black!65] at (1.3, 3.3) {Available data};
\node[font=\scriptsize\bfseries, text=blue!75!black] at (1.3, 1.5){$\mathcal D_s^{\operatorname{o}}
, \mathcal D_s^{\operatorname{i}}, \mathcal{M}_s$};

\node[rounded corners=5pt, draw=red!45!black, fill=red!6, line width=0.9pt, inner sep=6pt, font=\scriptsize]
   (dro) at (3.4, -0.7) {$\displaystyle \min_{\tau}\;\max_{\xi\in\mathcal{A}(\varepsilon)}\;
      \frac{1}{|\mathcal I|}\sum_{\iota\in\mathcal I}\mathsf F_\iota(\tau,\xi)$};

\draw[->, draw=red!65!black, line width=1.0pt] (3.4, -1.3) -- (3.4, -1.9);
\node[rounded corners=4pt, draw=gray!55, fill=white, thick, inner sep=4pt, font=\small] (qts)
   at (3.4, -2.2) {$Q_{\tau\# \textcolor{blue!95!black}{s}} = \mathbb{E}_{\tau\# \textcolor{blue!95!black}{\mathbb{P}_{s}}}
\left[Y \mid do(X)\right]$};
\draw[->, draw=red!65!black, line width=1.0pt] (qts.south) -- (3.4, -3.2);

\node[font=\scriptsize\bfseries, text=black!75] at (3.4, -3.5) {Certified interval for $Q_t$: $[\ell,L]$};
\draw[->, line width=0.8pt, black!55] (1.0, -4.3) -- (5.8, -4.3) node[right, font=\scriptsize, text=black!55, align=center] {query\\ value};
\draw[red!70!black, line width=1.8pt] (2.2, -4.3) -- (4.6, -4.3);
\draw[red!70!black, line width=1.2pt] (2.2, -4.0) -- (2.2, -4.64);
\draw[red!70!black, line width=1.2pt] (4.6, -4.0) -- (4.6, -4.64);
\node[font=\small\bfseries, text=red!70!black] at (2.2, -4.85) {$\ell$};
\node[font=\small\bfseries, text=red!70!black] at (4.6, -4.85) {$L$};
\fill[orange!75] (3.4, -4.3) circle (1.5pt);
\node[font=\scriptsize] at (3.4, -4.68) {$Q_{\tau\# \textcolor{blue!95!black}{s}}$};
\fill[orange!75] (3.95, -4.3) circle (1.2pt);
\node[font=\scriptsize, text=orange!85!black] at (4.1, -4.68) {$Q_t$};
\end{scope}

\draw[->, draw=red!65!black, line width=1.2pt, dashed]
    (6.45, 0.5) to[bend left=0] (9.2, -0.6);
\draw[->, draw=red!65!black, line width=1.2pt, dashed]
    (no.east) to[bend right=0] (9.2, -0.8);
\end{tikzpicture}
\caption{\emph{From query-level verdicts to model-level certified
transportability.} \textbf{Left:} \textit{Classical Transportability.} A query estimated in a source population (blue) need not stay valid in a target (orange) where some nodes' mechanisms shift. From the source domain we are given observational and interventional data
$(\mathcal D_s^{\operatorname{o}},\mathcal D_s^{\operatorname{i}})$, generated by the laws $\mathbb P_s$ and $\mathbb P_s^{\operatorname{do}(\cdot)}$ (interventions marked by \protect\inlinehammer), together with target observational data $\mathcal D_t^{\operatorname{o}}$ from law $\mathbb P_t$, entailed by the source and target SCMs $\mathcal M_s,\mathcal M_t$, which share graph and interventions and differ only at the shifted mechanism ($Y$). One asks whether $Q_t=\mathbb E_{\mathbb P_t}[Y\mid do(X)]$ is identifiable. The do-calculus returns a binary verdict (\emph{Yes}: exact formula; \emph{No}: none), leaving no answer in two regimes: \textbf{($\mathcal R_1$)} $Q_t$ is not identifiable; \textbf{($\mathcal R_2$)} no target data exist, so even a transportable formula cannot be evaluated. \textbf{Right:} \textit{Generalised Transportability via Causal Abstractions.} As source and target share variables, graph, and interventions and differ only at the shifted mechanisms, we model transport through a causal abstraction map $\tau$, with ambiguity concentrated around the shifted mechanisms. \textbf{TraCA:} We learn $\tau$ via a min-max problem that pushes $\mathbb P_s$ toward a neighbourhood $\mathcal A(\varepsilon)$ of admissible targets and hedges against the worst. Because $\tau$ is only approximate, the transported value $Q_{\tau\# s}=\mathbb E_{\tau_\#\mathbb P_s}[Y\mid do(X)]$ need not coincide with the true target value. Instead, it anchors the certified target interval $[\ell,L]$, shown in red, which contains $Q_t^\star$ for every admissible target.}
\label{fig:intro-overview}
\end{figure}
A clinical study in a metropolitan hospital network estimates the effect of a new respiratory therapy on patient recovery. Before deploying it in a rural region with a different age distribution and air quality characteristics, a natural question is whether the estimated population effect still applies, and if not, how it should be adjusted.  This is the classical question of \emph{external validity}: whether a causal conclusion established under the conditions in which it was studied continues to hold under different ones. In the experimental-design literature \citep{campbell2015experimental,reichardt2002experimental}, external validity is contrasted with \emph{internal validity}: generalisation beyond the study conditions versus correct identification within them. Two regimes of external validity are usually distinguished \citep{Degtiar2021ARO}: \emph{generalisability}, which involves transferring a finding from a sample to the broader source population from which it was drawn, and \emph{transportability}, which involves transferring it to a distinct target population that differs from the source. The theory of \emph{causal transportability} \citep{pearl2011transportability,bareinboim2016causalfusion} gives this second regime its modern graphical formalisation, as a form of formal criterion to transfer causal effects learned from experiments in a source population to a target where only observational data can be collected. Source and target domains are taken to share the same causal graph but differ through a known subset of mechanisms, marked by a selection diagram. The do-calculus then provides a complete query-level criterion \citep{bareinboim2013metatransportability}: for a given target query, it either derives an identification formula expressing it in terms of source experimental and target observational data or proves that no such formula exists. The theory has been extended to multiple source domains \citep{bareinboim2016causalfusion} and to finite-sample statistical questions \citep{correa2019statistical}. It has also been generalised to related data-fusion regimes in which experiments are available only on a restricted set of controllable variables, including
\(z\)-transportability \citep{lee2013ztransport}. More broadly, it connects to the problem of generalisation across environments \citep{pearl2014external,rojas2018invariant,magliacane2018domainadaptation}.

This query-level view is precise and powerful within its scope, but it is often uninformative in two practically important regimes. The first (\(\mathcal R_1\)) is when a target query is not identifiable. In the metropolitan study, suppose an unmeasured factor such as socioeconomic status influences both who receives the therapy and how well they recover. The do-calculus may then return ``not transportable'' for the effect of interest, not because nothing can be said about the target effect, but because there is no exact identifying formula to return. The second regime (\(\mathcal R_2\)) is when no target data are available at all: the rural region has not yet been sampled, or one wishes to commit to a deployment policy that remains valid across a whole family of plausible target regions. In that case, even queries that are classically transportable cannot be evaluated because their identifying formulas reference target-side quantities that have not been observed. In both regimes, a method that returns a single binary verdict per query leaves the practitioner with little actionable information.

We argue that in these regimes, the natural object of study is not the individual query but the relation between the source and target models. The transportability setting is highly structured: source and target share variables, graph, and intervention semantics and differ only through a known subset of mechanisms. Structure-preserving relations between causal models are exactly the object of Causal Abstraction (CA) theory \citep{rubenstein2017causal,beckers2018abstracting}, which formalises maps between causal models that describe the same system at different resolutions. Thus, the transportability setting is a special, same-resolution instance of it. Causal Abstraction Learning (CAL) \citep{zennaro2022computingoptimalabstractionstructural} aims to recover abstraction maps directly from data, enabling cross-scale representation learning and the integration of heterogeneous evidence. Early CAL methods established distinct computational frameworks for learning approximate abstractions from data: joint differentiable programming \citep{zennaro2023jointly} and multi-marginal optimal transport \citep{felekis2024causal}, with subsequent approaches broadening the class of abstractions and objectives \citep{kekić2023targeted,xia2024neural, dacunto2025causalabstractionlearningbased}. Causal abstractions have been applied across domains, from climate modelling \citep{chalupka2017causal} and neural-network interpretability \citep{geiger2021causal} to bandits \citep{zennarobandits}, surrogate modelling \citep{dyer2023interventionally}, and causal discovery \citep{massidda2024learningcausalabstractionslinear}.

Our setting is a constrained, same-resolution instance of the general CA picture: the relevant map is \emph{automorphic}, acting as the identity on the invariant parts of the model and only locally on the shifted mechanisms. There is no aggregation of variables, no change of granularity, and no nontrivial translation between interventions. Concretely, in our example, it leaves the recovery mechanism untouched and re-expresses only the shifted demographic and environmental mechanisms, so that a single map carries every interventional query from the metropolitan source to the rural target at once. We therefore ask whether the target model is an \emph{automorphic abstraction} of the source. When such a map exists, it transports the full source interventional family to the target and hence transports all causal queries simultaneously. The converse need not hold: several individual queries may be classically transportable while no single common map explains them all. We call this the \emph{compatibility gap}. When no exact map exists, a best approximate one always does, and its residual error is what we turn into certified bounds: the question then shifts from whether an automorphic abstraction exists to \emph{how reliably} the best one transports and what its approximation error certifies about the target queries.

The value of this perspective depends on the setting. In the Markovian case with target data, every query is already transportable by direct adjustment, whereas in the semi-Markovian case, some queries admit no identifying formula at all (\(\mathcal R_1\)). Precisely for these non-identifiable queries, the existence of an abstraction provides a mechanism-level route to transport, where the query-level criterion supplies none. However, the main payoff of the model-level perspective appears in the approximate case. When no exact map exists, the best approximate one still carries structural information; its residual error induces certified bounds on target queries. This places our contribution alongside the literature on partial identification \citep{Zhang_Bareinboim_2021,Duarte02072024,bellot2024boundingME} and partial transportability \citep{jalaldoust2024partial}, which seeks bounds when point identification fails. Our route differs on three axes. First, rather than optimising over the full class of SCMs compatible with the graph and observed distributions, re-solving per query, we learn a single transport map $\tau$ that carries the entire source interventional family to the target at once. Second, our ambiguity set is \emph{metric} rather than combinatorial: a structured neighbourhood around the source SCM, parameterised by bounded mechanism and environment perturbations, allows us to certify \emph{how far} a mechanism may shift rather than only \emph{which} mechanisms may differ. Third, we derive query intervals directly from the map's residual error through structural stability constants, rather than by sampling or optimising over admissible models. Together, these yield what the classical query-level theory cannot: a certified interval, not a binary verdict, in both regimes. In $\mathcal R_1$, the interval needs no identifying formula, so it is informative exactly where the do-calculus is silent; in $\mathcal R_2$, the map minimises worst-case transport error over the ambiguity set, so the interval holds uniformly over every admissible target. Close to our setting is also \citep{kostin2024achievable}, which studies distributional robustness under partial identifiability in the same reduced form linear Gaussian model, distinguishing shift directions seen during training from unseen ones, a distinction that parallels our $\mathcal R_1$/$\mathcal R_2$ regimes. Their shifts, however, are covariate shifts with an invariant mechanism, and their central object is a robust \emph{predictor} minimising prediction risk; we instead admit mechanism perturbations and learn a transport \emph{map} that certifies interventional queries, yielding query intervals rather than a risk bound.

Our robust formulation leverages Distributionally Robust Optimisation (DRO) \citep{kuhn2019wassersteindistributionallyrobustoptimization}, of a specific kind. In standard DRO, and in its use for domain generalisation \citep{wang2024wdrdg}, the ambiguity set models uncertainty about the data-generating distribution itself. Here, by contrast, the source SCM is treated as a trusted causal reference, and ambiguity is placed on the unseen target through perturbations of the shifted mechanisms and the target environment. The decision variable is a causal transport map, and robustness is over causal models rather than predictive distributions. The ambiguity set can further encode a directional prior, a declared sign and magnitude for an anticipated mechanism shift, which sharpens both the transport map and its certificate when the realised shift is consistent with it. Our formulation builds directly on recent work introducing distributional robustness into CAs \citep{felekis2026distributionallyrobustcausalabstractions}. The result is a formalism in which identifiability is no longer a binary gate but a special, best-case point in a continuum of certified approximate transport: \emph{transportability is the existence of a structural relation between two causal models, and where that relation fails to hold exactly, its best approximation still certifies what can be transported.} We summarise our contributions, and with them the structure of the paper, as follows.
\begin{itemize}[leftmargin=1.5em]
    \item \textbf{Transportability as automorphic causal abstraction} (Section~\ref{sec:structural-relation}). We show that the transportability setting, where the source and target share variables, graph, and intervention semantics, differing only through a known subset of mechanisms, is a same-resolution instance of causal abstraction. This connection has not been drawn before, and it is what allows per-query verdicts to be replaced by a single model-level object relating the two domains.
    
  \item \textbf{A hierarchy of model-level consistency notions} (Section~\ref{sec:structural-relation}). We define four source–target relations, from observational matching to a single constructive map transporting the full interventional family, each implying the one before, together with an orthogonal query-restricted relaxation. This hierarchy places classical \(\tau\)-\(\omega\) abstraction in the transportability setting and identifies the compatibility gap between query-level and model-level transport.

  \item \textbf{Exact existence characterisations} (Section~\ref{sec:characterizing}). We provide characterisations for the existence of a constructive abstraction for both Markovian and semi-Markovian cases. In finite state spaces, these reduce to linear feasibility conditions, with accompanying uniqueness criteria.

  \item \textbf{A target-agnostic robust formulation of approximate transport} (Section~\ref{sec:transport-via-causal-abstractions}). We introduce TraCA, a new framework that formulates approximate model-level transport as a distributionally robust optimisation problem over target-side mechanism and environment perturbations around a trusted source SCM.

  \item \textbf{An optimisation algorithm} (Sections~\ref{sec:lan} and~\ref{sec:algorithms}). For linear additive-noise models, we derive stability bounds for mechanism perturbations and instantiate TraCA with Gaussian and empirical objectives, yielding a practical alternating min-max learning algorithm.

  \item \textbf{Provable query-level certificates} (Section~\ref{sec:provable-robustness}). We show how model-level transport error induces certified intervals for downstream target queries, including the target-agnostic robust regime where the bounds hold uniformly over the ambiguity set. The certificate comes in a general form, valid for any target in the ambiguity set, and a sharper directional form that exploits a declared prior.

    \item \textbf{Empirical validation in the two motivating regimes} (Section~\ref{sec:experiments}). We validate the framework on synthetic Markovian and semi-Markovian benchmarks and on a real-world ecological dataset, focusing on certified transport for a query that is not classically transportable and on robust target-agnostic transport. Across all benchmarks, the certified intervals bracket the ground-truth query at the radius the true shift requires; the directional certificate is several times tighter than the general one where both are informative, and it remains the only informative bound where the general one is vacuous.

    \item \textbf{Practical guidance} (Section~\ref{sec:practitioner-guidelines}). We provide concrete guidance to practitioners on choosing the ambiguity geometry, encoding a directional prior, sizing the operating radius, and reading the two certificates together.
    
\end{itemize}

\section{Background}
\label{sec:background}

\subsection{Structural causal models}
\label{subsec:background-scm}
We briefly recall the Structural Causal Model (SCM) framework of
\citet{pearl2009causality}.

\begin{defn}\label{def:scm}
    A $d$-dimensional Structural Causal Model (SCM) is a pair $\scmsimple \coloneq (\cS, \rho)$, where $\cS = \langle \en, \ex, \structfuncset \rangle$ defines the deterministic \emph{causal basis}, consisting of a set of $d$ \emph{endogenous variables} $\en$, a set of \emph{exogenous variables} $\ex$, and a set of $d$ \emph{structural functions} $\structfuncset$, each defining the value of an endogenous variable as
    \begin{align}\label{eq:scm-definition}
    X_i=f_i(\operatorname{PA}(X_i), U_i) \quad \forall \; i \in [d],
    \end{align}
where $\operatorname{PA}(X_i) \subseteq \en \setminus {X_i}$ denotes the direct causes (parents) of $X_i$. For any variable or index set $V$, we write $\dom{V}$ for its \emph{domain} (state space), the set of values it may take; the environment $\rho$ is a joint probability distribution over $\ex$.
\end{defn}
The structural assignments induce a directed graph $\scmdag_\scmsimple$, called the \emph{causal graph} on nodes $X_1,\dots,X_d$ by placing an edge $X_k \to X_i$ whenever $X_k$ appears as an argument of $f_i$. Throughout the paper, we assume acyclic SCMs, so $\scmdag_\scmsimple$ is a directed acyclic graph (DAG). This allows us to recursively compose the structural functions into a single map $\mathbf{g}: \dom{\ex} \to \dom{\en}$, referred to as the \emph{mixing function}. This defines the SCM’s \emph{reduced form} $\en = \mathbf{g}(\ex)$, where endogenous variables are expressed purely in terms of exogenous noise. Consequently, the induced distribution is the pushforward $\mathbb{P}_{\scmsimple}(\en) = \mathbf{g}_{\#}(\rho)$, making explicit the generative process by which exogenous uncertainty propagates through the model. 

\subsection{Interventions and causal queries}
\label{subsec:background-interventions}
SCMs facilitate reasoning about \emph{interventions}. An exact intervention $\iota = \text{do}(\mathbf{A} = \textbf{a})$ fixes variables $\mathbf{A} \subseteq \en$ to values $\textbf{a}$, while allowing the rest of the system to evolve as usual. Graphically, this \emph{mutilates} $\scmdag_{\scmsimple}$ by removing incoming edges to $\mathbf{A}$. This yields a post-interventional SCM $\scmsimple_{\iota}$ with joint distribution $\prob_{\scmsimple_\iota}(\en)$. Whenever clear from the context, we write $\scmdo(\mathbf{a})$ as shorthand for $\scmdo(\mathbf{A}=\mathbf{a})$. For the purpose of our paper, we consider a \emph{set of relevant interventions} $\cI =\{ \iota_1,...\iota_k\}$ which restricts attention to interventions that are scientifically meaningful or practically feasible. To distinguish between the intervention itself and the variables whose post-intervention behaviour one ultimately cares about, we introduce the following object.
\begin{defn}[Query specification family]
Let $\cI$ be a set of relevant interventions. A \emph{query specification family} is an index set
\begin{align}\label{eq:query-spec-family}
  \Qcal=\{(\iota,O_\iota): \iota\in\cI,\; O_\iota \subseteq [d]\setminus J_\iota\},
\end{align}
where $J_\iota\subseteq [d]$ denotes the set of variables directly intervened upon by $\iota$, and $O_\iota$ specifies the post-intervention variables of interest under $\iota$. Throughout the paper, we assume $\Qcal$ is finite.
\end{defn}
When $O_\iota=[d]\setminus J_\iota$, this reduces to the full post-interventional case. The role of \(\Qcal\) is to specify which interventional marginals are relevant. It does not yet specify which numerical functional of those marginals will be evaluated. For each $(\iota,O_\iota)\in\Qcal$, the corresponding \emph{query-relevant interventional distribution} is the marginal
\begin{align}\label{eq:query-relevant-marginal}
  \pi_{O_\iota\#}\,\prob_{\scmsimple_\iota}(\en),
\end{align}
where $\pi_{O_\iota}$ denotes the projection onto the coordinates indexed by $O_\iota$.

\begin{defn}[Causal query]
A causal query is any measurable functional of a query-relevant interventional marginal. Concretely, if $\Phi$ is a measurable map on probability measures over $\dom{\en^{O_\iota}}$, then the causal query associated with $(\iota,O_\iota)\in\Qcal$ is
\begin{align}
  Q_{\scmsimple}^{(\iota,O_\iota)}
  :=
  \Phi\!\left(\pi_{O_\iota\#}\prob_{\scmsimple_\iota}(\en)\right).
\end{align}
\end{defn}
For instance, take $\iota=do(X{=}x)$ and $O_\iota=\{Y\}$. Then $\prob_{\scmsimple_\iota}(\en)$ is the joint post-interventional law over all variables, $\pi_{O_\iota\#}$ marginalises it to $Y$ alone, and choosing $\Phi$ to be the mean gives $Q_{\scmsimple}^{(\iota,O_\iota)}=\mathbb E[Y\mid do(X{=}x)]$. A different $\Phi$ on the same marginal, e.g. the median, the variance, gives a different query from the same interventional distribution, which is why the two are kept separate. This separation between \(\Qcal\) and the functional \(\Phi\) is important in the present work. The object we later optimise is not a single scalar query but the discrepancy between source and target \emph{interventional marginals} over a designated family \(\Qcal\). Query-level guarantees are then obtained by applying appropriate functionals \(\Phi_{(\iota,O_\iota)}\) to these marginals. Thus \(\Qcal\) specifies the interventional distributions that matter, while the query functional specifies how they are ultimately evaluated. Given a causal query, a central question in causal inference is identifiability: whether the quantity of interest can be re-expressed in terms of the distributions actually available. Specifically,
\begin{defn}[Identifiability]
Given a graph $\mathcal G$ and a class $\mathcal P$ of available distributions, a causal query $Q_{\scmsimple}^{(\iota,O_\iota)}$ is \emph{identifiable} from $\mathcal P$ if there is a functional $\Phi$ with $Q_{\scmsimple}^{(\iota,O_\iota)}=\Phi(\mathcal P)$ for every SCM consistent with $\mathcal G$ that induces the given distributions in $\mathcal P$.
\end{defn}
Here, \emph{consistent with $\mathcal G$} means the SCM has $\mathcal G$ as its causal graph and induces exactly the distributions in $\mathcal P$. Many such SCMs typically exist, and they may disagree on the interventional query; identifiability demands a single functional that returns the correct value for all of them, so the answer does not depend on which one generated the data. Thus, identifiability is a statement about whether the value of an interventional quantity of interest can be determined by the observational and interventional information available, together with the structural assumptions defining the model class. In the SCM framework, identifiability questions are typically addressed through \emph{do-calculus}, a collection of graphical rules that allows expressions involving interventional distributions to be reduced to functionals of observational or simpler interventional quantities when such a reduction is valid. A central result is that do-calculus is \emph{complete} for identifiability in nonparametric SCMs: whenever a causal query is identifiable from the available assumptions and data, there exists a derivation using the rules of do-calculus together with standard probability manipulations that recovers it. In this sense, do-calculus provides the canonical query-level language for deciding and deriving identifiability results \citep{shpitser2008complete}.

\subsection{Markov kernels}
\label{subsec:background-kernels}

Since our framework studies relations between interventional distributions, we need a general notion of transformation acting on probability measures. In general, such a transformation need not be deterministic. This is captured by the notion of a \emph{Markov kernel}.
\begin{defn}[Markov kernel]
Let $(\mathcal X,\mathcal A)$ and $(\mathcal Y,\mathcal B)$ be measurable spaces (each a set equipped with a $\sigma$-algebra). A Markov kernel from $\mathcal X$ to $\mathcal Y$ is a map $K:\mathcal X \times \mathcal B \to [0,1]$, such that:
\begin{enumerate}[label=(\roman*)]
    \item for every $x\in\mathcal X$, the map $B \mapsto K(x,B)$ is a probability measure on $(\mathcal Y,\mathcal B)$;
    \item for every $B\in\mathcal B$, the map $x \mapsto K(x,B)$ is $\mathcal A$-measurable.
\end{enumerate}
Equivalently, for each input $x\in\mathcal X$, the kernel $K(\cdot\mid x)$ specifies a probability distribution over outputs in $\mathcal Y$.
\end{defn}
A deterministic measurable map $\tau:\mathcal X\to\mathcal Y$ is a special case of a Markov kernel given by
\begin{align}
  K(\cdot\mid x)=\delta_{\tau(x)},
\end{align}
where $\delta_{\tau(x)}$ denotes the Dirac measure at $\tau(x)$. Thus, Markov kernels strictly generalise deterministic maps by allowing the output associated with a given input to be random rather than fixed.  If $\mu$ is a probability measure on $\mathcal X$, the image of $\mu$ under $K$ is the probability measure $\mu K$ on $\mathcal Y$ defined by
\begin{align}
  (\mu K)(B)=\int_{\mathcal X} K(x,B)\,\mu(dx),
  \qquad B\in\mathcal B.
\end{align}
When $K(\cdot\mid x)=\delta_{\tau(x)}$, this reduces to the usual pushforward $\tau_{\#}\mu$.

These preliminaries provide the language for the rest of the paper: SCMs specify the source and target domains, interventions generate the relevant post-interventional families, and Markov kernels provide the general transport operators relating them. We now use this language to reformulate transportability as a structural relation between models, rather than as a collection of query-specific formulas.

\section{Transportability as a structural relation between models}
\label{sec:structural-relation}
\subsection{Standard transportability setting}
\label{subsec:standard-transportability}
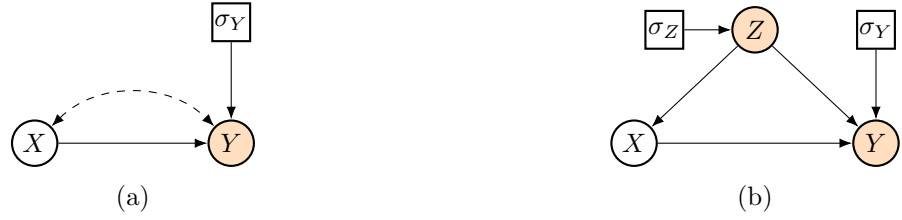
\begin{figure}[t]
\centering
\begin{subfigure}[b]{0.46\textwidth}
\centering
\begin{tikzpicture}[
    >=Latex,
    obs/.style={circle, draw, thick, minimum size=6mm, inner sep=0pt,
      font=\small},
    shifted/.style={circle, draw, thick, minimum size=6mm, inner sep=0pt,
      font=\small, fill=orange!25},
    sel/.style={rectangle, draw, thick, minimum size=5mm,
      inner sep=0pt, font=\small},
    every edge/.style={draw, ->, thick}
]
\node[obs]     (X)  at (-1.3,1) {$X$};
\node[shifted] (Y)  at ( 1.3,1) {$Y$};
\node[sel]     (Sy) at ( 1.3,2.6) {$\sigma_Y$};
\draw[->] (X) to (Y);
\draw[<->, dashed, bend left=45] (X) to (Y);
\draw[->] (Sy) to (Y);
\end{tikzpicture}
\subcaption{}
\label{fig:selection_ate}
\end{subfigure}
\hfill
\begin{subfigure}[b]{0.46\textwidth}
\centering
\begin{tikzpicture}[
    >=Latex,
    obs/.style={circle, draw, thick, minimum size=6mm, inner sep=0pt,
      font=\small},
    shifted/.style={circle, draw, thick, minimum size=6mm, inner sep=0pt,
      font=\small, fill=orange!25},
    sel/.style={rectangle, draw, thick, minimum size=5mm,
      inner sep=0pt, font=\small},
    every edge/.style={draw, ->, thick}
]
\node[shifted] (Z)  at ( 0   ,2.5) {$Z$};
\node[obs]     (X)  at (-1.6 ,1)   {$X$};
\node[shifted] (Y)  at ( 1.6 ,1)   {$Y$};
\node[sel]     (Sz) at (-1.2 ,2.5) {$\sigma_Z$};
\node[sel]     (Sy) at ( 1.6 ,2.5) {$\sigma_Y$};
\draw[->] (Z) to (X);
\draw[->] (Z) to (Y);
\draw[->] (X) to (Y);
\draw[->] (Sz) to (Z);
\draw[->] (Sy) to (Y);
\end{tikzpicture}
\subcaption{}
\label{fig:selection_atce}
\end{subfigure}
\caption{\emph{Selection diagrams}. Each panel shows one source-target pair: the two SCMs $\scmsimple_s$ and $\scmsimple_t$ share endogenous variables, causal graph, and intervention semantics, and differ only at the \emph{shifted} nodes (orange), where the mechanism, the exogenous noise, or both may vary across domains. A square selection node $\sigma_{(\cdot)}$ marks each. \textbf{(a)} A semi-Markovian example: the bidirected arc $X \leftrightarrow Y$ encodes an unobserved confounder, and the shifted effect node $Y$ varies in its mechanism $f_Y$ (the coefficient on its parent $X$) and/or its exogenous noise, while the mechanism of $X$ is invariant. \textbf{(b)} A Markovian example with two shifted nodes: the outcome $Y$ shifts through its mechanism $f_Y$, the coefficients on its parents $Z$ and $X$ and/or its exogenous noise, while the root $Z$, having no parents, shifts only through its exogenous law $f_Z(U_Z)$; the mechanism of $X$ is invariant.}
\label{fig:selection_diagrams}
\end{figure}
The standard setting of transportability imposes a precise shared structure on the two populations/domains: source and target are governed by SCMs $\scmsimple_s$ and $\scmsimple_t$ on the same set of endogenous variables $\en = (X_1,\dots,X_d)$, the same DAG $\scmdag_{\scmsimple_s} = \scmdag_{\scmsimple_t} =: \mathcal G$, and the same intervention semantics. They differ only in the mechanisms or exogenous distributions of a known subset of endogenous nodes $\mathcal K\subseteq[d]$, which we will call the \emph{shifted nodes}. Their local conditionals may therefore change across domains:
\begin{align}
  \prob_{\scmsimple_s}\!\bigl(X_i \mid \operatorname{PA}(X_i)\bigr)
  \neq
  \prob_{\scmsimple_t}\!\bigl(X_i \mid \operatorname{PA}(X_i)\bigr)
  \quad \text{for } i \in \mathcal K,
\end{align}
while for every \emph{invariant node} $i\notin\mathcal K$, the mechanisms coincide:
\begin{align}
  \prob_{\scmsimple_s}\!\bigl(X_i \mid \operatorname{PA}(X_i)\bigr)
  =
  \prob_{\scmsimple_t}\!\bigl(X_i \mid \operatorname{PA}(X_i)\bigr).
\end{align}
The shifted nodes are encoded graphically by a \emph{selection diagram} $\mathcal G^\sigma$ (Figure~\ref{fig:selection_diagrams}), which augments $\mathcal G$ with a selection variable $\sigma_i$ pointing into each $X_i$ with $i\in\mathcal K$. 

In the standard setting, the available information
consists of source observational data $ \mathcal D_s^{o}$, source interventional data $\mathcal D_s^{i}$, and target observational data $\mathcal D_t^{o}$. The \emph{transportability problem} then asks whether a target causal query can be identified as a functional of the available information \((\mathcal D_s^{o},\mathcal D_s^{i},\mathcal D_t^{o})\), together with the structural assumptions encoded by the selection diagram. For notational convenience, for each intervention $\iota\in\cI$, we write
\begin{align}
P_s^{(\iota)} := \prob_{(\scmsimple_s)_\iota}(\en),
  \qquad
  P_t^{(\iota)} := \prob_{(\scmsimple_t)_\iota}(\en).
\end{align}
In the classical full-joint formulation, a target causal query is typically written as:
\begin{align}
  Q_t^{(\iota)}
  := \Phi\!\left(P_t^{(\iota)}\right),
\end{align}
that is, as a functional of the full post-interventional distribution (or equivalently, of the distribution on the non-intervened variables). More generally, in the notation of Section~\ref{subsec:background-interventions}, one may regard this as the special case $O_\iota=[d]\setminus J_\iota$ of a query $Q_t^{(\iota,O_\iota)}$ based on a query-relevant marginal. The answer is given one query at a time: for a specific functional \(\Phi\), intervention \(\iota\in\cI\), and output set \(O_\iota\), either the target quantity can be written as an explicit expression in the available distributions \((\mathcal D_s^{o},\mathcal D_s^{i},\mathcal D_t^{o})\), a transport formula such as a back-door adjustment reweighted by the target covariate distribution, or no such expression exists. When such a formula exists, the query is said to be \emph{transportable}. The do-calculus on the selection diagram provides a complete graphical criterion for deciding transportability and deriving the formula when it exists \citep{pearl2014external}.

\subsection{A hierarchy of consistency notions}
\label{subsec:hierarchy}
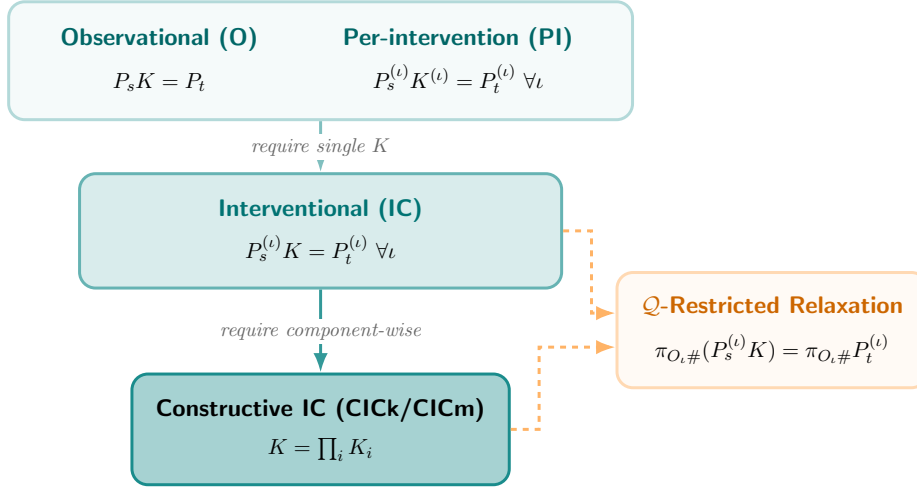
\begin{figure}[t]
\centering
\resizebox{0.8\columnwidth}{!}{%
\begin{tikzpicture}[>=Latex, thick, font=\sffamily]

  \tikzset{
    box/.style={rounded corners=8pt, align=center, inner sep=12pt, line width=1.5pt},
    lvl1/.style={box, draw=teal!30, fill=teal!4, minimum width=11.0cm},
    lvl2/.style={box, draw=teal!60, fill=teal!15, minimum width=8.5cm},
    lvl4/.style={box, draw=teal!90, fill=teal!35, minimum width=6.0cm},
    ortho/.style={box, draw=orange!30, fill=orange!4, minimum width=5.0cm},
    arrowLabel/.style={font=\small\itshape, text=gray!80!black, fill=white, inner sep=2pt}
  }

  
  \node[lvl1] (L1) at (0, 5.0) {
     \begin{tabular}{c@{\hspace{1.5cm}}c}
       \textcolor{teal!80!black}{\large\bfseries Observational (O)} & \textcolor{teal!80!black}{\large\bfseries Per-intervention (PI)} \\[2mm]
       $P_s K = P_t$ & $P_s^{(\iota)} K^{(\iota)} = P_t^{(\iota)}\ \forall\iota$
     \end{tabular}
  };

  \node[lvl2] (L2) at (0, 2.0) {
     \textcolor{teal!90!black}{\large\bfseries Interventional (IC)}\\[2mm]
     $P_s^{(\iota)} K = P_t^{(\iota)}\ \forall\iota$
  };

  \node[lvl4] (L4) at (0, -1.5) {
     \textcolor{black}{\large\bfseries Constructive IC (CICk/CICm)}\\[2mm]
     $K=\prod_i K_i$
  };

  \draw[->, teal!50, line width=1.5pt] (L1) -- node[arrowLabel]{require single $K$} (L2);
  \draw[->, teal!80, line width=1.5pt] (L2) -- node[arrowLabel]{require component-wise} (L4);

  
  \node[ortho, anchor=west] (Orth) at (5.2, 0.25) {
     \textcolor{orange!80!black}{\large\bfseries $\Qcal$-Restricted Relaxation}\\[3mm]
     \normalsize $\pi_{O_\iota\#}(P_s^{(\iota)}K)=\pi_{O_\iota\#}P_t^{(\iota)}$
  };

  \draw[->, orange!60, dashed, line width=1.5pt] (L2.east) -- ++(0.5,0) |- ([yshift=3mm]Orth.west);
  \draw[->, orange!60, dashed, line width=1.5pt] (L4.east) -- ++(0.5,0) |- ([yshift=-3mm]Orth.west);

\end{tikzpicture}%
}
\caption{The hierarchy of source-target consistency notions. Constraint strength increases downward along the spine (the narrower the box,
the more constraining the notion) and, orthogonally, from the $\Qcal$-restricted relaxation toward the full-family notions on the left. (O) and (PI) hold for any pair of models; the real constraint begins at interventional consistency (IC). Constructive interventional consistency (CICk/CICm) is the strongest and the central object of Section~\ref{sec:characterizing}. In an orthogonal dimension, $\mathcal{Q}$-restricted consistency (right) relaxes any of the spine notions to hold only on designated query marginals; it is informative precisely at IC and Constructive IC, since O and PI already hold for every model pair and their relaxations are therefore vacuous.}
\label{fig:hierarchy}
\end{figure}
This query-level viewpoint is natural from the perspective of identification, but it leaves open a more structural question. In the transportability setting, the source and target domains are not arbitrary unrelated SCMs. They already share the same endogenous variables, the same graph, and the same intervention semantics, differing only through the mechanisms indexed by the selection nodes, as illustrated in Figure~\ref{fig:selection_diagrams}. This means that the two models come equipped with a canonical variable-by-variable and intervention-by-intervention correspondence. Once this shared scaffold is fixed, it becomes natural to ask not only whether a given query on the target can be transported from the source, but also whether the two models themselves are related by a common structural relation that explains such transport as a consequence. This leads to a shift in perspective from \emph{query transport} to \emph{model-level transport}. Instead of asking for a separate transport formula for each target quantity, we ask whether there exists a single relation between $\scmsimple_s$ and $\scmsimple_t$ that aligns them in a meaningful way. The central question of the paper is therefore the following:
\begin{tcolorbox}[
  colback=purple!8,
  colframe=gray!50,
  borderline west={2pt}{0pt}{black},
  boxrule=0pt,
  sharp corners
]
\begin{center}
    \emph{When is transportability supported by a common, model-level,  structural relation between the source and target models themselves?}
\end{center}
\end{tcolorbox}
Equivalently, can the transport of target queries be understood as a consequence of a single model-level object rather than a collection of unrelated query-specific formulas? In the strongest form considered in this paper, this means asking whether there exists one common map sending the full source interventional family $\{P_s^{(\iota)}\}_{\iota\in\cI}$ to the target interventional family $\{P_t^{(\iota)}\}_{\iota\in\cI}$. This question is strictly stronger than the classical one: even if many individual queries are transportable, there may still be no single common structural transformation explaining them simultaneously.

Understanding when such a common relation exists, when it fails, and what it means for the source and target to be aligned at the model level is the starting point for the hierarchy developed next. We therefore define four increasingly structured notions of relation between the two models, together with an orthogonal relaxation. Along the main chain, each step imposes an additional structural requirement on how the source and target are allowed to be related. At the weakest level, one may ask only whether some operator relates the observational distributions, with no causal or mechanistic structure imposed.

\begin{defn}[Observational consistency]
$\scmsimple_s$ is observationally consistent with $\scmsimple_t$ if there exists a Markov kernel $K : \dom{\en} \times \mathcal B(\dom{\en}) \to [0,1]$ such that
\begin{align}
  P_s K = P_t.
\end{align}
\end{defn}
A stronger notion requires consistency not only at the observational level but also separately for each intervention in the family $\cI$.

\begin{defn}[Per-intervention consistency]
$\scmsimple_s$ is per-intervention consistent with $\scmsimple_t$ over $\cI$ if, for every $\iota \in \cI$, there exists a Markov kernel $K^{(\iota)} : \dom{\en} \times \mathcal B(\dom{\en}) \to [0,1]$ such that
\begin{align}
  P_s^{(\iota)} K^{(\iota)} = P_t^{(\iota)}.
\end{align}
\end{defn}
Here, each intervention is treated independently, and the operator may vary with $\iota$.

\begin{lemma}
\label{prop:pairwise-kernel-always-exists}
For any two probability measures $\mu$ and $\nu$ on $\dom{\en}$, there exists a Markov kernel $K$ with $\mu K = \nu$. Consequently, observational and per-intervention consistency always hold between $\scmsimple_s$ and $\scmsimple_t$ over any $\cI$.
\end{lemma}
\begin{proof}
Define the constant kernel
\begin{align}
K(x,B):=\nu(B),~\; x\in \dom{\en},~\; B\in \mathcal B(\dom{\en}).
\end{align}
For each fixed \(x\), the map \(B\mapsto K(x,B)\) is a probability measure on \(\dom{\en}\), and for each fixed measurable set \(B\), the map \(x\mapsto K(x,B)\) is constant; hence, it is measurable. Thus, \(K\) is a Markov kernel. Applying \(K\) to \(\mu\) gives
\begin{align}
(\mu K)(B)
=
\int_{\dom{\en}} K(x,B)\,\mu(dx)
=
\int_{\dom{\en}} \nu(B)\,\mu(dx)
=
\nu(B),
\end{align}
for every \(B\in \mathcal B(\dom{\en})\). Hence, \(\mu K=\nu\). Now it is straightforward to see that taking \(\mu=P_s\) and \(\nu=P_t\) proves observational consistency, and taking \(\mu=P_s^{(\iota)}\) and \(\nu=P_t^{(\iota)}\) for each \(\iota\in\cI\) proves per-intervention consistency.
\end{proof}

\begin{remark}[Deterministic maps]
\label{rem:deterministic-maps}
Deterministic pairwise transport \(\tau_{\#}\mu=\nu\) is not automatic for arbitrary pairs \((\mu,\nu)\): its existence requires additional structure on the source-target pair. On standard Borel spaces, nonatomicity of \(\mu\) is a sufficient condition: any atomless standard probability space is Borel-isomorphic to \(([0,1],\mathrm{Leb})\) \citep[Thm.~17.41]{alma991025533659706532}, and any target law \(\nu\) on a standard Borel space is the pushforward of Lebesgue measure under some Borel map. Hence, there exists a measurable \(\tau\) with \(\tau_{\#}\mu=\nu\). In Euclidean settings, stronger results are available from optimal transport: under quadratic cost, if \(\mu\) is absolutely continuous with respect to Lebesgue measure, Brenier's theorem yields an optimal deterministic transport map \citep{brenierthm, villani2009optimal}. Thus, unlike a common kernel, which always exists, a deterministic map exists only under such regularity conditions, so it already reflects real structure in the pair $(\mu,\nu)$.
\end{remark}
That said, observational consistency and per-intervention consistency remain pairwise, distribution-level notions. They allow the transport operator to vary freely across interventions and therefore do not yet express a single relation between the two models. The first genuinely model-level notion requires \emph{the same} operator to work simultaneously across the full interventional family.

\begin{defn}[Interventional consistency]\label{def:interventional-consistency}
$\scmsimple_s$ is \emph{interventionally consistent} with $\scmsimple_t$ over $\cI$ if there exists a single Markov kernel $K : \dom{\en} \times \mathcal B(\dom{\en}) \to [0,1]$, such that
\begin{align}
  P_s^{(\iota)} K = P_t^{(\iota)}
  \qquad \forall\, \iota \in \cI.
\end{align}
The deterministic version restricts to kernels of the form
\begin{align}
      K(\cdot \mid x) = \delta_{\tau(x)},
\end{align}
for some measurable map $\tau : \dom{\en} \to \dom{\en}$.
\end{defn}
Interventional consistency upgrades per-intervention consistency by replacing the family $\{K^{(\iota)}\}_{\iota\in\cI}$ with a single common operator. This is the first notion that is not automatically achievable: the existence of a common $K$ imposes a genuine compatibility constraint between the source and target interventional families and is precisely the central consistency requirement that appears in the Causal Abstraction (CA) literature \citep{rubenstein2017causal, beckers2018abstracting}. CA theory studies when one SCM $\cM^h$ can serve as a causally faithful higher-level representation of another SCM $\cM^\ell$, typically at a different level of granularity. The key requirement is that interventions and abstraction commute, so that abstracting after intervening gives the same result as intervening after abstracting. 

\begin{defn}[Exact transformation]
\label{def:exact-transformation}
Let $\scmbase$ and $\scmabst$ be SCMs with respective intervention sets $\intervsetbase$ and $\intervsetabst$, and let $\omega : \intervsetbase \to \intervsetabst$ be a surjective order-preserving map between interventions. A measurable map $\tau : \dom{\enbase} \to \dom{\enabst}$ is an exact transformation if
\begin{align}\label{eq:exact_tranformation}
  \tau_{\#}\bigl(\prob_{\scmbase_{\iota}}(\enbase)\bigr)
  =
  \prob_{\scmabst_{\omega(\iota)}}(\enabst)
  \qquad \forall\, \iota \in \intervsetbase.
\end{align}
\end{defn}
Definition~\ref{def:exact-transformation} is the abstraction-theoretic version of interventional consistency: a single deterministic map $\tau$ must align the entire family of interventional distributions, while the map $\omega$ specifies which intervention at the lower level corresponds to which intervention at the higher level. In other words, the abstraction must preserve not just observational behaviour but the full interventional semantics of the model via a single deterministic map $\tau$. 

Our transportability setting can now be understood as a special case of this general abstraction framework. Indeed, source and target SCMs are defined on the \emph{same} endogenous variables, with the \emph{same} graphical scaffold and the \emph{same} intervention semantics. Hence, there is no change of granularity, and no nontrivial intervention translation is needed. Formally, this means:
\begin{align}\label{eq:abstraction-correspondace}
    \dom{\enbase} = \dom{\enabst}, \qquad \intervsetbase = \intervsetabst, \qquad \omega = \mathrm{id}_{\cI}
\end{align}
Under Equations~\ref{eq:abstraction-correspondace}, the exact transformation condition (Equation~\ref{eq:exact_tranformation}) becomes $\tau_{\#} P_s^{(\iota)} = P_t^{(\iota)},~ \forall \iota\in\cI$, which is exactly the deterministic form of interventional consistency from Definition~\ref{def:interventional-consistency}. Thus, the transportability problem studied here is an \emph{automorphic} or \emph{same-level} instance of causal abstraction: rather than relating a fine-grained SCM to a coarser one, we relate two SCMs on the same variable space that differ only through the mechanisms indexed by the selection diagram. This shows that our model-level notion of transportability is not separate from causal abstraction theory, but rather a structurally specialised instance of it.

The correspondence above is stated for deterministic exact transformations. Our framework extends it by allowing a common Markov kernel in place of a single measurable map. This is also the level at which it connects to optimal-transport-based approaches to causal abstraction learning.
\begin{remark}[Connection to Optimal Transport couplings]
\label{rem:kernels-couplings-cota}
A Markov kernel may be viewed as the conditional form of a coupling. In particular, if \(\pi\) is a coupling of probability measures \(\mu\) and \(\nu\) on standard Borel spaces, then \(\pi\) admits a disintegration with respect to its first marginal:
\begin{align}
  \pi(dx,dy)=\mu(dx)\,K(x,dy),
\end{align}
for some Markov kernel \(K\), and therefore \(\nu=\mu K\). In the finite discrete case, this reduces to $P(x,y)=\mu(x)\,K(y\mid x)$, so row-normalising a transport plan yields a stochastic map. Thus, the stochastic abstraction maps learned from Optimal Transport ~\citep{villani2009optimal} plans in \citet{felekis2024causal} can be viewed as finite-sample instances of the kernel-valued transport operators considered here. The present paper uses such operators directly at the level of interventional families, rather than a single empirical coupling.
\end{remark}
In the general CA literature, the map \(\tau\) may merge several low-level variables into one high-level variable, split one low-level variable across several high-level ones, or otherwise change the level of representation. An important subclass is the \emph{constructive} one, in which each high-level variable depends only on a dedicated, non-overlapping subset of low-level variables.

\begin{defn}[Constructive exact transformation]
\label{def:constructive-exact-transformation-ca}
Let $\en^h=(X^h_1,\dots,X^h_m)$ be the endogenous variables of the high-level model. An exact transformation $(\tau,\omega)$ from $\scmsimple^\ell$ to $\scmsimple^h$ is called constructive if there exists a partition $\Pi=\{B_1,\dots,B_m,B_0\}$ of the low-level variable indices, with $B_1,\dots,B_m$ nonempty, such
that for each $k=1,\dots,m$ there exists an exact transformation
\begin{align}
  \tau_k:\dom{X_{B_k}^\ell}\to\dom{X_k^h},
  \qquad
  \text{where }
  \dom{X_{B_k}^\ell}:=\prod_{j\in B_k}\dom{X_j^\ell},
\end{align}
satisfying
\begin{align}
  \tau(x)
  =
  \bigl(
    \tau_1(x_{B_1}),
    \dots,
    \tau_m(x_{B_m})
  \bigr)
  \qquad \forall x\in\dom{\en^\ell}.
\end{align}
The block $B_0$ (possibly empty) collects low-level variables that are marginalised away.
\end{defn}
\begin{remark}
One may additionally require the intervention map $\omega$ to be constructive, in the sense that interventions on $X_k^h$ depend only on interventions acting on variables in $B_k$. This is automatic in our setting, where $\omega=\mathrm{id}_{\cI}$.
\end{remark}
Definition~\ref{def:constructive-exact-transformation-ca} captures the general many-to-one setting of causal abstraction, where a single high-level variable summarises an entire block of low-level variables. In the transportability setting, by contrast, constructiveness becomes the natural same-level specialisation of this idea: the partition reduces to singletons \(B_i=\{i\}\) for \(i\in[d]\), with \(B_0=\emptyset\), and \(\omega=\mathrm{id}_{\cI}\), so constructiveness is automatic. Since source and target differ only through the mechanisms indexed by \(\mathcal K\), the relevant transformation should act locally on the variables whose mechanisms may shift, while leaving invariant variables untouched. In this sense, constructiveness is not an external modelling choice but the canonical form that interventional consistency takes in the transportability setting.

\begin{remark}
\label{rem:constructive-not-collapse}
Constructiveness is the form that the transportability setting motivates, not the only one possible. Interventional consistency asks for some common $K$; that does not necessarily factorise; thus, non-constructive solutions can also exist in this setting, and therefore, the two notions remain distinct.
\end{remark}

\begin{defn}[Constructive interventional consistency]
$\scmsimple_s$ is constructively interventionally consistent with $\scmsimple_t$ over $\cI$ if it is interventionally consistent and the common kernel $K$ factorises component-wise:
\begin{align}
  K(d\mathbf{x}' \mid \mathbf{x})
  =
  \prod_{i=1}^d K_i(dx_i' \mid x_i),
\end{align}
with
\begin{align}
  K_i(\cdot \mid x_i)=\delta_{x_i}
  \qquad \text{for every invariant node } i\notin\mathcal K.
\end{align}
The deterministic version is the special case in which
\begin{align}
  K_i(\cdot \mid x_i)=\delta_{\tau_i(x_i)}
\end{align}
for measurable maps $\tau_i : \dom{X_i}\to\dom{X_i}$.
\end{defn}
We call such a $K$ constructively interventionally consistent kernel (CICk) and the deterministic equivalent constructively interventionally consistent map (CICm). Alongside this main hierarchy, there is also a query-family-restricted relaxation of interventional consistency. Instead of requiring a common operator to align the full post-interventional joint for each intervention, one may require it to align only the query-relevant marginals specified by a query specification family $\Qcal$. This restriction is orthogonal to the constructive or not distinction: a common operator may be $\Qcal$-restricted with or without being constructive.

\begin{defn}[$\Qcal$-restricted interventional consistency]
Let $\Qcal$ be a query specification family over $\cI$. We say that $\scmsimple_s$ is \emph{$\Qcal$-restricted interventionally consistent} with $\scmsimple_t$ if there exists a single Markov kernel $K:\dom{\en}\times\mathcal B(\dom{\en})\to[0,1]$ such that
\begin{align}
  \pi_{O_\iota\#}\bigl(P_s^{(\iota)}K\bigr)
  =
  \pi_{O_\iota\#}P_t^{(\iota)}
  \qquad \forall\,(\iota,O_\iota)\in\Qcal.
\end{align}
The deterministic version restricts to kernels of the form
\begin{align}
  K(\cdot\mid x)=\delta_{\tau(x)}
\end{align}
for some measurable map $\tau:\dom{\en}\to\dom{\en}$.
\end{defn}

A \emph{constructive $\Qcal$-restricted interventional consistency} condition is obtained by adding to this definition the same componentwise factorisation requirement as in constructive interventional consistency. Thus, $\Qcal$-restricted consistency weakens what must be matched, while constructiveness restricts how the common operator is allowed to act.

\begin{remark}
The two extremes of $\Qcal$-restricted interventional consistency recover two endpoints of interest. Taking $O_\iota = [d]\setminus J_\iota$ for every $\iota\in\cI$ induces the full post-interventional joint and recovers interventional consistency. At the other extreme, taking $\Qcal$ to be a singleton isolates the single marginal on which a query of interest lives.
\end{remark}

Constructive interventional consistency is the strongest notion in the main hierarchy considered here. Once such a common operator exists, it simultaneously aligns the full source interventional family with the target one; therefore, the transportability of all queries follows immediately.

\begin{remark}
\label{rem:common-transport-implies-all-queries}
If $P_s^{(\iota)} K = P_t^{(\iota)}, \forall \iota \in \cI$, then for every query family $\Qcal$, every measurable functional $\Phi$, and every $(\iota,O_\iota)\in\Qcal$, we have
\begin{align}
  \Phi\!\left(\pi_{O_\iota\#}P_t^{(\iota)}\right)
  =
  \Phi\!\left(\pi_{O_\iota\#}(P_s^{(\iota)}K)\right).
\end{align}
Thus, full interventional consistency implies the transport of all query-restricted causal queries simultaneously. More generally, if $K$ is only $\Qcal$-restricted interventionally consistent, then the same conclusion holds for all $(\iota,O_\iota)\in\Qcal$, but not necessarily outside $\Qcal$. However, the converse fails in general: even if all queries in a given family are transportable, there may be no single common operator that aligns the full interventional family. This failure is what we call the \emph{compatibility gap}: it separates query-level transportability from model-level transportability.
\end{remark}
Constructive interventional consistency is thus the strongest member of the hierarchy. In the following section, we study when this holds and how it can be characterised.

\section{Characterising Exact Model-level Transportability}
\label{sec:characterizing}
Section~\ref{sec:structural-relation} introduced the hierarchy of model-level transport notions and isolated constructive interventional consistency as the central exact notion. We now ask when such a common constructive transport operator exists, first for Markovian SCMs and then extending the analysis to the semi-Markovian case.

\subsection{The Markovian Setting}
\label{subsec:markovian-setting}
We first specialise to the case where both the source and target SCMs, \(\scmsimple_s=(\cS_s,\rho_s)\) and \(\scmsimple_t=(\cS_t,\rho_t)\), are \emph{Markovian}. Thus, both models are acyclic, and their exogenous variables are jointly independent; i.e., $\rho_\alpha=\bigotimes_{i=1}^d \rho_{\alpha,i},~ \alpha\in\{s,t\}$. Joint independence implies causal sufficiency, so there are no latent common causes among the endogenous variables. We assume that \(\scmsimple_s\) and \(\scmsimple_t\) are defined on the same endogenous variable set \(\en\) and entail the same DAG \(\scmdag\). Throughout, we index the endogenous variables $X_1,\dots,X_d$ in a fixed topological order of $\scmdag$ so that $\operatorname{PA}(X_i)\subseteq\{X_1, \dots,X_{i-1}\}$. Under this numbering, ``variable $X_i$'', ``coordinate $i$'', and ``stage $i$'' of the assembly all refer to the same index; we use the terms interchangeably. Markovianity yields the usual DAG factorisation:
\begin{equation}
  P_\alpha(\mathbf{x})
  =
  \prod_{i=1}^d
  \prob_{\scmsimple_\alpha}
  \bigl(x_i \mid \operatorname{PA}(X_i)=x_{\operatorname{PA}(X_i)}\bigr),
  \qquad \alpha\in\{s,t\}.
\end{equation}
For every invariant node \(i\notin\mathcal K\), $\prob_{\scmsimple_s}\!\bigl(X_i \mid \operatorname{PA}(X_i)\bigr)=\prob_{\scmsimple_t}\!\bigl(X_i \mid \operatorname{PA}(X_i)\bigr)$. Now fix an intervention \(\iota=\scmdo(\mathbf A=\mathbf a)\in\cI\), and let \(J_\iota\subseteq[d]\) denote the index set of the variables in \(\mathbf A\). In the Markovian case, the interventional distribution admits the \emph{truncated factorisation} \citep{pearl2009causality}:
\begin{equation}
  P_\alpha^{(\iota)}(\mathbf{x})
  =
  \prod_{i\notin J_\iota}
  \prob_{\scmsimple_\alpha}
  \bigl(x_i \mid \operatorname{PA}(X_i)=x_{\operatorname{PA}(X_i)}\bigr)
  \cdot
  \prod_{i\in J_\iota}\delta_{a_i}(x_i),
  \qquad \alpha\in\{s,t\},
\end{equation}
where \(a_i\) is the intervention value assigned to \(X_i\). Hence, under a given intervention, the only non-intervened factors that can differ between source and target are those associated with shifted nodes \(i\in\mathcal K\setminus J_\iota\).

\subsubsection{Relevant parent contexts}
\label{subsec:relevant-contexts-markovian}
Classical transportability reasons at the level of target queries of  a shifted node \(X_i\) rather than at the level of its mechanisms. Here we ask a different and stronger question: whether the \emph{mechanism} of \(X_i\), namely its conditional law \(\prob(X_i\mid\operatorname{PA}(X_i))\), can be aligned across domains by a single local transport component. Since a mechanism is indexed by parent values, this is not a statement about \(X_i\) in isolation: it must hold at every parent configuration on which the mechanism is actually evaluated. The intervention family determines exactly which parent values can actually occur while the mechanism remains active. An intervention on an ancestor of \(X_i\) can drive its parents into values never seen observationally, enlarging the set that must be matched; an intervention on \(X_i\) itself switches the mechanism off entirely. We therefore isolate exactly the parent configurations that matter.
\begin{defn}[Relevant parent contexts]
\label{def:relevant-contexts-markovian}
Fix a shifted node \(i\in\mathcal K\), and for each \(\iota\in\cI\), let \(J_\iota\) be its intervened set. The set of \emph{relevant parent contexts} for \(X_i\) is
\begin{align}
  \mathcal P_i^{\mathrm{rel}}
  :=
  \bigcup_{\substack{\iota\in\cI\\ i\notin J_\iota}}
  \left(
    \mathrm{supp}\!\Bigl(P_s^{(\iota)}\bigl(\operatorname{PA}(X_i)\bigr)\Bigr)
    \;\cup\;
    \mathrm{supp}\!\Bigl(P_t^{(\iota)}\bigl(\operatorname{PA}(X_i)\bigr)\Bigr)
  \right).
\end{align}
\end{defn}
The set \(\mathcal P_i^{\mathrm{rel}}\) collects all the parent configurations at which the native mechanism of \(X_i\) is queried under interventions in \(\cI\); this is the domain on which local source-target alignment of that mechanism must be checked. If \(i\in J_\iota\) the intervention fixes \(X_i\) and bypasses its mechanism, replacing it with a point mass at the intervention value, so such regimes impose no condition on the local component \(K_i\) or \(\tau_i\).

As a concrete instance, take a chain \(X_1\to X_2\) with \(X_2\) shifted. Comparing the observational conditionals \(\prob_{\scmsimple_s}(X_2\mid X_1)\) and \(\prob_{\scmsimple_t}(X_2\mid X_1)\) only on the naturally observed support of \(X_1\) can miss parent values that arise once one intervenes on \(X_1\); since constructive interventional consistency must hold over all of \(\cI\), the mechanism of \(X_2\) must be aligned on every such context while \(X_2\) remains active.

If every parent of a shifted node \(X_i\) is invariant, alignment is direct: one asks whether the source mechanism for \(X_i\), composed with a local component \(K_i\), matches the target mechanism on \(\mathcal P_i^{\mathrm{rel}}\). The difficulty lies when a parent of \(X_i\) is itself shifted. By the time the mechanism of \(X_i\) is reached, its parents no longer follow the source law, and they have been modified by the earlier components, so the stage-\(i\) condition must be imposed against the \emph{partially transformed} parent law rather than the original source conditionals. This is what makes the problem recursive. However, recursion alone is still not enough: a general componentwise kernel can destroy the Markov factorisation of the intermediate laws. Thus, besides recursive alignment, one must require that the partially transformed laws retain the relevant Markov structure.

\subsubsection{Mechanism-level characterisation and recursive assembly}
\label{subsec:mechanism-characterization-markovian}
The idea now is simple: by stage \(i-1\) all parents of \(X_i\) have already been transported, so if \(K_i\) is chosen to give \(X_i\) the correct target conditional given those transported parents, the first \(i\) variables match the target law. Iterating down the topological order assembles the full target joint, one mechanism at a time. Making this precise requires two pieces of bookkeeping: a notation for applying the local components one coordinate at a time, and a condition ensuring that each intermediate law still presents the next mechanism as a conditional on its parent tuple.

For each \(i\in[d]\), let \(\bar K_i\) denote the lifted kernel on the full state space that acts as \(K_i\) on coordinate \(i\) and as the identity on all other coordinates:
\begin{align}
  \bar K_i(d\mathbf{x}'\mid \mathbf{x})
  :=
  K_i(dx_i'\mid x_i)\prod_{j\neq i}\delta_{x_j}(dx_j').
\end{align}
Since these lifted kernels act on disjoint coordinates, they commute, and their composition is exactly the componentwise kernel $K(d\mathbf{x}'\mid \mathbf{x}) = \prod_{i=1}^d K_i(dx_i'\mid x_i)$.
\begin{defn}
\label{def:stagewise-markov-preserving}
A componentwise kernel $K(d\mathbf{x}'\mid \mathbf{x})=\prod_{i=1}^d K_i(dx_i'\mid x_i)$ is called \emph{stagewise Markov-preserving over \(\cI\)} if, for every intervention \(\iota\in\cI\), with
\begin{align*}
  R_0^{(\iota)} := P_s^{(\iota)},
  \qquad
  R_i^{(\iota)} := R_{i-1}^{(\iota)}\bar K_i,
  \qquad i=1,\dots,d,
\end{align*}
one has, for every node \(i\notin J_\iota\),
\begin{align}
  R_i^{(\iota)}\!\bigl(X_i \mid X_1,\dots,X_{i-1}\bigr)
  =
  R_i^{(\iota)}\!\bigl(X_i \mid X_{\operatorname{PA}(X_i)}\bigr)
  \qquad
  R_i^{(\iota)}\text{-a.s.}
\end{align}
\end{defn}

\begin{theorem}[Recursive global assembly]
\label{thm:mechanism-characterization-markovian}
Let $K$ be a stagewise Markov-preserving componentwise Markov kernel such that \(K_i(\cdot\mid x_i)=\delta_{x_i}\) for every invariant node \(i\notin\mathcal K\). For each intervention \(\iota\in\cI\), set $R_0^{(\iota)} := P_s^{(\iota)}$ and $R_i^{(\iota)} := R_{i-1}^{(\iota)}\bar K_i$ for $i=1,\dots,d$. Then \(K\) is a CICk from \(\scmsimple_s\) to \(\scmsimple_t\) over \(\cI\) if and only if, for every intervention \(\iota\in\cI\) and every node \(i\in[d]\), the partially transformed law \(R_i^{(\iota)}\) satisfies:
\begin{enumerate}[label=(\roman*)]
  \item if \(i\in J_\iota\), then $R_i^{(\iota)}(X_i)=\delta_{a_i}$;

  \item if \(i\notin J_\iota\), then $R_i^{(\iota)}\!\bigl(X_i \mid X_{\operatorname{PA}(X_i)}\bigr)
    =
    \prob_{\scmsimple_t}\!\bigl(X_i \mid X_{\operatorname{PA}(X_i)}\bigr)~~
    R_i^{(\iota)}\text{-a.s.}$
\end{enumerate}
The deterministic analogue holds with \(K_i(\cdot\mid x_i)=\delta_{\tau_i(x_i)}\).
\end{theorem}
\begin{figure}[t]
\centering
\resizebox{\columnwidth}{!}{%
\begin{tikzpicture}[
    >=Latex,
    font=\sffamily,
    node distance=1.4cm,
    src/.style   ={circle, draw, thick, fill=blue!10,   minimum size=8mm, inner sep=0pt, font=\small},
    tgt/.style   ={circle, draw, thick, fill=orange!25, minimum size=8mm, inner sep=0pt, font=\small},
    act/.style   ={circle, draw=red!70!black, line width=1.2pt, fill=orange!25, minimum size=8mm, inner sep=0pt, font=\small},
    stage/.style ={font=\bfseries\small, anchor=south},
    tag/.style   ={font=\scriptsize\itshape, align=center, anchor=north},
    every edge/.style={draw, ->, thick, >=Latex}
]

\begin{scope}[shift={(0,0)}]
  \node[stage] at (1, 1.45) {Stage 0};
  \node[src] (a1) at (0, 0.6) {$X_1$};
  \node[src] (a2) at (1, 0.6) {$X_2$};
  \node[src] (a3) at (2, 0.6) {$X_3$};
  \draw[->] (a1) -- (a2);
  \draw[->] (a2) -- (a3);
  \node[tag] at (1, 0.0) {$R_0^{(\iota)} = P_s^{(\iota)}$};
\end{scope}

\draw[->, line width=1pt] (2.55, 0.6) -- node[above, font=\scriptsize] {$\bar K_1$} (3.45, 0.6);

\begin{scope}[shift={(4,0)}]
  \node[stage] at (1, 1.45) {Stage 1};
  \node[act] (b1) at (0, 0.6) {$X_1$};
  \node[src] (b2) at (1, 0.6) {$X_2$};
  \node[src] (b3) at (2, 0.6) {$X_3$};
  \draw[->] (b1) -- (b2);
  \draw[->] (b2) -- (b3);
  \node[tag] at (1, 0.0) {$R_1^{(\iota)}$};
\end{scope}

\draw[->, line width=1pt] (6.55, 0.6) -- node[above, font=\scriptsize] {$\bar K_2$} (7.45, 0.6);

\begin{scope}[shift={(8,0)}]
  \node[stage] at (1, 1.45) {Stage 2};
  \node[tgt] (c1) at (0, 0.6) {$X_1$};
  \node[act] (c2) at (1, 0.6) {$X_2$};
  \node[src] (c3) at (2, 0.6) {$X_3$};
  \draw[->] (c1) -- (c2);
  \draw[->] (c2) -- (c3);
  \node[tag] at (1, 0.0) {$R_2^{(\iota)}$};
\end{scope}

\draw[->, line width=1pt] (10.55, 0.6) -- node[above, font=\scriptsize] {$\bar K_3$} (11.45, 0.6);

\begin{scope}[shift={(12,0)}]
  \node[stage] at (1, 1.45) {Stage 3};
  \node[tgt] (d1) at (0, 0.6) {$X_1$};
  \node[tgt] (d2) at (1, 0.6) {$X_2$};
  \node[act] (d3) at (2, 0.6) {$X_3$};
  \draw[->] (d1) -- (d2);
  \draw[->] (d2) -- (d3);
  \node[tag] at (1, 0.0) {$R_3^{(\iota)} = P_t^{(\iota)}$};
\end{scope}

\end{tikzpicture}%
}
\caption{\emph{Recursive global assembly (Theorem~\ref{thm:mechanism-characterization-markovian}).}
Schematic illustration of the induction behind the theorem on a chain \(X_1\to X_2\to X_3\). At stage \(i\), the lifted kernel \(\bar K_i\) acts on coordinate \(i\) while the others are held fixed. Nodes shaded in blue still carry the source-side law, nodes shaded in orange already agree with the target law, and the red outline marks the coordinate updated at the current stage. The partially transformed laws \(R_i^{(\iota)}\) satisfy the induction invariant that, after stage \(i\), the prefix \((X_1,\dots,X_i)\) agrees with the target interventional law \(P_t^{(\iota)}\); by stage \(d\), the full target joint is recovered.}
\label{fig:recursive-assembly}
\end{figure}
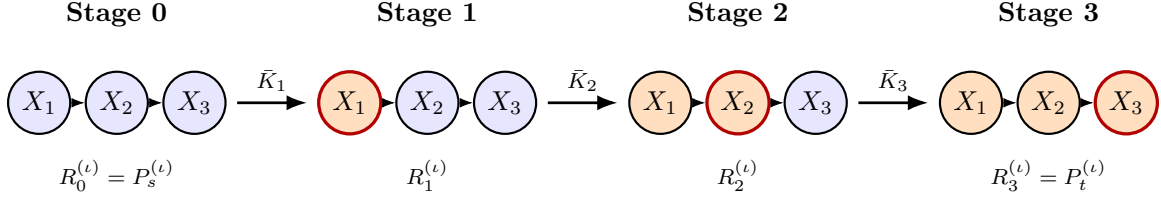
The proof formalises the mechanism-by-mechanism assembly described above, made visible in Figure~\ref{fig:recursive-assembly}: one proceeds by induction along the topological order, showing that matching the stagewise conditional at node $i$ aligns the first $i$ variables with the target law. We defer the full argument to Appendix~\ref{app:markovian-proofs}.

\begin{remark}
\label{rem:markov-preserving-free}
The stagewise Markov-preserving assumption excludes no genuine CICks. Indeed, if a componentwise kernel \(K\) is a CICk, then in the forward direction above one has
\begin{align}
  R_i^{(\iota)}(X_1,\dots,X_i)=P_t^{(\iota)}(X_1,\dots,X_i)
  \qquad \forall\,i,\iota,
\end{align}
and since \(P_t^{(\iota)}\) is Markov with respect to the DAG, \(R_i^{(\iota)}\) inherits the same node-\(i\) conditional independence. Thus, a CICk exists in the full componentwise class if and only if it exists in the stagewise Markov-preserving componentwise class.
\end{remark}

\begin{remark}
\label{rem:markov-preserving-sufficient}
The stagewise Markov-preserving property is automatic in several regimes. First, if no shifted node is an ancestor of another shifted node, then the earlier kernels act as the identity on the parent tuples of later shifted nodes. Second, in the deterministic case, it suffices that the earlier maps be injective (e.g. relabelling or permutation) on the relevant parent supports, so that no conditioning information is lost. More generally, any kernel family that preserves the information carried by the relevant parent tuple yields the required stagewise Markov property.
\end{remark}
\begin{remark}
\label{rem:one-mechanism-at-a-time}
When all parents of \(X_i\) are invariant, stagewise Markov-preservation is automatic and the stage-\(i\) condition collapses to the direct local alignment
\begin{align}
  \prob_{\scmsimple_s}\!\bigl(X_i\mid \operatorname{PA}(X_i)=p\bigr)\,K_i
  =
  \prob_{\scmsimple_t}\!\bigl(X_i\mid \operatorname{PA}(X_i)=p\bigr)
  \qquad \forall\,p\in\mathcal P_i^{\mathrm{rel}}.
\end{align}
Theorem~\ref{thm:mechanism-characterization-markovian} extends this picture by replacing the original parent law with its recursively transported counterpart.
\end{remark}

\paragraph{Finite-state reduction.}
When \(\dom{X_i}\) is finite, the stage-\(i\) compatibility condition is equivalent to the feasibility of a linear system for the row-stochastic kernel \(K_i\), with additional row-fixing constraints enforcing intervention preservation. In this regime, existence becomes a linear feasibility question, while uniqueness can be studied via the rank of the transformed source conditional family. We record these finite-state results, together with the deterministic specialisation and related remarks, in Appendix~\ref{app:markovian-finite-stagewise}.

\subsection{Semi-Markovian extension}
\label{subsec:semimarkovian-extension}
The Markovian case shows that exact model-level transport can be assembled recursively by aligning one active mechanism at a time along the DAG. We now ask how this picture changes when latent confounding destroys node-wise factorisation and the relevant transport objects are districts rather than individual variables. When latent confounding is present, the graph is an \emph{acyclic directed mixed graph} (ADMG): a DAG augmented with bidirected edges $\leftrightarrow$ marking pairs of variables sharing an unobserved common cause. Its \emph{districts} (also called c-components) are the connected components under bidirected edges; i.e., the maximal sets of variables linked, directly or indirectly, by latent confounding. Let \(\cD(\scmdag)\) denote the districts of the ADMG \(\scmdag\), and for \(D\in\cD(\scmdag)\) define
\begin{align}
  \operatorname{PA}(D)
  :=
  \Bigl(\bigcup_{i\in D}\operatorname{PA}(X_i)\Bigr)\setminus D,
  \qquad
  \dom{X_D}:=\prod_{i\in D}\dom{X_i}.
\end{align}
The corresponding district mechanism is
\begin{align}
  q_D^\alpha(\cdot\mid p)
  :=
  \prob_{\scmsimple_\alpha}\!\bigl(
    X_D \mid \scmdo(\operatorname{PA}(D)=p)
  \bigr),
  \qquad
  \alpha\in\{s,t\}.
\end{align}
In the Markovian special case, every district is a singleton and \(q_D^\alpha\) reduces to the usual node-wise conditional. Let $\mathcal K^{\cD} := \{D \in \cD(\scmdag) : D \cap \mathcal K \neq \emptyset\}$ denote the shifted districts, i.e., those containing at least one shifted node; $\mathcal K \subseteq [d]$ retains its meaning from Section~\ref{subsec:standard-transportability} as the set of shifted nodes. We assume that for every invariant district $D \notin \mathcal K^{\cD}$,
\begin{equation}
  q_D^s(\cdot\mid p)=q_D^t(\cdot\mid p)
  \qquad \forall\,p\in\dom{\operatorname{PA}(D)}.
  \label{eq:invariant-district-mechanisms}
\end{equation}
Constructive kernels factor over districts:
\begin{align}
  K(d\mathbf{x}'\mid \mathbf{x})
  =
  \prod_{D\in\cD(\scmdag)} K_D(d\mathbf{x}'_D\mid \mathbf{x}_D),
  \qquad
  K_D(\cdot\mid \mathbf{x}_D)=\delta_{\mathbf{x}_D}
  \text{ for every } D\notin\mathcal K^{\cD}.
\end{align}
The following assumption restricts the intervention family so that each shifted district is either left intact or intervened upon, but never split. Since a district is the unit coupled by latent confounding, an intervention cutting across it would break the districtwise factorisation the assembly relies on; the cost is that interventions partially targeting a confounded set fall outside the present theory.
\begin{assumption}[Shifted-district consistency]
\label{ass:shifted-district-consistency}
For every shifted district \(D\in\mathcal K^{\cD}\) and every \(\iota\in\cI\) with intervened set \(J_\iota\), $D\subseteq J_\iota~\text{or}~ D\cap J_\iota=\emptyset$.
\end{assumption}
Contracting each district to a single node yields a directed graph $\scmdag_{/\cD}$ on the districts, with directed edges lifted from $\scmdag$ (bidirected edges are internal to districts and are absorbed by the contraction). Contracting such multi-node blocks can create cycles even when $\scmdag$ is acyclic: with districts $D_X=\{X_1,X_2\}$ and $D_Y=\{Y_1,Y_2\}$ (from
$X_1\!\leftrightarrow\!X_2$, $Y_1\!\leftrightarrow\!Y_2$), the directed edges $X_1\!\to\!Y_1$ and $Y_2\!\to\!X_2$ lift to $D_X\!\to\!D_Y$ and $D_Y\!\to\!D_X$, a $2$-cycle. Following the standard restriction for Cluster DAGs \citep{anand2023causaleffectidentificationcluster}, we focus on models whose quotient $\scmdag_{/\cD}$ is acyclic, which holds whenever no two districts are joined by directed edges in both directions, giving a topological order $D_1,\dots,D_m$ for the assembly. This is a requirement of our \emph{constructive} route, not of transportability as such: a cyclic quotient removes the order the stagewise assembly relies on, but such a model need not be non-transportable. Whether a common transport operator exists in the cyclic case, plausibly, by a non-constructive argument for models with a unique solution, we leave to future work. Using this order, for each \(r\in[m]\), let \(\bar K_{D_r}\) denote the lifted kernel that acts as \(K_{D_r}\) on the block \(D_r\) and as the identity on all other coordinates, and define:
\begin{align}
  R_0^{(\iota)} := P_s^{(\iota)},
  \qquad
  R_r^{(\iota)} := R_{r-1}^{(\iota)}\bar K_{D_r},
  \qquad r=1,\dots,m.
\end{align}

\begin{defn}[Stagewise district-preserving kernel]
\label{def:stagewise-district-preserving}
A districtwise product kernel $K$ is called \emph{stagewise district-preserving over \(\cI\)} if, for every intervention \(\iota\in\cI\) and every district stage \(r\), one has, whenever \(D_r\cap J_\iota=\emptyset\),
\begin{align}
  R_r^{(\iota)}\!\bigl(
    X_{D_r}\mid X_{D_1},\dots,X_{D_{r-1}}
  \bigr)
  =
  R_r^{(\iota)}\!\bigl(
    X_{D_r}\mid X_{\operatorname{PA}(D_r)}
  \bigr)
  \qquad
  R_r^{(\iota)}\text{-a.s.}
\end{align}
\end{defn}

\begin{theorem}[Recursive district assembly]
\label{thm:mechanism-characterization-semimarkovian}
Under Assumption~\ref{ass:shifted-district-consistency}, let $K(d\mathbf{x}'\mid \mathbf{x}) = \prod_{D\in\cD(\scmdag)} K_D(d\mathbf{x}'_D\mid \mathbf{x}_D)$ be a stagewise district-preserving districtwise product kernel such that \(K_D(\cdot\mid \mathbf{x}_D)=\delta_{\mathbf{x}_D}\) for every invariant district \(D\notin\mathcal K^{\cD}\). Then \(K\) is a CICk from \(\scmsimple_s\) to \(\scmsimple_t\) over \(\cI\) if and only if, for every intervention \(\iota\in\cI\) and every district \(D_r\in\cD(\scmdag)\), the partially transformed law \(R_r^{(\iota)}\) satisfies:
\begin{enumerate}[label=(\roman*)]
  \item if \(D_r\subseteq J_\iota\), then $R_r^{(\iota)}(X_{D_r})=\delta_{a_{D_r}}$;

  \item if \(D_r\cap J_\iota=\emptyset\), then $ R_r^{(\iota)}\!\bigl(
      X_{D_r}\mid X_{\operatorname{PA}(D_r)}
    \bigr)
    =
    q_{D_r}^t\!\bigl(
      \cdot\mid X_{\operatorname{PA}(D_r)}
    \bigr)
    ~~
    R_r^{(\iota)}\text{-a.s.}$

\end{enumerate}
The deterministic analogue holds with \(K_D(\cdot\mid \mathbf{x}_D)=\delta_{\tau_D(\mathbf{x}_D)}\).
\end{theorem}

\begin{proof}
The proof uses the same inductive argument as in Theorem~\ref{thm:mechanism-characterization-markovian}, with nodes replaced by districts and Markov preservation replaced by district preservation.
\end{proof}

\begin{remark}
\label{rem:district-preserving-free}
As in the Markovian case, the stagewise district-preserving assumption excludes no genuine CICks: any districtwise CICk automatically induces partially transformed laws whose district-prefix marginals agree with the target interventional law and hence inherit the required district-level conditional structure.
\end{remark}

\paragraph{Finite-state district reduction.}
When the district state space \(\dom{X_{D_r}}\) is finite, the stage-\(r\) district compatibility problem again reduces to a linear feasibility problem for a row-stochastic district kernel \(K_{D_r}\), now with district state spaces replacing node state spaces. The corresponding finite-state existence, uniqueness, and deterministic-style refinements follow the same logic as in the Markovian case, with districts replacing nodes throughout. We defer these finite-discrete districtwise results, together with related computational and query-level remarks, to Appendix~\ref{app:semimarkovian-finite-stagewise}.

The exact characterisations above answer the question of when a common constructive transport operator exists. However, its real value is in the approximate case: when no exact map exists, the best approximate one still certifies query-level intervals \(\ell\le Q_t\le L\), turning approximation error into a quantitative notion of approximate transportability, which is informative even for non-identifiable queries (\(\mathcal R_1\)) and when learned from source data alone, with no target data (\(\mathcal R_2\)). This motivates the approximate notion introduced next.

\section{Approximate Model-level Transportability}
\label{sec:approximate-model-level-transportability}
The characterisation results of Section~\ref{sec:characterizing} are exact: they ask whether a single common kernel or map transports the relevant source interventional family to the target family \emph{exactly}. In many practically important settings, however, such exact consistency is too strong to be expected. We therefore relax exact equality to a discrepancy between the two sides of the desired transport relation. This yields a quantitative notion of model-level transport error, analogous to abstraction error in causal abstraction, but specialised to the transportability setting. Let \(\mathsf D\) be any nonnegative discrepancy on probability measures, with \(\mathsf D(\mu,\nu)=0\) iff \(\mu=\nu\).

\begin{defn}[Intervention-query transport error]
\label{def:intervention-query-transport-error}
For a common Markov kernel \(K\), an intervention-query pair \((\iota,O_\iota)\in\Qcal\), and discrepancy \(\mathsf D\), define the \emph{intervention-query transport error}
\begin{equation}
  e_K^{(\iota,O_\iota)}
  :=
  \mathsf D\!\left(
    \pi_{O_\iota\#}\bigl(P_s^{(\iota)}K\bigr),\;
    \pi_{O_\iota\#}P_t^{(\iota)}
  \right).
  \label{eq:pointwise-transport-error}
\end{equation}
\end{defn}

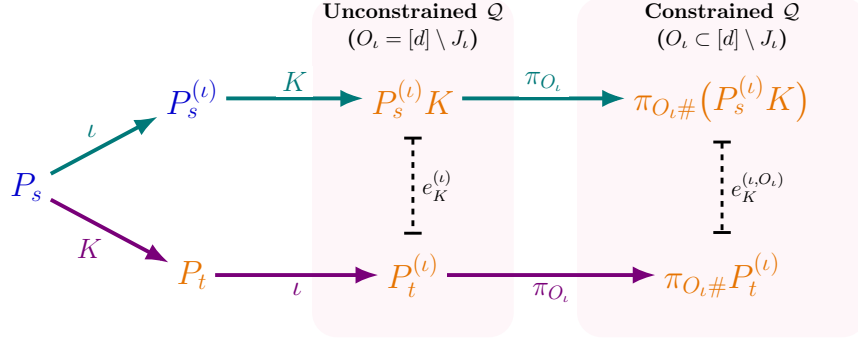
\begin{figure}[t]
\centering
\resizebox{0.75\columnwidth}{!}{%
\begin{tikzpicture}[>=Latex, thick, font=\sffamily]

  \tikzset{
    srcNode/.style={font=\LARGE, text=blue!80!black},
    tgtNode/.style={font=\LARGE, text=orange!90!black},
    transNode/.style={font=\LARGE, text=orange!90!black},
    edgeLabel/.style={font=\Large, fill=white, inner sep=3pt},
    gapLabel/.style={font=\large\bfseries, text=black, inner sep=4pt},
    blobLabel/.style={font=\normalsize\bfseries, align=center}
  }

  
  \fill[magenta!4, rounded corners=12pt] (5.2, -2.7) rectangle (8.8, 3.4);
  \draw[magenta!4, line width=1.5pt, rounded corners=12pt] (5.2, -2.7) rectangle (8.8, 3.4);
  \node[blobLabel, text=black] at (7, 2.7) {Unconstrained $\mathcal{Q}$\\ \normalsize ($O_{\iota} = [d]\setminus J_{\iota}$)\\[0.2mm]};

  \fill[magenta!4, rounded corners=12pt] (10.0, -2.7) rectangle (15.2, 3.4);
  \draw[magenta!4, line width=1.5pt, rounded corners=12pt] (10.0, -2.7) rectangle (15.2, 3.4);
  \node[blobLabel, text=black] at (12.6, 2.7) {Constrained $\mathcal{Q}$\\ \normalsize ($O_{\iota} \subset [d]\setminus J_{\iota}$)\\[0.2mm]};

  
  \node[srcNode] (Ps) at (0, 0) {$P_s$};

  \node[srcNode] (Ps1) at (3, 1.6) {$P_s^{(\iota)}$};
  \node[tgtNode] (Pt)  at (3, -1.6) {$P_t$};

  \node[transNode] (PsK1) at (7, 1.6) {$P_s^{(\iota)}K$};
  \node[tgtNode]   (Pt1)  at (7, -1.6) {$P_t^{(\iota)}$};

  \node[transNode] (ProjK1)  at (12.6, 1.6) {$\pi_{O_{\iota}\#}\bigl(P_s^{(\iota)}K\bigr)$};
  \node[tgtNode]   (ProjPt1) at (12.6, -1.6) {$\pi_{O_{\iota}\#}P_t^{(\iota)}$};

  
  \draw[->, line width=2pt, teal!95!black] (Ps) -- node[above left, edgeLabel] {$\iota$} (Ps1);
  \draw[->, line width=2pt, violet!95!black] (Ps) -- node[below left, edgeLabel] {$K$} (Pt);

  \draw[->, line width=2pt, teal!95!black]   (Ps1) -- node[above, edgeLabel] {$K$} (PsK1);
  \draw[->, line width=2pt, violet!95!black] (Pt)  -- node[below, edgeLabel] {$\iota$} (Pt1);

  \draw[|-|, dashed, black, line width=1.5pt] ([yshift=-2mm]PsK1.south) -- node[gapLabel, right] {$e_K^{(\iota)}$} ([yshift=2mm]Pt1.north);

  \draw[->, line width=2pt, teal!95!black]   (PsK1) -- node[above, edgeLabel] {$\pi_{O_{\iota}}$} (ProjK1);
  \draw[->, line width=2pt, violet!95!black] (Pt1)  -- node[below, edgeLabel] {$\pi_{O_{\iota}}$} (ProjPt1);

  \draw[|-|, dashed, black, line width=1.5pt] ([yshift=-2mm]ProjK1.south) -- node[gapLabel, right] {$e_K^{(\iota, O_{\iota})}$} ([yshift=2mm]ProjPt1.north);

\end{tikzpicture}%
}
\caption{\emph{Computation of the intervention-query error.} From the source observational law $P_s$, two pathways are compared: {\color{teal}(a) intervene with $\iota$ on the source, then map to the target via $K$} (yielding {\color{teal}$P_s^{(\iota)}K$}); and {\color{violet}(b) map via $K$, then apply $\iota$} (yielding {\color{violet}$P_t^{(\iota)}$}). Their gap is the full-joint error $e_K^{(\iota)}$ for the unconstrained regime $O_{\iota}=[d]\setminus J_{\iota}$. Projecting both laws onto the query coordinates via $\pi_{O_{\iota}}$ ($O_{\iota}\subset[d]\setminus J_{\iota}$) instead gives the pointwise error $e_K^{(\iota,O_{\iota})}$; averaging this over all pairs in $\Qcal$ yields the $\Qcal$-restricted transport error $\mathcal E_{\Qcal}$.}
\label{fig:commuting-horizontal}
\end{figure}

Thus \(e_K^{(\iota,O_\iota)}\) measures, for a fixed intervention and designated output set, how far the transported source interventional distribution is from the target one. Figure~\ref{fig:commuting-horizontal} depicts this comparison as a commuting diagram: the transported law \(P_s^{(\iota)}K\) and the true target \(P_t^{(\iota)}\) are compared in the target space, and the residual after projecting both onto the query coordinates \(O_\iota\) is exactly \(e_K^{(\iota,O_\iota)}\). Aggregating these pointwise errors over the relevant query family yields the model-level transport error.

\begin{defn}[\(\Qcal\)-transport error]
\label{def:q-transport-error}
Let \(K\) be a common Markov kernel and \(\Qcal\) a query specification family. The \(\Qcal\)-restricted transport error of \(K\) is defined as:
\begin{equation}
  \mathcal E_{\Qcal}(K)
  :=
  \frac{1}{|\Qcal|}
  \sum_{(\iota,O_\iota)\in\Qcal}
  e_K^{(\iota,O_\iota)}.
  \label{eq:abstract-q-transport-error}
\end{equation}
\end{defn}
\begin{remark}
In the full post-interventional case \(O_\iota=[d]\setminus J_\iota\) for every \(\iota\in\cI\), \(\mathcal E_{\Qcal}(K)\) reduces to the model-level average abstraction error~\citep{felekis2024causal, dyer2023interventionally, kekić2023targeted}.
\end{remark}
Restricting the admissible kernel class to deterministic kernels recovers the approximate abstraction map problem, while restricting it to the constructive class recovers the approximate constructive problem. Vanishing transport error is equivalent to exact $\Qcal$-restricted consistency, and the infimum is attained under standard compactness/lower-semicontinuity conditions; both facts are formalised in Appendix~\ref{app:attainment}.

In the present setting, \(\mathcal E_{\Qcal}(K)\) measures how far a single common transport operator is from transporting the relevant source interventional family to the target one. For this reason, it admits a structural interpretation as a \emph{model-level transport error}. The definition above is abstract and applies to arbitrary SCM classes. To make it useful, we next specialise in the concrete parametric setting of linear additive noise models.

\section{Modelling shifts in Linear Additive Noise models}
\label{sec:lan}

\subsection{Setting and Structural Forms}
\label{subsec:lan-setting}
We specifically consider Linear Additive Noise Models (LANs), where structural assignments take the form
\begin{equation}
  \en = W \en + \ex,
  \label{eq:lan-structural}
\end{equation}
where $\en \in\mathbb R^d$, $\ex \in\mathbb R^d$, and $W\in\mathbb R^{d\times d}$ is a weighted adjacency matrix strictly lower-triangular under a fixed topological ordering. Here $W_{ij}$ is the coefficient of $X_j$ in the structural equation for $X_i$, and it is nonzero only when $X_j\to X_i$; each row of $W$ therefore collects the incoming edges of one node, i.e., its mechanism. The reduced form of the SCM becomes $\en = A \ex$, with the mixing matrix $A \coloneq (I - W)^{-1}$, which is called the \emph{propagator}. Since $W$ is strictly lower triangular, it is nilpotent: $W^d=0$. Hence, $(I-W)$ is automatically invertible, and its inverse admits a finite polynomial expansion. Thus, the propagator can be written as:
\begin{equation}\label{eq:A_finite_series}
  A = (I-W)^{-1} = I + W + W^2 + \cdots + W^{d-1}
\end{equation}
Since $W$ is strictly lower triangular, each power $W^k$ is lower triangular, and therefore $A$ is lower triangular as well. The matrix $W$ encodes \emph{direct causal mechanisms} (local edge weights), whereas $A=(I-W)^{-1}$ encodes \emph{total causal effects} (global propagation along all directed paths).

Let $\iota\in\mathcal I$ denote a hard intervention acting on a set of nodes
$J_\iota\subset\{1,\dots,d\}$. As mentioned, a hard intervention disables the structural equations
of the intervened nodes (equivalently, it removes all incoming edges into $J_\iota$).
We represent this via a diagonal gating matrix $R_\iota\in\mathbb R^{d\times d}$:
\begin{align}
(R_\iota)_{jj}=
\begin{cases}
0,& j\in J_\iota,\\
1,& \text{otherwise}.
\end{cases}
\end{align}
The interventional propagator is then $A_\iota := (I - R_\iota W)^{-1}$. Since $R_\iota W$ remains strictly lower triangular, $A_\iota$ also admits a finite polynomial expansion analogous to \eqref{eq:A_finite_series}. The LAN reduced form \(\en=A\ex\) makes the two possible sources of domain shift clear: the environment $\ex$ may change, the mechanism \(A=(I-W)^{-1}\) may change, or both may change jointly. In the LAN instantiations below, we therefore model target uncertainty with separate mechanism and environment perturbations.

\subsection{Exact Perturbation Identities}
\label{subsec:resolvent}
Under a mechanism perturbation $\Delta W$, necessarily strictly lower triangular under the same fixed ordering, so that $W'=W+\Delta W$ remains a valid LAN structural matrix, the perturbed propagator $A'=(I-W')^{-1}$ always exists and remains lower triangular. The \emph{resolvent identity} $B^{-1}-C^{-1}=B^{-1}(C-B)C^{-1}$, applied with \(B=I-W-\Delta W\) and \(C=I-W\), gives
\begin{equation}
  A' - A = A'\Delta W A.
  \label{eq:resolvent}
\end{equation}
This is exact but \emph{implicit}: $A'$ still appears on the right-hand side. To obtain an explicit expansion involving only $A$ and $\Delta W$, we use the factorisation $A'=(I-A\Delta W)^{-1}A$. Note that $A\Delta W$ is strictly lower triangular because $A$ is lower triangular with unit diagonal, whereas $\Delta W$ is strictly lower triangular. Hence $(A\Delta W)^d=0$, so $A\Delta W$ is nilpotent, and we obtain the finite expansion:
\begin{equation}
A'=\left(\sum_{k=0}^{d-1} (A\Delta W)^k\right)A.
\label{eq:finite_expansion}
\end{equation}
No convergence condition is required since the series terminates after $d-1$ terms. Subtracting $A$ gives the exact identity
\begin{equation}
A' - A
=
\sum_{k=1}^{d-1} (A\Delta W)^k A.
\label{eq:A_difference_finite}
\end{equation}
Under intervention $\iota$, analogous identities hold with $A_\iota$ and $R_\iota\Delta W$.

\paragraph{First-order structure.}
The leading term in \eqref{eq:A_difference_finite} is
\begin{equation}
A' \;=\; A + A\Delta WA + \text{higher-order terms}.
\label{eq:A_first_order}
\end{equation}
Even if $\Delta W$ is sparse, it appears in the sandwiched form $A\Delta WA$. Since $A$ aggregates propagation along all directed paths in the DAG, a localised mechanism perturbation can induce a \emph{global} change in the propagator.

\subsection{Structured Stability Bounds}
\label{subsec:stability}
Throughout this subsection, let \(\|\cdot\|\) denote any submultiplicative matrix norm. Let \(\Delta W\) be an admissible mechanism perturbation preserving the fixed topological order, and for each intervention \(\iota\in\mathcal I\), define the perturbed interventional propagator
\begin{equation}
  A'_\iota := (I-R_\iota(W+\Delta W))^{-1}.
\end{equation}
Since \(R_\iota(W+\Delta W)\) remains strictly lower triangular, \(A'_\iota\) is always well defined. Let \(\mathcal A_W\) denote the admissible set of mechanism perturbations preserving the fixed acyclic structural class. To quantify the effect of mechanism perturbations after intervention, we introduce two intervention-specific quantities. The first is the \emph{amplification factor}, which measures the stability of the nominal propagator \(A_\iota\) under admissible mechanism perturbations:
\begin{equation}
  \gamma_\iota
  :=
  \sup_{\Delta W\in\mathcal A_W}
  \|A_\iota(R_\iota\Delta W)\|,
  \label{eq:def-gamma-iota}
\end{equation}
The second is the corresponding \emph{propagator stability modulus}, which controls the deviation between the perturbed and nominal propagators
\begin{equation}
  \alpha_\iota
  :=
  \sup_{\Delta W\in\mathcal A_W}
  \|A'_\iota-A_\iota\|,
  \label{eq:def-alpha-iota}
\end{equation}
The exact perturbation identity from Section~\ref{subsec:resolvent} gives, for every admissible \(\Delta W\),
\begin{equation}
  A'_\iota-A_\iota
  =
  \sum_{k=1}^{d-1}\bigl(A_\iota(R_\iota\Delta W)\bigr)^kA_\iota.
  \label{eq:Aiota-difference-finite}
\end{equation}
This immediately yields the following bound.

\begin{proposition}[Polynomial stability bound]
\label{prop:stability-polynomial}
For every intervention \(\iota\in\mathcal I\),
\begin{equation}
  \alpha_\iota
  \le
  \|A_\iota\|
  \sum_{k=1}^{d-1}\gamma_\iota^k.
  \label{eq:stability-polynomial}
\end{equation}
\end{proposition}

\begin{proof}
Fix \(\Delta W\in\mathcal A_W\). By \eqref{eq:Aiota-difference-finite}, subadditivity and submultiplicativity,
\begin{align}
  \|A'_\iota-A_\iota\|
  \le
  \sum_{k=1}^{d-1}
  \|A_\iota(R_\iota\Delta W)\|^k\,\|A_\iota\|.
\end{align}
Since \(\|A_\iota(R_\iota\Delta W)\|\le \gamma_\iota\) by definition of
\(\gamma_\iota\), taking sup over \(\Delta W\in\mathcal A_W\) gives
\eqref{eq:stability-polynomial}.
\end{proof}
The existence of $A'_\iota$ is already guaranteed by acyclicity, as noted above. Proposition~\ref{prop:stability-polynomial} adds the quantitative half, controlling its sensitivity through $\gamma_\iota$. Thus, structural perturbations are amplified polynomially in $\gamma_\iota$. When $\gamma_\iota$ is small, the dominant term is linear in $\|\Delta W\|$, yielding a local Lipschitz regime. A sharper control is available in the small-perturbation regime \(\gamma_\iota<1\), using the Neumann-series bound: for any matrix \(T\) with \(\|T\|<1\),
\begin{align}
  (I-T)^{-1}=\sum_{m=0}^{\infty}T^m,
  \qquad
  \|(I-T)^{-1}\|
  \le
  \sum_{m=0}^{\infty}\|T\|^m
  =
  \frac{1}{1-\|T\|}.
\end{align}
Applying this with \(T=A_\iota(R_\iota\Delta W)\) yields the following sharper control.

\begin{proposition}[Neumann-regime stability bound]
\label{prop:stability-neumann}
If \(\gamma_\iota<1\), then
\begin{align}
  \|A'_\iota\| \le
  \frac{\|A_\iota\|}{1-\gamma_\iota},~~\text{and}~~
  \alpha_\iota \le
  \frac{\|A_\iota\|^2}{1-\gamma_\iota}
  \sup_{\Delta W\in\mathcal A_W}\|R_\iota\Delta W\|.
\end{align}
\end{proposition}

\begin{proof}
Using the factorisation $A'_\iota = \bigl(I-A_\iota(R_\iota\Delta W)\bigr)^{-1}A_\iota$, the assumption \(\gamma_\iota<1\) implies \(\|A_\iota(R_\iota\Delta W)\|<1\) for every admissible \(\Delta W\). Hence, the Neumann-series bound gives
\begin{align}
  \bigl\|
    \bigl(I-A_\iota(R_\iota\Delta W)\bigr)^{-1}
  \bigr\|
  \le
  \frac{1}{1-\gamma_\iota}\quad \implies \|A'_\iota\|
  \le
  \frac{\|A_\iota\|}{1-\gamma_\iota}.
\end{align}
Moreover,
\begin{align}
  A'_\iota-A_\iota
  =
  \bigl(I-A_\iota(R_\iota\Delta W)\bigr)^{-1}
  A_\iota(R_\iota\Delta W)A_\iota \quad \implies
  \|A'_\iota-A_\iota\|
  \le
  \frac{1}{1-\gamma_\iota}\,
  \|A_\iota\|^2\,\|R_\iota\Delta W\|.
\end{align}
Taking the sup over \(\Delta W\in\mathcal A_W\) yields the second bound.
\end{proof}
The Neumann regime is therefore not needed for invertibility, which is already guaranteed by acyclicity but only for sharper quantitative control. In that regime, perturbation amplification is governed by the geometric factor \((1-\gamma_\iota)^{-1}\). The preceding bounds hold for any admissible mechanism ambiguity set \(\mathcal A_W\) within the fixed acyclic structural class. In the optimisation section, we instantiate \(\mathcal A_W\) using several geometries: Frobenius, row-wise, column-wise, and entrywise; and derive the corresponding explicit controls of \(\gamma_\iota\) and \(\alpha_\iota\).

These stability bounds provide the quantitative control needed for robust learning: they turn local perturbations of the structural matrix into explicit bounds on the induced perturbations of the post-interventional propagators. The next section uses this control to formulate TraCA as a target-agnostic robust optimisation problem.

\section{Transportation via Causal Abstractions: TraCA}
\label{sec:transport-via-causal-abstractions}
\begin{figure}[t]
\centering
\resizebox{0.97\textwidth}{!}{%
\begin{tikzpicture}[
    >=Latex, font=\sffamily, line width=0.6pt,
    ballc/.style={draw=blue!50!black, dashed, fill=blue!5, line width=0.7pt},
    ballt/.style={draw=orange!65!black, dashed, fill=orange!7, line width=0.7pt},
    ctr/.style={circle, draw=black!55, fill=gray!35, line width=0.6pt,
                minimum size=2.6mm, inner sep=0pt},
    swc/.style={diamond, draw=blue!30!black, fill=blue!60!white,
                  minimum size=3.4mm, inner sep=0pt},
    twc/.style={diamond, draw=orange!45!black, fill=orange!75!white,
                  minimum size=3.4mm, inner sep=0pt},
    strue/.style={star, star points=5, draw=blue!30!black, fill=blue!60!white,
                minimum size=3.6mm, inner sep=0pt},
    ttrue/.style={star, star points=5, draw=orange!45!black, fill=orange!75!white,
                minimum size=3.6mm, inner sep=0pt},
    slbl/.style={font=\scriptsize, text=gray!55!black},
    ptitle/.style={font=\normalsize\bfseries},
    vlbl/.style={font=\scriptsize\itshape, text=black!65},
    popbox/.style={rounded corners=4pt, dotted, line width=0.8pt},
]

\begin{scope}[shift={(0,0)}]
\node[draw=gray!4, rounded corners=7pt, line width=0pt, fill=gray!4,
      minimum width=6.6cm, minimum height=6.8cm] at (3.1,-0.35) {};
\node[ptitle, text=black!80]  at (3.1,2.72) {(a) Classical DRO};
\node[vlbl]                   at (3.1,2.28) {internal validity / single population};

\node[popbox, draw=blue!40!black, fill=blue!3,
      minimum width=5.4cm, minimum height=3.9cm] at (3.1,0.05) {};
\node[font=\scriptsize\bfseries, text=blue!45!black] at (3.1,1.72)
     {Source};

\node[ballc, circle, minimum size=2.9cm] at (3.1,-0.05) {};

\node[ctr] (emp) at (3.1,-0.05) {};
\node[font=\scriptsize, text=blue!60!black, anchor=north east] at (3.4,0.7)
     {$\widehat{\mathbb P}_s$};

\node[strue] (tA) at (3.72,0.56) {};
\node[font=\scriptsize, text=blue!45!black, anchor=west] at (3.45,0.9)
     {$\mathbb P_s^\star$};

\node[swc] (wA) at (2.20,-0.64) {};
\node[font=\scriptsize, text=blue!45!black, anchor=east] at (2.65,-0.2)
     {$\mathbb P_s^{\mathrm{wc}}$};
\draw[->, blue!45!black, line width=0.7pt, shorten >=3pt, shorten <=2pt]
     (emp) -- (wA);

\draw[blue!45!black, line width=0.5pt, shorten >=3pt, shorten <=2pt]
     (emp) to node[above, font=\small\bfseries, text=blue!45!black, pos=0.52] {$\varepsilon$} (4.5, -0.7);

\node[slbl, text=blue!60!black, align=center] at (3.1,-2.6)
     {The ambiguity set is:\\ i. centered on the empirical source distribution;\\ ii. hedges against \textbf{sampling error}};
\end{scope}

\begin{scope}[shift={(7.6,0)}]
\node[draw=red!3, rounded corners=7pt, line width=0pt, fill=red!3,
      minimum width=8.4cm, minimum height=6.8cm] at (4.0,-0.35) {};
\node[ptitle, text=red!72!black] at (4.0,2.72) {(b) TraCA};
\node[vlbl]                      at (4.0,2.28) {external validity / \emph{transportability}};

\node[popbox, draw=blue!40!black, fill=blue!3,
      minimum width=2.2cm, minimum height=2.8cm] at (1.25,-0.05) {};
\node[font=\scriptsize\bfseries, text=blue!45!black] at (1.25,1.1) {Source};
\node[strue] (srcM) at (1.25,-0.22) {};
\node[font=\small, text=blue!45!black, anchor=south] at (1.25,0.0)
     {$\mathbb P_s^\star$};
\node[slbl, text=blue!60!black, align=center] at (1.25,-1.08)
     {trusted SCM};

\node[popbox, draw=orange!60!black, fill=orange!3,
      minimum width=4.7cm, minimum height=3.9cm] at (5.45,0.05) {};
\node[font=\scriptsize\bfseries, text=orange!62!black] at (5.45,1.72)
     {Target};

\node[ballt, circle, minimum size=3.0cm] at (5.45,-0.05) {};

\node[strue] (ref) at (5.45,-0.05) {};

\draw[->, draw=red!68!black, line width=1.0pt]
     (srcM.east) to[bend left=20]
     node[above, font=\small\bfseries, text=red!72!black, pos=0.52] {$\tau$}
     (4.0, -0.05);

\node[ttrue] (tB) at (6.07,0.54) {};
\node[font=\scriptsize, text=orange!62!black, anchor=west] at (5.75,0.9){$\mathbb P_t^\star$};

\node[twc] (wB) at (4.49,-0.64) {};
\node[font=\scriptsize, text=orange!62!black, anchor=east] at (4.9,-0.2)
     {$\mathbb P_t^{\mathrm{wc}}$};
\draw[->, orange!45!black, line width=0.7pt, shorten >=3pt, shorten <=2pt]
     (ref) -- (wB);

\draw[orange!45!black, line width=0.5pt, shorten >=3pt, shorten <=2pt]
     (ref) to node[above, font=\small\bfseries, text=orange!45!black, pos=0.52] {$\varepsilon$} (6.9, -0.7);

\node[slbl, text=orange!72!black, align=center] at (4.0,-2.6)
     {The ambiguity set:\\ i. centered on the available source distribution;\\ ii. hedges against \textbf{domain shift}};
\end{scope}

\end{tikzpicture}%
}
\caption{\emph{Two uses of distributional robustness.} (a) \textbf{Classical DRO} and (b) \textbf{TraCA} share a minimax form, but the ambiguity set $\mathbb B_\varepsilon(\cdot)$ (dashed ball) plays different roles. In Classical DRO (internal validity): the set is centered on the empirical source $\widehat{\mathbb P}_s$ and hedges against \emph{sampling error}; the true law $\mathbb P_s^\star$ is assumed to lie in it, and the minimax is evaluated at the worst case $\mathbb P_s^{\mathrm{wc}}$. In TraCA (external validity): the set is centered on the trusted source $\mathbb P_s^\star$, and each admissible target perturbs it by a mechanism and an environment shift; the true target $\mathbb P_t^\star$ is assumed to lie in it, and $\tau$ is evaluated against the worst case $\mathbb P_t^{\mathrm{wc}}$. Markers: \protect\glyphcentre{} center, \protect\glyphwc{} worst case, \protect\glyphtrue{} true law.}
\label{fig:dro-contrast}
\end{figure}
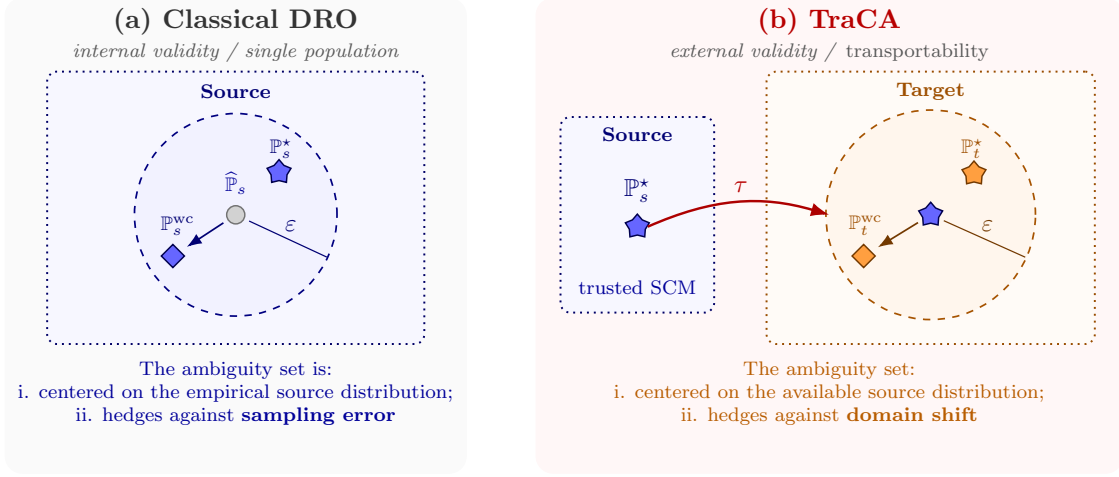

Section~\ref{sec:approximate-model-level-transportability} introduced
approximate model-level transportability through the
\(\Qcal\)-transport error \(\mathcal E_{\Qcal}(K)\), which measures how far a single common operator is from transporting the relevant source interventional family
to the target one. Meanwhile, Section~\ref{sec:lan} provided the linear-additive setting and the stability quantities needed to control mechanism perturbations. We now combine these two ingredients in the practically important regime of domain generalisation, where the target domain is not observed ($\mathcal R_2$): we move beyond the classical transportability setting \((\mathcal D_s^{o},\mathcal D_s^{i},\mathcal D_t^{o})\) to a target-agnostic regime in which \(\mathcal D_t^{o}\) is unavailable at training time. Given a source SCM, structural knowledge of which mechanisms may vary across domains, and an ambiguity model for the target environment, the goal is to learn a single transport map that remains reliable across a class of plausible target domains. This naturally suggests a distributionally robust formulation, but in a different sense from classical DRO. In DRO, the ambiguity set models uncertainty about the \emph{true data-generating distribution} itself, typically around an empirical approximation of that same distribution. 

In TraCA, by contrast, we take the source SCM as the trusted reference model and use DRO for \emph{domain generalisation}: the ambiguity set does not represent uncertainty about the observed source domain but rather uncertainty about the \emph{unseen target domain} to which the transport map will later be applied. Thus, the robust problem is not to hedge against the misspecification of the source data generating process but against structurally plausible shifts away from it. Since the target domain is unobserved, a candidate map cannot be evaluated against the true target family. In the same spirit as DiRoCA~\citep{felekis2026distributionallyrobustcausalabstractions}, we therefore replace the target term in the transport error with its worst case over an ambiguity set, but where DiRoCA's set captures uncertainty about a single system's own environment, ours captures uncertainty about a distinct, unobserved population. This yields the target-agnostic robust counterpart of the earlier model-level transport error: instead of evaluating a candidate map against one fixed target family, we evaluate it against the most adverse target family consistent with the assumed structural and environmental uncertainty. This ambiguity set is not fixed once and for all. Prior domain knowledge constrains its \emph{shape}: which mechanisms may shift, which entries may change, and what patterns of shift are plausible. Partial target information, when available, can instead shrink the \emph{radius} of the ambiguity set around a better localised target. In TraCA, this ambiguity has two distinct components: uncertainty in the target environment (exogenous distribution) and uncertainty in the target mechanisms. The distinction is essential since a change in the reduced-form mechanism cannot, in general, be reproduced by keeping the mechanism fixed and changing only the exogenous law; see Appendix~\ref{app:lan-aux} for a formal counterexample. We therefore model both sources of target-side uncertainty explicitly.

\subsection{Unified minimax formulation}
\label{subsec:unified-minimax}

The general DRO template is
\begin{equation}
  \inf_{x\in\mathcal X}\sup_{\mathbb Q\in\mathbb B}
  \mathbb E_{\xi\sim\mathbb Q}[f(x,\xi)],
  \label{eq:dro-template-traca}
\end{equation}
where \(x\) is a decision variable, \(\xi\) is an uncertain quantity, \(f\) is a loss, and \(\mathbb B\) is an ambiguity set of plausible distributions. In TraCA, the decision variable is the transport map \(\lintau\), and the uncertainty lies entirely on the target side. Accordingly, TraCA learns a transport map \(\lintau\) by minimising a worst-case transport loss over a target-side ambiguity set:
\begin{equation}
  \min_{\lintau}\max_{\xi\in\mathcal A}
  \frac{1}{|\mathcal I|}
  \sum_{\iota\in\mathcal I}
  \mathsf F_\iota(\lintau,\xi),
  \label{eq:generic-minimax-traca}
\end{equation}
where \(\xi\) collects the adversarial target-side variables and \(\mathcal A\) denotes the corresponding ambiguity set. The term \(\mathsf F_\iota\) denotes the same loss family evaluated under intervention \(\iota\), with intervention-specific propagators and target-side adversarial variables; it thereby instantiates the pointwise transport error \(e_K^{(\iota,O_\iota)}\). This objective should be read as the target-agnostic robust version of the earlier \(\Qcal\)-transport error: in the absence of target observations, the unknown target SCM is replaced by the worst admissible target model.  We denote the optimal value of \eqref{eq:generic-minimax-traca} by $\mathfrak R^\star(\mathcal A)$. We focus on the LAN setting from Section~\ref{sec:lan}.  In particular, we introduce a \emph{mechanism adversary} \(\Delta W\), which, for each intervention \(\iota\in\mathcal I\), perturbs the target propagator from \(A_\iota\) to $A'_\iota(\Delta W)$:
\begin{equation}
  A_\iota := (I-R_\iota W)^{-1},
  \qquad
  A'_\iota(\Delta W):=(I-R_\iota(W+\Delta W))^{-1}.
\end{equation}
The matrix \(A_\iota\) is the nominal source propagator under intervention \(\iota\), while \(A'_\iota(\Delta W)\) is its perturbed target counterpart induced by a mechanism shift \(\Delta W\).

To encode the assumption that only a designated set of mechanisms may vary across domains, we define the shifted node row mask $M_{\mathcal K}:=\diag\bigl(\mathbf 1\{1\in\mathcal K\},\dots,\mathbf 1\{d\in\mathcal K\}\bigr)$,  where \(\mathcal K\subseteq[d]\) is the set of shifted nodes (Section~\ref{subsec:standard-transportability}). All admissible mechanism perturbations are required to satisfy
\begin{equation}
  \Delta W = M_{\mathcal K}\Delta W.
\end{equation}
Hence, the ambiguity is not arbitrary: it is structurally restricted to the shifted mechanisms and may be further refined if prior knowledge suggests that only certain magnitudes or patterns of shift are plausible. Furthermore, we specialise the generic stability quantities of Section~\ref{subsec:stability} to the spectral norm and write
\begin{equation}
  \gamma_\iota
  :=
  \sup_{\Delta W\in\mathcal A_W}
  \|A_\iota(R_\iota\Delta W)\|_2,
  \qquad
  \alpha_\iota
  :=
  \sup_{\Delta W\in\mathcal A_W}
  \|A'_\iota(\Delta W)-A_\iota\|_2.
\end{equation}
Here \(\gamma_\iota\) and \(\alpha_\iota\) are the spectral-norm specialisations of the general stability quantities of Section~\ref{subsec:stability}; the bounds of Corollary~\ref{cor:gamma-alpha-plug-in-traca} make them explicit for each ambiguity geometry. The optimisation objectives and robustness bounds below use these quantities
\begin{equation}
  \|A'_\iota(\Delta W)-A_\iota\|_2 \le \alpha_\iota,
  \qquad
  \|A'_\iota(\Delta W)\|_2 \le \|A_\iota\|_2+\alpha_\iota,
\end{equation}
with explicit control of \(\alpha_\iota\) supplied by the structured stability bounds of Section~\ref{subsec:stability}. 

\begin{remark}
\label{rem:information-monotonicity}
The robust value
\(\mathfrak R^\star(\mathcal A):=\inf_{\lintau}\sup_{\xi\in\mathcal A}
\tfrac{1}{|\mathcal I|}\sum_{\iota\in\mathcal I}\mathsf F_\iota(\lintau,\xi)\) scales monotonically with target-side information. Shrinking the ambiguity set can only lower the robust objective and tighten the induced certificate: if \(\mathcal A_1\subseteq\mathcal A_2\) then \(\mathfrak R^\star(\mathcal A_1)\le\mathfrak R^\star(\mathcal A_2)\), and whenever the true target perturbation \(\xi^\circ\) lies in \(\mathcal A_1\), the \(\mathcal A_1\)-robust map already certifies the true target loss, \(\tfrac{1}{|\mathcal I|}\sum_\iota \mathsf F_\iota(\lintau_1^\star,\xi^\circ)\le\mathfrak R^\star(\mathcal A_1)\). When the losses are Lipschitz in \(\xi\), the gap to the oracle value \(\mathfrak R^\star(\{\xi^\circ\})\) is controlled by the ambiguity diameter, so the robust problem approaches the full-information oracle as the ambiguity shrinks to \(\xi^\circ\). These are direct consequences of the minimax structure; we provide them in full detail in Appendix~\ref{app:information-monotonicity}.
\end{remark}
The minimax formulation above is intentionally generic: it specifies the target-side adversary and the robust objective, but not yet how the target ambiguity is parameterised in practice. Two natural choices arise from the structure of available source-side knowledge. When the source environment is summarised by sufficient statistics (mean and covariance), a moment-based Gaussian formulation matches the natural geometry. When the source environment is summarised by samples, a perturbation-based empirical formulation matches the natural geometry. We treat both since they share an algorithmic backbone but differ in the ascent objective and projection operators.

\section{Optimisation}
\label{sec:algorithms}
Section~\ref{sec:transport-via-causal-abstractions} formulated TraCA as a generic target-agnostic minimax problem. We now instantiate that framework in two concrete forms: a Gaussian moment-based version and an empirical perturbation-based version. Both are realisations of the same robust objective~\eqref{eq:generic-minimax-traca} and differ only in how the unseen target environment is parameterised. The min-max optimisation backbone common to both: the alternating projected gradient descent-ascent scheme and, in the empirical case, the Frobenius-budgeted noise perturbations follows the logic of DiRoCA \citep{felekis2026distributionallyrobustcausalabstractions}. A note on their relation can be found in Appendix~\ref{app:relation-diroca}.

\paragraph{Gaussian case.}
The adversary is
\begin{align}
  \xi=(\Delta W,\mu_t,\Sigma_t)\in \mathcal A_W\times\mathcal A_\rho(\varepsilon),
\end{align}
and the intervention-wise loss is
\begin{align}
  \mathsf F_\iota^{\rho}(\lintau,\Delta W,\mu_t,\Sigma_t)
  :=
  \mathcal W_2^2\!\Bigl(
    \mathcal N(\lintau A_\iota\mu_s,\;
               \lintau A_\iota\Sigma_sA_\iota^\top\lintau^\top),\;
    \mathcal N(A'_\iota(\Delta W)\mu_t,\;
               A'_\iota(\Delta W)\Sigma_tA'_\iota(\Delta W)^\top)
  \Bigr).
  \label{eq:F-iota-gau}
\end{align}

\paragraph{Empirical case.}
The adversary is
\begin{align}
  \xi=(\Delta W,\Theta)\in \mathcal A_W\times\mathcal A_U(\varepsilon),
\end{align}
and the intervention-wise loss is
\begin{align}
  \mathsf F_\iota^{U}(\lintau,\Delta W,\Theta)
  :=
  \bigl\|
    \lintau A_\iota U^s
    -
    A'_\iota(\Delta W)(U^s+\Theta)
  \bigr\|_F^2.
  \label{eq:F-iota-emp}
\end{align}
The admissible transport maps $\lintau$ are restricted to the constructive class: diagonal in the Markovian case and block-diagonal by district in the semi-Markovian case. No analogue of the intervened-value pinning $K_i(\cdot\mid a)=\delta_a$ of Theorem~\ref{thm:finite-discrete-compatibility-markovian} is required here. Under $\iota$, the rows of $A'_\iota(\Delta W)$ indexed by $J_\iota$ equal the corresponding standard basis vectors for every admissible $\Delta W$, so the intervened coordinates carry the intervention value and nothing else. Since $O_\iota\subseteq[d]\setminus J_\iota$, these coordinates are excluded from every $\Qcal$-restricted loss, and the corresponding rows of $\lintau$ do not enter any certified quantity.

\paragraph{Min-Max optimisation architecture.}
Both formulations use the same algorithmic backbone. At each outer iteration, one first updates \(\lintau\) by projected gradient descent, holding the adversarial variables fixed, and then updates the target-side adversarial block by projected/proximal ascent, followed by the feasibility corrections required by the corresponding ambiguity set. The Gaussian and empirical formulations, therefore, differ only in the adversarial variables, the ascent objective used for them, and the associated projection/proximal operators.

\subsection{Mechanism ambiguity sets and projections}
\label{subsec:mechanism-ambiguity}
We begin with the mechanism ambiguity set \(\mathcal A_W\), since it is shared by both optimisation problems; the environment-side ambiguity sets \(\mathcal A_\rho(\varepsilon)\) and \(\mathcal A_U(\varepsilon)\) are introduced later in the Gaussian and empirical subsections, respectively. The set \(\mathcal A_W\) constrains admissible perturbations \(\Delta W\) of the target structural matrix. Throughout, these perturbations are understood to preserve the fixed acyclic support of \(W\) (equivalently, the same topological ordering/triangular pattern), although we suppress this structural constraint from the notation for readability. We consider four geometries for \(\mathcal A_W\), each paired with its natural projection operator for the projected-gradient ascent update in \(\Delta W\).

\paragraph{Frobenius ball.}
\begin{align}
  \mathcal A_W^{\mathrm F}(\eta)
  :=
  \Bigl\{
    \Delta W:\Delta W=M_{\mathcal K}\Delta W,\ \|\Delta W\|_F\le \eta
  \Bigr\}.
\end{align}
Its Euclidean projection is
\begin{equation}
  \proj_{\mathcal A_W^{\mathrm F}(\eta)}(\widetilde\Delta)
  =
  \widetilde\Delta
  \min\!\left(1,\frac{\eta}{\|\widetilde\Delta\|_F}\right).
  \label{eq:proj-frob-traca}
\end{equation}

\paragraph{Row-wise budgets.}
Given row budgets $\rho=(\rho_j)_{j\in\mathcal K}$,
\begin{align}
  \mathcal A_W^{\mathrm{row}}(\rho)
  :=
  \Bigl\{
    \Delta W:\Delta W=M_{\mathcal K}\Delta W,\
    \sum_{k=1}^d |\Delta W_{jk}|\le \rho_j\ \forall j\in\mathcal K
  \Bigr\}.
\end{align}
Projection is row-wise $\ell_1$ projection:
\begin{equation}
  \proj_{\mathcal A_W^{\mathrm{row}}(\rho)}(\widetilde\Delta)_{j,:}
  =
  \proj_{\|\cdot\|_1\le \rho_j}(\widetilde\Delta_{j,:}).
  \label{eq:proj-row-traca}
\end{equation}

\paragraph{Column-wise budgets.}
Given column budgets $c=(c_k)_{k=1}^d$,
\begin{align}
  \mathcal A_W^{\mathrm{col}}(c)
  :=
  \Bigl\{
    \Delta W:\Delta W=M_{\mathcal K}\Delta W,\
    \sum_{j=1}^d |\Delta W_{jk}|\le c_k\ \forall k
  \Bigr\}.
\end{align}
Projection is column-wise $\ell_1$ projection:
\begin{equation}
  \proj_{\mathcal A_W^{\mathrm{col}}(c)}(\widetilde\Delta)_{:,k}
  =
  \proj_{\|\cdot\|_1\le c_k}(\widetilde\Delta_{:,k}).
  \label{eq:proj-col-traca}
\end{equation}

\paragraph{Entrywise box.}
Given entrywise radii $B=(b_{jk})$ with $b_{jk}=0$ for $j\notin\mathcal K$,
\begin{align}
  \mathcal A_W^{\mathrm{box}}(B)
  :=
  \Bigl\{
    \Delta W:\Delta W=M_{\mathcal K}\Delta W,\
    |\Delta W_{jk}|\le b_{jk}\ \forall j,k
  \Bigr\}.
\end{align}
Projection is entrywise clipping:
\begin{equation}
  \proj_{\mathcal A_W^{\mathrm{box}}(B)}(\widetilde\Delta)_{jk}
  =
  \sign(\widetilde\Delta_{jk})\,
  \min\bigl(|\widetilde\Delta_{jk}|,\,b_{jk}\bigr).
  \label{eq:proj-box-traca}
\end{equation}

\paragraph{Directional entrywise box.}
The entrywise box above is centered at the source, encoding symmetric uncertainty $\Delta W_{jk}\in[-b_{jk},b_{jk}]$ about each coefficient. When the practitioner has \emph{directional} prior knowledge; i.e., a belief that a coefficient shifts predominantly in a known direction, this is encoded by recentering the box at an offset $\delta=(\delta_{jk})$ with $\delta_{jk}=0$ for $j\notin\mathcal K$:
\begin{align}
  \mathcal A_W^{\mathrm{box}}(\delta,B)
  :=
  \Bigl\{
    \Delta W:\Delta W=M_{\mathcal K}\Delta W,\
    |\Delta W_{jk}-\delta_{jk}|\le b_{jk}\ \forall j,k
  \Bigr\},
\end{align}
so that $\Delta W_{jk}\in[\delta_{jk}-b_{jk},\,\delta_{jk}+b_{jk}]$: the admissible targets are now those near an anticipated shifted mechanism rather than those near the source one. The offset indicates which way the coefficient is expected to move and by how much, and $b_{jk}$ indicates how sure one is of that. The projection is the shifted clipping
\begin{align}
  \proj_{\mathcal A_W^{\mathrm{box}}(\delta,B)}(\widetilde\Delta)_{jk}
  =
  \delta_{jk}
  + \sign(\widetilde\Delta_{jk}-\delta_{jk})\,
    \min\bigl(|\widetilde\Delta_{jk}-\delta_{jk}|,\,b_{jk}\bigr),
\end{align}
and setting $\delta=0$ recovers the symmetric box \eqref{eq:proj-box-traca} exactly. The offset $\delta$ enters the optimisation only through this recentering; the adversary, the transport update, and the feasibility corrections are otherwise unchanged. This recentering changes the solution qualitatively. At $\delta=0$, the box is symmetric about the origin, so the opposing vertices $\pm B$ contribute equal and opposite transport gradients, and the identity map remains stationary; at $\delta\neq 0$, the vertex at $\delta+B$ lies farther from the origin, the contributions no longer cancel, and $\lintau$ is driven away from the identity. The directional prior therefore yields a nontrivial point correction by construction, independently of the loss. The practical role of the directional prior is discussed in Section~\ref{sec:practitioner-guidelines}.

\begin{remark}
\label{rem:validity-ambiguity-traca}
The radius constraints above control only the \emph{size} of the mechanism perturbation. The validity of the perturbed target model comes from the structural restriction that admissible \(\Delta W\) preserve the fixed acyclic support class of \(W\). Thus, \(W+\Delta W\) remains a valid linear acyclic SCM, and hence for every intervention \(\iota\), $A'_\iota(\Delta W)$ is well defined.
\end{remark}

\paragraph{Interpretive guide to mechanism ambiguity sets.}
The choice of \(\mathcal A_W\) should reflect the type of prior knowledge available about the target-side mechanism shift. The table below summarises the four geometries used in TraCA.

\begin{center}
\small
\renewcommand{\arraystretch}{1.35}
\begin{tabular}{@{}p{6.4cm}p{4.2cm}p{3.6cm}@{}}
\toprule
\textbf{Modelling intent}
& \textbf{Ambiguity set}
& \textbf{Budget parameter} \\
\midrule
The \emph{total} magnitude of the mechanism perturbation is globally bounded
& Frobenius 
& scalar $\eta$ \\

Each shifted structural equation may change by at most a prescribed amount
& Row-wise  
& vector $\rho=(\rho_j)_{j\in\mathcal K}$ \\

The total change in the outgoing influence of each parent variable is bounded
& Column-wise  
& vector $c=(c_k)_{k=1}^d$ \\

A separate bound is known for each individual structural coefficient
& Entrywise 
& matrix $B=(b_{jk})$ \\
\bottomrule
\end{tabular}
\end{center}
Once a geometry for \(\mathcal A_W\) is fixed, the stability analysis of Section~\ref{subsec:stability} yields an explicit bound on the amplification factor \(\gamma_\iota\), and hence, via the polynomial or Neumann bounds from Propositions~\ref{prop:stability-polynomial} and~\ref{prop:stability-neumann} an explicit bound on the propagator stability modulus \(\alpha_\iota\). 
\begin{corollary}
\label{cor:gamma-alpha-plug-in-traca}
For every intervention \(\iota\in\mathcal I\), the following bounds hold:
\begin{align}
  \Delta W\in\mathcal A_W^{\mathrm F}(\eta)
  &\quad\Longrightarrow\quad
  \gamma_\iota \le \|A_\iota\|_2\,\eta,
  \label{eq:gamma-frob-traca}\\
  \Delta W\in\mathcal A_W^{\mathrm{row}}(\rho)
  &\quad\Longrightarrow\quad
  \gamma_\iota \le \sqrt d\,
  \max_{\ell}\sum_{j\in\mathcal K}|(A_\iota)_{\ell j}|\,\rho_j,
  \label{eq:gamma-row-traca}\\
  \Delta W\in\mathcal A_W^{\mathrm{col}}(c)
  &\quad\Longrightarrow\quad
  \gamma_\iota \le \sqrt d\,\|A_\iota\|_1\max_k c_k,
  \label{eq:gamma-col-traca}\\
  \Delta W\in\mathcal A_W^{\mathrm{box}}(B)
  &\quad\Longrightarrow\quad
  \gamma_\iota \le
  \bigl\||A_\iota|R_\iota B\bigr\|_F,
  \label{eq:gamma-box-traca}\\
  \Delta W\in\mathcal A_W^{\mathrm{box}}(\delta,B)
  &\quad\Longrightarrow\quad
  \gamma_\iota \le
  \bigl\||A_\iota|R_\iota(|\delta|+B)\bigr\|_F,
  \label{eq:gamma-box-dir-traca}
\end{align}
where $|\delta|+B$ collects the effective half-widths $|\delta_{jk}|+b_{jk}$, reducing to \eqref{eq:gamma-box-traca} at $\delta=0$. Substituting any of \eqref{eq:gamma-frob-traca}-\eqref{eq:gamma-box-dir-traca} into Propositions~\ref{prop:stability-polynomial} and~\ref{prop:stability-neumann} yields explicit control of the propagator stability modulus \(\alpha_\iota\). The proof is given in Appendix~\ref{app:gamma-alpha-plug-in}.
\end{corollary}

\paragraph{Unified alternating solver.}
Both the Gaussian and empirical formulations use the same high-level optimisation pattern: alternating descent in the transport map \(\lintau\) and ascent in the target-side adversarial variables. The difference between the two formulations lies only in how the target-side environment perturbation is parameterised and projected back to its feasible set. Algorithm~\ref{alg:traca-unified} is the common optimisation backbone of TraCA. The Gaussian and empirical formulations differ only in the parameterisation of the target-side environment block, the ascent objective used for that block, and the associated feasibility correction. The analytical algorithms can be found in Appendices~\ref{app:gaussian-optimization-details} and \ref{app:empirical-optimization-details}.

\begin{algorithm}[t]
\caption{Unified TraCA optimisation}
\label{alg:traca-unified}
\begin{algorithmic}[1]
\STATE Initialise \(\lintau^{(0)}\), \(\Delta W^{(0)}=0\), and the target-side adversarial variables.
\REPEAT
  \STATE Update \(\lintau\) by projected gradient descent on the current objective.
  \STATE Update \(\Delta W\) by gradient ascent, followed by masking and projection onto the mechanism ambiguity set \(\mathcal A_W\).
  \STATE Update the target-side environment variables by ascent, followed by the corresponding feasibility correction/projection.
\UNTIL{convergence}
\STATE \textbf{return} \(\lintau\)
\end{algorithmic}
\end{algorithm}

\subsection{Gaussian adversarial learning}
\label{subsec:gaussian-implementation}
Assume the source environment is Gaussian, $\rho^s \sim \mathcal N(\mu_s,\Sigma_s)$, with $\Sigma_s$ diagonal in the Markovian case. The Gaussian intervention-wise loss \eqref{eq:F-iota-gau} compares the transported source law and the adversarial target law under intervention $\iota$. Concretely,
\begin{align}
  \tau_{\#}P_s^{(\iota)}
  =
  \mathcal N\bigl(
    \lintau A_\iota\mu_s,\;
    \lintau A_\iota\Sigma_sA_\iota^\top\lintau^\top
  \bigr),
\end{align}
while
\begin{align}
  P_t^{(\iota)}(\Delta W,\mu_t,\Sigma_t)
  =
  \mathcal N\bigl(
    A'_\iota(\Delta W)\mu_t,\;
    A'_\iota(\Delta W)\Sigma_tA'_\iota(\Delta W)^\top
  \bigr).
\end{align}
Under $\iota=do(X_{J_\iota}=v_{J_\iota})$, the coordinates in $J_\iota$ are fixed deterministically to the intervention values $v_{J_\iota}$ in both the source and target; therefore, the displayed Gaussian laws describe the random post-interventional coordinates, which are equivalent to the full post-interventional law with the deterministic intervened coordinates suppressed. Concretely, the intervention value enters as a deterministic offset in the exogenous mean at the coordinates in $J_\iota$, whose own exogenous contribution is removed; the propagator then carries it forward as usual. The effective exogenous mean under $\iota$, which we write $\mu_s^{(\iota)}$, therefore holds the source noise mean at free coordinates and the intervention value at clamped ones.

\paragraph{Environment ambiguity.}
Let $\mathcal A_\rho(\varepsilon)$ denote the target-side Gaussian Wasserstein ambiguity set centered at $(\mu_s,\Sigma_s)$, the \emph{Gelbrich ball} \citep{gelbrich1990formula}:
\begin{align}
  \mathcal A_\rho(\varepsilon)
  :=
  \Bigl\{
    (\mu_t,\Sigma_t):
    \mathcal W_2^2\!\bigl(
      \mathcal N(\mu_t,\Sigma_t),\,
      \mathcal N(\mu_s,\Sigma_s)
    \bigr)
    \le \varepsilon^2
  \Bigr\},
\end{align}
together with the invariance pinning $\mu_i^t=\mu_{s,i}$ for $i\notin\mathcal K$, and $\Sigma_{ij}^t=(\Sigma_s)_{ij}$ whenever $i\notin\mathcal K$ or $j\notin\mathcal K$. Accordingly, the Gaussian objective specialises \eqref{eq:generic-minimax-traca} to the average intervention-wise loss \eqref{eq:F-iota-gau} over $\mathcal I$:
\begin{align}
  F(\lintau,\Delta W,\mu_t,\Sigma_t)
  :=
  \frac{1}{|\mathcal I|}
  \sum_{\iota\in\mathcal I}
  \mathsf F_\iota^{\rho}(\lintau,\Delta W,\mu_t,\Sigma_t).
\end{align}
and the robust Gaussian abstraction error is
\begin{equation}
  \mathcal E_\rho^{\mathrm{joint}}(\lintau)
  :=
  \sup_{\substack{\Delta W\in\mathcal A_W\\
                  (\mu_t,\Sigma_t)\in\mathcal A_\rho(\varepsilon)}}
  F(\lintau,\Delta W,\mu_t,\Sigma_t).
  \label{eq:gaussian-adversarial-objective-traca}
\end{equation}

The Wasserstein distance between Gaussians has a closed form:
\begin{align}
  \mathcal W_2^2\bigl(\mathcal N(m_1,C_1),\mathcal N(m_2,C_2)\bigr)
  =
  \|m_1-m_2\|_2^2
  + \Tr(C_1)+\Tr(C_2)
  -2\Tr\bigl((C_1^{1/2}C_2C_1^{1/2})^{1/2}\bigr)
\end{align}
Let $C_{\iota,s}(\lintau)
  := \lintau A_\iota \Sigma_s A_\iota^\top \lintau^\top,~
  C_{\iota,t}(\Delta W,\Sigma_t)
  := A'_\iota(\Delta W)\Sigma_t A'_\iota(\Delta W)^\top$. Then,
\begin{align}
  F(\lintau,\Delta W,\mu_t,\Sigma_t)
  &=
  \underbrace{
    \frac{1}{|\mathcal I|}
    \sum_{\iota\in\mathcal I}
    \Big[
      \|\lintau A_\iota\mu_s - A'_\iota(\Delta W)\mu_t\|_2^2
      + \Tr\!\bigl(C_{\iota,s}(\lintau)\bigr)
      + \Tr\!\bigl(C_{\iota,t}(\Delta W,\Sigma_t)\bigr)
    \Big]
  }_{F^\rho_s\ \text{(smooth term)}}
  \nonumber\\
  &\quad
  \underbrace{
    -\frac{2}{|\mathcal I|}
    \sum_{\iota\in\mathcal I}
    \Tr\!\Bigl(
      \bigl(
        C_{\iota,s}(\lintau)^{1/2}
        C_{\iota,t}(\Delta W,\Sigma_t)
        C_{\iota,s}(\lintau)^{1/2}
      \bigr)^{1/2}
    \Bigr)
  }_{F^\rho_n\ \text{(Bures cross-term)}}.
  \label{eq:traca-gaussian-exact}
\end{align}
\paragraph{Smooth surrogate for the ascent step.}
As in the DiRoCA optimiser \citep{felekis2026distributionallyrobustcausalabstractions}, we replace the Bures cross-term with a smooth lower surrogate. Since
\begin{align}
  \Tr\!\Bigl(
    \bigl(
      C_{\iota,s}^{1/2}C_{\iota,t}C_{\iota,s}^{1/2}
    \bigr)^{1/2}
  \Bigr)
  =
  \|C_{\iota,s}^{1/2}C_{\iota,t}^{1/2}\|_*,
\end{align}
the variational upper bound \(\|XY\|_*\le \|X\|_F\|Y\|_F\) \citep{NIPS2004_e0688d13}, yields:
\begin{equation}
  \widetilde F^\rho(\lintau,\Delta W,\mu_t,\Sigma_t)
  :=
  F_s^\rho(\lintau,\Delta W,\mu_t,\Sigma_t)
  -\frac{2}{|\mathcal I|}
  \sum_{\iota\in\mathcal I}
  \|C_{\iota,s}(\lintau)^{1/2}\|_F\,
  \|C_{\iota,t}(\Delta W,\Sigma_t)^{1/2}\|_F,
  \label{eq:traca-gaussian-surrogate}
\end{equation}
and \(\widetilde F^\rho\le F^\rho\). Since the Bures cross-term is upper bounded by the Frobenius product and appears with a minus sign in \eqref{eq:traca-gaussian-exact}, this replacement yields a conservative lower bound on the true Gaussian abstraction error, suitable for gradient-based ascent. This suggests performing the covariance update at the level of the transformed target covariance roots $C_{\iota,t}(\Delta W,\Sigma_t)^{1/2}$, rather than directly at the level of $\Sigma_t$.

\paragraph{Min-Max updates.}
Following the shared alternating architecture from Section~\ref{subsec:unified-minimax}, the Gaussian formulation updates $\lintau$ by projected gradient descent on the exact objective \eqref{eq:traca-gaussian-exact}. With $\lintau$ fixed, the adversarial block $(\Delta W,\mu_t,\Sigma_t)$ is then updated by ascent on the smooth surrogate \eqref{eq:traca-gaussian-surrogate}: $\Delta W$ by projected gradient ascent onto the chosen mechanism ambiguity set $\mathcal A_W$, $\mu_t$ by gradient ascent, and $\Sigma_t$ by a proximal-gradient update acting on the square roots of the transformed target covariances $C_{\iota,t}(\Delta W,\Sigma_t)$. After each adversarial update, the invariance constraints outside $\mathcal K$ are re-imposed via $\Pi_{\mathrm{inv}}(\mu,\Sigma):=(\tilde\mu,\tilde\Sigma)$, where $\tilde\mu_i=\mu_{s,i}$ for $i\notin\mathcal K$ and $\tilde\mu_i=\mu_i$ for $i\in\mathcal K$, while $\tilde\Sigma_{ij}=(\Sigma_s)_{ij}$ whenever $i\notin\mathcal K$ or $j\notin\mathcal K$, and $\tilde\Sigma_{ij}=\Sigma_{ij}$ otherwise; and the pair $(\mu_t,\Sigma_t)$ is projected back to the Gelbrich ball $\mathcal A_\rho(\varepsilon)$. The full sequence of updates is given in Algorithm~\ref{alg:traca-gaussian-adversarial} of Appendix~\ref{app:gaussian-optimization-details}.

\begin{remark}
\label{rem:gaussian-implementation-notes-traca}
The Gaussian optimiser follows the logic of the Gaussian DiRoCA optimiser, but with two TraCA-specific additions: the mechanism adversary \(\Delta W\) and the invariance projection \(\Pi_{\mathrm{inv}}\). Gradients with respect to \(\Delta W\) are computed using the standard derivative of an inverse identity \citep{Petersen2008}: $\mathrm D(X^{-1})[H]
  =
  -\,X^{-1}(\mathrm DX[H])X^{-1}$. Applying this to $X(\Delta W):=I-R_\iota(W+\Delta W),$ so that: $A'_\iota(\Delta W)=X(\Delta W)^{-1}$, and noting that $\mathrm DX(\Delta W)[H]=-R_\iota H,$ yields
\begin{equation}
  \mathrm D A'_\iota(\Delta W)[H]
  =
  A'_\iota(\Delta W)\,R_\iota H\,A'_\iota(\Delta W),
  \label{eq:resolvent-derivative-traca}
\end{equation}
for every matrix direction \(H\). The remaining implementation details follow the Gaussian DiRoCA template. First, the Bures cross-term is replaced in the ascent step by the same smooth lower surrogate \(\widetilde F\). Second, the covariance block is handled through the transformed target covariances
\begin{align}
  C_{\iota,t}(\Delta W,\Sigma_t)
  :=
  A'_\iota(\Delta W)\Sigma_tA'_\iota(\Delta W)^\top,
  \qquad \iota\in\mathcal I,
\end{align}
rather than directly through \(\Sigma_t\) alone. The proximal update is applied to the square roots \(C_{\iota,t}(\Delta W,\Sigma_t)^{1/2}\) via the Frobenius-norm proximal map
\begin{align}
  \prox_{\lambda\|\cdot\|_F}(A)
  =
  \begin{cases}
    \left(1-\dfrac{\lambda}{\|A\|_F}\right)A,
    & \|A\|_F>\lambda,\\[1ex]
    0, & \text{otherwise},
  \end{cases}
\end{align}
which induces the covariance update \(\mathrm{CovProx}_{\lambda_t}\): for each intervention, one applies the proximal map to the square root of the transformed covariance, pulls the result back to the exogenous covariance space, and then aggregates across interventions. Thus the proximal step acts on the square-root/Frobenius structure that remains in the surrogate \eqref{eq:traca-gaussian-surrogate}, while the exact Bures cross-term itself has already been replaced by the smooth lower bound \(\widetilde F\). Finally, after the adversarial covariance and mean updates, the pair \((\mu_t,\Sigma_t)\) is re-pinned by \(\Pi_{\mathrm{inv}}\) and projected onto the Gelbrich ball in \((\mu,\Sigma^{1/2})\)-space.
\end{remark}

\subsection{Empirical adversarial learning}
\label{subsec:empirical-implementation}
Let $U^s\in\mathbb R^{d\times N}$ be the abducted source exogenous samples. The target empirical adversary perturbs them through  $U^t = U^s + \Theta$, where $\Theta\in\mathbb R^{d\times N}$ is constrained to lie in the target-side ambiguity set
\begin{equation}
  \mathcal A_U(\varepsilon)
  :=
  \Bigl\{
    \Theta:
    \Theta=M_{\mathcal K}\Theta,\;
    \|\Theta\|_F\le \varepsilon\sqrt N
  \Bigr\}.
\end{equation}
The factor $\sqrt N$ normalises for sample size, so $\varepsilon$ controls the per-sample perturbation magnitude rather than the total. By the finite-dimensional reduction of \citet{kuhn2019wassersteindistributionallyrobustoptimization}, this Frobenius-ball constraint is analytically equivalent to restricting the target empirical environment to a Wasserstein ball of radius $\varepsilon$ around the source. For each intervention $\iota\in\mathcal I$, the intervention-wise empirical loss from \eqref{eq:F-iota-emp} takes the form
\begin{equation}
  \mathsf F_\iota^{U}(\lintau,\Delta W,\Theta)
  :=
  \bigl\|
    \lintau A_\iota U^s
    -
    A'_\iota(\Delta W)(U^s+\Theta)
  \bigr\|_F^2.
\end{equation}
Accordingly, the empirical objective specialises
\eqref{eq:generic-minimax-traca} to
\begin{equation}
  F(\lintau,\Delta W,\Theta)
  :=
  \frac{1}{|\mathcal I|}
  \sum_{\iota\in\mathcal I}
  \mathsf F_\iota^{U}(\lintau,\Delta W,\Theta).
\end{equation}
and the corresponding robust empirical abstraction error is
\begin{equation}
  \mathcal E_U^{\mathrm{joint}}(\lintau)
  :=
  \sup_{\substack{\Delta W\in\mathcal A_W\\
                  \Theta\in\mathcal A_U(\varepsilon)}}
   F(\lintau,\Delta W,\Theta).
  \label{eq:empirical-adversarial-objective-traca}
\end{equation}
In the implementation, we work with the normalised empirical objective \(N^{-1}\mathcal E_U^{\mathrm{joint}}(\lintau)\), equivalently \(N^{-1}\mathsf F_\iota^{U}\) at the intervention level. This does not change the minimiser in \(\lintau\), but keeps the empirical losses \(\mathcal{O}(1)\) as the sample size \(N\) grows.

\paragraph{Min-Max updates.}
Following the shared alternating architecture from Section~\ref{subsec:unified-minimax}, the empirical formulation updates $\lintau$ by projected gradient descent on the exact empirical objective $F(\lintau,\Delta W,\Theta)$, followed by projection back to the constructive class. With $\lintau$ fixed, the adversarial variables $\Delta W$ and $\Theta$ are updated by projected gradient ascent on the same objective: $\Delta W$ is masked and projected onto the chosen mechanism ambiguity set $\mathcal A_W$, while $\Theta$ is masked and projected onto the empirical ambiguity set $\mathcal A_U(\varepsilon)$. The full update sequence is given in Algorithm~\ref{alg:traca-empirical-adversarial} of Appendix~\ref{app:empirical-optimization-details}.

\begin{remark}
\label{rem:empirical-implementation-notes-traca}
The empirical optimiser is the direct projected-gradient analogue of the empirical DiRoCA scheme, but now with a single target perturbation block \(\Theta\) together with the mechanism block \(\Delta W\). The empirical perturbation is constrained by $\mathcal A_U(\varepsilon)$ so the ascent step for \(\Theta\) is followed by masking and projection onto a Frobenius ball. Concretely, for an intermediate update \(\widetilde\Theta\), one first enforces the shifted-node structure via \(\widetilde\Theta\leftarrow M_{\mathcal K}\widetilde\Theta\), and then applies the standard Frobenius rescaling
\begin{align}
  \proj_{\mathcal A_U(\varepsilon)}(\widetilde\Theta)
  =
  \widetilde\Theta\,
  \min\!\Bigl(1,\frac{\varepsilon\sqrt N}{\|\widetilde\Theta\|_F}\Bigr).
\end{align}
Thus, the empirical adversary remains feasible after every ascent step. Gradients with respect to the mechanism perturbation \(\Delta W\) again propagate through the perturbed interventional propagator via the resolvent identity \eqref{eq:resolvent-derivative-traca}.
\end{remark}

\subsection{Query-restricted relaxation}
\label{subsec:q-restricted-objectives}
The full-joint objectives remain the conceptual center of TraCA, as they aim to align the full post-interventional family. 

When only a designated output family matters, however, it is natural to restrict the loss to those coordinates. Recall the query family from Eq.~\eqref{eq:query-spec-family},
\begin{align}
  \Qcal=\{(\iota,O_\iota):\iota\in\mathcal I,\; O_\iota\subseteq[d]\setminus J_\iota\},
\end{align}
with \(J_\iota\) the intervened coordinates and \(O_\iota\) the outputs of interest. For every nonempty \(O\subseteq[d]\), let
\begin{align}
  S_O\in\{0,1\}^{|O|\times d}
\end{align}
denote the coordinate-selection matrix associated with the projection \(\pi_O\), so that \(\pi_O(x)=S_Ox\) for \(x\in\mathbb R^d\), and \(\pi_O(X)=S_OX\) for \(X\in\mathbb R^{d\times N}\). Query restriction changes only the \emph{loss target}: the map \(\lintau\) and the adversarial variables remain global.

\paragraph{Gaussian query-restricted loss.}
In the Gaussian formulation, the full post-interventional source and target laws under intervention \(\iota\) are
\begin{align}
  \lintau_{\#}P_s^{(\iota)}
  =
  \mathcal N\bigl(
    \lintau A_\iota\mu_s,\;
    \lintau A_\iota\Sigma_sA_\iota^\top\lintau^\top
  \bigr),
\end{align}
and
\begin{align}
  P_t^{(\iota)}(\Delta W,\mu_t,\Sigma_t)
  =
  \mathcal N\bigl(
    A'_\iota(\Delta W)\mu_t,\;
    A'_\iota(\Delta W)\Sigma_tA'_\iota(\Delta W)^\top
  \bigr).
\end{align}
Restricting attention to the coordinates in \(O_\iota\) means pushing both laws forward through \(\pi_{O_\iota}\), equivalently left- and bi-sided multiplication by \(S_{O_\iota}\). This yields the intervention-wise \(\Qcal\)-restricted Gaussian loss
\begin{align}
  \mathsf F_{(\iota,O_\iota)}^{\rho,\Qcal}
  (\lintau,\Delta W,\mu_t,\Sigma_t)
  &:=
  \mathcal W_2^2\!\Bigl(
    \mathcal N\!\bigl(
      S_{O_\iota}\lintau A_\iota\mu_s,\;
      S_{O_\iota}\lintau A_\iota\Sigma_sA_\iota^\top\lintau^\top S_{O_\iota}^\top
    \bigr), \nonumber\\
    &\hspace{3.2cm}
    \mathcal N\!\bigl(
      S_{O_\iota}A'_\iota(\Delta W)\mu_t,\;
      S_{O_\iota}A'_\iota(\Delta W)\Sigma_tA'_\iota(\Delta W)^\top S_{O_\iota}^\top
    \bigr)
  \Bigr).
  \label{eq:q-restricted-gaussian-loss-explicit}
\end{align}
Averaging over \(\Qcal\), define
\begin{equation}
  F_\Qcal^\rho(\lintau,\Delta W,\mu_t,\Sigma_t)
  :=
  \frac{1}{|\Qcal|}
  \sum_{(\iota,O_\iota)\in\Qcal}
  \mathsf F_{(\iota,O_\iota)}^{\rho,\Qcal}
  (\lintau,\Delta W,\mu_t,\Sigma_t),
  \label{eq:q-restricted-gaussian-average}
\end{equation}
and the corresponding robust query-restricted Gaussian abstraction error
\begin{equation}
  \mathcal E_{\rho,\Qcal}^{\mathrm{joint}}(\lintau)
  :=
  \sup_{\substack{\Delta W\in\mathcal A_W\\
                  (\mu_t,\Sigma_t)\in\mathcal A_\rho(\varepsilon)}}
  F_\Qcal^\rho(\lintau,\Delta W,\mu_t,\Sigma_t).
  \label{eq:q-restricted-gaussian-objective}
\end{equation}

\begin{remark}
\label{rem:q-restricted-gaussian-order}
In the Gaussian case, query restriction must be imposed at the level of the marginal Gaussian itself. If \(S_{O_\iota}\) is the selector onto the coordinates in \(O_\iota\), then the relevant source and target laws are
\begin{align}
\mathcal N(S_{O_\iota}\mu_1,\;S_{O_\iota}\Sigma_1S_{O_\iota}^\top),
\qquad
\mathcal N(S_{O_\iota}\mu_2,\;S_{O_\iota}\Sigma_2S_{O_\iota}^\top),
\end{align}
and the Wasserstein loss must be computed between these marginals. The reason is that the Gaussian \(\mathcal W_2^2\) formula contains the nonlinear Bures term, so coordinate restriction does not commute with the matrix square root; i.e. $(S_{O_\iota}CS_{O_\iota}^\top)^{1/2} \neq S_{O_\iota}C^{1/2}S_{O_\iota}^\top$ in general. Hence, one must first form the marginal covariance \(S_{O_\iota}CS_{O_\iota}^\top\), and then compute its square root.
\end{remark}

\paragraph{Empirical query-restricted loss.}
In the empirical formulation, the full residual under intervention \(\iota\) is
\begin{align}
  R_\iota(\lintau,\Delta W,\Theta)
  :=
  \lintau A_\iota U^s
  -
  A'_\iota(\Delta W)(U^s+\Theta)
  \in\mathbb R^{d\times N}.
\end{align}
Query restriction simply selects the rows indexed by \(O_\iota\). Hence, the intervention-wise \(\Qcal\)-restricted empirical loss is
\begin{equation}
  \mathsf F_{(\iota,O_\iota)}^{U,\Qcal}
  (\lintau,\Delta W,\Theta)
  :=
  \bigl\|
    S_{O_\iota}R_\iota(\lintau,\Delta W,\Theta)
  \bigr\|_F^2
  =
  \bigl\|
    S_{O_\iota}\lintau A_\iota U^s
    -
    S_{O_\iota}A'_\iota(\Delta W)(U^s+\Theta)
  \bigr\|_F^2.
  \label{eq:q-restricted-empirical-loss}
\end{equation}
Averaging over \(\Qcal\), define
\begin{equation}
  F_\Qcal^U(\lintau,\Delta W,\Theta)
  :=
  \frac{1}{|\Qcal|}
  \sum_{(\iota,O_\iota)\in\Qcal}
  \mathsf F_{(\iota,O_\iota)}^{U,\Qcal}
  (\lintau,\Delta W,\Theta),
  \label{eq:q-restricted-empirical-average}
\end{equation}
and the corresponding robust query-restricted empirical abstraction error
\begin{equation}
  \mathcal E_{U,\Qcal}^{\mathrm{joint}}(\lintau)
  :=
  \sup_{\substack{\Delta W\in\mathcal A_W\\
                  \Theta\in\mathcal A_U(\varepsilon)}}
  F_\Qcal^U(\lintau,\Delta W,\Theta).
  \label{eq:q-restricted-empirical-objective}
\end{equation}

\begin{remark}[Extreme cases of the \(\Qcal\)-restricted objective]
\label{rem:q-restricted-reductions}
The \(\Qcal\)-restricted formulation contains both the full-joint and single-query settings as special cases. If, for each intervention \(\iota\), the selected output set \(O_\iota\) is the full set of non-intervened coordinates, then \(\pi_{O_\iota\#}\) imposes no further restriction and one recovers the full post-interventional objective. At the other extreme, if \(\Qcal\) contains only one pair \((\iota_0,\{k\})\), then the objective reduces to a single intervention/output loss: a robust transport guarantee for one target query, the query-level end of the framework. Thus \(\Qcal\) simply specifies how much of the post-interventional family the optimisation is required to match.
\end{remark}
The \(\Qcal\)-restricted formulation does not require a different optimisation architecture: one replaces the full post-interventional losses with their \(\Qcal\)-restricted counterparts and applies the same alternating ascent-descent schemes. In the Gaussian case, the only additional care is that the Bures term and its smooth surrogate must be evaluated on the \(|O_\iota|\times |O_\iota|\) marginal covariances for each \((\iota,O_\iota)\in\Qcal\).

These optimisation problems are not merely algorithmic surrogates; their structure also permits explicit worst-case guarantees. We next show how the learned map and the stability moduli \(\alpha_\iota\) yield certified bounds on both model-level transport error and downstream queries.

\section{Provable Robustness and Query-Level Transport}
\label{sec:provable-robustness}
The optimisation problems of Section~\ref{sec:algorithms} also yield explicit worst-case guarantees for the abstraction error. We first state the more general \(\Qcal\)-restricted certificates. For each \((\iota,O_\iota)\in\Qcal\), let \(S_{O_\iota}\in\{0,1\}^{|O_\iota|\times d}\) denote the selector for \(\pi_{O_\iota}\). Full post-interventional guarantees are recovered by taking \(O_\iota=[d]\setminus J_\iota\), and the single-query case when \(|\Qcal|=1\). Each certificate decomposes into transport, mechanism, and environment terms. Hence, the bounds are small when \(\lintau\) is close to the identity on the relevant outputs, the intervention amplifies mechanism perturbations only mildly (small \(\alpha_\iota\)), and the admissible target environment is close to the source.
\begin{theorem}[Query-restricted Gaussian certificate]
\label{thm:q-restricted-gaussian-certificate}
Consider the Gaussian setting, with source law \((\mu_s,\Sigma_s)\), linear map \(\lintau\), and mechanism ambiguity set \(\mathcal A_W\). For each $(\iota,O_\iota)\in\Qcal$, define
\begin{align}
  \delta_{\iota,O_\iota}^{\rho}(\lintau)^2
  :=\;&
  \underbrace{
    4\|S_{O_\iota}(\lintau-I)A_\iota\mu_s\|_2^2
    +4\|\Sigma_s\|_2\,\|S_{O_\iota}(\lintau-I)A_\iota\|_F^2
  }_{\text{transport mismatch}}
  \nonumber\\
  &\quad+
  \underbrace{
    4\alpha_\iota^2\|\mu_s\|_2^2
    +4|O_\iota|\,\|\Sigma_s\|_2\,\alpha_\iota^2
  }_{\text{mechanism perturbation}}
  +
  \underbrace{
    2(\|S_{O_\iota}A_\iota\|_2+\alpha_\iota)^2\varepsilon^2
  }_{\text{environment ambiguity}}.
  \label{eq:q-restricted-gaussian-delta}
\end{align}
Then for every $\lintau$,
\begin{equation}
  \mathcal E_{\rho,\Qcal}^{\mathrm{joint}}(\lintau)
  \le
  \frac{1}{|\Qcal|}
  \sum_{(\iota,O_\iota)\in\Qcal}
  \delta_{\iota,O_\iota}^{\rho}(\lintau)^2.
  \label{eq:q-restricted-gaussian-certificate}
\end{equation}
In particular, any minimiser $\lintau^\star$ of the $\Qcal$-restricted Gaussian adversarial objective satisfies the same bound evaluated at $\lintau^\star$.
\end{theorem}
\begin{proof}\textit{(sketch)}
Fix \((\iota,O_\iota)\in\Qcal\). We interpolate between the transported source law and the target law through an intermediate Gaussian that carries the perturbed target mechanism \(A'_\iota(\Delta W)\) but retains the source environment \((\mu_s,\Sigma_s)\); applying the triangle inequality in \(\mathcal W_2\) then splits the error into two parts: one measuring the mismatch between the transported source law and the same-environment perturbed law, and one measuring the effect of changing the environment from \((\mu_s,\Sigma_s)\) to \((\mu_t,\Sigma_t)\). The first term is controlled by coupling both laws through the same source Gaussian and writing
\[
  \lintau A_\iota-A'_\iota(\Delta W)
  =
  (\lintau-I)A_\iota + (A_\iota-A'_\iota(\Delta W)).
\]
This separates transport mismatch from mechanism perturbation. The transport part gives the terms involving \((\lintau-I)A_\iota\), while the perturbation part is controlled by the stability bound \(\|A'_\iota(\Delta W)-A_\iota\|_2\le \alpha_\iota\).

The second term is controlled by the fact that pushforward by a fixed linear map is Lipschitz in \(\mathcal W_2\). Since \((\mu_t,\Sigma_t)\in\mathcal A_\rho(\varepsilon)\), this contributes a factor \(\varepsilon\), multiplied by the operator norm of the perturbed projected propagator, which is bounded by \(\|S_{O_\iota}A_\iota\|_2+\alpha_\iota\). Combining the two bounds yields the stated pointwise certificate, and averaging over \((\iota,O_\iota)\in\Qcal\) gives the result.
\end{proof}
The full analytical proof is deferred to Appendix~\ref{app:certificate-proofs}. Concretely, the mechanism part scales with \(\alpha_\iota\), so it grows when intervention \(\iota\) amplifies admissible mechanism perturbations. The environment part scales with the ambiguity radius \(\varepsilon\) and the projected propagator \(\|S_{O_\iota}A_\iota\|_2+\alpha_\iota\). Thus, the certificate grows with transport mismatch, intervention instability, and target-side uncertainty.

\begin{theorem}[Query-restricted empirical certificate]
\label{thm:q-restricted-empirical-certificate}
For each $(\iota,O_\iota)\in\Qcal$, define
\begin{equation}
  \delta_{\iota,O_\iota}^{U}(\lintau)
  :=
  \underbrace{\|S_{O_\iota}(\lintau-I)A_\iota U^s\|_F}_{\text{transport mismatch}}
  +\underbrace{\alpha_\iota\|U^s\|_F}_{\text{mechanism perturbation}}
  +\underbrace{\varepsilon\sqrt N\,(\|S_{O_\iota}A_\iota\|_2+\alpha_\iota)}_{\text{environment ambiguity}}.
\end{equation}
Then for every $\lintau$,
\begin{equation}
  \mathcal E_{U,\Qcal}^{\mathrm{joint}}(\lintau)
  \le
  \frac{1}{|\Qcal|}
  \sum_{(\iota,O_\iota)\in\Qcal}
  \delta_{\iota,O_\iota}^{U}(\lintau)^2.
  \label{eq:q-restricted-empirical-certificate}
\end{equation}
In particular, any minimiser $\lintau^\star$ of the $\Qcal$-restricted empirical adversarial objective satisfies the same bound evaluated at $\lintau^\star$.
\end{theorem}

\begin{proof}
By the same three-term decomposition (transport, mechanism, environment) in Frobenius norm; see Appendix~\ref{app:certificate-proofs}.
\end{proof}
The empirical certificate has the same three-part structure in the Frobenius norm. The first term measures the residual transport error of \(\lintau\) on the source sample; the second captures mechanism instability through \(\alpha_\iota\), and the third measures the effect of an admissible environment perturbation of radius \(\varepsilon\sqrt N\) after propagation through \(A_\iota\). As in the Gaussian case, the bound is larger when transport mismatch, mechanism instability, or target ambiguity are larger.
\begin{proposition}[Full post-interventional recovery]
\label{prop:recovery-full-joint-certificates}
Let $\Qcal_{\mathrm{full}}:=\{(\iota,[d]\setminus J_\iota):\iota\in\mathcal I\}$. Then the query-restricted certificates of Theorems~\ref{thm:q-restricted-gaussian-certificate} and \ref{thm:q-restricted-empirical-certificate} specialise to the full post-interventional output family. In particular, for every $\lintau$,
\begin{align}
  \mathcal E_{\rho,\Qcal_{\mathrm{full}}}^{\mathrm{joint}}(\lintau)
  \le
  \frac{1}{|\mathcal I|}
  \sum_{\iota\in\mathcal I}
  \Big[
    &4\|S_{[d]\setminus J_\iota}(\lintau-I)A_\iota\mu_s\|_2^2
    +4\|\Sigma_s\|_2\,\|S_{[d]\setminus J_\iota}(\lintau-I)A_\iota\|_F^2
    \nonumber\\
    &+4\alpha_\iota^2\|\mu_s\|_2^2
    +4(d-|J_\iota|)\|\Sigma_s\|_2\,\alpha_\iota^2
    +2\bigl(\|S_{[d]\setminus J_\iota}A_\iota\|_2+\alpha_\iota\bigr)^2\varepsilon^2
  \Big],
  \label{eq:gaussian-postinterventional-certificate-traca}
\end{align}
and
\begin{equation}
  \mathcal E_{U,\Qcal_{\mathrm{full}}}^{\mathrm{joint}}(\lintau)
  \le
  \frac{1}{|\mathcal I|}
  \sum_{\iota\in\mathcal I}
  \Bigl(
    \|S_{[d]\setminus J_\iota}(\lintau-I)A_\iota U^s\|_F
    +\alpha_\iota\|U^s\|_F
    +\varepsilon\sqrt N\,\bigl(\|S_{[d]\setminus J_\iota}A_\iota\|_2+\alpha_\iota\bigr)
  \Bigr)^2.
  \label{eq:empirical-postinterventional-certificate-traca}
\end{equation}
Moreover, these imply the simpler selector-free full-joint bounds
\begin{align}
  \mathcal E_{\rho,\Qcal_{\mathrm{full}}}^{\mathrm{joint}}(\lintau)
  \le
  \frac{1}{|\mathcal I|}
  \sum_{\iota\in\mathcal I}
  \Big[
    &4\|(\lintau-I)A_\iota\mu_s\|_2^2
    + 4\|\Sigma_s\|_2\,\|(\lintau-I)A_\iota\|_F^2
    \nonumber\\
    &+ 4\alpha_\iota^2\|\mu_s\|_2^2
    + 4d\,\|\Sigma_s\|_2\,\alpha_\iota^2
    + 2(\|A_\iota\|_2+\alpha_\iota)^2\varepsilon^2
  \Big],
  \label{eq:gaussian-joint-certificate-traca-recovered}
\end{align}
and
\begin{equation}
  \mathcal E_{U,\Qcal_{\mathrm{full}}}^{\mathrm{joint}}(\lintau)
  \le
  \frac{1}{|\mathcal I|}
  \sum_{\iota\in\mathcal I}
  \Bigl(
    \|(\lintau-I)A_\iota U^s\|_F
    + \alpha_\iota\|U^s\|_F
    + \varepsilon\sqrt N\,(\|A_\iota\|_2+\alpha_\iota)
  \Bigr)^2.
  \label{eq:empirical-joint-certificate-traca-recovered}
\end{equation}
\end{proposition}

\begin{proof}
Set $O_\iota=[d]\setminus J_\iota, \forall\,\iota\in\mathcal I$. Then $\Qcal=\Qcal_{\mathrm{full}}$, so the averages in Theorems~\ref{thm:q-restricted-gaussian-certificate} and \ref{thm:q-restricted-empirical-certificate} run over the full post-interventional output family. Substituting this choice of $O_\iota$ into the query-restricted Gaussian and empirical certificates gives \eqref{eq:gaussian-postinterventional-certificate-traca} and \eqref{eq:empirical-postinterventional-certificate-traca}. The selector-free bounds \eqref{eq:gaussian-joint-certificate-traca-recovered} and \eqref{eq:empirical-joint-certificate-traca-recovered} then follow from the monotonicity of coordinate restriction:
\begin{align}
  \|S_{[d]\setminus J_\iota}M\|_2\le \|M\|_2,
  \qquad
  \|S_{[d]\setminus J_\iota}M\|_F\le \|M\|_F,
  \qquad
  d-|J_\iota|\le d,
\end{align}
applied to the corresponding terms in \eqref{eq:gaussian-postinterventional-certificate-traca} and \eqref{eq:empirical-postinterventional-certificate-traca}.
\end{proof}

\begin{corollary}[Single-query certificates]
\label{cor:single-query-pair-certificates}
Let $\Qcal_{\mathrm{single}}:=\{(\iota_0,O_0)\}$ for some $\iota_0\in\mathcal I$ and $O_0\subseteq[d]\setminus J_{\iota_0}$. Then for every $\lintau$,
\begin{equation}
  \mathcal E_{\rho,\Qcal_{\mathrm{single}}}^{\mathrm{joint}}(\lintau)
  \le
  \delta_{\iota_0,O_0}^{\rho}(\lintau)^2,
  \label{eq:gaussian-single-query-certificate}
\end{equation}
and
\begin{equation}
  \mathcal E_{U,\Qcal_{\mathrm{single}}}^{\mathrm{joint}}(\lintau)
  \le
  \delta_{\iota_0,O_0}^{U}(\lintau)^2.
  \label{eq:empirical-single-query-certificate}
\end{equation}
In particular, any minimiser of the corresponding $\Qcal_{\mathrm{single}}$-restricted objective satisfies both bounds at that minimiser.
\end{corollary}
\begin{proof}
Immediate from Theorems~\ref{thm:q-restricted-gaussian-certificate} and~\ref{thm:q-restricted-empirical-certificate} by taking \(|\Qcal|=1\).
\end{proof}

\begin{remark}
\label{rem:q-restricted-sharpening}
Let
\begin{align}
  O_\iota^{\mathrm{post}}:=[d]\setminus J_\iota,
  \qquad
  \delta_{\iota,\mathrm{post}}^{\rho}(\lintau)
  :=
  \delta_{\iota,O_\iota^{\mathrm{post}}}^{\rho}(\lintau),
  \qquad
  \delta_{\iota,\mathrm{post}}^{U}(\lintau)
  :=
  \delta_{\iota,O_\iota^{\mathrm{post}}}^{U}(\lintau).
\end{align}
The query-restricted certificates have two immediate consequences. First, for every included query $(\iota,O_\iota)\in\Qcal$ and every
$\lintau$
\begin{equation}
  \delta_{\iota,O_\iota}^{\rho}(\lintau)^2
  \le
  \delta_{\iota,\mathrm{post}}^{\rho}(\lintau)^2,
  \qquad
  \delta_{\iota,O_\iota}^{U}(\lintau)
  \le
  \delta_{\iota,\mathrm{post}}^{U}(\lintau).
  \label{eq:q-restricted-sharper}
\end{equation}
Thus, on any query explicitly included in $\Qcal$, the restricted certificate is never worse than the corresponding full post-interventional one, though the gap may be small when the mechanism and environment terms dominate the bound. The selector-free bounds \eqref{eq:gaussian-joint-certificate-traca-recovered} and \eqref{eq:empirical-joint-certificate-traca-recovered} are looser still, but they are often convenient when no query family is specified. Second, if a pair $(\iota,O)$ does not belong to $\Qcal$, then the restricted objective contains no term associated with that query. Consequently, no certificate of the form \eqref{eq:q-restricted-gaussian-certificate} or \eqref{eq:q-restricted-empirical-certificate} is induced for that excluded query by the restricted objective alone. In this sense, query-restricted training sharpens robustness guarantees on the queries that are explicitly targeted, but does not provide analogous guarantees for outputs left out of $\Qcal$.
\end{remark}

\subsection{Specialisation to query-level transport}
\label{subsec:q-restricted-query-transport}
The certificates above control the worst-case discrepancy between the transported source and the target at the level of the
post-interventional marginals selected by $\Qcal$. We now specialise these model-level guarantees to downstream causal queries. The key observation is that once a query functional is Lipschitz with respect to the discrepancy used in the certificate, the abstraction-error bound transfers directly to the query value itself. In the Gaussian case, the relevant discrepancy is $\mathcal W_2$, while in the empirical case, it is the Frobenius norm. This yields certified target-query intervals centered at the transported-source query value for each designated query pair $(\iota,O_\iota)\in\Qcal$.

\begin{theorem}
\label{thm:q-restricted-query-level-bounds}
Let $(\iota,O_\iota)\in\Qcal$ and let $Q_t^{(\iota,O_\iota)}:=\Phi_{(\iota,O_\iota)}\!\left(\pi_{O_\iota\#}P_t^{(\iota)} \right)$ be a causal query. Then,
\begin{enumerate}[label=(\roman*)]
  \item \textbf{Gaussian case.}
  Suppose $\Phi_{(\iota,O_\iota)}$ is $L_{(\iota,O_\iota)}$-Lipschitz with respect to $\mathcal W_2$ on probability measures over $\mathbb R^{|O_\iota|}$. Then for every admissible $(\Delta W,\mu_t,\Sigma_t)$,
  \begin{align}
    \Bigl|
      \Phi_{(\iota,O_\iota)}\!\left(
        \pi_{O_\iota\#}P_t^{(\iota)}(\Delta W,\mu_t,\Sigma_t)
      \right)
      -
      \Phi_{(\iota,O_\iota)}\!\left(
        \pi_{O_\iota\#}(\lintau_{\#}P_s^{(\iota)})
      \right)
    \Bigr|
    \le
    L_{(\iota,O_\iota)}\,\delta_{\iota,O_\iota}^{\rho}(\lintau).
    \label{eq:q-restricted-query-gaussian}
  \end{align}
  \item \textbf{Empirical case.}
  Let $\widehat\Phi_{(\iota,O_\iota)}$ be a functional on $|O_\iota|\times N$ data matrices that is $\widehat L_{(\iota,O_\iota)}$-Lipschitz with respect to the Frobenius norm. Then for every admissible $(\Delta W,\Theta)$,
  \begin{align}
    \Bigl|
      \widehat\Phi_{(\iota,O_\iota)}\!\left(
        S_{O_\iota}A'_\iota(\Delta W)(U^s+\Theta)
      \right)
      -
      \widehat\Phi_{(\iota,O_\iota)}\!\left(
        S_{O_\iota}\lintau A_\iota U^s
      \right)
    \Bigr|
    \le
    \widehat L_{(\iota,O_\iota)}\,\delta_{\iota,O_\iota}^{U}(\lintau).
    \label{eq:q-restricted-query-empirical}
  \end{align}
\end{enumerate}
\end{theorem}

\begin{proof}
For part (i), the Lipschitz property of $\Phi_{(\iota,O_\iota)}$ implies that for any two probability measures $\nu_1,\nu_2$ on $\mathbb R^{|O_\iota|}$,
\begin{align}
  \bigl|
    \Phi_{(\iota,O_\iota)}(\nu_1)
    -
    \Phi_{(\iota,O_\iota)}(\nu_2)
  \bigr|
  \le
  L_{(\iota,O_\iota)}\,\mathcal W_2(\nu_1,\nu_2).
\end{align}
Apply this with
\begin{align}
  \nu_1
  =
  \pi_{O_\iota\#}P_t^{(\iota)}(\Delta W,\mu_t,\Sigma_t),
  \qquad
  \nu_2
  =
  \pi_{O_\iota\#}(\lintau_{\#}P_s^{(\iota)}).
\end{align}
By the pointwise bound established in the proof of Theorem~\ref{thm:q-restricted-gaussian-certificate},
\begin{align}
  \mathcal W_2\!\Bigl(
    \pi_{O_\iota\#}P_t^{(\iota)}(\Delta W,\mu_t,\Sigma_t),\;
    \pi_{O_\iota\#}(\lintau_{\#}P_s^{(\iota)})
  \Bigr)
  \le
  \delta_{\iota,O_\iota}^{\rho}(\lintau),
\end{align}
which gives \eqref{eq:q-restricted-query-gaussian}. For part (ii), use the Lipschitz property of $\widehat\Phi_{(\iota,O_\iota)}$ with respect to the Frobenius norm:
\begin{align}
  \bigl|
    \widehat\Phi_{(\iota,O_\iota)}(Y_1)
    -
    \widehat\Phi_{(\iota,O_\iota)}(Y_2)
  \bigr|
  \le
  \widehat L_{(\iota,O_\iota)}\,\|Y_1-Y_2\|_F
\end{align}
for any two $|O_\iota|\times N$ matrices $Y_1,Y_2$. Apply this with
\begin{align}
  Y_1
  =
  S_{O_\iota}A'_\iota(\Delta W)(U^s+\Theta),
  \qquad
  Y_2
  =
  S_{O_\iota}\lintau A_\iota U^s.
\end{align}
The pointwise estimate underlying Theorem~\ref{thm:q-restricted-empirical-certificate} gives
\begin{align}
  \bigl\|
    S_{O_\iota}\bigl(
      \lintau A_\iota U^s
      -
      A'_\iota(\Delta W)(U^s+\Theta)
    \bigr)
  \bigr\|_F
  \le
  \delta_{\iota,O_\iota}^{U}(\lintau),
\end{align}
and substituting this bound yields
\eqref{eq:q-restricted-query-empirical}.
\end{proof}
Equivalently, writing $\ell:=Q_{\tau \# s}^{(\iota,O_\iota)} - L_{(\iota,O_\iota)}
\delta_{\iota,O_\iota}^{\rho}(\lintau)$ and $L:=Q_{\tau \# s}^{(\iota,O_\iota)} +
L_{(\iota,O_\iota)}\delta_{\iota,O_\iota}^{\rho}(\lintau)$, every admissible
target satisfies $\ell \le Q_t^{(\iota,O_\iota)} \le L$: the certified interval, centered at the transported-source value. These certificates above are general: they bound arbitrary Lipschitz query functionals by first controlling a discrepancy between full
query-relevant marginals. This generality is useful, but conservative
for linear and mean queries. The modulus $\alpha_\iota$ controls the
worst admissible mechanism perturbation in every input and output
direction, whereas a linear query reads out only one direction of the
selected marginal. In particular, for
\begin{align}
\Phi_{(\iota,O_\iota)}(\nu) = \mathbb E_{X\sim\nu}[c^\top X], \qquad c\in\mathbb R^{|O_\iota|},
\end{align}
define the lifted readout
\begin{align}
q := S_{O_\iota}^\top c \in \mathbb R^d.
\end{align}
Then the query depends on the full post-interventional vector only through the scalar $q^\top X$.  Concretely, $q$ is a vector of weights recording which coordinates the query reads and with what coefficients: for the single-outcome queries of our benchmarks, it is the indicator of the outcome node, so that $q^\top X$ is that coordinate and nothing else. This motivates replacing the global mechanism modulus with a directional one.

\begin{defn}[Directional mechanism modulus]
\label{def:directional-mechanism-modulus}
For an intervention $\iota\in\mathcal I$, a readout direction
$q\in\mathbb R^d$, and a vector $v\in\mathbb R^d$, define
\begin{align}
m_{\iota,q}(v) := \sup_{\Delta W\in\mathcal A_W} \left| q^\top\bigl(A_\iota'(\Delta W)-A_\iota\bigr)v \right|.
\end{align}
\end{defn}
The contrast with $\alpha_\iota$ is that the latter is a spectral norm, hence a supremum over the input and output directions as well as over $\Delta W$, whereas $m_{\iota,q}(v)$ fixes the input to $v$ and the output to $q$, leaving only the mechanism adversarial.
\begin{corollary}[Directional certificate for linear queries]
\label{cor:directional-linear-certificate}
Let $(\iota,O_\iota)\in\Qcal$, and let
\begin{align}
\Phi_{(\iota,O_\iota)}(\nu) = \mathbb E_{X\sim\nu}[c^\top X]
\end{align}
be a linear query on $\mathbb R^{|O_\iota|}$. Set $q:=S_{O_\iota}^\top c$, and define the transported-source query value
\begin{align}
Q_{\tau \# s}^{(\iota,O_\iota)} := \Phi_{(\iota,O_\iota)} \left( \pi_{O_\iota\#}\lintau_{\#}P_s^{(\iota)} \right).
\end{align}
Then every admissible target satisfies
\begin{align}
\left| Q_t^{(\iota,O_\iota)} - Q_{\tau \# s}^{(\iota,O_\iota)} \right| 
\le\; \underbrace{ \left|q^\top(\lintau-I)A_\iota v\right| }_{\textnormal{transport mismatch}} + \underbrace{ m_{\iota,q}(v) }_{\textnormal{mechanism shift}}  + \underbrace{ \varepsilon \sup_{\Delta W\in\mathcal A_W} \left\| M_{\mathcal K}A_\iota'(\Delta W)^\top q \right\|_2 }_{\textnormal{environment shift}},
\label{eq:directional-certificate}
\end{align}
where $v=\mu_s^{(\iota)}$ in the Gaussian case and $v=\bar U^{s,(\iota)} := \frac{1}{N} U^{s,(\iota)}\mathbf 1$ in the empirical case: the effective exogenous mean under $\iota$ of Section~\ref{subsec:gaussian-implementation}, which carries the intervention value at the clamped coordinates. The first-order part of $m_{\iota,q}(v)$ admits explicit support-function expressions for each mechanism ambiguity geometry, and the full directional modulus is controlled by this first-order part plus a finite higher-order remainder; see Appendix~\ref{app:directional-certificates}.
\end{corollary}

\begin{proof}
This is the directional specialisation of Theorem~\ref{thm:q-restricted-query-level-bounds}. Since the query is a rank-one readout $q^\top X$, the residual decomposition can be projected onto $q$ before taking norms. The mechanism term is then controlled by Definition~\ref{def:directional-mechanism-modulus}, while the environment term follows from Cauchy-Schwarz and the fact that admissible environment perturbations are supported only on the shifted coordinates. The full proof and the support-function expressions for $m_{\iota,q}$ are given in Appendix~\ref{app:directional-certificates}.
\end{proof}
We refer to the bounds of Theorems~\ref{thm:q-restricted-gaussian-certificate} and~\ref{thm:q-restricted-empirical-certificate} as the \emph{general} certificate since they hold for any Lipschitz functional and any admissible target, and to Corollary~\ref{cor:directional-linear-certificate} as the \emph{directional} certificate; these are the two intervals reported in Section~\ref{sec:experiments}.

\begin{remark}[Why the directional certificate is tighter]
\label{rem:directional-tightening}
The general certificates use the global perturbation modulus $\alpha_\iota$, which controls the entire operator $(A_\iota'(\Delta W)-A_\iota)$. The directional certificate removes two sources of conservatism. First, the mechanism term controls only the scalar perturbation seen by the query direction $q$, and in the empirical case, the query reads out the sample mean $\bar U^s$ rather than the full sample cloud $U^s$; for centered data, this alone is a substantial reduction. Second, the environment term uses the masked operator
\begin{align}
\left\| M_{\mathcal K}A_\iota'(\Delta W)^\top q \right\|_2
\end{align}
in place of the unmasked full-output quantity $(\|S_{O_\iota}A_\iota\|_2+\alpha_\iota)$, because admissible environment shifts are pinned outside the shifted coordinates. Thus, the broad certificates protect all Lipschitz functionals of the selected marginal, while the directional certificate exploits the exact rank-one readout used by mean queries. In the extreme case, the mechanism term vanishes outright: if the signal reaching a shifted coefficient is zero under a given intervention, no admissible perturbation of it can move the query, and $m_{\iota,q}(v)=0$ while $\alpha_\iota$ remains at its full value. This is the mechanism behind the coincident directional widths on $\mathbb E[Y\mid do(X{=}0)]$ in Figure~\ref{fig:ate-width}.
\end{remark}

Theorem~\ref{thm:q-restricted-query-level-bounds} turns the preceding model-level certificates into query-specific bounds for arbitrary Lipschitz functionals, while Corollary~\ref{cor:directional-linear-certificate} sharpens this transfer for the linear and mean queries that dominate our benchmarks. More broadly, for a fixed transport map $\lintau$, the ambiguity sets $\mathcal A_W$, $\mathcal A_\rho(\varepsilon)$, and $\mathcal A_U(\varepsilon)$ determine the admissible target marginals associated with the queries in $\Qcal$; in the Gaussian case, these are exactly
\begin{align}
\left\{
\left(
\pi_{O_\iota\#}P_t^{(\iota)}(\Delta W,\mu_t,\Sigma_t)
\right)_{(\iota,O_\iota)\in\Qcal}
:
\Delta W\in\mathcal A_W,
(\mu_t,\Sigma_t)\in\mathcal A_\rho(\varepsilon)
\right\},
\end{align}
and analogously in the empirical case with $S_{O_\iota}A_\iota'(\Delta W)(U^s+\Theta)$ and $\Theta\in\mathcal A_U(\varepsilon)$. In this sense, the robust objectives $\mathcal E_{\rho,\Qcal}^{\mathrm{joint}}(\lintau)$ and $\mathcal E_{U,\Qcal}^{\mathrm{joint}}(\lintau)$ measure worst-case abstraction discrepancy over these admissible target families, while the Gaussian and empirical certificates provide the corresponding model-level bounds. The Lipschitz theorem converts those model-level bounds into certified intervals for arbitrary designated queries, and the directional corollary provides a sharper specialisation when the query is a linear readout. This is close in spirit to the target-class viewpoint in partial-transportability analyses~\citep{jalaldoust2024partial}, but here the relevant family is induced by worst-case robust ambiguity sets rather than by the set of target domains compatible with the source distributions and graphical assumptions. Once a single transport map $\lintau$ is learned, the logic of the framework is:
\begin{align}
\Delta W
\longrightarrow
\gamma_\iota
\longrightarrow
\alpha_\iota
\longrightarrow
\delta_{\iota,O_\iota}(\lintau)
\longrightarrow
\text{certified interval for }
Q_t^{(\iota,O_\iota)}.
\end{align}

\section{Experiments}
\label{sec:experiments}
We evaluate TraCA on three synthetic benchmark families and one real-world benchmark. The synthetic families isolate the two regimes that the framework targets: a query that classical transportability cannot handle ($\mathcal R_1$), and a target-agnostic setting in which no target data are available at all ($\mathcal R_2$). The real-world benchmark fixes a single observed target domain with invariant mechanisms and a shifted covariate distribution, testing the certificate on data we did not generate. Details on the full structural equations, parameters, intervention sets, graphs, and the remaining configurations are deferred to Appendix~\ref{app:benchmarks}. Across the four benchmarks, we ask three questions.
\begin{enumerate}[label=\textbf{Q\arabic*.}, leftmargin=*]
\item \emph{Does the certificate hold, and is it informative?} This is the claim the theory makes: the certified interval $[\ell,L]$ contains the true query value for every admissible target. A guarantee of this form is trivially satisfiable by a sufficiently wide interval, so coverage must be read jointly with width. We call a certificate \emph{informative} when its width is below $10\times$ the mean absolute query value (the query scale, QS), and \emph{vacuous} otherwise, and we compare widths only where they are informative. The factor of ten is a deliberately loose order-of-magnitude convention rather than a tuned threshold: an interval ten times wider than the typical query value carries almost no information about that query, whatever the units. Normalising by the query scale makes the criterion scale-free, so it transfers across benchmarks whose queries differ by orders of magnitude. The certificates themselves are, of course, scale-dependent, since they bound $|Q_t-Q_{\tau\#s}|$, which carries the units of the query; this is precisely why widths are reported and compared in QS units rather than in absolute terms.
\item \emph{Does a directional prior improve the learned map?} A belief about the direction of the target shift is encoded by recentering the mechanism box at an offset $\delta$ (Section~\ref{subsec:mechanism-ambiguity}).
\item \emph{Can the operating radius be calibrated without interventional target data?} Where the target can be observed but not intervened upon, the radius must be chosen from observational quantities alone.
\end{enumerate}
Throughout, we call a run \emph{symmetric} when $\delta=0$ and \emph{offset} when $\delta\neq 0$, and reserve \emph{general} and \emph{directional} for the two certificates of Section~\ref{sec:provable-robustness}; both certificates are computed under both variants.
The codebase is available at \url{https://github.com/yfelekis/traca}.

\subsection{Benchmarks}
\label{subsec:benchmarks}
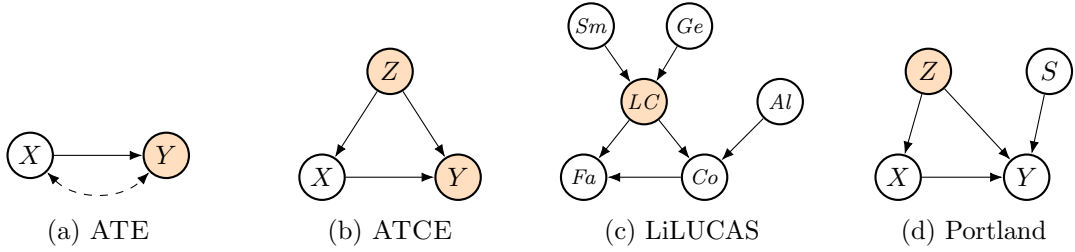
\begin{figure}[t]
\centering
\begin{subfigure}[b]{0.24\textwidth}
\centering
\begin{tikzpicture}[
    >=Latex,
    obs/.style={circle, draw, thick, minimum size=6mm, inner sep=0pt, font=\small},
    shifted/.style={circle, draw, thick, minimum size=6mm, inner sep=0pt,
      font=\small, fill=orange!25},
    sel/.style={rectangle, draw, thick, minimum size=5mm, inner sep=0pt, font=\small},
    every edge/.style={draw, ->, thick}
]
\node[obs]     (X)  at (-0.9,0)   {$X$};
\node[shifted] (Y)  at ( 0.9,0)   {$Y$};
\draw[->] (X) to (Y);
\draw[<->, dashed, bend right=45] (X) to (Y);
\end{tikzpicture}
\subcaption{\textsc{ATE}}
\label{fig:sel-ate}
\end{subfigure}
\hfill
\begin{subfigure}[b]{0.24\textwidth}
\centering
\begin{tikzpicture}[
    >=Latex,
    obs/.style={circle, draw, thick, minimum size=6mm, inner sep=0pt, font=\small},
    shifted/.style={circle, draw, thick, minimum size=6mm, inner sep=0pt,
      font=\small, fill=orange!25},
    sel/.style={rectangle, draw, thick, minimum size=5mm, inner sep=0pt, font=\small},
    every edge/.style={draw, ->, thick}
]
\node[shifted] (Z)  at ( 0   ,1.4) {$Z$};
\node[obs]     (X)  at (-0.9 ,0)   {$X$};
\node[shifted] (Y)  at ( 0.9 ,0)   {$Y$};
\draw[->] (Z) to (X);
\draw[->] (Z) to (Y);
\draw[->] (X) to (Y);
\end{tikzpicture}
\subcaption{\textsc{ATCE}}
\label{fig:sel-atce}
\end{subfigure}
\hfill
\begin{subfigure}[b]{0.24\textwidth}
\centering
\begin{tikzpicture}[
    >=Latex,
    obs/.style={circle, draw, thick, minimum size=6mm, inner sep=0pt, font=\scriptsize},
    shifted/.style={circle, draw, thick, minimum size=6mm, inner sep=0pt,
      font=\scriptsize, fill=orange!25},
    sel/.style={rectangle, draw, thick, minimum size=5mm, inner sep=0pt, font=\scriptsize},
    every edge/.style={draw, ->, thick}
]
\node[obs]     (Sm) at (-1.0, 2.1) {$\mathit{Sm}$};
\node[obs]     (Ge) at ( 0.3, 2.1) {$\mathit{Ge}$};
\node[obs]     (Al) at ( 1.5, 1.1) {$\mathit{Al}$};
\node[shifted] (LC) at (-0.3, 1.1) {$\mathit{LC}$};
\node[obs]     (Co) at ( 0.5, 0.1) {$\mathit{Co}$};
\node[obs]     (Fa) at (-1.1, 0.1) {$\mathit{Fa}$};
\draw[->] (Sm) to (LC);
\draw[->] (Ge) to (LC);
\draw[->] (LC) to (Co);
\draw[->] (Al) to (Co);
\draw[->] (LC) to (Fa);
\draw[->] (Co) to (Fa);
\end{tikzpicture}
\subcaption{LiLUCAS}
\label{fig:sel-lilucas}
\end{subfigure}
\hfill
\begin{subfigure}[b]{0.24\textwidth}
\centering
\begin{tikzpicture}[
    >=Latex,
    obs/.style={circle, draw, thick, minimum size=6mm, inner sep=0pt, font=\small},
    shifted/.style={circle, draw, thick, minimum size=6mm, inner sep=0pt,
      font=\small, fill=orange!25},
    sel/.style={rectangle, draw, thick, minimum size=5mm, inner sep=0pt, font=\small},
    every edge/.style={draw, ->, thick}
]
\node[shifted] (Z)  at (-0.6, 1.4) {$Z$};
\node[obs]     (S)  at ( 1.0, 1.4) {$S$};
\node[obs]     (X)  at (-1.0, 0)   {$X$};
\node[obs]     (Y)  at ( 0.7, 0)   {$Y$};
\draw[->] (Z) to (X);
\draw[->] (Z) to (Y);
\draw[->] (S) to (Y);
\draw[->] (X) to (Y);
\end{tikzpicture}
\subcaption{Portland}
\label{fig:sel-portland}
\end{subfigure}
\caption{Causal graphs of the four benchmarks. Shaded nodes are shifted across domains; the dashed arc in (a) is latent confounding.}
\label{fig:benchmark-graphs}
\end{figure}

Figure~\ref{fig:benchmark-graphs} shows the causal graph of each benchmark and its shifted nodes.
\paragraph{ATE ($\mathcal R_1$).}
ATE is a two-variable semi-Markovian graph $X\to Y$ with latent confounding $X\leftrightarrow Y$ and a domain shift on the mechanism of $Y$. Because the confounder is hidden, there is no valid adjustment set, so $P(Y\mid do(X))$ is not transportable by the classical do-calculus transport formula (Figure 6b from \citet{pearl2014external}). It is therefore our $\mathcal R_1$ benchmark: the certified interval is informative precisely where classical transportability is silent. We report the certified interval for the two queries $\mathbb E_t[Y\mid do(X{=}0)]$ and $\mathbb E_t[Y\mid do(X{=}1)]$.

\paragraph{ATCE ($\mathcal R_2$).}
ATCE is a three-variable graph $Z\to X\to Y$ with an additional $Z\to Y$ edge and is one of our $\mathcal R_2$ benchmarks: it is target-agnostic, and no target data enters at any stage. As in ATE, we evaluate the query family $\{\mathbb E_t[Y\mid do(X{=}0)],\,\mathbb E_t[Y\mid do(X{=}1)]\}$, whose queries remain sensitive to mechanism perturbation. 

\paragraph{LiLUCAS ($\mathcal R_2$).}
LiLUCAS is a six-node Markovian DAG based on the original LUCAS dataset,\footnote{\href{https://www.causality.inf.ethz.ch/data/LUCAS.html}{www.causality.inf.ethz.ch/data/LUCAS.html}} modelling lung cancer with roots \emph{Smoking} (\textit{Sm}), \emph{Genetics} (\textit{Ge}), and \emph{Allergy} (\textit{Al}), and downstream \emph{LungCancer} (\textit{LC}), \emph{Coughing} (\textit{Co}), and \emph{Fatigue} (\textit{Fa}); the shift is placed on the \textit{LC} mechanism. As with ATCE, this benchmark is target-agnostic. We evaluate the query family $\{\mathbb E_t[\mathit{LC}\mid do(\mathit{Sm}{=}v)],\ \mathbb E_t[\mathit{LC}\mid do(\mathit{Ge}{=}v)]\}$ for $v\in\{0,1\}$.

\paragraph{Portland (real-world).}
The Portland benchmark \citep{Tabell2026} has been constructed to ask whether riparian tree-canopy cover causally affects dissolved oxygen in urban streams and whether that effect transports from the observed source watersheds to a held-out target, Fanno Creek, whose environmental conditions differ from the source. We use the dataset and causal graph of the original study directly. \citet{Tabell2026} model the system as a linear-Gaussian SCM over four variables: canopy cover $X$, dissolved oxygen $Y$, a one-dimensional environmental summary $Z$ (Appendix~\ref{app:portland}), and season $S$, with graph $Z\to X$, $Z\to Y$, $S\to Y$, $X\to Y$ and structural coefficients $w_{ZX}, w_{ZY}, w_{SY}, w_{XY}$, and treat the structural mechanisms as invariant across watersheds while the environmental context shifts. Each structural equation is fit by ordinary least squares, without intercept, on the source observational data, after standardising all variables to zero mean and unit variance using source statistics; the exogenous-noise variances are set to the unbiased residual variance of the corresponding regression. Fanno Creek is the target environment for which we have $N_t=100$ observational samples, and the remaining watersheds form the source ($N_s=247$).

We share the modelling assumptions of the original study, the linear-Gaussian form, and the same four-variable graph but differ in how the interventional query is obtained. \citet{Tabell2026} estimate $\mathbb E_t[Y\mid do(X{=}x)]$ by back-door adjustment, regressing $Y$ on $X, Z, S$ and averaging over the target covariate distribution; because TraCA transports a structural model rather than a single estimate, we instead realise each $do(X{=}x)$ on the fitted SCM, fixing $X$ and propagating the exogenous noise through the resulting interventional propagator. For this graph, the two coincide exactly since both give $\mathbb E_t[Y\mid do(X{=}x)] = w_{ZY}\,\mathbb E_t[Z] + w_{SY}\,\mathbb E_t[S] + w_{XY}\,x$ under the linear-Gaussian model, as the only back-door path $X\leftarrow Z\to Y$ is blocked by $Z$ and $S$ opens none; the choice of representation therefore does not change the query, only the object that the transport map and certificate act on. We evaluate interventions $x\in\{45,50,55,60\}\%$ canopy cover, the values considered by the original study.

\subsection{Setup and evaluation protocol}
\label{subsec:setup}
\paragraph{Models and objective.}
The three synthetic benchmarks are generated from linear additive-noise SCMs with Gaussian exogenous noise; Portland is fitted to real data as described above and is likewise a linear-Gaussian SCM. The optimisation objective differs by benchmark: ATE, ATCE, and Portland use the Gaussian objective, while LiLUCAS is presented under the empirical residual-based objective, with a Gaussian objective variant retained as a cross-check.

\paragraph{Shift geometry and budget.}
Although the certificates of Section~\ref{sec:provable-robustness} hold for all mechanism geometries, the three synthetic benchmarks use entrywise (box) constraints: each shifted mechanism coefficient $W_{jk}$ is given its own interval budget $\Delta W_{jk}\in[\delta_{jk}-\eta,\ \delta_{jk}+\eta]$, so the mechanism radius $\eta$ is the per-coefficient perturbation half-width; symmetric about zero when $\delta{=}0$ and recentered at the declared offset $\delta$ in the offset case. In the notation of Section~\ref{subsec:mechanism-ambiguity}, this is $\mathcal A_W^{\mathrm{box}}(\delta, B)$ with $b_{jk}=\eta$ on the budgeted coefficients and $b_{jk}=0$ elsewhere, so the applicable amplification bound is the entrywise one, Eq.~\eqref{eq:gamma-box-traca}, evaluated at the effective half-widths $|\delta|+\eta$. The entrywise form is more interpretable than a global norm bound and matches how domain shifts are typically specified in practice. The budgeted mechanism coefficients are the incoming edges of each benchmark's shifted node: the $X\to Y$ coefficient $w_{X\to Y}$ in ATE; the two incoming edges of $Y$, $w_{Z\to Y}$, and $w_{X\to Y}$ in ATCE; and the two incoming edges of \emph{LC}, $w_{\mathit{Sm}\to\mathit{LC}}$ and $w_{\mathit{Ge}\to\mathit{LC}}$, in LiLUCAS. In ATCE, the root $Z$ additionally carries an environmental (exogenous) shift, and in Portland, the sole shifted node $Z$ is a root with no incoming edges, so no mechanism coefficient moves ($\eta{=}0$), and the entire budget falls on the environment of $Z$: a Gelbrich ball $\mathcal A_\rho(\varepsilon)$ of radius $\varepsilon$ on $Z$'s exogenous Gaussian law, centered at the source. Table~\ref{tab:benchmark-configs} in Appendix~\ref{app:benchmarks} summarises these choices across all four benchmarks.

The training radius is swept over $\{0,\,0.2,\,0.5,\,1.0,\,2.0,\,4.0\}$; for ATCE and LiLUCAS, the mechanism and environment radii are tied ($\varepsilon=\eta$); for ATE, they vary independently over the full $(\varepsilon,\eta)$ grid, and for Portland, only the environment radius $\varepsilon$, the Gelbrich-ball radius on $Z$, is swept ($\eta=0$), since the shift is purely environmental. Throughout, we write $r_{\mathrm{train}}$ for the ambiguity radius at which a map is trained (equal to $\varepsilon{=}\eta$ on the tied-radius benchmarks, and to $\varepsilon$ alone on Portland), and $r_{\mathrm{test}}$ for the magnitude of the target shift a map is evaluated against. The two are independent: a map trained against $\mathcal A(r_{\mathrm{train}})$ can be evaluated against shifts drawn from $\mathcal A(r_{\mathrm{test}})$ both inside and beyond the ball for which it was trained, the two sets coinciding only when $r_{\mathrm{train}}=r_{\mathrm{test}}$.

\paragraph{Training protocol.}
In all cases, the transport map $\lintau$ is a diagonal rescaling of the shifted nodes (block-diagonal in the semi-Markovian case), learned by minimising the worst-case transport loss over the ambiguity set. Throughout, we train and certify against the $\Qcal$-restricted objective of Section~\ref{subsec:q-restricted-objectives}, with $\Qcal$ being the query family listed for each benchmark above. We use $5$-fold cross-validation: for each fold and training radius, we learn $\lintau$ on the source split alone and store it; a separate evaluation pass loads the stored maps and scores them against targets. The identity map ($r_{\mathrm{train}}=0$) acts as the internal baseline throughout. In no case does target interventional information enter fitting or radius selection; the target interventional query values are held out entirely and revealed only at scoring time, as the ground-truth values against which coverage of the certified interval is checked.

\paragraph{Benchmarks with a single observed target.}
For ATE and Portland, a single real target is available: the synthetic ATE target and the Fanno Creek watershed, respectively. Only the target's observational distribution enters the pipeline, and only to select the operating radius: we sweep $r_{\mathrm{train}}$ and retain the one minimising the observational transport distance $\mathcal W_2^2\!\bigl(\tau_\#P_s^{\mathrm{obs}},P_t^{\mathrm{obs}}\bigr)$, a criterion that uses no interventional information. Section~\ref{subsec:results} examines when this selector succeeds and when it does not. The sweep here is over the training radius alone: we vary the ambiguity radius and, at each value, check whether the certified interval brackets the real target's interventional query.

\paragraph{Benchmarks with no target data.}
For the target-agnostic benchmarks ATCE and LiLUCAS, no target data exist at all. We instead draw $K=100$ admissible targets from the ambiguity geometry itself, so each map is scored against synthetic shifts it was never shown, and we probe how the maps and certificates respond across shift sizes by sweeping $r_{\mathrm{test}}$. At each test magnitude, the $K$ admissible targets are drawn from the ambiguity geometry rescaled to that radius, so the sampled shifts are precisely those that the certificate claims to cover. In the symmetric variant, the mechanism perturbation $\Delta W$ is drawn from an entrywise box centered at zero; in the offset variant, the same box is recentered at a declared offset $\delta$, fixed at $\delta = 0.5$ throughout our experiments. We also examine how the offset results degrade when the declared direction is wrong. Both the mechanism and the exogenous noise perturbation are applied to the shifted node and propagated structurally through the perturbed propagator, yielding a consistent target SCM. The mechanism perturbation is drawn uniformly from the box, while the exogenous perturbation is added to the same held-out source realisations, paired sample for sample, and normalised to lie exactly on the boundary of $\mathcal A_U(\varepsilon)$ (or of the Gelbrich ball under the Gaussian objective). Every sampled target is therefore admissible by construction, and the environment component is drawn at the extremal magnitude that the certificate must cover. Sampling is seeded independently of the fold, so every trained variant is scored against the same $K$ targets at each magnitude, and differences between variants are attributable to the map rather than to the draw.

\paragraph{Scoring and aggregation.}
For every evaluation instance, indexed by radius, target, fold, and variant, we record the transport error $\mathcal E_{\Qcal}$ from Definition~\ref{def:q-transport-error}; the true query value; whether the certified interval, fixed at the variant's own training radius, contains it; and the interval width. The discrepancy $\mathsf D$ entering $\mathcal E_{\Qcal}$ is the one used for training, $\mathcal W_2^2$ under the Gaussian objective, and $N^{-1}\|\cdot\|_F^2$ under the empirical one. All query-level quantities are computed on the benchmark's query family alone, never pooled over coordinates that the map was not trained to transport, and coverage is reported per query rather than requiring every query in the family to be contained simultaneously. For the $\mathcal R_2$ benchmarks, coverage is aggregated in two steps: the per-fold fraction of the $K$ contained targets, followed by the mean across folds, while transport error is averaged over all sampled targets and folds jointly; for the single-target benchmarks (ATE, Portland), the corresponding quantities are averaged over folds alone. All comparisons in the transport-error tables use a two-sided paired $t$-test across the $K$ perturbation seeds, with the seed as the pairing unit. The tabulated standard deviations are marginal, dominated at large $r_{\mathrm{test}}$ by each seed's target difficulty; variation shared across methods and cancelled by pairing: the paired standard deviation is typically $3$-$12\%$ of the marginal one. In Tables~\ref{tab:dro-lilucas-ew} and~\ref{tab:dro-lilucas-gau}, bold marks the column minimum together with every entry not significantly worse than it ($p>0.05$); in Table~\ref{tab:dro-crossover}, each radius is instead compared against the identity baseline, and bold marks the best entry that improves on it significantly ($p<0.05$).

\subsection{Results \& Analysis}
\label{subsec:results}

\subsubsection{Synthetics}
\label{subsubsec:results_synth}
\begin{figure}[t]
  \centering
  \includegraphics[width=\textwidth]{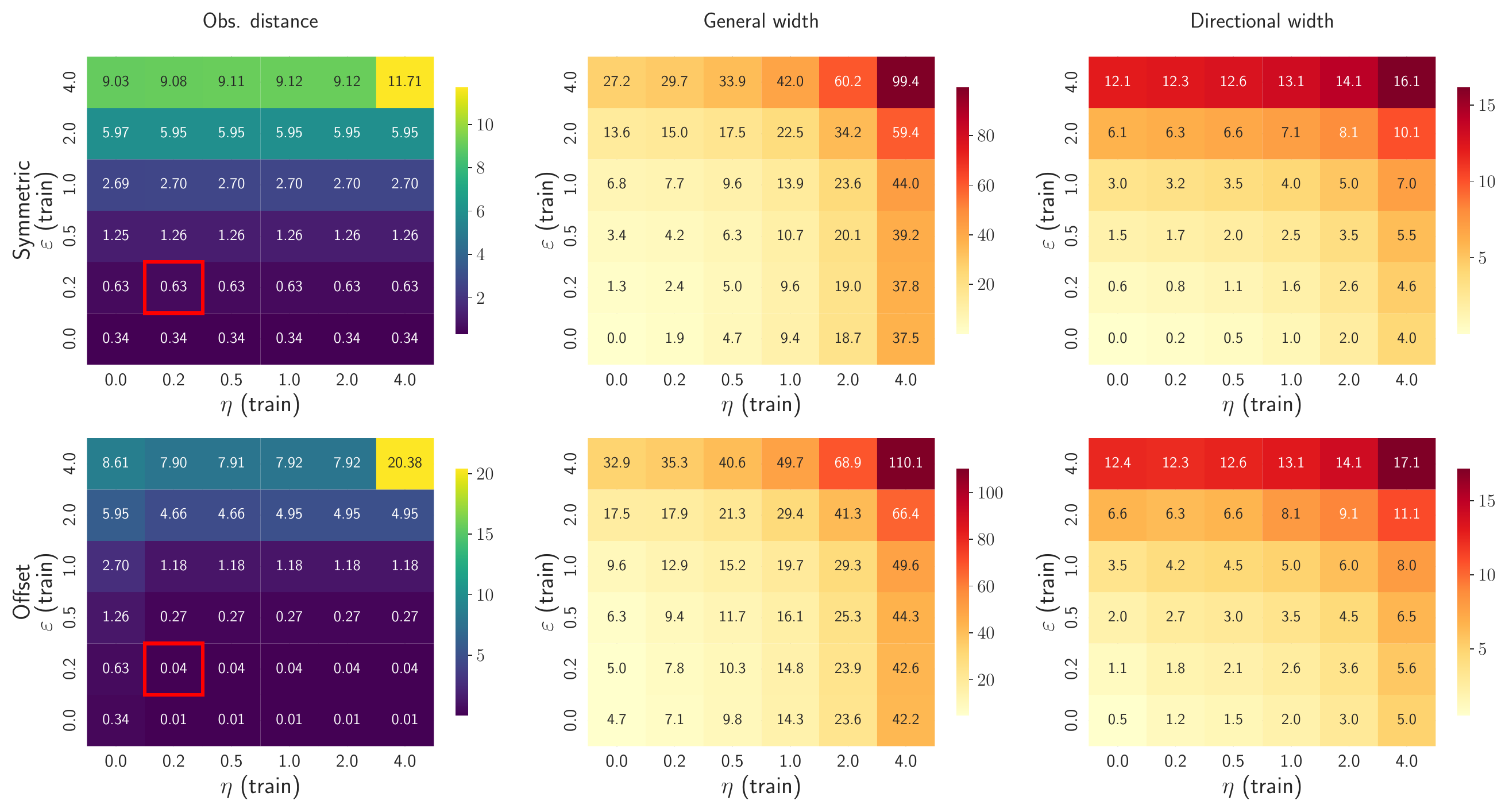}
  \caption{ATE: observational transport distance (left column) and certificate widths (center: general, right: directional) over the $(\varepsilon,\eta)$ grid, for the symmetric (top) and offset (bottom) variants. The observational distance uses no interventional information; the selected cell $(\varepsilon{=}0.2,\,\eta{=}0.2)$, boxed, sits at $\mathcal W_2^2 = 0.040$ under the offset variant.}
  \label{fig:ate-heatmap}
\end{figure}

\paragraph{ATE.}
\emph{The observational selector finds a covering cell.} The first question we take up is (Q3): whether a practitioner can select an operating radius without any interventional target data. Sweeping $(\varepsilon,\eta)$ on the ATE benchmark, we select the cell minimising the observational transport distance $\mathcal W_2^2\!\bigl(\tau_\#P_s^{\mathrm{obs}},P_t^{\mathrm{obs}}\bigr)$, subject to $\varepsilon,\eta>0$. Under the offset variant, the selector lands on $(\varepsilon{=}0.2,\,\eta{=}0.2)$ at $\mathcal W_2^2 = 0.040$ (Figure~\ref{fig:ate-heatmap}). The offset prior recenters the box at $\delta=0.5$, informing the optimiser that the shifted coefficient moves upward rather than in an unknown direction. The learned map $\lintau$ can therefore commit to a correction in that direction, and the transported source lands much closer to the observed target: the same cell under the symmetric variant reads $0.63$, an order of magnitude worse. This tightening is what makes the observational selector land on a covering cell rather than on the identity. The minimiser is not isolated: along $\varepsilon{=}0.2$, the offset distance is $0.63$ at $\eta{=}0$, where the mechanism adversary is inactive and the offset prior cannot yet act, and then $0.04$ at all five values $\eta\ge 0.2$, a five-way tie that we break to the smallest $\eta$ (Figure~\ref{fig:ate-heatmap}).

\begin{figure}[t]
  \centering
  \includegraphics[width=\textwidth]{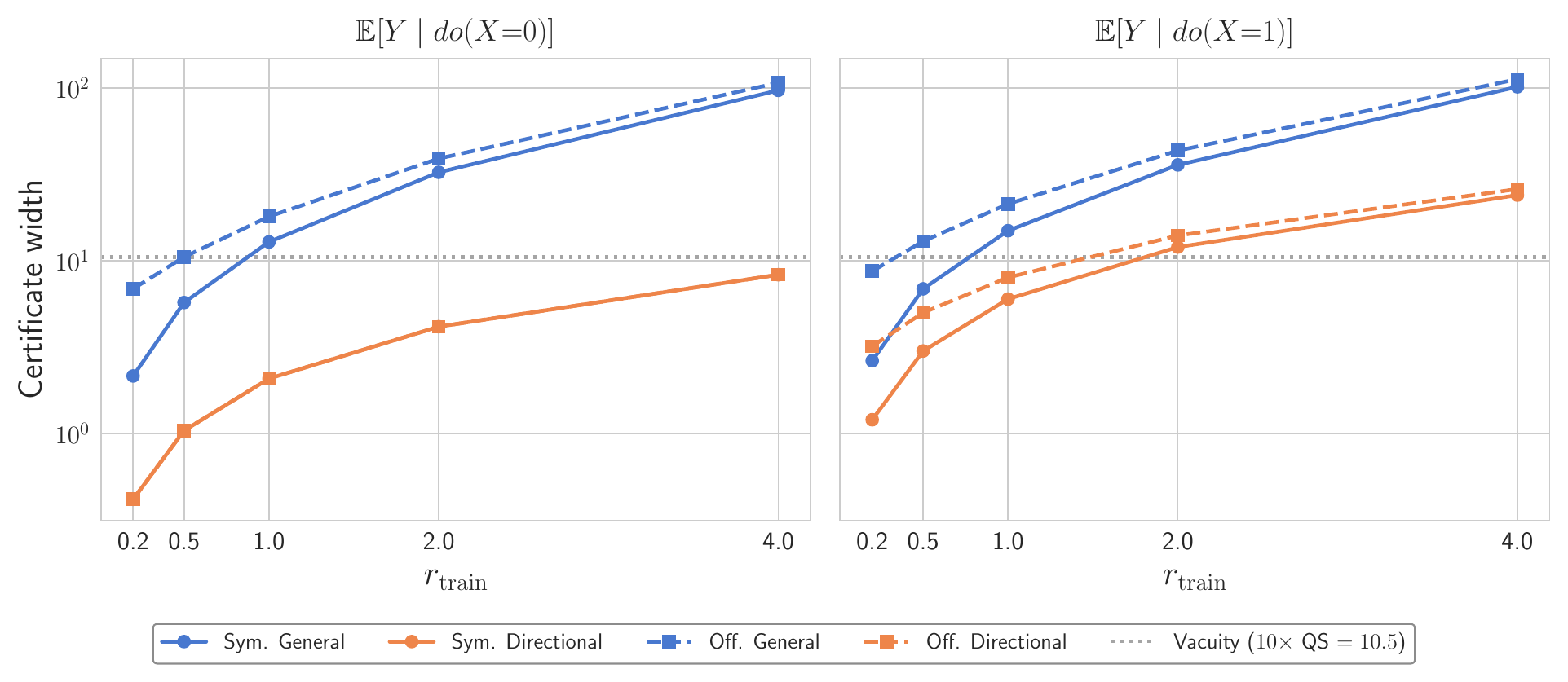}
  \caption{ATE: certificate width along the diagonal $r_{\mathrm{train}} = \varepsilon{=}\eta$ for both variants and both certificates (log scale). The horizontal line is the vacuity threshold ($10\times\text{QS} = 10.5$). Note that in the left panel the two directional curves, symmetric (solid, circles) and offset (dashed, squares), coincide exactly and are drawn on top of one another: the intervention sets $X$ to zero, so the shifted $X\to Y$ coefficient contributes nothing to this query and the offset prior has nothing to act on; the width is then set by the environment term alone. This is the vanishing-mechanism case of Remark~\ref{rem:directional-tightening}.}
  \label{fig:ate-width}
\end{figure}

\emph{The directional bound remains informative where the general becomes vacuous.} Considering \textbf{(Q1)}, Figure~\ref{fig:ate-width} shows how the two certificate widths grow with $r_{\mathrm{train}} = \varepsilon{=}\eta$ along the diagonal. Under the symmetric variant, the general certificate becomes vacuous (width $> 10\times\text{QS} = 10.5$) at $r_{\mathrm{train}}\ge 1.0$, while the directional certificate remains informative across the entire grid on $\mathbb{E}[Y\mid do(X{=}0)]$ and crosses into vacuity only at $r_{\mathrm{train}}{=}2.0$ on $\mathbb{E}[Y\mid do(X{=}1)]$. Under the offset variant, the general widths are larger still ($6.88$ and $8.70$ at $r_{\mathrm{train}}{=}0.2$), while the directional widths remain moderate ($0.42$ and $3.20$). The widening is mostly because the offset raises the effective mechanism bound from $\eta$ to $|\delta|+\eta$, so $\gamma_\iota$ and $\alpha_\iota$ inflate by a factor of $3.5$ and enter the general certificate quadratically. Holding $\lintau$ at its symmetric optimum and changing only the amplification constants reproduces about three quarters of the increase.
\begin{figure}[t]
  \centering
  \includegraphics[width=\textwidth]{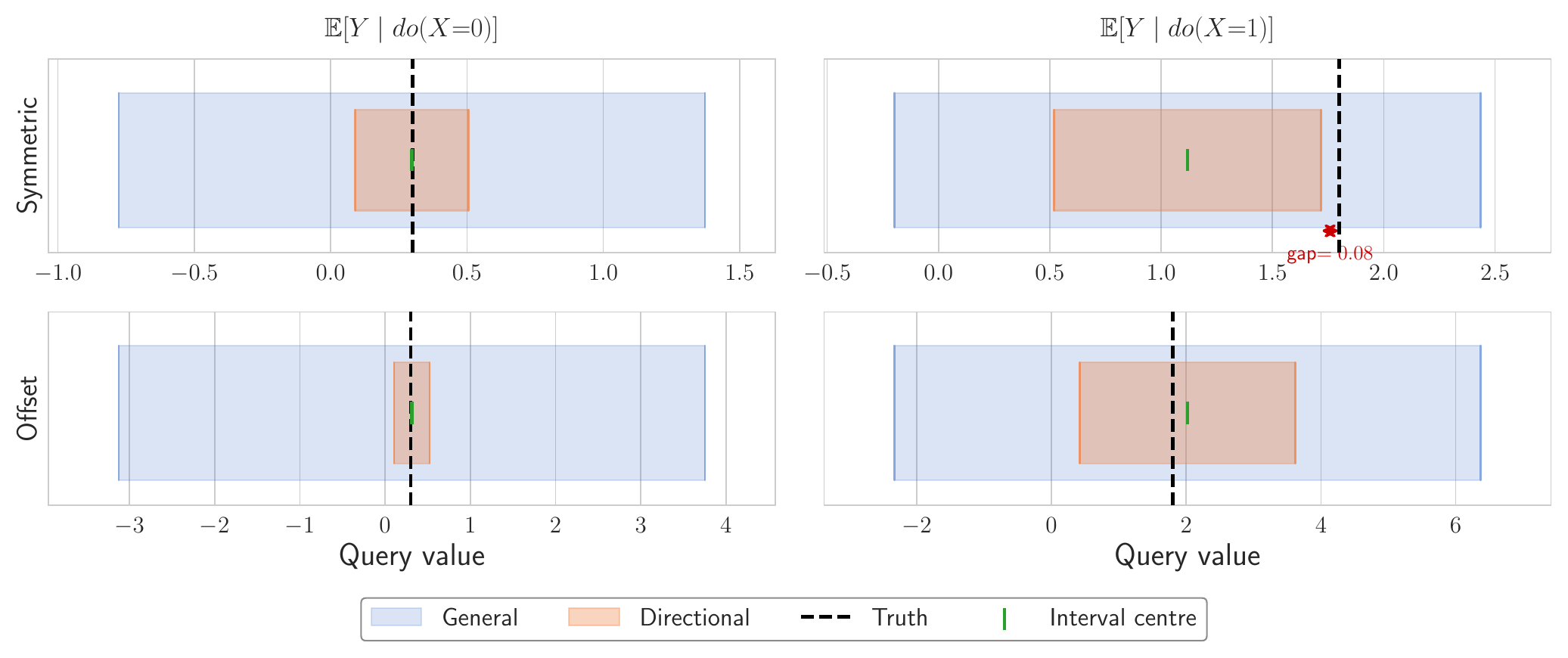}
  \caption{ATE: certified intervals at the observationally selected cell $r_{\mathrm{train}}{=}0.2$, for both variants and both queries. Under the symmetric variant the directional certificate misses $\mathbb{E}[Y\mid do(X{=}1)]$ by a gap of $0.08$ (annotated); the offset variant covers both queries.}
  \label{fig:ate-intervals}
\end{figure}
\emph{The offset prior is what makes small-radius selection work.} On \textbf{(Q2)}, the coverage contrast between the two variants is sharp. Figure~\ref{fig:ate-intervals} shows the certified intervals at the observationally selected cell $r_{\mathrm{train}}{=}0.2$. The offset variant covers both queries in all folds. The symmetric variant covers $\mathbb{E}[Y\mid do(X{=}0)]$ but not $\mathbb{E}[Y\mid do(X{=}1)]$: the truth ($1.80$) lies $0.08$ above the directional upper bound ($1.72$), missing coverage in all five folds. The zero-centered ambiguity set at this radius simply does not contain the realised shift on that query, and only at $r_{\mathrm{train}}\ge 0.5$ does the directional certificate recover coverage under the symmetric variant, with a correspondingly wider interval. A symmetric prior therefore requires a larger ambiguity set to obtain the same guarantee, and it is the offset prior that makes observational radius selection viable at a small radius. We show how the intervals expand with $r_{\mathrm{train}}$ in Figure~\ref{fig:ate-shaded} (Appendix~\ref{app:supplementary-results}): there, the general certificate grows rapidly, becoming vacuous by $r_{\mathrm{train}}{=}1.0$, while the directional certificate grows more slowly and remains informative longer. 

\begin{figure}[h]
  \centering
  \includegraphics[width=.95\textwidth]{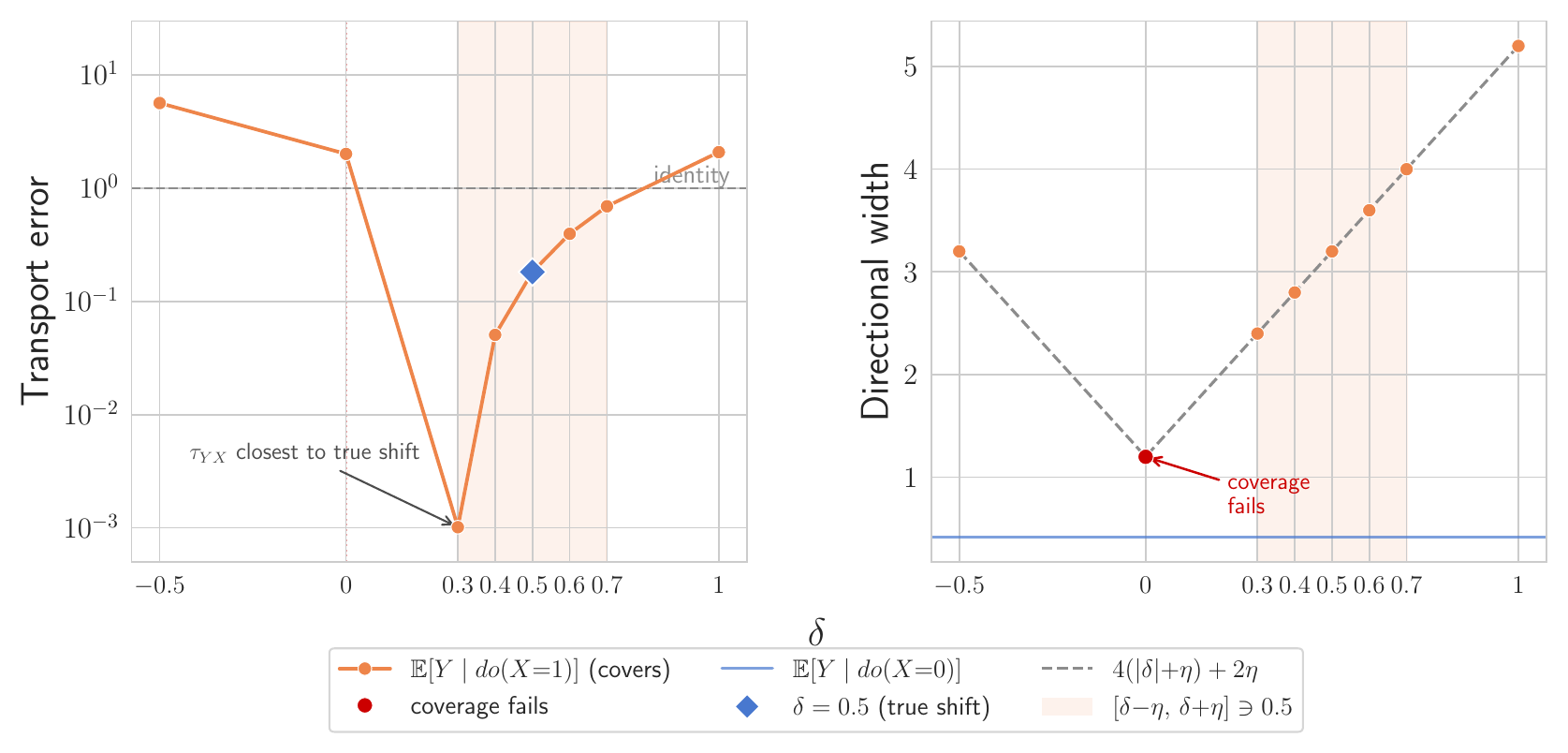}
    \caption{ATE: directional misspecification. The target is fixed and the offset prior~$\delta$ varies, at $r_{\mathrm{train}} = \varepsilon{=}\eta{=}0.2$. Shading marks the $\delta$ for which the box $[\delta{-}\eta,\,\delta{+}\eta]$ contains the true shift $\beta_t-\beta_s = 0.5$. \textbf{Left:} transport error relative to the identity (log scale); the minimum falls at $\delta = 0.3$, where the learned coefficient $\tau_{YX}\approx 0.465$ is closest to the truth on the grid. \textbf{Right:} directional width, following $4(|\delta|{+}\eta) + 2\eta$ (dashed) on $do(X{=}1)$ and constant at~$0.415$ on $do(X{=}0)$. Coverage fails only at $\delta = 0$ (red).}
  \label{fig:ate-misspec}
\end{figure}

\emph{What happens when the offset prior is wrong?} The offset results above declare $\delta = 0.5$, which, on this benchmark, is exactly the realised shift $\beta_t-\beta_s$, so they represent the favourable case for a directional prior. Figure~\ref{fig:ate-misspec} holds the target fixed and varies the offset. The radius is held at $r_{\mathrm{train}}=0.2$, the value the observational selector returns above, so that the effect of a wrong offset is isolated from re-selection. Two things hold: (a) The learned map improves on the identity exactly when the declared box contains the realised shift; that is, when $|\delta - 0.5|\le\eta$, and (b) it is worse than not transporting outside that window; within it, the best offset lies \emph{below} the truth because the map overshoots its declared center ($\tau_{YX}\approx\delta+0.165$ throughout the sweep), though how much better depends on where the grid happens to fall. The certificate, however, does not become tighter as the offset becomes more incorrect: the mechanism budget is the effective half-width $|\delta|+\eta$, so a wrong offset of large magnitude \emph{widens} the interval. Its width decomposes exactly, with transport and mechanism each contributing $|\delta|+\eta$ to the half-width and the environment contributing $\eta$; the $do(X{=}0)$ width is flat because the intervention zeroes the shifted coefficient (Remark~\ref{rem:directional-tightening}). Coverage, therefore, survives at $\delta = -0.5$ and $\delta = 1.0$, whose boxes both miss the truth, and fails at the single point $\delta = 0$, the symmetric variant, whose $0.08$ miss is the one already reported above. Thus, a small and incorrect offset is the failure mode; one that is large and incorrect is not.

\begin{figure}[h]
  \centering
  \includegraphics[width=0.7\textwidth]{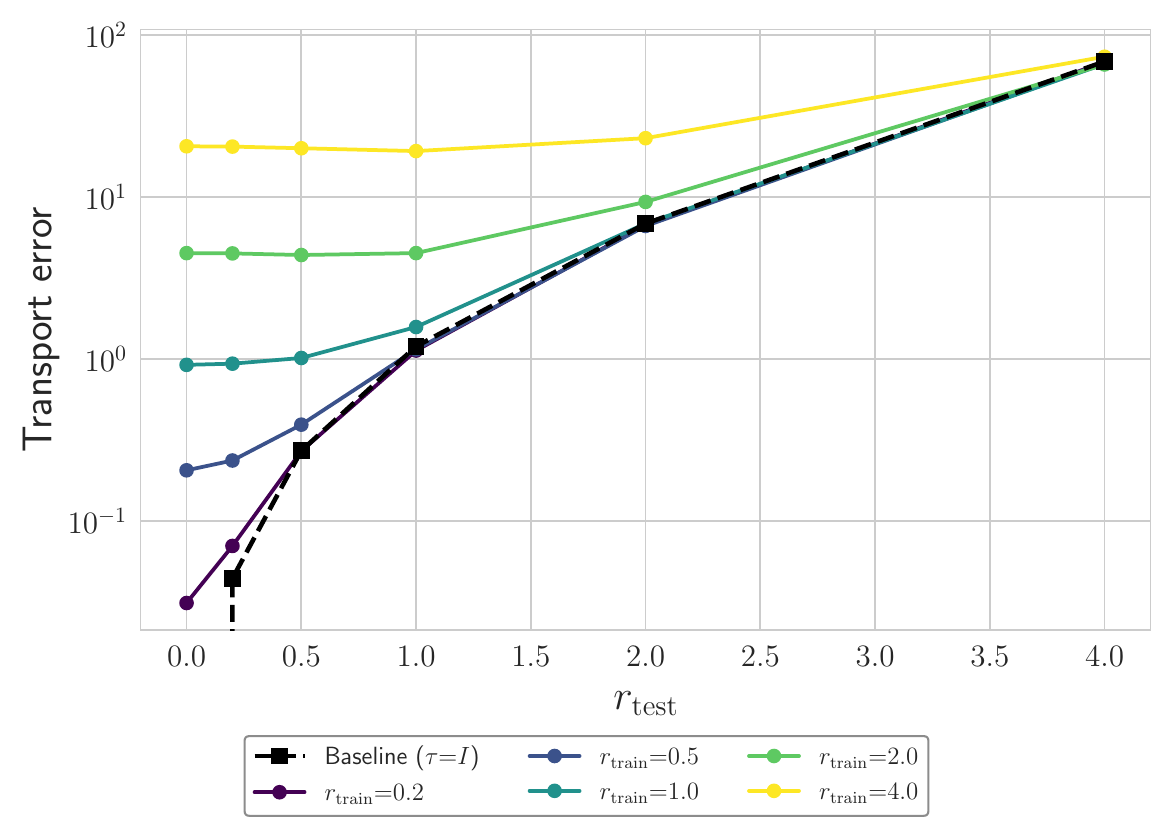}
  \caption{ATCE: transport error against $r_{\mathrm{test}}$, one line per $r_{\mathrm{train}}$ (log scale). The identity baseline (dashed) grows monotonically. The lines separate most clearly at moderate $r_{\mathrm{test}}$; at $r_{\mathrm{test}}{=}4.0$ the errors are large and their differences small in relative terms, so the large-shift comparison is quantified in Table~\ref{tab:dro-crossover}.}
  \label{fig:atce-error}
\end{figure}

\paragraph{ATCE.}
\emph{A DRO crossover: no single radius dominates.} Figure~\ref{fig:atce-error} shows how transport error evolves as the test perturbation magnitude $r_{\mathrm{test}}$ increases for maps trained at each $r_{\mathrm{train}}$. The identity baseline ($\tau{=}I$, dashed) increases monotonically, confirming that any improvement comes from the learned map rather than from the evaluation geometry. A DRO crossover is visible. At small $r_{\mathrm{test}}$, the baseline wins: any robustification adds unnecessary distortion, and Table~\ref{tab:dro-crossover} confirms this at every training radius. At $r_{\mathrm{test}}{=}4.0$, the picture reverses: maps trained at $r_{\mathrm{train}}\in\{0.2,0.5,1.0\}$ significantly reduce error relative to the baseline ($r_{\mathrm{train}}{=}1.0$ gives $66.38$ against the baseline's $68.79$), while the more aggressive radii $\{2.0,4.0\}$ show no significant improvement. Note
that the best radius at $r_{\mathrm{test}}{=}4.0$ is not $r_{\mathrm{train}}{=}4.0$ but $r_{\mathrm{train}}{=}1.0$: training hedges against the single worst perturbation in the ambiguity ball, whereas the sampled targets are spread across it, so a radius matched to the test magnitude buys protection in directions the typical target does not take, at the cost of accuracy in the ones it does. No single training radius, therefore, dominates everywhere: this is the classic DRO trade-off, and its lesson is that the operating radius should be sized to the anticipated shift, not to the worst imaginable one.


\begin{table}[h]
\centering
\caption{DRO crossover on ATCE: transport error (mean $\pm$ std) at small
($r_{\mathrm{test}}{=}0.2$) and large ($r_{\mathrm{test}}{=}4.0$) perturbation
magnitudes. For each perturbation seed $k$, we average the error over $5$ folds,
yielding $n{=}100$ paired observations; the pairing unit is $k$, so the folds do not
inflate the sample. Paired $t$-tests compare each $r_{\mathrm{train}}$ against the
baseline ($\tau{=}I$). At small $r_{\mathrm{test}}$, the baseline wins
significantly; i.e. no robustification is needed. At large $r_{\mathrm{test}}$, TraCA wins significantly.}
\label{tab:dro-crossover}
\small
\begin{tabular}{l c c}
\toprule
& $r_{\mathrm{test}} = 0.2$ & $r_{\mathrm{test}} = 4.0$ \\
\cmidrule(lr){2-2} \cmidrule(lr){3-3}
Method & Error & Error \\
\midrule
Baseline ($\tau{=}I$)
  & $\mathbf{0.044 \pm 0.033}$
  & $68.794 \pm 71.240$ \\
$r_{\mathrm{train}} = 0.2$
  & $0.070 \pm 0.068$
  & $68.213 \pm 71.017$ \\
$r_{\mathrm{train}} = 0.5$
  & $0.236 \pm 0.152$
  & $67.426 \pm 70.739$ \\
$r_{\mathrm{train}} = 1.0$
  & $0.935 \pm 0.314$
  & $\mathbf{66.383 \pm 70.464}$ \\
$r_{\mathrm{train}} = 2.0$
  & $4.487 \pm 0.689$
  & $65.928 \pm 70.981$ \\
$r_{\mathrm{train}} = 4.0$
  & $20.500 \pm 1.468$
  & $73.625 \pm 76.890$ \\
\bottomrule
\end{tabular}
\end{table}

\begin{figure}[h]
  \centering
  \includegraphics[width=\textwidth]{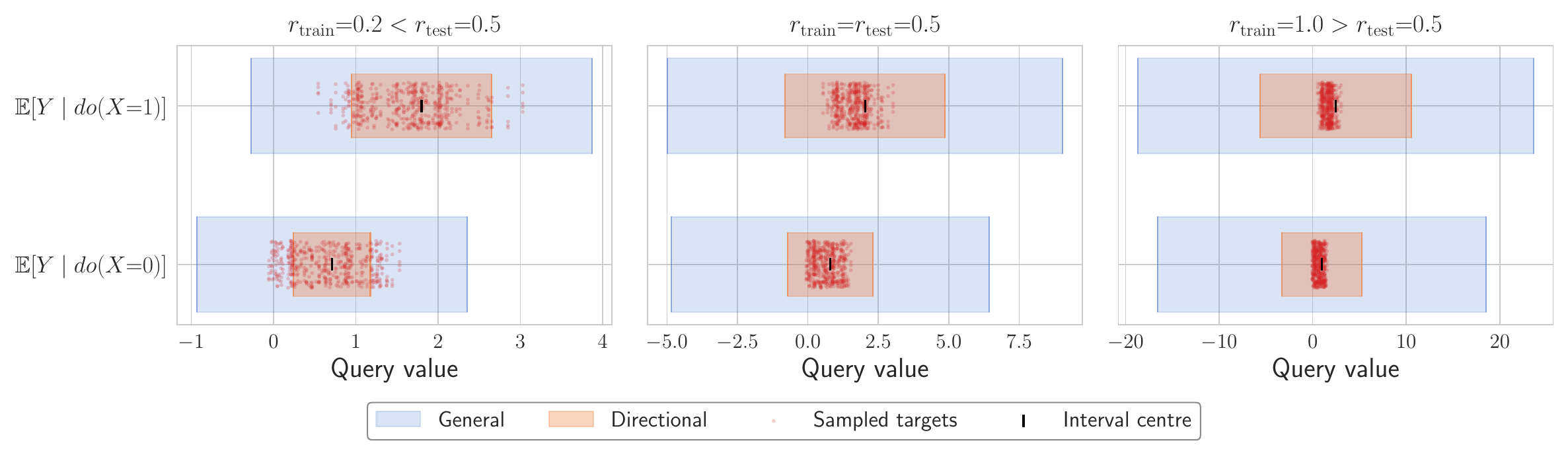}
  \caption{ATCE: certified intervals at three operating points. Left: $r_{\mathrm{train}} < r_{\mathrm{test}}$ (out-of-ball). Center: matched. Right: $r_{\mathrm{train}} > r_{\mathrm{test}}$ (in-ball with margin). Blue = general, orange = directional; each red point is one sampled target's true query value, so coverage is the fraction falling inside the band.}
  \label{fig:atce-brackets}
\end{figure}

\emph{Tighter and a more faithful diagnostic.} Moving to \textbf{(Q1)} again, both queries remain sensitive to mechanism perturbation, which is what makes the certificate non-trivial here: the intervention removes the edge $Z\to X$, so $Z$ no longer reaches $Y$ through the treatment, but the direct path $Z\to Y$ is untouched, and $Z$ remains random. A perturbation of the $Z\to Y$ coefficient therefore moves the query at both intervention values, and, by linearity, a perturbation of the $X\to Y$ coefficient moves it whenever $x\neq 0$. Figure~\ref{fig:atce-brackets} shows the certified intervals at three representative operating points, with $r_{\mathrm{test}}{=}0.5$ fixed and $r_{\mathrm{train}}$ varied: below the test magnitude ($r_{\mathrm{train}}{=}0.2$, out-of-ball), matched ($r_{\mathrm{train}}{=}0.5$), and above ($r_{\mathrm{train}}{=}1.0$, in-ball with margin). At the matched point, both certificates are informative and contain the sampled targets, with the directional interval $3.7\times$ tighter than the general one on $\mathbb{E}[Y\mid do(X{=}0)]$ and $2.5\times$ on $\mathbb{E}[Y\mid do(X{=}1)]$ ($2.9\times$ family-averaged). Below the matched point, the directional certificate under-covers ($0.61$ and $0.89$ across the two queries), while the general certificate still covers all targets; the tighter bound correctly reports that the target lies outside the ball for which it was built. Above the matched point, both provide coverage, and the directional certificate remains tighter throughout.

\begin{figure}[t]
  \centering
  \includegraphics[width=\textwidth]{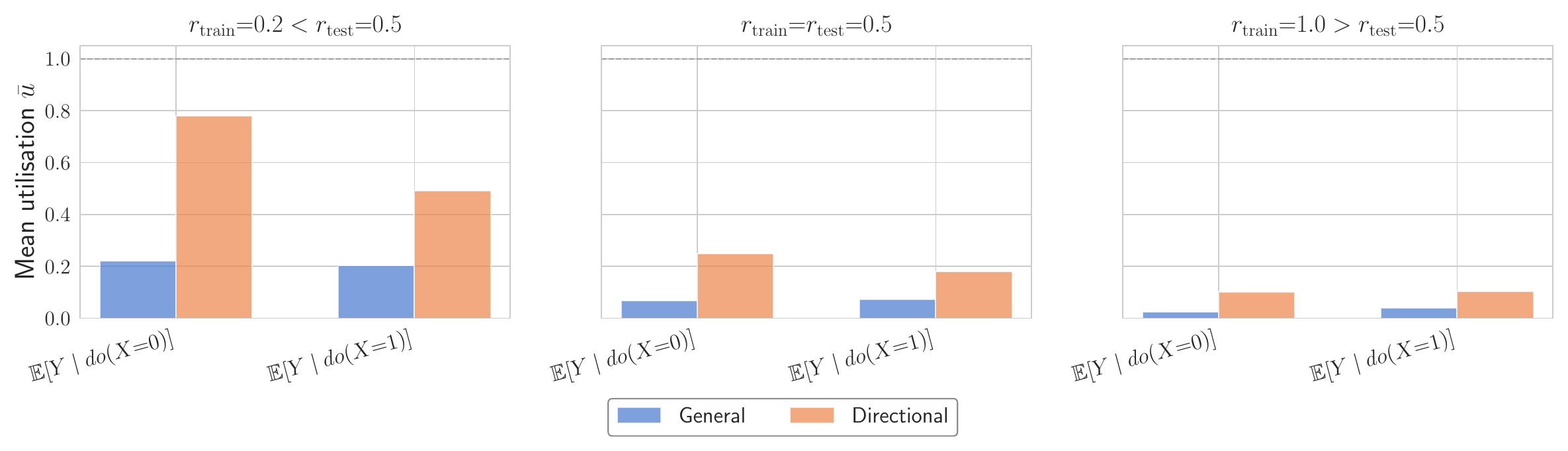}
  \caption{ATCE: mean utilisation ratio $\bar u$ at the same three operating points. The dashed line marks $\bar u{=}1$, the coverage boundary; higher is tighter, and a mean above the line indicates the average target lies outside the interval (out-of-ball regime).}
  \label{fig:atce-tightness}
\end{figure}

Figure~\ref{fig:atce-tightness} quantifies this through the \emph{utilisation ratio}:
\begin{align}
  u := \frac{\bigl|Q_t - Q_{\tau\# s}\bigr|}{\delta_{\iota,O_\iota}(\lintau)},
\end{align}
the distance from the interval's center to the true query value, expressed as a fraction of its half-width. Because the certified interval is centered at the transported-source value $Q_{\tau \# s}$, a target is covered exactly when $u\le 1$. Therefore, $u$ measures tightness on a scale where coverage is the unit: values near zero mean the interval is far wider than the target requires, values just below one mean it is as tight as it can be while still covering, and values above one mean the target lies outside it. Higher is therefore better, up to one. We report the mean $\bar u$ over targets and folds; note that $\bar u<1$ does not imply full coverage, since individual targets may exceed the interval while the mean does not, which is why coverage is reported separately. At the matched point $\bar u<1$ for both certificates, the directional bound consumes the larger fraction of its interval: tighter at no cost to coverage. At the out-of-ball point, the directional $\bar u$ exceeds one, correctly reporting that the average target now lies beyond the guarantee's intended scope. The tighter bound is thus also the more faithful diagnostic. Fixing $r_{\mathrm{test}}$ and increasing $r_{\mathrm{train}}$, directional coverage rises to one exactly as $r_{\mathrm{train}}$ reaches $r_{\mathrm{test}}$, tracking the in-ball boundary, whereas the wider general certificate already covers below it, admitting targets outside the ball it was built for. We show this coverage transition in Figure~\ref{fig:atce-coverage} (Appendix~\ref{app:supplementary-results}).

\begin{figure}[t]
  \centering
  \includegraphics[width=\textwidth]{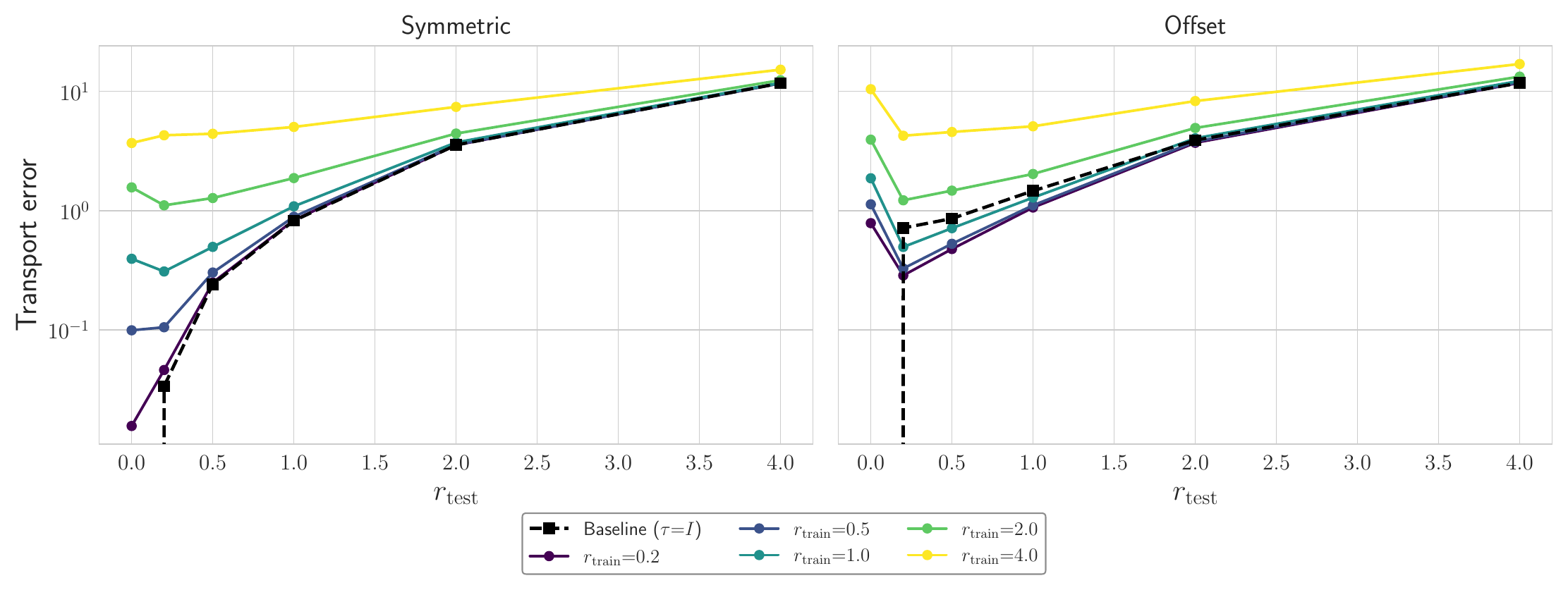}
  \caption{LiLUCAS (empirical): transport error against $r_{\mathrm{test}}$ for the symmetric (left) and offset (right) variants, one line per $r_{\mathrm{train}}$ (log scale), against the identity baseline (dashed).}
  \label{fig:lilucas-transport}
\end{figure}

\paragraph{LiLUCAS.}
LiLUCAS is our six-variable benchmark and the only one under the empirical objective. We run three configurations: empirical symmetric, empirical offset, and Gaussian symmetric. Figure~\ref{fig:lilucas-transport} contrasts the two empirical variants: transport error against the test magnitude $r_{\mathrm{test}}$, one curve per $r_{\mathrm{train}}$, against the identity baseline. Under the symmetric variant, no trained map ever \emph{significantly} improves on the identity: at small $r_{\mathrm{test}}$, every trained map is significantly worse than the baseline, and at large $r_{\mathrm{test}}$, the smaller radii are statistically indistinguishable from it until over-robustification degrades them at $r_{\mathrm{train}}\ge 2.0$ (Table~\ref{tab:dro-lilucas-ew}). Under the offset variant, the picture changes: at small $r_{\mathrm{test}}$, the trained maps reduce error by up to $60\%$ ($r_{\mathrm{train}}{=}0.2$: $0.286$ against the baseline's $0.716$, $p<0.001$, with the improvement between $59.7\%$ and $60.7\%$ across all five folds; $54\%$ and $31\%$ at $r_{\mathrm{train}}{=}0.5$ and $1.0$). The offset declares a direction for the shift, and the learned map exploits it, anticipating a systematic displacement that the identity ignores. The two baselines differ ($0.034$ symmetric against $0.716$ offset) because the two ambiguity sets are centered differently: the symmetric box is centered at zero, so perturbations partially cancel, and the identity performs well, while the offset box is centered at $\delta{=}0.5$, so every target carries a consistent positive displacement that the identity cannot absorb. Each map is therefore compared against the identity on its own prior's target distribution, and the $60\%$ is a within-variant claim. The offset advantage is also confined to small $r_{\mathrm{test}}$, and at every $r_{\mathrm{test}}$, over-robustification at $r_{\mathrm{train}}\ge 2.0$ degrades both variants.

\begin{table}[t]
\centering
\caption{LiLuCaS transport error under symmetric and offset selectors.
Each cell shows mean $\pm$ std over $n{=}100$ perturbation seeds $k$
(fold-averaged). Bold marks entries not significantly worse than the
column minimum (paired $t$-test, $p > 0.05$).
The baselines differ because the offset selector centres
its ambiguity set at $\delta{=}0.5$.}
\label{tab:dro-lilucas-ew}
\footnotesize
\begin{tabular}{l cc cc}
\toprule
& \multicolumn{2}{c}{Symmetric}
& \multicolumn{2}{c}{Offset} \\
\cmidrule(lr){2-3} \cmidrule(lr){4-5}
$r_{\mathrm{train}}$ & $r_{\mathrm{test}}{=}0.2$ & $r_{\mathrm{test}}{=}4.0$ & $r_{\mathrm{test}}{=}0.2$ & $r_{\mathrm{test}}{=}4.0$ \\
\midrule
  Baseline ($\tau{=}I$) & $\mathbf{0.034 \pm 0.027}$ & $\mathbf{11.715 \pm 9.315}$ & $0.716 \pm 0.266$ & $\mathbf{11.762 \pm 10.353}$ \\
  $r_{\mathrm{train}}{=}0.2$ & $0.046 \pm 0.032$ & $\mathbf{11.689 \pm 9.312}$ & $\mathbf{0.286 \pm 0.085}$ & $\mathbf{11.791 \pm 9.379}$ \\
  $r_{\mathrm{train}}{=}0.5$ & $0.105 \pm 0.049$ & $\mathbf{11.691 \pm 9.323}$ & $0.327 \pm 0.052$ & $\mathbf{11.923 \pm 9.320}$ \\
  $r_{\mathrm{train}}{=}1.0$ & $0.309 \pm 0.084$ & $\mathbf{11.800 \pm 9.383}$ & $0.496 \pm 0.035$ & $\mathbf{12.249 \pm 9.334}$ \\
  $r_{\mathrm{train}}{=}2.0$ & $1.108 \pm 0.159$ & $12.408 \pm 9.658$ & $1.222 \pm 0.152$ & $13.293 \pm 9.793$ \\
  $r_{\mathrm{train}}{=}4.0$ & $4.286 \pm 0.311$ & $15.205 \pm 10.753$ & $4.250 \pm 0.410$ & $16.964 \pm 12.142$ \\
\bottomrule
\end{tabular}
\end{table}

\begin{figure}[t]
  \centering
  \includegraphics[width=0.6\textwidth]{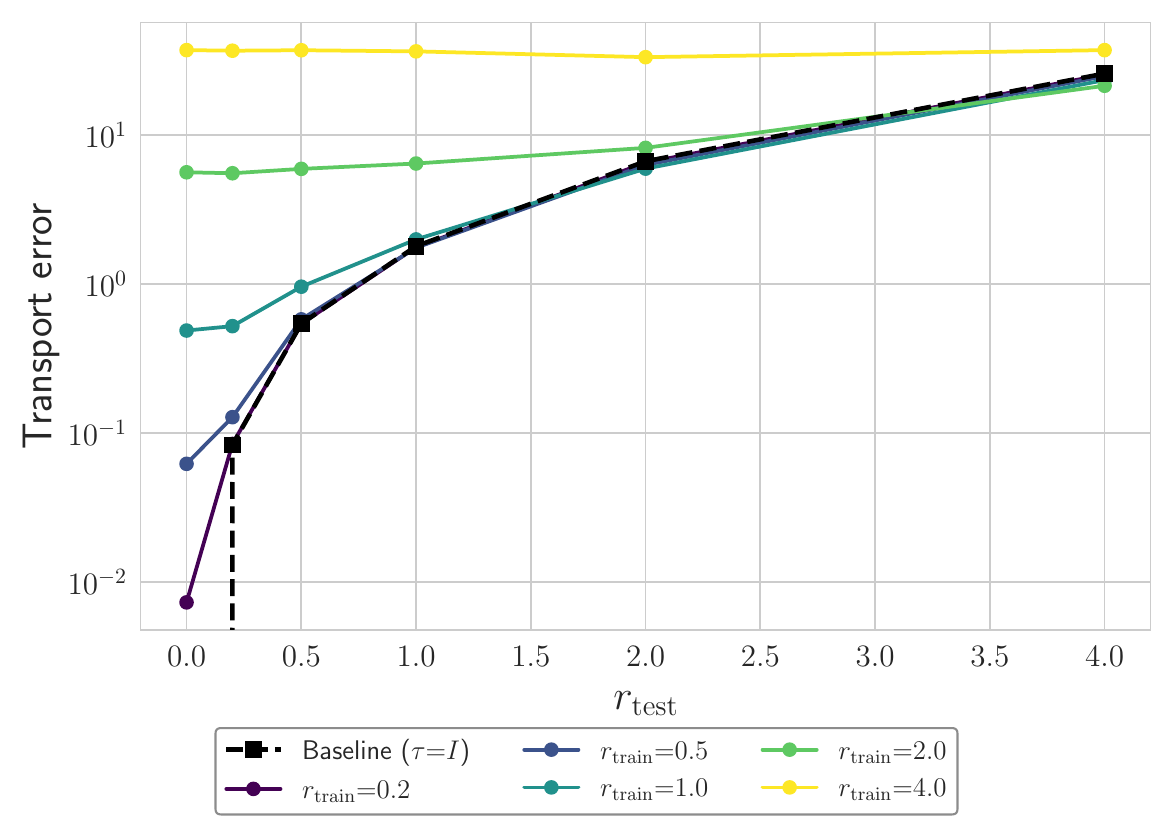}
  \caption{LiLUCAS (Gaussian): transport error against $r_{\mathrm{test}}$, one line per $r_{\mathrm{train}}$ (log scale), against the identity baseline (dashed).}
  \label{fig:lilucas-gau-error}
\end{figure}

\emph{Robustification also helps at large shifts under the Gaussian objective.} The Gaussian symmetric configuration shows the complementary crossover (Figure~\ref{fig:lilucas-gau-error}, Table~\ref{tab:dro-lilucas-gau}): at small $r_{\mathrm{test}}$, the identity is again near-optimal, and only the smallest trained radius matches it. However, at the largest test magnitude $r_{\mathrm{test}}{=}4.0$, the trained maps significantly reduce error relative to the baseline: by $1.4\%$ at $r_{\mathrm{train}}{=}0.2$, growing monotonically to $17\%$ at $r_{\mathrm{train}}{=}2.0$ (all $p<0.001$, fold-stable to within $0.1$ percentage points). As with ATCE, this is an interior optimum: robustification helps up to $r_{\mathrm{train}}{=}2.0$ and then over-robustifies, with $r_{\mathrm{train}}{=}4.0$ falling well below the baseline. Where the empirical offset earns its advantage at \emph{small} shifts through the directional prior, the Gaussian symmetric map earns a smaller advantage at \emph{large} shifts through robustification alone; two faces of the same shift-sizing trade-off seen on ATCE.

\begin{table}[t]
\centering
\caption{LiLuCaS (Gaussian): transport error at small and large test
perturbations. Each cell shows mean $\pm$ std over $n{=}100$ perturbation
seeds $k$ (fold-averaged). Bold marks entries not significantly worse than the
column minimum (paired $t$-test, $p > 0.05$).}
\label{tab:dro-lilucas-gau}
\small
\begin{tabular}{l c c}
\toprule
& $r_{\mathrm{test}} = 0.2$
& $r_{\mathrm{test}} = 4.0$ \\
\cmidrule(lr){2-2} \cmidrule(lr){3-3}
Method & Error & Error \\
\midrule
  Baseline ($\tau{=}I$) & $\mathbf{0.083 \pm 0.042}$ & $25.916 \pm 15.466$ \\
  $r_{\mathrm{train}} = 0.2$ & $\mathbf{0.085 \pm 0.048}$ & $25.560 \pm 15.188$ \\
  $r_{\mathrm{train}} = 0.5$ & $0.128 \pm 0.098$ & $24.919 \pm 14.666$ \\
  $r_{\mathrm{train}} = 1.0$ & $0.522 \pm 0.271$ & $23.437 \pm 13.328$ \\
  $r_{\mathrm{train}} = 2.0$ & $5.551 \pm 0.936$ & $\mathbf{21.468 \pm 10.324}$ \\
  $r_{\mathrm{train}} = 4.0$ & $36.909 \pm 2.421$ & $37.246 \pm 18.358$ \\
\bottomrule
\end{tabular}
\end{table}

\emph{The directional certificate is the informative bound.}
On the certificate side, the results mirror ATCE: as the radius grows, the general certificate quickly becomes vacuous, while the directional certificate remains informative longer. For the Gaussian configuration, the general width crosses the vacuity threshold ($10\times\text{QS}$) between $r_{\mathrm{train}}{=}0.2$ and $0.5$, while the directional width stays roughly an order of magnitude below it and remains informative through $r_{\mathrm{train}}{=}0.5$. The directional certificate is thus the only usable bound at $r_{\mathrm{train}}{=}0.5$: at $r_{\mathrm{train}}{=}0.2$, both are informative, and by $r_{\mathrm{train}}{=}1.0$, neither is. Under the empirical symmetric variant, the general certificate is already vacuous at the smallest radius, so only the
directional certificate is ever informative. This is a query scale effect rather than a wider interval: the empirical general certificate is, in fact, narrower than the Gaussian one ($7.46$ against $10.14$ at $r_{\mathrm{train}}{=}0.2$), but the smaller empirical query scale lowers the vacuity threshold below the width. The full width-versus-radius profiles for both configurations are given in Appendix~\ref{app:supplementary-results} (Figures~\ref{fig:lilucas-gau-width} and~\ref{fig:lilucas-ewsym-width}).

\begin{figure}[t]
  \centering
  \includegraphics[width=\textwidth]{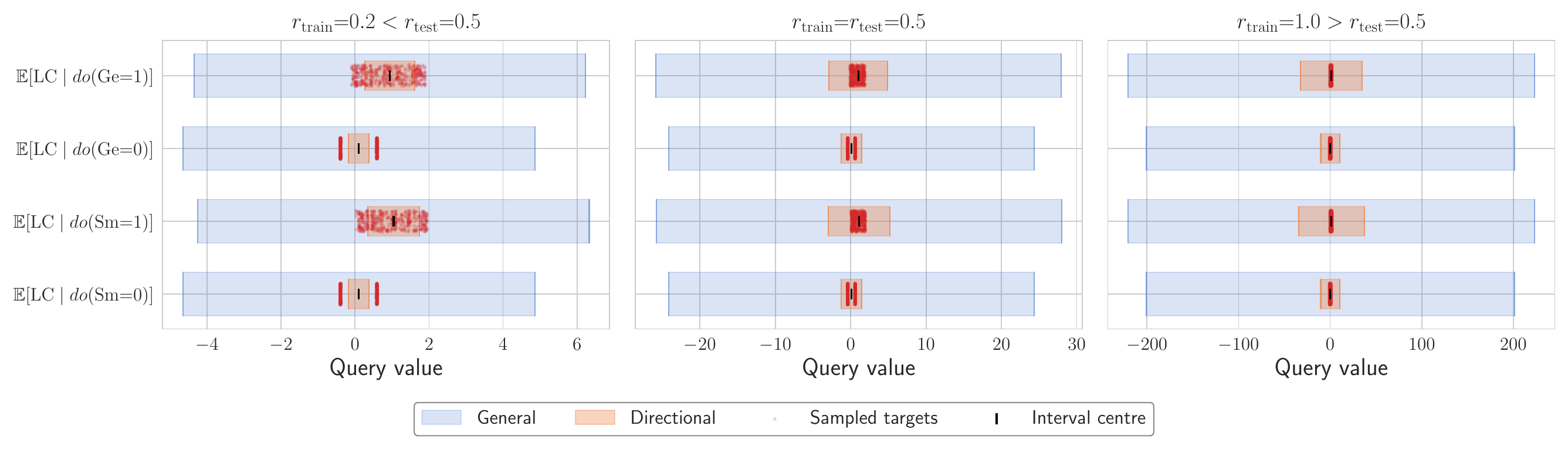}
  \caption{LiLUCAS (Gaussian): certified intervals at three operating points, $r_{\mathrm{test}}{=}0.5$ fixed. Left ($r_{\mathrm{train}}{=}0.2$, out-of-ball): both informative, directional $7.6$-$17.2\times$ tighter per query, but the directional interval covers only $30\%$ as the shift exceeds the training radius. Center (matched): directional informative and fully covering, general vacuous. Right (margin): both vacuous. Blue = general, orange = directional, red = sampled targets.}
  \label{fig:lilucas-gau-brackets}
\end{figure}

\begin{figure}[t]
  \centering
  \includegraphics[width=\textwidth]{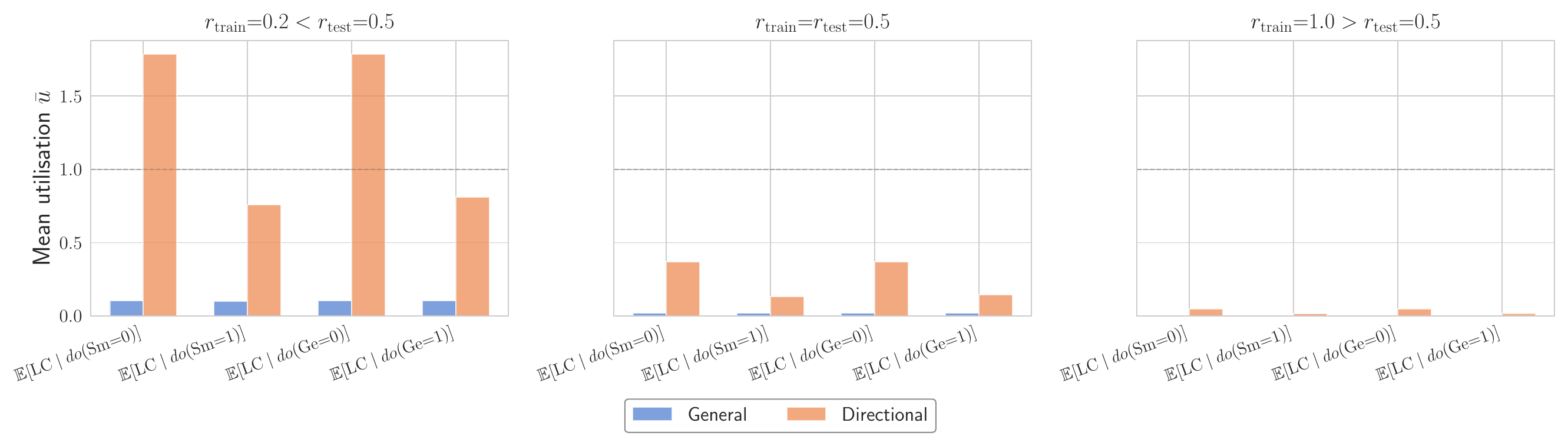}
  \caption{LiLUCAS (Gaussian): mean utilisation ratio at the three operating points of Figure~\ref{fig:lilucas-gau-brackets}. At the out-of-ball point the directional utilisation exceeds one for two queries, consistent with the $30\%$ coverage there.}
  \label{fig:lilucas-gau-tightness}
\end{figure}

\begin{figure}[t]
  \centering
  \includegraphics[width=\textwidth]{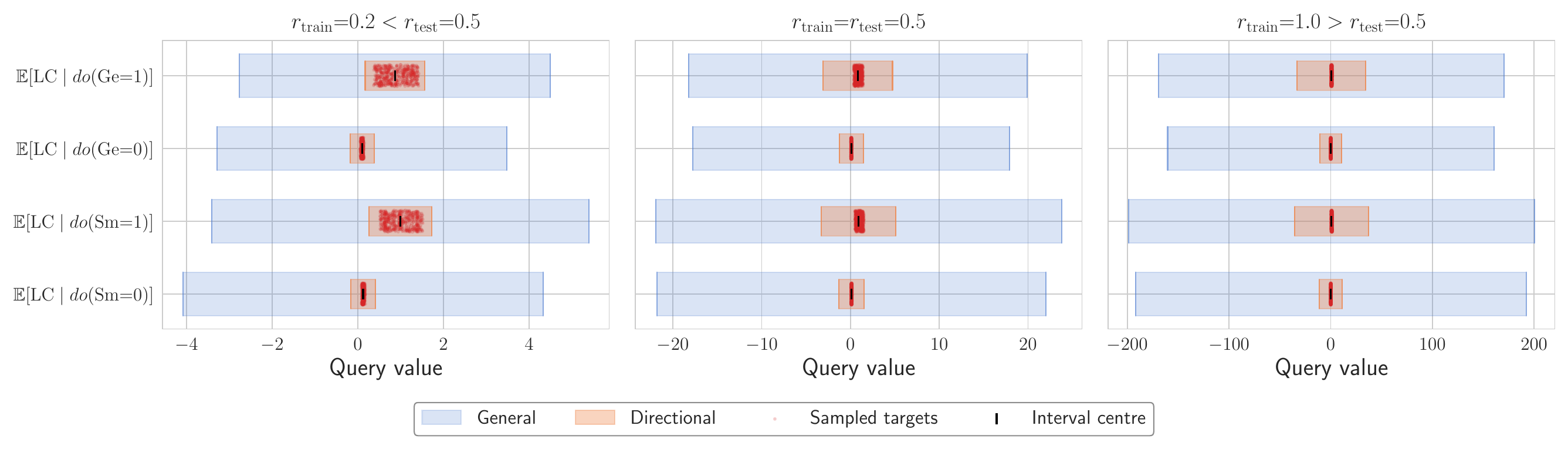}
  \caption{LiLUCAS (empirical symmetric): certified intervals at three operating points, $r_{\mathrm{test}}{=}0.5$ fixed. The directional certificate is informative and fully covering where the general is vacuous (left, center); both are vacuous at the margin (right).}
  \label{fig:lilucas-ewsym-brackets}
\end{figure}

\begin{figure}[t]
  \centering
  \includegraphics[width=\textwidth]{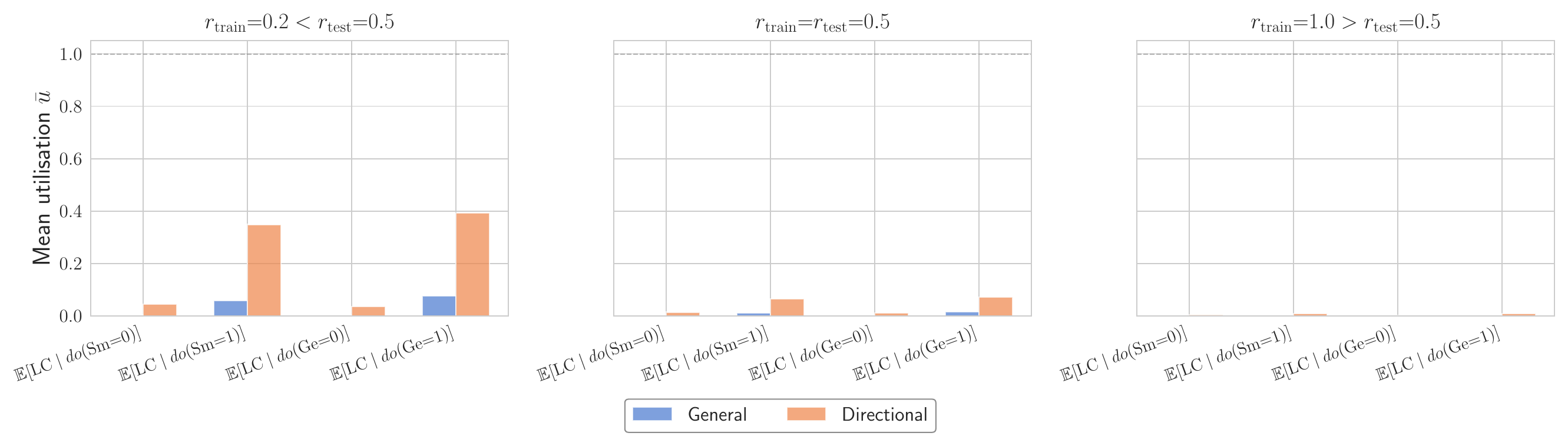}
  \caption{LiLUCAS (empirical symmetric): mean utilisation ratio at the three operating points of Figure~\ref{fig:lilucas-ewsym-brackets}.}
  \label{fig:lilucas-ewsym-tightness}
\end{figure}

Figure~\ref{fig:lilucas-gau-brackets} shows the certified intervals at three operating points for the Gaussian symmetric configuration, $r_{\mathrm{test}}{=}0.5$ fixed. At the matched radius $r_{\mathrm{train}}{=}0.5$, the directional certificate is informative and covers all four queries in full, while the general certificate is already  vacuous: the regime where the directional bound is the only usable one. The out-of-ball panel ($r_{\mathrm{train}}{=}0.2$) illustrates the tightness/coverage trade-off. Here, both certificates are informative, and the directional certificate is $7.6$-$17.2\times$ narrower than the general per query ($10.4\times$ family-averaged), but because the test shift exceeds the training radius, the directional interval covers only $30\%$ of targets. The directional utilisation indeed exceeds one for two of the four queries (Figure~\ref{fig:lilucas-gau-tightness}), confirming that the average target lies outside the interval: the tightness is real, but the guarantee holds only within the trained ball. The general certificate covers here, but only because it is wide. At the margin panel ($r_{\mathrm{train}}{=}1.0$) both certificates are vacuous. The empirical symmetric configuration tells the same story, except that the general certificate is never informative: the directional certificate is informative and fully covering at the two smallest radii while the general one is vacuous throughout, and both become vacuous by $r_{\mathrm{train}}{=}1.0$ (Figures~\ref{fig:lilucas-ewsym-brackets} and~\ref{fig:lilucas-ewsym-tightness}).

\subsubsection{Real-world}
\label{subsubsec:results_real}
\begin{figure}[t]
  \centering
  \includegraphics[width=0.6\textwidth]{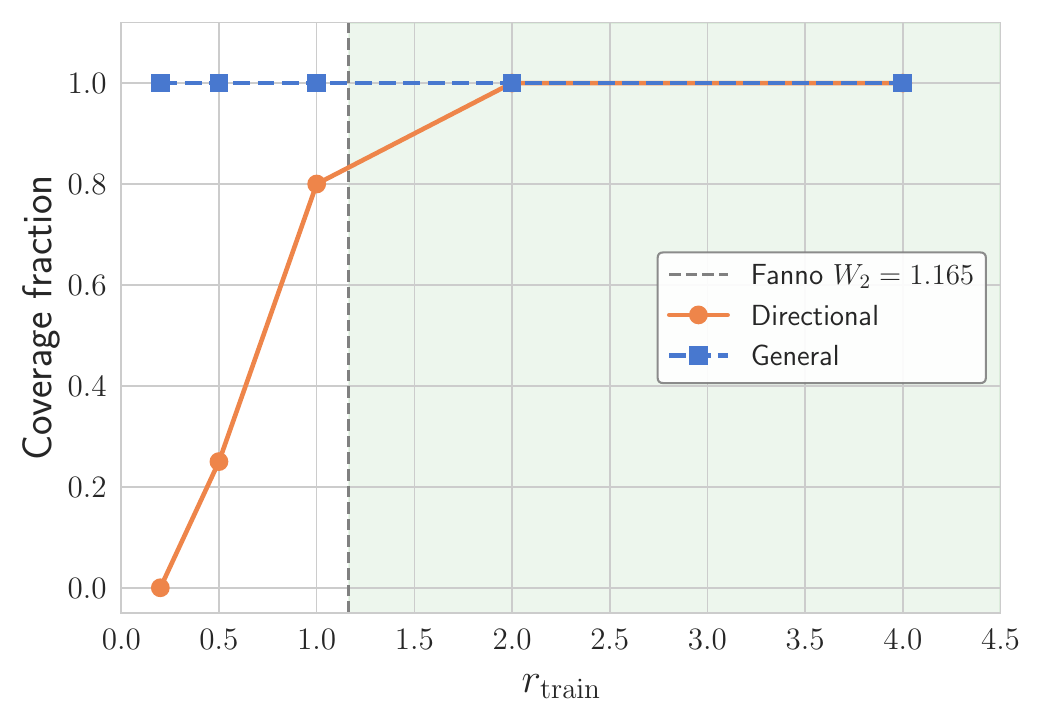}
  \caption{Portland: coverage against $r_{\mathrm{train}}$. The dashed line marks the Wasserstein distance of the realised Fanno shift ($\mathcal W_2 = 1.165$); the green region indicates $\varepsilon\ge \mathcal W_2$, where the target lies inside the ball.}
  \label{fig:portland-coverage}
\end{figure}

\paragraph{The certificate covers at the radius the real shift requires.}
On the Portland dataset, the Wasserstein distance between source and target exogenous distributions on the shifted coordinate $Z$ of Fanno Creek is $\mathcal W_2 = 1.165$, placing the target inside the Gelbrich ball only for $\varepsilon\ge 2$. At and beyond that radius, the directional certificate attains full coverage on all four canopy
interventions (Figure~\ref{fig:portland-coverage}). Across the grid, the general certificate becomes vacuous ($>10\times\text{QS} = 1.63$) from $\varepsilon{=}0.5$ onward (width $1.75$), while the directional certificate remains informative throughout, reaching $1.02$ ($6.3\times\text{QS}$) only at $\varepsilon{=}4.0$ (Figure~\ref{fig:portland-width}, Appendix~\ref{app:supplementary-results}). Since $Z$ is a root, the mechanism set is empty throughout ($\eta{=}0$) and the whole certificate is carried by the environment term; the directional tightening derives entirely from that term's direction-specificity.

\emph{Observational calibration can point the wrong way.} Unlike ATE, choosing the operating radius here from observational data is not entirely safe. The observational transport distance $\mathcal W_2^2\!\bigl(\tau_\#P_s^{\mathrm{obs}},P_t^{\mathrm{obs}}\bigr)$ increases monotonically in $r_{\mathrm{train}}$: enlarging the ambiguity set pushes the learned map further from the identity, degrading the fit to the observed target distribution, even though a larger radius is exactly what coverage of the interventional query requires. Therefore, the selector returns the smallest admissible radius, $\varepsilon{=}0.2$, at which directional coverage is zero. On ATE, the same selector succeeded because there the
shifted mechanism gives the observational distance a nontrivial profile over the $(\varepsilon,\eta)$ grid with an interior minimum; on Portland, the shift is purely environmental and only $\varepsilon$ is swept, leaving no interior optimum for the selector to find. Whether a purely environmental shift always yields a monotone observational profile, or whether this is specific to the present single-radius sweep, we leave for future work. We report the observational-distance profile in Figure~\ref{fig:portland-obs} (Appendix~\ref{app:supplementary-results}).

\begin{figure}[h]
  \centering
  \includegraphics[width=\textwidth]{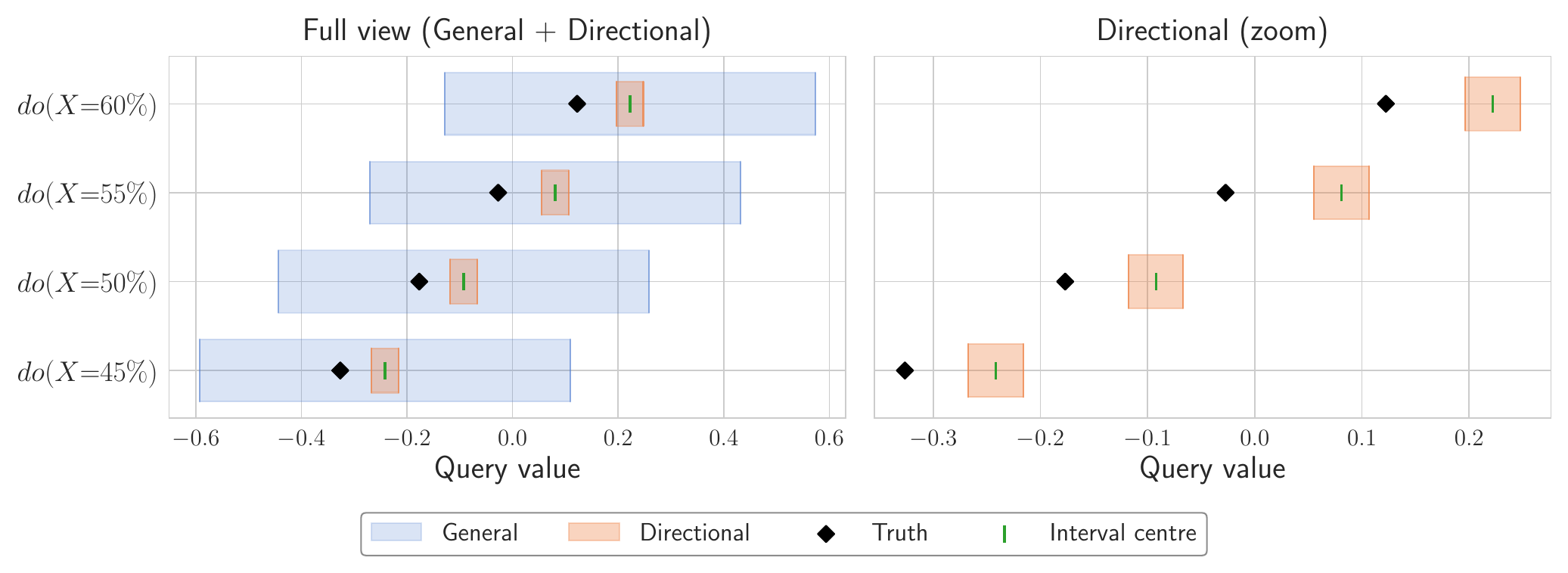}
  \caption{Portland: certified intervals at $r_{\mathrm{train}}{=}0.2$ for four canopy interventions. Left: full view (general + directional). Right: directional zoom. The directional interval is tight but does not yet cover the truth at this radius.}
  \label{fig:portland-bracket}
\end{figure}

\begin{figure}[h]
  \centering
  \includegraphics[width=.9\textwidth]{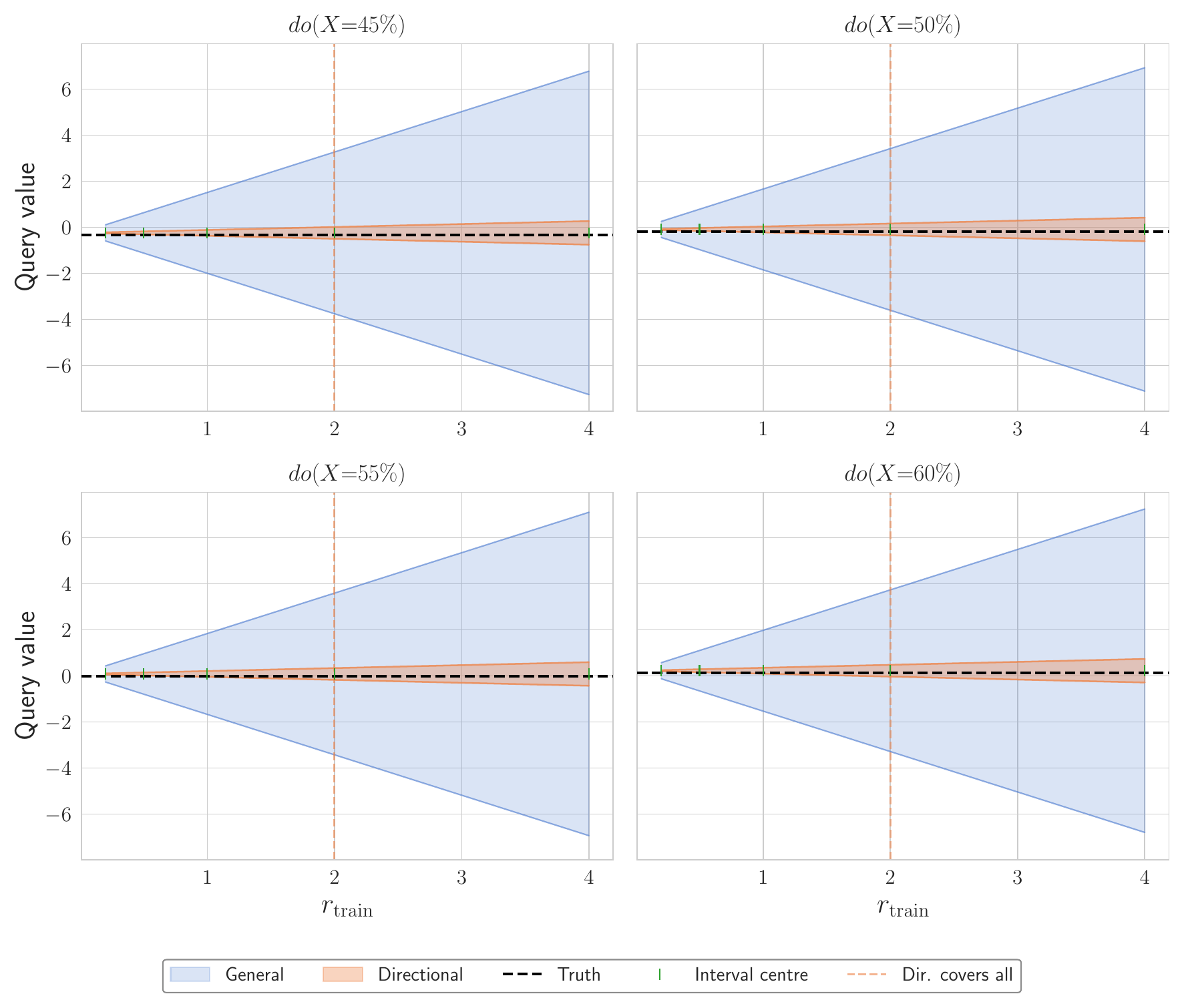}
  \caption{Portland: certified intervals across $r_{\mathrm{train}}$ for each canopy intervention. Blue = general, orange = directional, dashed = truth. The vertical line marks $r_{\mathrm{train}}{=}2.0$, where directional coverage reaches one for all queries.}
  \label{fig:portland-shaded}
\end{figure}

The consequence is visible at the selected radius. At $\varepsilon{=}0.2$, the directional interval is tight ($0.05$, i.e., $0.3\times\text{QS}$), but does not yet cover the truth, which lies outside the narrow band; the general interval does contain the truth, but only because it is wide enough ($0.70$) to span it (Figure~\ref{fig:portland-bracket}). Increasing $r_{\mathrm{train}}$ resolves this: the directional certificate covers all four queries once $r_{\mathrm{train}}$ reaches $2.0$, the radius at which the ball contains the realised shift, while the general certificate widens without becoming more informative (Figure~\ref{fig:portland-shaded}). Where observational calibration is uninformative, the certificate still supplies the guarantee, at the price of a conservative but valid interval, and coverage is recovered once $r_{\mathrm{train}}$ reaches the shift magnitude.

\section{Guidelines for Practitioners}
\label{sec:practitioner-guidelines}
We distil the preceding results into concrete guidance on how to apply TraCA in practice, encode prior knowledge, and interpret the guarantees it returns.

\paragraph{Choosing the ambiguity geometry.}
The geometry of the ambiguity set should mirror the form of the prior knowledge available about how the target mechanism differs from the source. If only a global bound on the total mechanism shift is known, the Frobenius ball suffices; if per-equation or per-parent bounds are available, the row- or column-wise budgets are appropriate; and if a separate bound is known for each coefficient, the entrywise box is the tightest and
most informative choice. The entrywise box is the most useful default since structural knowledge about a causal system is usually localised to specific edges.

\paragraph{Encoding a directional prior.}
When the practitioner believes a coefficient shifts in a known direction; i.e., that a treatment effect strengthens or that an environmental driver intensifies in the target population, this belief is encoded through the offset $\delta$ of the directional box $\mathcal A_W^{\mathrm{box}}(\delta,B)$. The sign of $\delta_{jk}$ gives the direction of the anticipated shift, and its magnitude indicates the expected size, while the radius $b_{jk}$ expresses the residual uncertainty around that expectation. Crucially, $\delta$ is a \emph{declared prior}, not a quantity estimated from target data: it encodes assumed domain knowledge, and the guarantees TraCA returns are conditional on that knowledge. A symmetric box ($\delta=0$) makes no directional commitment and is the appropriate choice when the direction of the shift is genuinely unknown. A wrongly declared offset of large magnitude widens the certificate rather than narrowing it, so the guarantee is robust to a confidently wrong direction only when the declared magnitude is also large; the risky declaration is a small one that points in the wrong direction.

\paragraph{What to expect from the map.}
Whether the learned map $\lintau$ improves on the identity depends on the information supplied. Under a \emph{symmetric} ambiguity set, the map departs from the identity but improves on the baseline only in specific regimes: in our experiments at large shifts under the Gaussian objective, where robustification alone yields a significant gain, but not at small shifts, where the identity is already near-optimal. The practical value of a
symmetric run, therefore, lies as much in its \emph{certified intervals} as in its point correction. A \emph{directional} prior changes this. By recentering the ambiguity set, it gives the map a genuine correction in the declared direction, present by construction and independent of the loss, which can substantially outperform the identity: up to a $60\%$ error reduction in our experiments, when the realised shift is both small and aligned with the declared direction. The improvement is not unconditional. If the ambiguity radius is chosen much larger than the shift that actually occurs, the map over-robustifies, and a smaller radius would have served better. Practitioners should, therefore, size the radius to their genuine uncertainty rather than inflating it for perceived safety: larger radii cover more scenarios at the cost of a wider interval and a less precise point estimate.

\paragraph{Calibrating the radius from observational data.}
When the target system can be observed under natural conditions but not intervened upon, one may attempt to select the operating radius from observational quantities alone: sweep the training radius, evaluate the observational transport distance $\mathcal W_2^2(\tau_\#P_s^{\mathrm{obs}}, P_t^{\mathrm{obs}})$ at each point, and select the radius that best matches the observed target. This can succeed when the observational distance has an interior optimum over the radius grid. On ATE, where the shift acts on a mechanism, the selector lands on an operating point whose certified interval contains the true interventional query, even though the interventional value was never used. It is not, however, guaranteed to work: on Portland, where the shift is purely environmental, the observational distance increases monotonically with the radius, so enlarging the ambiguity set degrades the observational fit, even though a larger radius is exactly what interventional coverage requires, and the selector returns the smallest radius at which the directional certificate does not yet cover. The practical rule is, therefore, that observational calibration is a useful heuristic when the observational distance is informative about the radius, but it can point the wrong way. Where it fails, the certificate still supplies a valid guarantee, at the price of a conservative radius sized from domain knowledge rather than from the observational fit. When even observational access to the target is unavailable, the radius must be set from domain knowledge alone, and the certificate read against the declared ambiguity set.

\paragraph{Reading the certificates.}
TraCA returns two certified intervals per query: a general certificate, valid for any target in the ambiguity set, and a sharper directional certificate that exploits the declared prior. The directional interval is several times tighter than the general one, and on some benchmarks, it is the only informative bound at all, where the general certificate is already vacuous. Within the training ball, it covers the truth at no cost
to validity. Outside the ball, when the realised shift is larger than the radius the map was trained for, the directional interval can fail to cover; a feature rather than a defect, since the resulting under-coverage signals that the declared uncertainty was too small for the shift that occurred. The general certificate, being wider, tolerates such misspecification more readily. A practitioner should therefore read the two together: a
directional interval that covers indicates the shift lies within the declared ball, while directional under-coverage is a diagnostic that the radius should be enlarged.

\section{Conclusion}
\label{sec:conclusion}
We recast transportability as a problem of \emph{model-level} causal abstraction: rather than asking whether individual queries transport, we ask whether a single map carries the entire source interventional family to the target. From this reframing the paper's contributions follow. We define a hierarchy of source-target consistency notions, from observational matching to a single constructive map transporting the full
interventional family, and isolate the \emph{compatibility gap} that separates query-level from model-level transport. We give exact existence characterisations for a common constructive transport operator in both the Markovian and semi-Markovian cases, reducing, in finite state spaces, to linear feasibility conditions. For linear additive noise models, we formulate approximate transport as a distributionally robust optimisation over target-side mechanism and environment perturbations around a trusted source, derive the stability bounds it rests on, and instantiate it as a practical min-max algorithm. When no exact map exists, the best approximate one still certifies query intervals $\ell\le Q_t\le L$ from its residual error, informative precisely in the two regimes where classical transportability is silent: non-transportable queries ($\mathcal R_1$) and absent target data ($\mathcal R_2$).

These certificates are the practical payoff, and our experiments bear them out. Across a non-transportable effect, two target-agnostic benchmarks, and a real ecological dataset, the certified intervals bracket the ground-truth query at the radius the true shift requires. Where both certificates are informative, the directional one is several times tighter: $2.9\times$ on ATCE and as much as $17\times$ per query on the higher-dimensional LiLUCAS. Furthermore, where the general bound is already vacuous, it remains informative, including on the real ecological data, where at the radius the shift requires, it is the only usable interval. This is also a difference in kind from the classical theory: do-calculus returns a formula when one exists and leaves estimation to the practitioner, whereas TraCA returns the transported estimate together with an interval containing the truth. Transportability, on this view, is not a binary gate but a special, best-case point in a continuum of certified approximate same-level abstraction; i.e., transportability is the existence of a structural relation between two causal models, and where that relation fails to hold exactly, its best approximation still certifies what can be transported.

\paragraph{Limitations.}
The framework is currently limited to linear abstraction maps and linear structural causal models. The stability bounds of Section~\ref{subsec:stability} and the closed forms underlying both certificates rest on the reduced form $\en = A\ex$ and the 
resolvent identity for $A=(I-W)^{-1}$, neither of which applies in a nonlinear setting. It further assumes access to the true causal DAG or an accurate estimate obtained by causal discovery: the graph fixes which mechanisms may shift, which transport components are free, and, in the semi-Markovian case, the districts over which the map factorises. A second assumption concerns the source. TraCA treats the source SCM as a trusted causal reference rather than an object to be estimated and places the entire ambiguity on the target side. This is the right division when the source is a well-studied population and the target the unobserved one; the setting the framework is built for, but it means the guarantees inherit whatever error is present in the source model: sampling and estimation errors in the reference are not accounted to the final certified interval.

\paragraph{Future work.}
Several directions follow naturally. When the source SCM cannot be treated as a trusted reference, the natural extension is a doubly distributionally robust formulation: a standard DRO ambiguity set on the source, absorbing sampling and estimation errors in the reference model, alongside the target-side set we already use for domain generalisation; robustness over both the model we start from and the shift we transport to. A second direction relaxes the acyclicity of the district quotient, where transport may still hold by non-constructive means beyond the present existence theory. A third concerns the ambiguity set itself: we currently declare the shift geometry, which mechanisms may move, and in the directional case, the offset $\delta$; a natural next step is to learn this geometry from whatever partial target information is available, shrinking the ambiguity radius around a better-localised target and, with it, tightening the certificate. Finally, our certificates bound arbitrary Lipschitz query functionals and sharpen for linear readouts; extending the directional analysis to richer functionals and deriving tighter stability constants would narrow the gap between the certified interval and the true worst case.

\acks{\textbf{YF} acknowledges support by the Onassis Foundation - Scholarship ID: F ZR 063-1/2021-2022. \textbf{TD} acknowledges support from a UKRI Turing AI acceleration Fellowship [EP/V02678X/1].}

\bibliography{references}

\appendix
\section*{Appendix overview}
The Appendix collects the following supporting material. First, it contains the full proofs and finite-state refinements of the exact Markovian and semi-Markovian existence results from
Section~\ref{sec:characterizing} (Appendix~\ref{app:exact-transportability-details}), together with the attainment results for the transport error (Appendix~\ref{app:attainment}). Second, it records auxiliary LAN and minimax results deferred from Sections~\ref{sec:lan} and \ref{subsec:unified-minimax}: a structural counterexample separating mechanism from environment shifts (Appendix~\ref{app:lan-aux}), the proof of the geometry-specific amplification bounds (Appendix~\ref{app:gamma-alpha-plug-in}), and the information-monotonicity and oracle-convergence properties of the robust value (Appendix~\ref{app:information-monotonicity}). Third, it contains the full analytical proofs of the robustness certificates from Section~\ref{sec:provable-robustness} (Appendix~\ref{app:certificate-proofs}), including the directional certificates for linear queries and their computable support-function bounds (Appendix~\ref{app:directional-certificates}). Fourth, it provides the formulation-specific optimisation details omitted from the unified presentation in Section~\ref{sec:algorithms}, including the Gaussian and empirical adversarial algorithms (Appendix~\ref{app:optimization-details}). Fifth, it records the structural equations and intervention sets for all the benchmarks of Section~\ref{sec:experiments} (Appendix~\ref{app:benchmarks}) and the implementation and optimisation details (Appendix~\ref{app:implementation}). Sixth, it clarifies the relationship between TraCA and DiRoCA \citep{felekis2026distributionallyrobustcausalabstractions}, separating the shared optimisation machinery from TraCA's distinct problem and contributions (Appendix~\ref{app:relation-diroca}). Finally, it collects the supplementary experimental results deferred from Section~\ref{sec:experiments}, organised by benchmark (Appendix~\ref{app:supplementary-results}).

\section{Exact transportability existence results}
\label{app:exact-transportability-details}
This section gathers the exact existence results deferred from Section~\ref{sec:characterizing}. It mirrors the structure of the main text: we provide the full recursive proofs in the Markovian and semi-Markovian settings, as well as the finite-state reductions and uniqueness refinements that make these results algorithmically checkable.

\subsection{Markovian characterisation}\label{app:markovian-proofs} 
This subsection contains the full proof of the Markovian recursive assembly theorem from Section~\ref{subsec:mechanism-characterization-markovian}. The argument formalises the stagewise prefix-matching intuition used in the main text.
\vspace{3mm}

\begin{namedproof}{Theorem~\ref{thm:mechanism-characterization-markovian}}
(\(\Rightarrow\)) Assume \(K\) is a CICk, so $P_s^{(\iota)}K=P_t^{(\iota)},~\forall\,\iota\in\cI$. Hence, $R_d^{(\iota)}=P_s^{(\iota)}K=P_t^{(\iota)}$. For each \(i\), the later kernels \(\bar K_{i+1},\dots,\bar K_d\) do not alter the first \(i\) coordinates, so
\begin{align*}
  R_i^{(\iota)}(X_1,\dots,X_i)
  =
  R_d^{(\iota)}(X_1,\dots,X_i)
  =
  P_t^{(\iota)}(X_1,\dots,X_i).
\end{align*}
If \(i\in J_\iota\), then \(P_t^{(\iota)}(X_i)=\delta_{a_i}\), hence \(R_i^{(\iota)}(X_i)=\delta_{a_i}\). If \(i\notin J_\iota\), then by truncated factorisation in the target model, $P_t^{(\iota)}\!\bigl(X_i \mid X_{\operatorname{PA}(X_i)}\bigr) = \prob_{\scmsimple_t}\!\bigl(X_i \mid X_{\operatorname{PA}(X_i)}\bigr)$, and the same conditional therefore holds under \(R_i^{(\iota)}\), since \(R_i^{(\iota)}\) and \(P_t^{(\iota)}\) agree on the prefix \((X_1,\dots,X_i)\).

\((\Leftarrow)\)
Assume now that conditions (i), (ii) hold for every \(\iota\in\cI\) and every node \(i\in[d]\). We prove by induction on \(i\) that
\begin{align*}
  R_i^{(\iota)}(X_1,\dots,X_i)
  =
  P_t^{(\iota)}(X_1,\dots,X_i)
  \qquad \forall\,\iota\in\cI.
\end{align*}
The case \(i=0\) is trivial. Suppose the claim holds for \(i-1\). Since \(\operatorname{PA}(X_i)\subseteq\{X_1,\dots,X_{i-1}\}\), the induction hypothesis implies that \(R_{i-1}^{(\iota)}\) and \(P_t^{(\iota)}\) agree on the parent tuple of \(X_i\). If \(i\in J_\iota\), then by condition (i), \(R_i^{(\iota)}(X_i)=\delta_{a_i}\), which matches the intervened factor in \(P_t^{(\iota)}\). If \(i\notin J_\iota\), then by condition (ii) $R_i^{(\iota)}\!\bigl(X_i \mid X_{\operatorname{PA}(X_i)}\bigr) = \prob_{\scmsimple_t}\!\bigl(X_i \mid X_{\operatorname{PA}(X_i)}\bigr)$. Since \(K\) is stagewise Markov-preserving,
\begin{align*}
  R_i^{(\iota)}\!\bigl(X_i \mid X_1,\dots,X_{i-1}\bigr)
  =
  R_i^{(\iota)}\!\bigl(X_i \mid X_{\operatorname{PA}(X_i)}\bigr)
  =
  \prob_{\scmsimple_t}\!\bigl(X_i \mid X_{\operatorname{PA}(X_i)}\bigr).
\end{align*}
By the target truncated factorisation,
\begin{align}
  P_t^{(\iota)}\!\bigl(X_i \mid X_1,\dots,X_{i-1}\bigr)
  =
  \prob_{\scmsimple_t}\!\bigl(X_i \mid X_{\operatorname{PA}(X_i)}\bigr).
\end{align}
Thus,
\begin{align}
  R_i^{(\iota)}\!\bigl(X_i \mid X_1,\dots,X_{i-1}\bigr)
  =
  P_t^{(\iota)}\!\bigl(X_i \mid X_1,\dots,X_{i-1}\bigr).
\end{align}
Combining this with equality of the prefix law on \((X_1,\dots,X_{i-1})\) yields
\begin{align}
  R_i^{(\iota)}(X_1,\dots,X_i)
  =
  P_t^{(\iota)}(X_1,\dots,X_i).
\end{align}
This closes the induction. Taking \(i=d\) gives
\begin{align}
  R_d^{(\iota)}=P_t^{(\iota)}.
\end{align}
Since \(R_d^{(\iota)}=P_s^{(\iota)}K\), it follows that \(K\) is a CICk over \(\cI\). The deterministic case is obtained by restricting to kernels of the form \(K_i(\cdot\mid x_i)=\delta_{\tau_i(x_i)}\).
\end{namedproof}

\subsection{Markovian finite-state stagewise compatibility, uniqueness, and deterministic specialisation}
\label{app:markovian-finite-stagewise}
This subsection records the finite-state consequences of the Markovian recursive assembly theorem: a linear-feasibility formulation of stagewise existence, a rank-based uniqueness criterion, and the deterministic specialisation.

We now specialise the stage-\(i\) alignment condition to the case where the state space of a node is finite. Fix a node \(i\in[d]\) and suppose \(\dom{X_i}=\{1,\dots,n\}\). For each intervention \(\iota\in\cI\) with \(i\notin J_\iota\), and each parent context $p \in \mathrm{supp}\!\Bigl(R_{i-1}^{(\iota)}\bigl(\operatorname{PA}(X_i)\bigr) \Bigr)$, define
\begin{align*}
  \mu_{\iota,p}^{(i-1)}
  :=
  R_{i-1}^{(\iota)}\!\bigl(
    X_i \mid \operatorname{PA}(X_i)=p
  \bigr)
  \in \Delta(\dom{X_i}),
  \nu_p
  :=
  \prob_{\scmsimple_t}\!\bigl(
    X_i \mid \operatorname{PA}(X_i)=p
  \bigr)
  \in \Delta(\dom{X_i}),
\end{align*}
where \(\Delta(\dom{X_i})\) denotes the probability simplex. Let
\begin{align}
  \mathcal A_i
  :=
  \left\{
    (\iota,p):
    \iota\in\cI,\ i\notin J_\iota,\ 
    p\in \mathrm{supp}\!\Bigl(
      R_{i-1}^{(\iota)}\bigl(\operatorname{PA}(X_i)\bigr)
    \Bigr)
  \right\}
\end{align}
denote the set of admissible stage-\(i\) contexts.

\begin{theorem}
\label{thm:finite-discrete-compatibility-markovian}
At stage \(i\), the following are equivalent:
\begin{enumerate}[label=(\roman*)]
  \item There exists a row-stochastic matrix \(K_i\in[0,1]^{n\times n}\) such that
  \begin{align}
    \mu_{\iota,p}^{(i-1)} K_i = \nu_p,~ \forall\,(\iota,p)\in\mathcal A_i~\text{ and }~ K_i(\cdot\mid a)=\delta_a
  \end{align}
  for every intervention value \(a\) assigned to node \(i\) by some intervention in \(\cI\).

  \item There exist probability vectors \(k_1,\dots,k_n \in \Delta(\dom{X_i})\), one for each source state \(v\in\dom{X_i}\), such that \(k_a=\delta_a\) for every intervention value \(a\) assigned to node \(i\) by some intervention in \(\cI\), and for every \((\iota,p)\in\mathcal A_i\) and every \(y\in\dom{X_i}\),
  \begin{align}
    \nu_p(y)
    =
    \sum_{v=1}^n \mu_{\iota,p}^{(i-1)}(v)\,k_v(y).
  \end{align}

  \item The linear system
  \begin{align}
    \sum_{v=1}^n \mu_{\iota,p}^{(i-1)}(v)\,K_i(y\mid v)
    =
    \nu_p(y)
    \qquad
    \forall\,(\iota,p)\in\mathcal A_i,\ \forall\,y\in\dom{X_i},
  \end{align}
  together with the row-fixing constraints $K_i(\cdot\mid a)=\delta_a$ for every intervention value \(a\) assigned to node \(i\) by some intervention in \(\cI\), and the row-stochasticity constraints $K_i(y\mid v)\ge 0 \ \forall\,v,y,~ \sum_{y=1}^n K_i(y\mid v)=1 \ \forall\,v$,  is feasible.
\end{enumerate}
\end{theorem}

\begin{proof}
The equivalence of (i), (ii), and (iii) is only a matter of notation. A local kernel on the finite state space \(\dom{X_i}=\{1,\dots,n\}\) is the same thing as a row-stochastic matrix \(K_i\in[0,1]^{n\times n}\). If we write $k_v := K_i(\cdot\mid v)\in\Delta(\dom{X_i})$ for its \(v\)-th row, then for every admissible pair \((\iota,p)\in\mathcal A_i\) and every \(y\in\dom{X_i}\),
\begin{align}
  (\mu_{\iota,p}^{(i-1)}K_i)(y)
  =
  \sum_{v=1}^n \mu_{\iota,p}^{(i-1)}(v)\,K_i(y\mid v)
  =
  \sum_{v=1}^n \mu_{\iota,p}^{(i-1)}(v)\,k_v(y).
\end{align}
Thus, condition (i) written entrywise is exactly condition (iii), and rewriting the rows of \(K_i\) as probability vectors gives condition (ii).
\end{proof}
Theorem~\ref{thm:finite-discrete-compatibility-markovian} is the finite-state version of the stage-\(i\) alignment condition from Theorem~\ref{thm:mechanism-characterization-markovian}. Once the earlier kernels \(K_1,\dots,K_{i-1}\) have been fixed, the search for \(K_i\) reduces to a linear feasibility problem. The remaining assembly issue is whether these stagewise compatible kernels also satisfy the stagewise Markov-preserving constraint; this is itself a stagewise property of \(K_i\), given the earlier kernels.

Once stagewise existence has been reduced to a linear system, uniqueness can also be studied node by node: at stage \(i\), the kernel \(K_i\) is determined by the transformed source conditional family \(\{\mu_{\iota,p}^{(i-1)}\}_{(\iota,p)\in\mathcal A_i}\), together with the requirement that intervention values be fixed.

\begin{proposition}[Uniqueness of the stagewise kernel]
\label{prop:uniqueness-common-kernel-markovian}
Fix a node \(i\in[d]\) with a finite state space \(\dom{X_i}=\{1,\dots,n\}\), and suppose earlier kernels \(K_1,\dots,K_{i-1}\) have been fixed. Assume there exists a row-stochastic kernel \(K_i\) satisfying the stagewise compatibility conditions of Theorem~\ref{thm:finite-discrete-compatibility-markovian}. Let \(M_i\) be the matrix whose rows are the transformed source conditionals \(\mu_{\iota,p}^{(i-1)}\), one for each \((\iota,p)\in\mathcal A_i\), and one additional row \(\delta_a\) for each intervention value \(a\) assigned to node \(i\) by some intervention in \(\cI\). Let \(N_i\) be the matrix whose corresponding rows are the target conditionals \(\nu_p\), and the same point masses \(\delta_a\). Then:
\begin{enumerate}[label=(\roman*)]
  \item If \(\operatorname{rank}(M_i)=n\), then the compatible row-stochastic kernel \(K_i\) is unique.

  \item If \(\operatorname{rank}(M_i)<n\), then the equation \(M_iK_i=N_i\) does not uniquely determine \(K_i\). Its solution set in \(\mathbb R^{n\times n}\) is an affine space of positive dimension. Hence, uniqueness, if it still occurs, can only come from the row-stochasticity constraints.

  \item Uniqueness requires that the rows of \(M_i\) contain at least \(n\) linearly independent vectors. In particular, the total number of admissible stage-\(i\) contexts, together with the number of distinct intervention values assigned to node \(i\), must be at least \(n\).
\end{enumerate}
\end{proposition}

\begin{proof}
By finite discrete stagewise compatibility, the stage-\(i\) constraints are exactly the equations
\begin{align}
  \mu_{\iota,p}^{(i-1)}K_i=\nu_p
  \qquad \forall\,(\iota,p)\in\mathcal A_i,
\end{align}
together with the intervention-preservation constraints $\delta_a K_i=\delta_a$ for every intervention value \(a\) assigned to node \(i\) by some intervention in \(\cI\). Stacking these equations gives the matrix equation $M_iK_i=N_i$.

For each output state \(y\in\dom{X_i}\), let \(c_y\in\mathbb R^n\) with \(c_y(v):=K_i(y\mid v)\) be the \(y\)-th column of \(K_i\), and let \(\nu_{\cdot y}\) denote the \(y\)-th column of \(N_i\). Then \(M_iK_i=N_i\) is equivalent to the family of column systems
\begin{align}
  M_ic_y=\nu_{\cdot y}
  \qquad \forall\,y\in\dom{X_i}.
\end{align}
If \(\operatorname{rank}(M_i)=n\), then \(\ker(M_i)=\{0\}\), so each column system has at most one solution. Hence, every column \(c_y\) is uniquely determined, and therefore \(K_i\) is unique. This proves (i).

If \(\operatorname{rank}(M_i)<n\), then \(\ker(M_i)\neq\{0\}\). Thus, if \(c_y\) solves \(M_ic_y=\nu_{\cdot y}\), then so does \(c_y+\delta\) for every \(\delta\in\ker(M_i)\), since $M_i(c_y+\delta)=M_ic_y+M_i\delta=\nu_{\cdot y}$. Therefore the unconstrained solution set of \(M_iK_i=N_i\) has positive dimension. This proves (ii).

Finally, full column rank \(n\) requires at least \(n\) rows. Thus, the rows of \(M_i\) must contain at least \(n\) linearly independent vectors, which proves (iii).
\end{proof}

\begin{remark}[Global from stagewise uniqueness]
\label{cor:global-uniqueness-markovian}
Suppose a CICk \(K=\prod_{i=1}^d K_i\) exists. If, at each stage \(i\in[d]\), the corresponding compatible kernel \(K_i\) is unique, then the CICk \(K\) is unique since, by constructiveness, the global kernel factorises componentwise \(K=\prod_{i=1}^d K_i\), and the invariant-node components are fixed to the identity.
\end{remark}

\begin{remark}
The row-stochasticity constraints cut out a convex polytope inside \(\mathbb R^{n\times n}\). Hence, the set of stagewise compatible stochastic kernels is the intersection of the affine space \(\{K_i:M_iK_i=N_i\}\) with that polytope. In particular, when \(\operatorname{rank}(M_i)<n\), the unconstrained linear system is never unique, but the stochasticity constraints may still reduce the admissible set to a singleton.
\end{remark}

\begin{remark}
\label{rem:interventions-drive-identification-markovian}
The matrix \(M_i\) is controlled by the intervention family \(\cI\) through the admissible stage-\(i\) contexts \(\mathcal A_i\) and through the intervention values that node \(i\) is required to fix. Interventions on ancestors of \(X_i\) alter the partially transformed law \(R_{i-1}^{(\iota)}\), create new admissible contexts \((\iota,p)\), and may increase the rank of \(M_i\). Direct interventions on \(X_i\) add the corresponding row-fixing constraints. Thus, richer intervention families do not merely make the stagewise compatibility condition more stringent; they can also make the compatible kernel \(K_i\) more tightly determined and, in favourable cases, uniquely determined.
\end{remark}

\begin{remark}[Deterministic specialisation]
At stage \(i\), the recursive compatibility condition is deterministic if and only if there exists a measurable map \(\tau_i:\dom{X_i}\to\dom{X_i}\) such that
\begin{align}
  (\tau_i)_{\#}\mu_{\iota,p}^{(i-1)}=\nu_p,~ \forall\,(\iota,p)\in\mathcal A_i ~\text{ and }~\tau_i(a)=a
\end{align}
for every intervention value \(a\) assigned to node \(i\) by some intervention in \(\cI\). Equivalently, the stagewise kernel can be chosen in the form \(K_i(\cdot\mid x_i)=\delta_{\tau_i(x_i)}\). This is strictly more restrictive than stochastic stagewise compatibility: one must find a single deterministic state map, independent of the intervention and the parent context, that transports every transformed source mechanism slice to the corresponding target slice while preserving all intervention values. In particular, a global CICm can exist only if this deterministic stagewise condition holds at every node and the resulting deterministic kernel is stagewise Markov-preserving.
\end{remark}

\subsection{Semi-Markovian finite-state district compatibility, uniqueness, and related remarks}
\label{app:semimarkovian-finite-stagewise}
This subsection records the finite-state districtwise reductions of the semi-Markovian theory. These are the exact district-level analogues of the Markovian finite-state compatibility and uniqueness results.

\paragraph{Finite discrete district compatibility and uniqueness.}
Suppose now that \(\dom{X_{D_r}}=\{1,\dots,N_r\}\) is finite. For every intervention \(\iota\in\cI\) with \(D_r\cap J_\iota=\emptyset\), and every parent context $p\in
  \mathrm{supp}\!\Bigl(
    R_{r-1}^{(\iota)}(\operatorname{PA}(D_r))
  \Bigr)$, define
\begin{align*}
  \mu_{\iota,p}^{(r-1)}
  :=
  R_{r-1}^{(\iota)}\!\bigl(
    X_{D_r}\mid \operatorname{PA}(D_r)=p
  \bigr)
  \in\Delta(\dom{X_{D_r}}),
  \qquad
  \nu_p
  :=
  q_{D_r}^t(\cdot\mid p)
  \in\Delta(\dom{X_{D_r}}),
\end{align*}
and let
\begin{align}
  \mathcal A_{D_r}
  :=
  \left\{
    (\iota,p):
    \iota\in\cI,\ D_r\cap J_\iota=\emptyset,\ 
    p\in\mathrm{supp}\!\Bigl(
      R_{r-1}^{(\iota)}(\operatorname{PA}(D_r))
    \Bigr)
  \right\}.
\end{align}

With these substitutions, the finite discrete stagewise compatibility and uniqueness results from the Markovian case carry over verbatim:
\begin{align}
  i \rightsquigarrow D_r,\qquad
  \dom{X_i}\rightsquigarrow \dom{X_{D_r}},\qquad
  n\rightsquigarrow N_r,\qquad
  \mu_{\iota,p}^{(i-1)},\nu_p,K_i
  \rightsquigarrow
  \mu_{\iota,p}^{(r-1)},\nu_p,K_{D_r}.
\end{align}
In particular, the existence of a compatible stagewise district kernel is, once again, a linear feasibility problem. Exact global CICk existence then follows, provided these stagewise district kernels can be assembled into a stagewise district-preserving districtwise product kernel.

\begin{remark}
\label{rem:semi-practical-difference}
The district-based semi-Markovian specialisation preserves the logical structure of the Markovian theory, but the local state space is now \(\dom{X_D}\) rather than \(\dom{X_i}\). Thus, the stagewise linear program at a shifted district has \(N_r^2\) unknowns, where $N_r=\prod_{i\in D_r}|\dom{X_i}|$, so the computational cost grows rapidly with district size. This is the main practical price of moving from nodes to districts.
\end{remark}

\begin{remark}
\label{rem:semi-query-payoff}
The query-level consequence of a common constructive kernel has exactly the same formal shape as in the Markovian case:
\begin{align}
  \Phi\!\left(\pi_{O_\iota\#}P_t^{(\iota)}\right)
  =
  \Phi\!\left(\pi_{O_\iota\#}\bigl(P_s^{(\iota)}K\bigr)\right)
  \qquad \forall\,(\iota,O_\iota)\in\Qcal.
\end{align}
The difference is that, in the semi-Markovian setting, this identity is no longer merely a structural re-expression of already transportable queries. It can provide a genuine universal transport operator in a regime where direct transportability fails.
\end{remark}

\section{Attainment of the transport error}\label{app:attainment} 
\begin{proposition}
\label{prop:consistency-transport-error}
Suppose \(\mathsf D(\mu,\nu)=0\) iff \(\mu=\nu\). Then:
\begin{enumerate}[label=(\roman*)]
    \item For any admissible common kernel \(K\), \(\mathcal E_{\Qcal}(K)=0\) if and only if \(K\) realises exact \(\Qcal\)-restricted consistency.

    \item Given a class \(\mathfrak K\) of admissible common kernels, let $\mathcal E_{\Qcal}^\star(\mathfrak K):=\inf_{K\in\mathfrak K}\mathcal E_{\Qcal}(K)$. If the infimum is attained, then \(\mathcal E_{\Qcal}^\star(\mathfrak K)=0\) if and only if there exists \(K\in\mathfrak K\) realising exact \(\Qcal\)-restricted consistency.
\end{enumerate}
\end{proposition}

\begin{proof}
For (i), \(\mathcal E_{\Qcal}(K)\) is a finite average of nonnegative terms \(e_K^{(\iota,O_\iota)}\). Hence
\(\mathcal E_{\Qcal}(K)=0
\iff e_K^{(\iota,O_\iota)}=0,~\forall\,(\iota,O_\iota)\in\Qcal\). By the defining property of \(\mathsf D\), each pointwise error vanishes if and only if \(\pi_{O_\iota\#}\bigl(P_s^{(\iota)}K\bigr) = \pi_{O_\iota\#}P_t^{(\iota)},~\forall\,(\iota,O_\iota)\in\Qcal,\) which is exactly \(\Qcal\)-restricted consistency.

For (ii), if some \(K\in\mathfrak K\) realises exact \(\Qcal\)-restricted consistency, then by (i) \(\mathcal E_{\Qcal}(K)=0\), so
\(\mathcal E_{\Qcal}^\star(\mathfrak K)=0\). Conversely, if the infimum is attained and \(\mathcal E_{\Qcal}^\star(\mathfrak K)=0\), let \(K^\star\in\mathfrak K\) be a minimiser; then \(\mathcal E_{\Qcal}(K^\star)=0\), and part (i) implies that \(K^\star\) realises exact \(\Qcal\)-restricted consistency.
\end{proof}

\begin{corollary}
\label{cor:attainment-transport-error}
Suppose \(\mathfrak K\) is compact in a topology under which \(\mathcal E_{\Qcal}\) is lower semicontinuous. Then the infimum
\[
  \mathcal E_{\Qcal}^\star(\mathfrak K)
  :=
  \inf_{K\in\mathfrak K}\mathcal E_{\Qcal}(K)
\]
is attained. Consequently, $\mathcal E_{\Qcal}^\star(\mathfrak K)=0$ if and only if $ \exists\,K\in\mathfrak K$ realising exact \(\Qcal\)-restricted consistency.
\end{corollary}
\begin{proof}
By the generalised Weierstrass theorem, every lower semicontinuous real-valued function on a compact set attains its minimum. Hence
\(\mathcal E_{\Qcal}^\star(\mathfrak K)\) is attained under the stated hypotheses. The final equivalence then follows from Proposition~\ref{prop:consistency-transport-error}(ii).
\end{proof}
Accordingly, whenever the admissible class \(\mathfrak K\) satisfies these compactness and lower semicontinuity conditions, vanishing transport error is equivalent to exact realisability within \(\mathfrak K\).

\section{Auxiliary structural results}\label{app:lan-aux}                   
\begin{proposition}
\label{prop:mech_not_noise}
Let two SCMs share the same endogenous space \(\mathcal X\) and exogenous space \(\mathcal U\), but have different reduced-form maps \(g,g':\mathcal U\to\mathcal X\). Fix an environment \(\rho\) on \(\mathcal U\), and define the induced observable distributions
\begin{align}
  P := g_{\#}\rho,
  \qquad
  P' := g'_{\#}\rho.
\end{align}
In general, there need not exist any exogenous law \(\rho^\star\) on \(\mathcal U\) such that
\begin{align}
  g_{\#}\rho^\star = P'.
\end{align}
Equivalently, the distribution generated by changing the reduced-form mechanism from \(g\) to \(g'\) cannot in general be reproduced by keeping \(g\) fixed and changing only the exogenous law.
\end{proposition}
\begin{proof}
Take $\mathcal U=\mathcal X=\{0,1\}$, let \(\rho\) be the uniform distribution on \(\{0,1\}\), and define the source reduced-form map \(g\) by $g(0)=g(1)=0$. Then the source SCM outputs the endogenous value \(0\) almost surely, regardless of the environment. Hence, for every probability measure \(\rho^\star\) on \(\mathcal U\), $g_{\#}\rho^\star=\delta_0$. Now define the target reduced-form map \(g'\) by $g'(y)=y$. Under the same environment \(\rho\),
\begin{align}
  g'_{\#}\rho=\tfrac12\delta_0+\tfrac12\delta_1.
\end{align}
This distribution cannot be written as \(g_{\#}\rho^\star\) for any \(\rho^\star\), because every pushforward through \(g\) is equal to \(\delta_0\) and therefore assigns zero mass to the state \(1\).
\end{proof}

\subsection{Corollary~\ref{cor:gamma-alpha-plug-in-traca}}
\label{app:gamma-alpha-plug-in}

\begin{namedproof}{Corollary~\ref{cor:gamma-alpha-plug-in-traca}}
The result follows by bounding \(\|A_\iota(R_\iota\Delta W)\|_2\) in a way adapted to each ambiguity geometry. Different geometries naturally control different matrix norms: row budgets control the induced \(\infty\)-norm, column budgets the induced \(1\)-norm, and entrywise bounds give direct entrywise control. Since \(\gamma_\iota\) is defined through the spectral norm, we convert these to \(\|\cdot\|_2\) via standard norm inequalities.

\begin{itemize}[itemsep=0.4em,topsep=0.4em]
    \item \emph{Frobenius ball.} Submultiplicativity gives
    \begin{align}
    \|A_\iota(R_\iota\Delta W)\|_2
    \le \|A_\iota\|_2\|R_\iota\Delta W\|_2
    \le \|A_\iota\|_2\|\Delta W\|_F
    \le \|A_\iota\|_2\eta.
    \end{align}

    \item \emph{Row-wise budgets.} For each row \(\ell\),
    \begin{align}
    \sum_k |(A_\iota(R_\iota\Delta W))_{\ell k}|
    \le \sum_j |(A_\iota)_{\ell j}| \sum_k |(R_\iota\Delta W)_{jk}|.
    \end{align}
    Since \(R_\iota\) is diagonal it can only zero out rows, and admissible perturbations satisfy \(\Delta W_{j\cdot}=0\) for \(j\notin\mathcal K\), so
    \begin{align}
    \sum_k |(A_\iota(R_\iota\Delta W))_{\ell k}|
    \le \sum_{j\in\mathcal K}|(A_\iota)_{\ell j}|\sum_k |\Delta W_{jk}|
    \le \sum_{j\in\mathcal K}|(A_\iota)_{\ell j}|\rho_j.
    \end{align}
    Hence \(\|A_\iota(R_\iota\Delta W)\|_\infty \le \max_\ell \sum_{j\in\mathcal K}|(A_\iota)_{\ell j}|\rho_j\), and by \(\|M\|_2\le \sqrt d\,\|M\|_\infty\),
    \begin{align}
    \|A_\iota(R_\iota\Delta W)\|_2
    \le \sqrt d\,\max_\ell \sum_{j\in\mathcal K}|(A_\iota)_{\ell j}|\rho_j.
    \end{align}

    \item \emph{Column-wise budgets.} For each column \(k\),
    \begin{align}
    \sum_\ell |(A_\iota(R_\iota\Delta W))_{\ell k}|
    \le \sum_j \Bigl(\sum_\ell |(A_\iota)_{\ell j}|\Bigr)
        |(R_\iota\Delta W)_{jk}|.
    \end{align}
    Again \(R_\iota\) can only zero out rows, so
    \begin{align}
    \sum_\ell |(A_\iota(R_\iota\Delta W))_{\ell k}|
    \le \|A_\iota\|_1\sum_j |\Delta W_{jk}|
    \le \|A_\iota\|_1 c_k.
    \end{align}
    Hence \(\|A_\iota(R_\iota\Delta W)\|_1 \le \|A_\iota\|_1\max_k c_k\), and by \(\|M\|_2\le \sqrt d\,\|M\|_1\),
    \begin{align}
    \|A_\iota(R_\iota\Delta W)\|_2
    \le \sqrt d\,\|A_\iota\|_1\max_k c_k.
    \end{align}

    \item \emph{Entrywise box.} The bound \(|\Delta W_{jk}|\le b_{jk}\) gives \(|(R_\iota\Delta W)_{jk}|\le (R_\iota)_{jj} b_{jk}\), so entrywise \(|A_\iota(R_\iota\Delta W)| \le |A_\iota|R_\iota B\), and therefore
    \begin{align}
    \|A_\iota(R_\iota\Delta W)\|_2
    \le \|A_\iota(R_\iota\Delta W)\|_F
    \le \bigl\||A_\iota|R_\iota B\bigr\|_F.
    \end{align}
\end{itemize}
For the directional box $\mathcal A_W^{\mathrm{box}}(\delta,B)$ the same argument applies with $b_{jk}$ replaced by the effective half-width $|\delta_{jk}|+b_{jk}$, since admissible entries satisfy $|\Delta W_{jk}|\le|\delta_{jk}|+b_{jk}$, giving \eqref{eq:gamma-box-dir-traca}. Taking the supremum over \(\Delta W\) in each case yields \eqref{eq:gamma-frob-traca}-\eqref{eq:gamma-box-dir-traca}.
\end{namedproof}

\section{Information monotonicity of the robust value}
\label{app:information-monotonicity}
This section records the monotonicity and oracle-convergence properties of the robust transport value referenced in Remark~\ref{rem:information-monotonicity}. Both are direct consequences of the minimax structure of \eqref{eq:generic-minimax-traca}.

\begin{proposition}[Information monotonicity and ambiguity tightening]
\label{prop:information-monotonicity-traca}
Let \((\Xi,d_{\mathcal A})\) denote the space of admissible target-side perturbation parameters, and for any nonempty \(\mathcal A\subseteq\Xi\) let \(\mathrm{diam}_{d_{\mathcal A}}(\mathcal A):= \sup_{\xi,\xi'\in\mathcal A} d_{\mathcal A}(\xi,\xi')\). Define the robust transport value associated with an admissible target-side ambiguity set \(\mathcal A\) by
\begin{align}
  \mathfrak R^\star(\mathcal A)
  :=
  \inf_{\lintau}\sup_{\xi\in\mathcal A}
  \frac{1}{|\mathcal I|}
  \sum_{\iota\in\mathcal I}
  \mathsf F_\iota(\lintau,\xi).
\end{align}
Fix a reference target perturbation \(\xi^\circ\in\Xi\), interpreted as the true target-side perturbation, so that \(\mathfrak R^\star(\{\xi^\circ\})\) is the \emph{oracle value} attained under full knowledge of the target. Then:
\begin{enumerate}[label=(\roman*)]
  \item \emph{(Monotonicity and certificate tightening.)}
  If \(\mathcal A_1\subseteq\mathcal A_2\), then \(\mathfrak R^\star(\mathcal A_1)\le\mathfrak R^\star(\mathcal A_2)\). If, in addition, \(\xi^\circ\in\mathcal A_1\) and \(\lintau_1^\star\in\displaystyle\argmin_{\lintau}\sup_{\xi\in\mathcal A_1} \tfrac{1}{|\mathcal I|}\sum_{\iota\in\mathcal I}\mathsf F_\iota(\lintau,\xi)\), then
  \begin{align}
    \frac{1}{|\mathcal I|} \sum_{\iota\in\mathcal I} \mathsf F_\iota(\lintau_1^\star,\xi^\circ) \le \mathfrak R^\star(\mathcal A_1) \le \mathfrak R^\star(\mathcal A_2).
  \end{align}

  \item \emph{(Quantitative tightening.)}
  Suppose each \(\mathsf F_\iota(\lintau,\xi)\) is \(L_\iota\)-Lipschitz in \(\xi\) with respect to \(d_{\mathcal A}\), uniformly over \(\lintau\), and set \(\bar L:=\tfrac{1}{|\mathcal I|}\sum_{\iota\in\mathcal I}L_\iota\). Then for any nonempty bounded \(\mathcal A\subseteq\Xi\) with \(\xi^\circ\in\mathcal A\),
  \begin{align}
    0
    \le
    \mathfrak R^\star(\mathcal A) - \mathfrak R^\star(\{\xi^\circ\})
    \le
    \bar L\,\mathrm{diam}_{d_{\mathcal A}}(\mathcal A).
  \end{align}
  If \(\mathcal A_{k+1}\subseteq\mathcal A_k\), \(\xi^\circ\in\mathcal A_k~\forall k\), and \(\mathrm{diam}_{d_{\mathcal A}}(\mathcal A_k)\to0\), then \(\mathfrak R^\star(\mathcal A_k)\to\mathfrak R^\star(\{\xi^\circ\})\).
\end{enumerate}
\end{proposition}

\begin{proof}
(i) For every fixed \(\lintau\), \(\mathcal A_1\subseteq\mathcal A_2\) gives \(\sup_{\xi\in\mathcal A_1}\tfrac{1}{|\mathcal I|}\sum_\iota \mathsf F_\iota(\lintau,\xi)\le\sup_{\xi\in\mathcal A_2} \tfrac{1}{|\mathcal I|}\sum_\iota\mathsf F_\iota(\lintau,\xi)\); taking \(\inf_{\lintau}\) yields the first inequality. When \(\xi^\circ\in\mathcal A_1\), evaluating the supremum at \(\xi^\circ\)
gives
\[
  \tfrac{1}{|\mathcal I|}\sum_\iota\mathsf F_\iota(\lintau_1^\star,\xi^\circ)
  \le\sup_{\xi\in\mathcal A_1}\tfrac{1}{|\mathcal I|}\sum_\iota
  \mathsf F_\iota(\lintau_1^\star,\xi)
  =\mathfrak R^\star(\mathcal A_1),
\]
where the equality holds because \(\lintau_1^\star\) attains the outer infimum; chaining with the first inequality completes the bound.

(ii) Fix \(\lintau\) and \(\xi\in\mathcal A\). Since \(\xi^\circ\in\mathcal A\), we have \(d_{\mathcal A}(\xi,\xi^\circ)\le \mathrm{diam}_{d_{\mathcal A}}(\mathcal A)\), so Lipschitz continuity gives
\[
  \tfrac{1}{|\mathcal I|}\sum_\iota\mathsf F_\iota(\lintau,\xi)
  \le\tfrac{1}{|\mathcal I|}\sum_\iota\mathsf F_\iota(\lintau,\xi^\circ)
  +\bar L\,\mathrm{diam}_{d_{\mathcal A}}(\mathcal A).
\]
Taking \(\sup_{\xi\in\mathcal A}\) and then \(\inf_{\lintau}\) yields the upper bound; the lower bound is part~(i) applied to \(\{\xi^\circ\}\subseteq\mathcal A\). The convergence claim follows from the two-sided bound and \(\mathrm{diam}_{d_{\mathcal A}}(\mathcal A_k)\to0\).
\end{proof}
The proposition formalises the scaling-with-information property of TraCA. Part~(i) is qualitative: if additional target information shrinks the ambiguity set, then the robust objective can only decrease, and the resulting certificate on the true target loss can only improve. Part~(ii) is quantitative: if the loss varies smoothly with the target parameter, then the gap to the oracle value is controlled by the size of the ambiguity set. Thus, whenever the ambiguity diameter vanishes, the robust problem approaches the full-information oracle problem.

\begin{remark}
\label{rem:oracle-convergence-compactness}
A stronger oracle-convergence statement can be obtained without Lipschitz continuity, under additional regularity assumptions. Let \(\{\mathcal A_k\}_{k\ge1}\) be a nested sequence of compact ambiguity sets with \(\bigcap_{k\ge1}\mathcal A_k=\{\xi^\circ\}\), let the admissible class of maps be compact, and let \(\mathsf F_\iota\) be jointly continuous in \((\lintau,\xi)\). Define
\[
  G_k(\lintau)
  :=
  \sup_{\xi\in\mathcal A_k}\tfrac{1}{|\mathcal I|}
    \sum_{\iota\in\mathcal I}\mathsf F_\iota(\lintau,\xi),
  \qquad
  G_\infty(\lintau)
  :=
  \tfrac{1}{|\mathcal I|}
    \sum_{\iota\in\mathcal I}\mathsf F_\iota(\lintau,\xi^\circ).
\]
For each fixed \(k\), \(G_k\) is continuous in \(\lintau\) because it maximises a jointly continuous function of \((\lintau,\xi)\) over the compact set \(\mathcal A_k\); this is an instance of Berge's maximum theorem \citep[Thm.~17.31]{aliprantis2006infinite}. Since the \(\mathcal A_k\) are nested and shrink to \(\{\xi^\circ\}\), continuity in \(\xi\) implies \(G_k(\lintau)\downarrow G_\infty(\lintau)\) pointwise for every \(\lintau\), and \(G_\infty\) is continuous in \(\lintau\). Because the admissible class of maps is compact, Dini's theorem \citep[Thm.~7.13]{rudin1976principles} upgrades this monotone pointwise convergence to uniform convergence. Hence
\[
  \mathfrak R^\star(\mathcal A_k)
  =\inf_{\lintau}G_k(\lintau)
  \longrightarrow
  \inf_{\lintau}G_\infty(\lintau)
  =\mathfrak R^\star(\{\xi^\circ\}).
\]
\end{remark}

\section{Proofs of the robustness certificates}\label{app:certificate-proofs} This section contains the full analytical proofs of the Gaussian and empirical robustness certificates stated in Section~\ref{sec:provable-robustness}. In both cases, the proof strategy is to decompose the discrepancy into transport, mechanism, and environment terms, and to control each term separately.
\vspace{3mm}

\begin{namedproof}{Theorem~\ref{thm:q-restricted-gaussian-certificate}} Fix \((\iota,O_\iota)\in\Qcal\) and write \(S:=S_{O_\iota}\). Our goal is to bound the Gaussian query-restricted loss
\begin{align}
  \mathcal W_2^2\!\Bigl(
    \pi_{O_\iota\#}(\lintau_{\#}P_s^{(\iota)}),\;
    \pi_{O_\iota\#}P_t^{(\iota)}(\Delta W,\mu_t,\Sigma_t)
  \Bigr).
\end{align}
We first introduce the intermediate Gaussian measure
\begin{align}
  \widetilde\nu_{\iota,O_\iota}(\Delta W)
  :=
  \mathcal N\!\Bigl(
    SA'_\iota(\Delta W)\mu_s,\;
    SA'_\iota(\Delta W)\Sigma_sA'_\iota(\Delta W)^\top S^\top
  \Bigr).
\end{align}
This is exactly the projected target law one would obtain if the mechanism had already changed from \(A_\iota\) to \(A'_\iota(\Delta W)\), but the environment were still the source one. We now compare the true source and target laws by passing through this intermediate Gaussian. By the triangle inequality in \(\mathcal W_2\),
\begin{align}
  \mathcal W_2\bigl(
    \pi_{O_\iota\#}(\lintau_{\#}P_s^{(\iota)}),\;
    \pi_{O_\iota\#}P_t^{(\iota)}
  \bigr)
  \le
  \Xi_{1,\iota,O_\iota}+\Xi_{2,\iota,O_\iota},
\end{align}
where
\begin{align}
  \Xi_{1,\iota,O_\iota}
  :=
  \mathcal W_2\bigl(
    \pi_{O_\iota\#}(\lintau_{\#}P_s^{(\iota)}),\;
    \widetilde\nu_{\iota,O_\iota}(\Delta W)
  \bigr),
\end{align}
\begin{align}
  \Xi_{2,\iota,O_\iota}
  :=
  \mathcal W_2\bigl(
    \widetilde\nu_{\iota,O_\iota}(\Delta W),\;
    \pi_{O_\iota\#}P_t^{(\iota)}(\Delta W,\mu_t,\Sigma_t)
  \bigr).
\end{align}
Hence
\begin{align}
  \mathcal W_2^2\bigl(
    \pi_{O_\iota\#}(\lintau_{\#}P_s^{(\iota)}),\;
    \pi_{O_\iota\#}P_t^{(\iota)}
  \bigr)
  \le
  2\Xi_{1,\iota,O_\iota}^2+2\Xi_{2,\iota,O_\iota}^2.
\end{align}

We first bound \(\Xi_{1,\iota,O_\iota}\). Since both measures are linear pushforwards of the same Gaussian \(U\sim\mathcal N(\mu_s,\Sigma_s)\), we couple them by taking
\begin{align}
  X:=S\lintau A_\iota U,
  \qquad
  Y:=SA'_\iota(\Delta W)U.
\end{align}
Then \(X\) and \(Y\) have the required marginals, so \((X,Y)\) is a coupling of \(\pi_{O_\iota\#}(\lintau_{\#}P_s^{(\iota)})\) and \(\widetilde\nu_{\iota,O_\iota}(\Delta W)\). By the coupling characterisation of \(\mathcal W_2\),
\begin{align}
  \mathcal W_2^2(\mu,\nu)\le \mathbb E\|X-Y\|_2^2,
\end{align}
hence
\begin{align}
  \Xi_{1,\iota,O_\iota}^2
  \le
  \mathbb E\Bigl\|
    S\lintau A_\iota U-SA'_\iota(\Delta W)U
  \Bigr\|_2^2.
\end{align}
Now factor the difference:
\begin{align}
  S\lintau A_\iota U-SA'_\iota(\Delta W)U
  =
  S\bigl(\lintau A_\iota-A'_\iota(\Delta W)\bigr)U.
\end{align}
The key identity is
\begin{align}
  \lintau A_\iota-A'_\iota(\Delta W)
  =
  (\lintau-I)A_\iota + (A_\iota-A'_\iota(\Delta W)).
\end{align}
This separates the error into a \emph{transport mismatch} term \((\lintau-I)A_\iota\), and a \emph{mechanism perturbation} term \(A_\iota-A'_\iota(\Delta W)\). Substituting this and using \((x+y)^2\le 2x^2+2y^2\), we obtain
\begin{align*}
  \Xi_{1,\iota,O_\iota}^2
  \le\
  2\,\mathbb E\|S(\lintau-I)A_\iota U\|_2^2 +
  2\,\mathbb E\|S(A_\iota-A'_\iota(\Delta W))U\|_2^2.
\end{align*}
For a Gaussian vector \(U\sim\mathcal N(\mu_s,\Sigma_s)\) and any matrix \(M\), writing \(U=\mu_s+(U-\mu_s)\) and using \(\mathbb E[U-\mu_s]=0\), we get $ \mathbb E\|MU\|_2^2 = \|M\mu_s\|_2^2+\mathbb E\|M(U-\mu_s)\|_2^2 = \|M\mu_s\|_2^2+\Tr(M\Sigma_sM^\top)$. Applying this identity with \(M=S(\lintau-I)A_\iota\) and \(M=S(A_\iota-A'_\iota(\Delta W))\) yields
\begin{align*}
  \Xi_{1,\iota,O_\iota}^2
  \le\;&
  2\|S(\lintau-I)A_\iota\mu_s\|_2^2
  +2\Tr\!\Bigl(
    S(\lintau-I)A_\iota\Sigma_sA_\iota^\top(\lintau-I)^\top S^\top
  \Bigr) \\
  &\quad+
  2\|S(A_\iota-A'_\iota(\Delta W))\mu_s\|_2^2 \\
  &\quad+
  2\Tr\!\Bigl(
    S(A_\iota-A'_\iota(\Delta W))
    \Sigma_s
    (A_\iota-A'_\iota(\Delta W))^\top S^\top
  \Bigr).
\end{align*}
Since \(\Sigma_s\) is symmetric positive semidefinite, we have \(\Sigma_s\preceq \|\Sigma_s\|_2 I\). Hence $\Tr(M\Sigma_s M^\top)=\Tr(\Sigma_s M^\top M)\le \|\Sigma_s\|_2\,\Tr(M^\top M)=\|\Sigma_s\|_2\,\|M\|_F^2$. Thus, 
\begin{align*}
  \Xi_{1,\iota,O_\iota}^2
  \le\;&
  2\|S(\lintau-I)A_\iota\mu_s\|_2^2
  +2\|\Sigma_s\|_2\,\|S(\lintau-I)A_\iota\|_F^2 \\
  &\quad+
  2\|S(A_\iota-A'_\iota(\Delta W))\|_2^2\|\mu_s\|_2^2 \\
  &\quad+
  2\|\Sigma_s\|_2\,\|S(A_\iota-A'_\iota(\Delta W))\|_F^2.
\end{align*}

Now use the propagator perturbation control. Since $\|A'_\iota(\Delta W)-A_\iota\|_2\le \alpha_\iota$, and \(S\) is a coordinate selector; i.e. $\|S(A_\iota-A'_\iota(\Delta W))\|_2\le \alpha_\iota$. Also, an \(|O_\iota|\times d\) matrix has Frobenius norm bounded by \(\sqrt{|O_\iota|}\) times its spectral norm, so
\begin{align}
  \|S(A_\iota-A'_\iota(\Delta W))\|_F^2
  \le |O_\iota|\,\alpha_\iota^2.
\end{align}
Substituting these bounds gives
\begin{align}
  \Xi_{1,\iota,O_\iota}^2
  \le
  2\|S(\lintau-I)A_\iota\mu_s\|_2^2
  +2\|\Sigma_s\|_2\,\|S(\lintau-I)A_\iota\|_F^2
  +2\alpha_\iota^2\|\mu_s\|_2^2
  +2|O_\iota|\,\|\Sigma_s\|_2\,\alpha_\iota^2.
\end{align}
We next bound \(\Xi_{2,\iota,O_\iota}\). Here the linear map \(SA'_\iota(\Delta W)\) is the same in both distributions; only the input Gaussian changes from \(\mathcal N(\mu_s,\Sigma_s)\) to \(\mathcal N(\mu_t,\Sigma_t)\). More generally, for a fixed linear map \(L\) its pushforward is \(\|L\|_2\)-Lipschitz in \(\mathcal W_2\): if \(\pi\in\Pi(\mu,\nu)\), then its image under \((x,y)\mapsto(Lx,Ly)\) is a coupling of \(L_\#\mu\) and \(L_\#\nu\), and \(\|Lx-Ly\|_2\le\|L\|_2\|x-y\|_2\) yields, after taking the infimum over \(\pi\),
\begin{align}
  \mathcal W_2(L_\#\mu,L_\#\nu)
  \le
  \|L\|_2\,\mathcal W_2(\mu,\nu).
\end{align}
Thus,
\begin{align}
  \Xi_{2,\iota,O_\iota}
  \le
  \|SA'_\iota(\Delta W)\|_2\,
  \mathcal W_2\bigl(
    \mathcal N(\mu_s,\Sigma_s),\,
    \mathcal N(\mu_t,\Sigma_t)
  \bigr).
\end{align}
Since \((\mu_t,\Sigma_t)\in\mathcal A_\rho(\varepsilon)\),
\begin{align}
  \mathcal W_2\bigl(
    \mathcal N(\mu_s,\Sigma_s),\,
    \mathcal N(\mu_t,\Sigma_t)
  \bigr)\le \varepsilon.
\end{align}
Moreover,
\begin{align}
  SA'_\iota(\Delta W)=SA_\iota+S(A'_\iota(\Delta W)-A_\iota),
\end{align}
so by the triangle inequality,
\begin{align}
  \|SA'_\iota(\Delta W)\|_2
  \le
  \|SA_\iota\|_2+\|S(A'_\iota(\Delta W)-A_\iota)\|_2
  \le
  \|SA_\iota\|_2+\alpha_\iota.
\end{align}
Hence,
\begin{align}
  \Xi_{2,\iota,O_\iota}
  \le
  (\|SA_\iota\|_2+\alpha_\iota)\varepsilon.
\end{align}

Finally, combine the bounds on \(\Xi_{1,\iota,O_\iota}\) and \(\Xi_{2,\iota,O_\iota}\) in
\begin{align}
  \mathcal W_2^2
  \le 2\Xi_1^2+2\Xi_2^2.
\end{align}
This yields the pointwise estimate
\begin{align}
  \sup_{\substack{\Delta W\in\mathcal A_W\\
                  (\mu_t,\Sigma_t)\in\mathcal A_\rho(\varepsilon)}}
  \mathcal W_2^2\!\Bigl(
    \pi_{O_\iota\#}(\lintau_{\#}P_s^{(\iota)}),\;
    \pi_{O_\iota\#}P_t^{(\iota)}
  \Bigr)
  \le
  \delta_{\iota,O_\iota}^{\rho}(\lintau)^2.
\end{align}
Averaging over \(\Qcal\) gives \eqref{eq:q-restricted-gaussian-certificate}.
\end{namedproof}
\vspace{3mm}

\begin{namedproof}{Theorem~\ref{thm:q-restricted-empirical-certificate}} Fix $(\iota,O_\iota)\in\Qcal$ and write \(S:=S_{O_\iota}\). The projected empirical residual can be decomposed as
\begin{align}
  S\bigl(
    \lintau A_\iota U^s
    -
    A'_\iota(\Delta W)(U^s+\Theta)
  \bigr)
  =
  S(\lintau-I)A_\iota U^s
  +
  S(A_\iota-A'_\iota(\Delta W))U^s
  -
  SA'_\iota(\Delta W)\Theta.
\end{align}
These are exactly the three contributions we need to control: transport-map error, mechanism error, and environment perturbation. Taking Frobenius norms and using the triangle inequality gives
\begin{align*}
  \bigl\|
    S\bigl(
      \lintau A_\iota U^s
      -
      A'_\iota(\Delta W)(U^s+\Theta)
    \bigr)
  \bigr\|_F
  &\le
  \|S(\lintau-I)A_\iota U^s\|_F \\
  &\quad+
  \|S(A_\iota-A'_\iota(\Delta W))\|_2\,\|U^s\|_F \\
  &\quad+
  \|SA'_\iota(\Delta W)\|_2\,\|\Theta\|_F.
\end{align*}
Now use the uniform bounds
\begin{align}
  \|S(A_\iota-A'_\iota(\Delta W))\|_2\le \alpha_\iota,
\qquad
  \|SA'_\iota(\Delta W)\|_2
  \le
  \|SA_\iota\|_2+\alpha_\iota,
\qquad
  \|\Theta\|_F\le \varepsilon\sqrt N.
\end{align}
Substituting these into the previous display yields
\begin{align}
  \bigl\|
    S\bigl(
      \lintau A_\iota U^s
      -
      A'_\iota(\Delta W)(U^s+\Theta)
    \bigr)
  \bigr\|_F
  \le
  \delta_{\iota,O_\iota}^{U}(\lintau).
\end{align}
This is a pointwise bound valid for every admissible \((\Delta W,\Theta)\). Squaring, averaging over \(\Qcal\), and then taking the supremum proves \eqref{eq:q-restricted-empirical-certificate}.
\end{namedproof}

\subsection{Directional certificates for linear queries}
\label{app:directional-certificates}

This subsection proves Corollary~\ref{cor:directional-linear-certificate} and shows how the directional mechanism modulus $m_{\iota,q}(v)$ is computed. Throughout, fix $(\iota,O_\iota)\in\Qcal$, write $q:=S_{O_\iota}^\top c$ for the query direction, and abbreviate the perturbed propagator $A_\iota'(\Delta W)$ to $A_\iota'$. The input signal is $v=\mu_s^{(\iota)}$ in the Gaussian case and $v=\bar U^{s,(\iota)}:=\tfrac1N U^{s,(\iota)}\mathbf 1$ in the empirical one. Since $\iota$ is fixed, we drop the superscript $(\iota)$ and write $\bar U^s$ and $\mu_s$.

\paragraph{Proof of the directional certificate.}
Take the empirical case and write $\bar\Theta:=\tfrac1N\Theta\mathbf 1$ for the mean environment perturbation. The transported-source and target query values are $Q_{\tau \# s}^{(\iota,O_\iota)}=q^\top\lintau A_\iota\bar U^s$ and $Q_t^{(\iota,O_\iota)}=q^\top A_\iota'(\bar U^s+\bar\Theta)$. These differ in two ways: the mechanism and the environment, so we compare them using an intermediate quantity carrying the target mechanism but the source environment. Inserting $q^\top A_\iota'\bar U^s$ splits the discrepancy into three contributions:
\begin{align}
Q_t^{(\iota,O_\iota)}-Q_{\tau \# s}^{(\iota,O_\iota)}
=
\underbrace{q^\top(I-\lintau)A_\iota\bar U^s}_{\text{transport}}
+\underbrace{q^\top(A_\iota'-A_\iota)\bar U^s}_{\text{mechanism}}
+\underbrace{q^\top A_\iota'\bar\Theta}_{\text{environment}}.
\end{align}
The identity is exact; the triangle inequality bounds the error by the sum of the three absolute values. The transport term is computable directly since $\lintau$, $A_\iota$, $q$, and $\bar U^s$ are all known. The mechanism term is bounded by $m_{\iota,q}(\bar U^s)$ by Definition~\ref{def:directional-mechanism-modulus}, and the rest of this subsection is devoted to computing it.

For the environment term, recall that the perturbation is confined to the shifted coordinates, $\Theta=M_{\mathcal K}\Theta$, hence $\bar\Theta=M_{\mathcal K}\bar\Theta$. The inner product is therefore unchanged if we mask the other factor as well, which discards the coordinates that the environment cannot reach:
\begin{align}
q^\top A_\iota'\bar\Theta
=\bigl(M_{\mathcal K}A_\iota'{}^\top q\bigr)^\top\bar\Theta .
\end{align}
The size of $\bar\Theta$ follows from the ambiguity set: since $\|\Theta\|_F\le\varepsilon\sqrt N$ and $\|\mathbf 1\|_2=\sqrt N$,
\begin{align}
\|\bar\Theta\|_2\le\tfrac1N\|\Theta\|_F\|\mathbf 1\|_2
=\tfrac1{\sqrt N}\|\Theta\|_F\le\varepsilon,
\end{align}
so the $\sqrt N$ in the definition of $\mathcal A_U(\varepsilon)$ cancels, and the mean perturbation has a size of at most $\varepsilon$ regardless of sample size. Cauchy-Schwarz then gives $|q^\top A_\iota'\bar\Theta|\le\varepsilon\|M_{\mathcal K}A_\iota'{}^\top q\|_2$.

The Gaussian case is identical with $\bar U^s$ replaced by $\mu_s$ and $\bar\Theta$ by $h:=\mu_t-\mu_s$: the Wasserstein constraint gives $\|h\|_2\le\varepsilon$, the mean pinning gives $h=M_{\mathcal K}h$, and the same step applies. No covariance term appears because a mean query depends on the target law only through its mean. Taking the supremum over $\Delta W\in\mathcal A_W$ yields \eqref{eq:directional-certificate}.

\paragraph{The mechanism term reduces to first order.}
The remaining task is to control $m_{\iota,q}(v)$, the worst change that the perturbed mechanism can induce in the query direction. The obstacle is that $\Delta W$ sits inside a matrix inverse. The finite perturbation expansion
\eqref{eq:Aiota-difference-finite}:
\begin{align}
A_\iota'(\Delta W)-A_\iota
=
\sum_{r=1}^{d-1}(A_\iota R_\iota\Delta W)^r A_\iota,
\label{eq:directional-resolvent-expansion}
\end{align}
terminates because $A_\iota R_\iota\Delta W$ is strictly lower triangular under the fixed acyclic ordering, hence nilpotent. For the geometries used here, the sum collapses to a single term. Suppose the budget lies on the incoming edges of a single shifted node $j$. Only row $j$ of $W$ can move, so $\Delta W=e_j w^\top$, where $w$ collects that row's entries and is supported on $\operatorname{PA}(X_j)$. Hence, $A_\iota R_\iota\Delta W = u\,w^\top$ with $u:=A_\iota R_\iota e_j$: an outer product, and therefore rank one. For any such matrix, $(u w^\top)^2 = (w^\top u)\,u w^\top$, so it squares to zero exactly when $w^\top u=0$. Here
\begin{align}
w^\top u=\textstyle\sum_k w_k\,(A_\iota)_{kj}(R_\iota)_{jj}=0,
\end{align}
because $w_k$ is nonzero only for parents of $j$, which precede $j$ in the ordering, while $(A_\iota)_{kj}=0$ for $k<j$ by lower triangularity. Structurally: the perturbed edges all point into $j$, whereas $j$ propagates only forward, so the perturbation can never feed back into itself. Every term with $r\ge2$ therefore vanishes.

\paragraph{The first-order term and what it means.}
Projecting the surviving term onto $q$ and $v$, and grouping the factors on either side of $\Delta W$,
\begin{align}
q^\top A_\iota R_\iota\Delta W A_\iota v=a^\top\Delta W\,b,
\qquad
a:=R_\iota A_\iota^\top q,
\quad
b:=A_\iota v,
\label{eq:first-order-bilinear}
\end{align}
using $R_\iota^\top=R_\iota$, since $R_\iota$ is diagonal. Here $a$ and $b$ are ordinary vectors, fixed by the model and the query, and independent of $\Delta W$, so they can be computed before the supremum over $\mathcal A_W$ is taken. They form a forward and a backward pass: $b$ propagates the source signal forward to the point where the mechanism acts, and $a$ propagates the query readout backward to the same point. Thus, $b_n$ is the value arriving at node $n$, and $a_m$ is the total derivative of the query with respect to a perturbation at node $m$. Their outer product $ab^\top$, with entries $a_mb_n$, is the \emph{sensitivity matrix}: entry $(m,n)$ is the derivative of the query with respect to the coefficient $\Delta W_{mn}$; i.e., how much the query moves per unit change in the coefficient $\Delta W_{mn}$. Under the rank-one structure above, we therefore have $m_{\iota,q}(v)=\beta_{\iota,q}(v)$ exactly, where
\begin{align}
\beta_{\iota,q}(v):=\sup_{\Delta W\in\mathcal A_W}\bigl|a^\top\Delta W\,b\bigr| .
\end{align}
The general case, where the budget spans several shifted nodes and higher-order terms survive, is treated at the end of this subsection.

\paragraph{Computing $\beta_{\iota,q}(v)$.}
Written out entrywise, the first-order term is
\begin{align}
a^\top\Delta W\,b=\sum_{m,n}\Delta W_{mn}\,a_mb_n,
\end{align}
a sum of products; that is, the Frobenius inner product $\langle\Delta W,\,ab^\top\rangle_F$. So $\beta_{\iota,q}(v)$ is the maximum of a \emph{linear} functional over $\mathcal A_W$, which is, by definition, the
support function of $\mathcal A_W$ at the sensitivity matrix $ab^\top$. This makes the computation a budgeting problem. Each coefficient $\Delta W_{mn}$ pays $|a_mb_n|$ per unit spent on it, and $\mathcal A_W$ specifies how much may be spent and where. The optimum spends the budget where the payoff is largest; the four geometries differ only in what spending is permitted, and each rule has a closed-form answer. Write $P\in\{0,1\}^{d\times d}$ for the support of $\mathcal A_W$, the entries of $\Delta W$ permitted to vary. Every admissible $\Delta W$ vanishes off $P$, so only $P\odot(ab^\top)$ can contribute, and for the four geometries of Section~\ref{subsec:mechanism-ambiguity},
\begin{align}
\mathcal A_W^{\mathrm F}(\eta):\;
\beta_{\iota,q}(v)\le\eta\|P\odot(ab^\top)\|_F,~~\mathcal A_W^{\mathrm{box}}(\delta,B):\; \beta_{\iota,q}(v)\le\textstyle\sum_{m,n}P_{mn}\bigl(|\delta_{mn}|+B_{mn}\bigr)|a_m||b_n|,
\nonumber\\
\mathcal A_W^{\mathrm{row}}(\rho):\;
\beta_{\iota,q}(v)\le\textstyle\sum_m\rho_m|a_m|\,\|P_{m,:}\odot b\|_\infty,~~\mathcal A_W^{\mathrm{col}}(c):\;
\beta_{\iota,q}(v)\le\textstyle\sum_n c_n|b_n|\,\|P_{:,n}\odot a\|_\infty.
\end{align}
Each line is the optimal spending rule for its budget. The Frobenius ball allows one shared allowance measured in Euclidean length, so the entries compete, and the best $\Delta W$ points along $ab^\top$; Cauchy-Schwarz gives the value. The box gives every entry its own allowance, independent of the others, so each can be pushed to its limit with the sign chosen to make its contribution positive. The row and column budgets are $\ell_1$ within each row or column, and an $\ell_1$ allowance is always best spent entirely on the single largest entry, hence the $\ell_1$-$\ell_\infty$ pairing. For a shifted box, an entry ranges over $[\delta_{mn}-B_{mn},\,\delta_{mn}+B_{mn}]$, whose largest magnitude is the \emph{effective half-width} $|\delta_{mn}|+B_{mn}$; at $\delta=0$, this is $B_{mn}$ and recovers the symmetric bound. It is slightly conservative relative to the exact shifted support function $\delta_{mn}a_mb_n+B_{mn}|a_mb_n|$, which keeps the sign of the center and is smaller whenever the declared offset points against the query direction. We use the effective half-width because it is sign-agnostic and needs no assumption relating $\delta$ to $q$.

Because the objective is linear in the free coefficients, its maximum over the box is attained at a vertex, and the expressions above give that maximum in closed form. The certificates we report are therefore exact rather than the output of a numerical search. We verified the rank-one identity numerically across all benchmarks, interventions, and admissible $\Delta W$, with residual at machine precision.

\paragraph{The environment supremum.}
The environment term of \eqref{eq:directional-certificate} carries its own supremum, $\sup_{\Delta W\in\mathcal A_W}\|M_{\mathcal K}A_\iota'(\Delta W)^\top q\|_2$, because the perturbed propagator also determines how an environment shift reaches the query. Splitting $A_\iota'=A_\iota+(A_\iota'-A_\iota)$, applying the triangle inequality, and using that $M_{\mathcal K}$ is a projection together with $\|A_\iota'-A_\iota\|_2\le\alpha_\iota$ gives
\begin{align}
\sup_{\Delta W\in\mathcal A_W}
\bigl\|M_{\mathcal K}A_\iota'(\Delta W)^\top q\bigr\|_2
\;\le\;
\bigl\|M_{\mathcal K}A_\iota^\top q\bigr\|_2+\alpha_\iota\|q\|_2 .
\label{eq:directional-env-bound}
\end{align}
When the mechanism set is trivial (as on the Portland benchmark, where the shifted node is a root and $\mathcal A_W=\{0\}$), there is a single admissible $\Delta W$: $\alpha_\iota=0$. The mechanism terms drop out of the certificate altogether, and the supremum reduces to the single evaluation $\|M_{\mathcal K}A_\iota^\top q\|_2$. The directional certificate is then carried entirely by the environment term, which is why its tightness there comes from the direction-specificity of that term alone.

\paragraph{The general case.}
The rank-one argument required the budget to sit on the incoming edges of a single node. When it spans the mechanisms of several shifted nodes, $\Delta W$ has more than one nonzero row, $A_\iota R_\iota\Delta W$ is no longer an outer product, and the higher-order terms of \eqref{eq:directional-resolvent-expansion} survive. These terms describe the perturbation compounding on itself: a change at one shifted mechanism now propagates into another, which propagates further, and so on. The $r$-th such term is $q^\top(A_\iota R_\iota\Delta W)^r A_\iota v$. Peeling off one factor from each end,
\begin{align}
\bigl|q^\top(A_\iota R_\iota\Delta W)^r A_\iota v\bigr|
\;\le\;
\underbrace{\|q^\top A_\iota R_\iota\Delta W\|_2}_{\le\,\lambda_{\iota,q}}
\;\underbrace{\|A_\iota R_\iota\Delta W\|_2^{\,r-1}}_{\le\,\gamma_\iota^{\,r-1}}
\;\underbrace{\|A_\iota v\|_2}_{\text{known}},
\end{align}
where $\gamma_\iota$ is the amplification factor of \eqref{eq:def-gamma-iota} and
\begin{align}
\lambda_{\iota,q}:=\sup_{\Delta W\in\mathcal A_W}\|a^\top\Delta W\|_2
\end{align}
is the query-side modulus: the same object as $\beta_{\iota,q}(v)$ but with the input vector $b$ removed, so it measures the worst effect on the query direction irrespective of the signal arriving. Summing over $r\ge2$ gives
\begin{align}
m_{\iota,q}(v)
\le
\underbrace{\beta_{\iota,q}(v)}_{\text{first order}}
+\underbrace{\lambda_{\iota,q}\|A_\iota v\|_2\textstyle\sum_{r=1}^{d-2}\gamma_\iota^r}_{\text{higher-order remainder}}.
\label{eq:directional-remainder}
\end{align}
The sum is finite by nilpotency; when $\gamma_\iota<1$ one may replace it with the geometric bound $\gamma_\iota/(1-\gamma_\iota)$. This step is conservative: it bounds each factor by its own worst case, allowing a different $\Delta W$ for each, whereas the true supremum is over a single $\Delta W$, making all of them large at once. No adversary can do that, so the bound overestimates the true supremum, but it is finite and computable, and the first-order term dominates whenever the perturbation is small.

The modulus $\lambda_{\iota,q}$ is also a support function of the same set at the vector $a$, rather than the matrix $ab^\top$, and admits bounds parallel to those mentioned above:
\begin{align}
\text{Frobenius:}\;\lambda_{\iota,q}\le\eta\|a\|_2,
\quad
\text{box:}\;\lambda_{\iota,q}\le\Bigl[\textstyle\sum_n\bigl(\sum_m P_{mn}(|\delta_{mn}|+B_{mn})|a_m|\bigr)^2\Bigr]^{1/2},
& \nonumber\\
\text{row:}\;\lambda_{\iota,q}\le\textstyle\sum_m\rho_m|a_m|,
\quad
\text{col.:}\;\lambda_{\iota,q}\le\Bigl[\textstyle\sum_n c_n^2\|P_{:,n}\odot a\|_\infty^2\Bigr]^{1/2}.
\end{align}
Finally, even without the rank-one structure, the map $\Delta W\mapsto q^\top(A'_\iota(\Delta W)-A_\iota)v$ remains a polynomial of degree at most $d-1$ in the free coefficients since the expansion terminates. With one or two free coefficients, its supremum can therefore still be found exactly by enumerating the endpoints and the critical points of a low-degree polynomial.

\emph{Applicability to the semi-Markovian case.}
The certificate depends only on $q$, $\lintau$, and $A_\iota,A_\iota'$. The argument therefore applies whether $\lintau$ is diagonal (Markovian) or block-diagonal (semi-Markovian).

\section{Formulation-specific optimisation details}
\label{app:optimization-details}
The two algorithms below instantiate the unified backbone of Algorithm~\ref{alg:traca-unified}.

\subsection{Gaussian specialisation} 
\label{app:gaussian-optimization-details}
We first record the Gaussian specialisation, where the adversarial environment is parameterised by target moments, and the ascent step uses the smooth surrogate from Section~\ref{subsec:gaussian-implementation}.
\begin{algorithm}[h]
\caption{Gaussian TraCA}
\label{alg:traca-gaussian-adversarial}
\begin{algorithmic}[1]
\STATE Initialise
$\lintau^{(0)}$, $\Delta W^{(0)}=0$, $\mu_t^{(0)}=\mu_s$,
$\Sigma_t^{(0)}=\Sigma_s$.
\REPEAT
\STATE // \textit{Projected gradient descent}
  \FOR{$k_{\tau}$ steps}
    \STATE $\lintau^{(t+1)} \leftarrow \lintau^{(t)} - \mathrm{lr}_{\lintau}\,
    \nabla_{\lintau}
    F\bigl(\lintau^{(t)},\Delta W^{(t)},\mu_t^{(t)},\Sigma_t^{(t)}\bigr)$
    \STATE $\lintau^{(t+1)} \leftarrow \proj_{\mathrm{const}}\!\bigl(\lintau^{(t+1)}\bigr)$
    //\textit{projection to constructive class}
  \ENDFOR
  \STATE // \textit{Projected proximal-gradient ascent}
  \FOR{$k_{\mathrm{adv}}$ steps } 
    \STATE $\Delta W^{(\frac{t+1}{2})} \leftarrow \Delta W^{(t)} + \mathrm{lr}_{\mathrm{adv}}\,
    \nabla_{\Delta W}
    \widetilde F\bigl(\lintau^{(t+1)},\Delta W^{(t)},\mu_t^{(t)},\Sigma_t^{(t)}\bigr)$
    \STATE $\Delta W^{(t+1)} \leftarrow \proj_{\mathcal A_W}\!\bigl(M_{\mathcal K}\Delta W^{(\frac{t+1}{2})}\bigr)$
    //\textit{mask + projection to the ambiguity set}
    \STATE $\mu_t^{(t+1)} \leftarrow \mu_t^{(t)} + \mathrm{lr}_{\mathrm{adv}}\,
    \nabla_{\mu_t}
    \widetilde F\bigl(\lintau^{(t+1)},\Delta W^{(t+1)},\mu_t^{(t)},\Sigma_t^{(t)}\bigr)$
    \STATE $\Sigma_t^{(\frac{t+1}{2})} \leftarrow \Sigma_t^{(t)} + \mathrm{lr}_{\mathrm{adv}}\,
    \nabla_{\Sigma_t}
    \widetilde F\bigl(\lintau^{(t+1)},\Delta W^{(t+1)},\mu_t^{(t)},\Sigma_t^{(t)}\bigr)$
    \STATE $\Sigma_t^{(t+1)} \leftarrow
\mathrm{CovProx}_{\lambda_t}\!\Bigl(
  \Sigma_t^{(\frac{t+1}{2})};\Delta W^{(t+1)}
\Bigr)$ //\textit{proximal-gradient ascent on covariance}
    \STATE $(\mu_t^{(t+1)},\Sigma_t^{(t+1)}) \leftarrow
    \proj_{\mathcal A_\rho(\varepsilon)}
    \Bigl(
      \Pi_{\mathrm{inv}}\bigl(\mu_t^{(t+1)},\Sigma_t^{(t+1)}\bigr)
    \Bigr)$
    //\textit{invariance + projection to ball}
  \ENDFOR
\UNTIL{convergence}
\end{algorithmic}
\end{algorithm}

\subsection{Empirical specialisation}
\label{app:empirical-optimization-details}
We then record the empirical specialisation, where the adversarial environment is represented directly by sample perturbations, and both adversarial blocks are updated by projected ascent.
\begin{algorithm}[h]
\caption{Empirical TraCA}
\label{alg:traca-empirical-adversarial}
\begin{algorithmic}[1]
\STATE Initialise
$\lintau^{(0)}$, $\Delta W^{(0)}=0$, $\Theta^{(0)}=0$.
\REPEAT
\STATE // \textit{Projected gradient descent}
  \FOR{$k_{\tau}$ steps}
    \STATE $\lintau^{(t+1)} \leftarrow \lintau^{(t)} - \mathrm{lr}_{\lintau}\,
    \nabla_{\lintau}
    F\bigl(\lintau^{(t)},\Delta W^{(t)},\Theta^{(t)}\bigr)$
    \STATE $\lintau^{(t+1)} \leftarrow \proj_{\mathrm{const}}\!\bigl(\lintau^{(t+1)}\bigr)$
    //\textit{projection to constructive class}
  \ENDFOR
  \STATE // \textit{Projected gradient ascent}
  \FOR{$k_{\mathrm{adv}}$ steps}
    \STATE $\Delta W^{(\frac{t+1}{2})} \leftarrow \Delta W^{(t)} + \mathrm{lr}_{\mathrm{adv}}\,
    \nabla_{\Delta W}
    F\bigl(\lintau^{(t+1)},\Delta W^{(t)},\Theta^{(t)}\bigr)$
    \STATE $\Delta W^{(t+1)} \leftarrow \proj_{\mathcal A_W}\!\bigl(M_{\mathcal K}\Delta W^{(\frac{t+1}{2})}\bigr)$
    //\textit{mask + projection to ambiguity set}
    \STATE $\Theta^{(\frac{t+1}{2})} \leftarrow \Theta^{(t)} + \mathrm{lr}_{\mathrm{adv}}\,
    \nabla_{\Theta}
    F\bigl(\lintau^{(t+1)},\Delta W^{(t+1)},\Theta^{(t)}\bigr)$
    \STATE $\Theta^{(t+1)} \leftarrow \proj_{\mathcal A_U(\varepsilon)}\!\bigl(M_{\mathcal K}\Theta^{(\frac{t+1}{2})}\bigr)$
    //\textit{mask + projection to ambiguity set}
  \ENDFOR
\UNTIL{convergence}
\end{algorithmic}
\end{algorithm}

\section{Benchmark datasets}
\label{app:benchmarks}

This appendix presents the structural causal models and intervention sets of the three synthetic benchmarks of Section~\ref{sec:experiments}, together with the construction of the Portland real-world benchmark. Throughout, $W$ is the strictly lower triangular structural matrix in topological order, with $W_{ij}$ as the weight on the edge $j\to i$; equivalently, row $i$ holds the incoming edges of node $i$. Further, Table~\ref{tab:benchmark-configs} summarises the configuration choices across all four benchmarks.

\subsection{ATE}
\label{app:ate}
ATE is a two-variable semi-Markovian graph $X\to Y$ with a bidirected edge $X\leftrightarrow Y$ (latent confounding) and a selection node on $Y$:
\begin{equation}
  X = U_X,
  \qquad
  Y = \beta X + U_Y,
  \qquad
  \mathrm{Cov}(U_X,U_Y)=\rho\neq 0 ,
\end{equation}
with source coefficient $\beta_s=1.0$, $\rho=0.6$, exogenous means $\mu_{U_X}=0.5$ and $\mu_{U_Y}=0.3$, and unit exogenous standard deviations $\sigma_{U_X}=\sigma_{U_Y}=1.0$. The confounded pair $\{X,Y\}$ forms a single district, so the constructive map acts as a free $2\times2$ block on that district rather than as a diagonal map. The designated
shifted node is $Y$, whose mechanism is the incoming coefficient $\beta=w_{X\to Y}$ together with its exogenous noise. The target differs from the source in this coefficient alone, with $\beta_t=1.5$; all other parameters are shared. The two reported queries are therefore $\mathbb E[Y\mid do(X{=}0)]=\mu_{U_Y}=0.30$, identical in both domains since the shifted coefficient multiplies zero, and $\mathbb E[Y\mid do(X{=}1)]=\beta_t+\mu_{U_Y}=1.80$ on the target against $1.30$ on the source. The intervention set is:
\begin{equation}
  \mathcal I_{\mathrm{ATE}}
  =
  \bigl\{\emptyset,\ \mathrm{do}(X{=}0),\ \mathrm{do}(X{=}1)\bigr\}.
\end{equation}

\subsection{ATCE}
\label{app:atce}
ATCE is a three-variable confounded-treatment graph with $Z$ an age-like covariate, $X$ a treatment, and $Y$ an outcome:
\begin{equation}
  Z = U_Z,
  \qquad
  X = \alpha Z + U_X,
  \qquad
  Y = \beta X + \gamma Z + U_Y,
\end{equation}
with $\alpha=0.3$, $\beta=1.0$, $\gamma=0.5$ and standard Gaussian exogenous noise ($\mu=0$, $\sigma=1$ for all three nodes). The intervention set is
\begin{equation}
  \mathcal I_{\mathrm{ATCE}}
  =
  \bigl\{\emptyset,\ \mathrm{do}(X{=}0),\ \mathrm{do}(X{=}1)\bigr\}.
\end{equation}
The designated shifted nodes are the root $Z$ and the outcome $Y$. The root \(Z\) is shifted only through its environment since it has no incoming structural mechanism. The non-root outcome \(Y\) is shifted through both its incoming structural coefficients $w_{Z\to Y}, w_{X\to Y}$ and its exogenous noise.

\subsection{LiLUCAS}
\label{app:lilucas}
The LUng CAncer dataset (LUCAS)\footnote{\href{http://www.causality.inf.ethz.ch/data/LUCAS.html}{http://www.causality.inf.ethz.ch/data/LUCAS.html}} was originally designed to simulate realistic causal relationships in lung cancer diagnosis. For our purposes, we leverage the linear variant of it: Linearised LUCAS
(LiLUCAS) used in \citep{felekis2026distributionallyrobustcausalabstractions}. LiLUCAS is a six-variable Markovian DAG with roots \emph{Smoking}, \emph{Genetics}, and \emph{Allergy}, and downstream variables \emph{LungCancer}, \emph{Coughing}, and
\emph{Fatigue}:
\begin{equation}
\begin{aligned}
  &\text{Smoking}=U_S,\qquad
  \text{Genetics}=U_G,\qquad
  \text{Allergy}=U_A,\\
  &\text{LungCancer}=0.9\,\text{Smoking}+0.8\,\text{Genetics}+U_L,\\
  &\text{Coughing}=0.6\,\text{LungCancer}+0.4\,\text{Allergy}+U_C,\\
  &\text{Fatigue}=0.9\,\text{LungCancer}+0.5\,\text{Coughing}+U_F,
\end{aligned}
\end{equation}
with Gaussian exogenous noise of means $\mu=[0,0,0.1,0.1,0.3,0.2]$ and standard deviations $\sigma=[0.5,2.0,1.5,1.0,0.8,1.2]$ in topological order. The designated shifted node is \emph{LungCancer}, whose mechanism comprises its incoming edges from \emph{Smoking} and \emph{Genetics} ($w_{\mathit{Sm}\to\mathit{LC}}$, $w_{\mathit{Ge}\to\mathit{LC}}$) together with its noise. The intervention set consists of the observational regime, the four single-node interventions $\mathrm{do}(\mathit{Sm}{=}v)$ and $\mathrm{do}(\mathit{Ge}{=}v)$ for
$v\in\{0,1\}$, and the four joint interventions $\mathrm{do}(\mathit{Sm}{=}v_1, \mathit{Ge}{=}v_2)$ for $v_1,v_2\in\{0,1\}$, giving $\lvert\mathcal I_{\mathrm{LiLUCAS}}\rvert=9$.

\subsection{Portland urban streams}
\label{app:portland}
The Portland benchmark \citep{Tabell2026} is a real-world domain-shift problem drawn from urban stream ecology. The scientific question is whether riparian tree-canopy cover ($X$) causally affects dissolved oxygen ($Y$) in urban streams, and whether an effect estimated on observed source watersheds transports to a held-out target watershed, Fanno Creek, whose environmental context differs from that of the source. The causal mechanisms are assumed to be invariant across watersheds while the distribution of environmental covariates shifts, making this a natural test of transport under a real, unmanufactured covariate shift.

\paragraph{Data sources.}
The benchmark is built by joining two public datasets. Seasonal water-quality field measurements from the Environmental Data Initiative (EDI) provide dissolved oxygen, water temperature, and site and season identifiers for urban stream sites in Portland, Oregon. Site-level landscape covariates from USGS ScienceBase provide elevation, mean annual precipitation, mean annual temperature, and riparian tree-canopy cover from the 2021 canopy layer. The two sources are merged by the shared site identifier, attaching the static landscape covariates to each site and season  observation. The original study and its analysis code are publicly available.\footnote{\url{https://github.com/ottotabell/transportability-ecology}}

\paragraph{Cleaning.}
We reproduce the original study's cleaning pipeline: rows missing key variables are removed ($-1$); the Columbia Slough watershed is dropped as hydrologically distinct from the remaining sites ($-60$); sites with zero canopy cover are removed ($-3$); duplicate site season records are collapsed to the first observation ($-2$); and a single extreme dissolved-oxygen outlier is removed ($-1$). The final benchmark contains \textbf{347 observations} across four seasonal rounds and $87$ unique sites.

\paragraph{Source/target split.}
Following the original study, Fanno Creek is the designated target domain ($N=100$) and all remaining watersheds form the source domain ($N=247$). The source provides the full joint distribution $p(y,x,s,z\mid\text{source})$, while the target contributes only its observational covariate distribution $p^{\ast}(z)$.

\paragraph{Causal structure.}
The modelled variables are a one-dimensional environmental summary $Z=\mathrm{PC}_1(\text{elevation, precipitation, mean temperature})$, the season $S$, the canopy cover $X$, and the dissolved oxygen $Y$; all are standardised
using source statistics, with the PCA fitted on source data only. The back-door view is the linear-Gaussian SCM
\begin{equation}
Z = U_Z,\quad
S = U_S,\quad
X = w_{ZX}Z + U_X,\quad
Y = w_{ZY}Z + w_{SY}S + w_{XY}X + U_Y,
\end{equation}
with edge weights fitted by ordinary least squares on source data. The designated shifted node is the root $Z$, carrying the environmental covariate shift between watersheds; as a root, it has no incoming edges, so its mechanism budget is structurally empty, and the shift is carried entirely by $Z$'s exogenous distribution. The constructive map is Markovian with a single free entry $\tau_Z$. The intervention set consists of the observational regime together with four canopy-cover interventions.
\begin{equation}
  \mathcal I_{\mathrm{Portland}}
  =
  \bigl\{\emptyset,\ \mathrm{do}(X{=}45),\ \mathrm{do}(X{=}50),\
  \mathrm{do}(X{=}55),\ \mathrm{do}(X{=}60)\bigr\},
\end{equation}
with canopy values in percent, standardised to source units before injection into the SCM.

\begin{table}[h]
\centering
\small
\begin{tabular}{@{}lllllp{2.4cm}@{}}
\toprule
& \textbf{Objective} & \textbf{Shifted} & \textbf{Budgeted coeff.} & \textbf{Radii swept} & \textbf{Target} \\
\midrule
ATE     & Gaussian              & $Y$        & $w_{X\to Y}$                                                 & $(\varepsilon,\eta)$ grid & one real \\
        &                       &            &                                                & & obs.\ only \\
ATCE    & Gaussian              & $Z$, $Y$   & $w_{Z\to Y}$, $w_{X\to Y}$                                     & $\varepsilon=\eta$        & $K{=}100$ \\
        &                       &            &                                                & & sampled \\
LiLUCAS & Empirical + Gaussian& \emph{LC}  & $w_{\mathit{Sm}\to\mathit{LC}}$, $w_{\mathit{Ge}\to\mathit{LC}}$ & $\varepsilon=\eta$      & $K{=}100$ \\
        &                       &            &                                                & & sampled \\
Portland& Gaussian              & $Z$ (root) & None ($\eta{=}0$)                                          & $\varepsilon$ only        & one real \\
        &                       &            &                                                & & obs.\ only \\
\bottomrule
\end{tabular}
\caption{Benchmark configurations. Shifted nodes carry the mechanism and/or environment ambiguity; budgeted coefficients are the incoming edges of each shifted node, and there are none for Portland, since its shifted node is a root. Targets are either a single observed domain, from which only the observational distribution is used, or $K{=}100$ shifts sampled from the ambiguity geometry.}
\label{tab:benchmark-configs}
\end{table}

\section{Implementation and optimisation details}
\label{app:implementation}
All models were trained on a MacBook Air (13-inch, M1, 2020) with an Apple M1 chip and 16\,GB of unified memory, running macOS~12.3. The optimiser and evaluation pipeline are implemented in Python with NumPy and PyTorch. For Gaussian benchmarks, the production runs use the \texttt{autograd} backend: the \(\tau\)-update differentiates the exact
Gaussian \(\mathcal W_2^2\) objective, while the adversarial mean and mechanism updates differentiate the smooth Bures surrogate. The covariance adversary is updated by the CovProx step. The empirical LiLUCAS runs do not use the Gaussian gradient backend; they use the closed-form empirical NumPy gradients.

\paragraph{Cross-validation and sample sizes.}
Performance is evaluated under a $k$-fold cross-validation scheme with $k=5$. For each intervention, we generate $N=5{,}000$ samples for ATE and $N=10{,}000$ samples for ATCE and LiLUCAS; the Portland real-world benchmark uses the observed sample sizes reported in Appendix~\ref{app:portland} ($N=247$ source, $N=100$ target). Fold assignments are fixed once per benchmark and reused verbatim across all training radii and both objectives, so that differences across configurations are not confounded by resampling. Each fold's map is trained on that fold's source split alone and scored only against its held-out split; targets are never seen during fitting or radius selection.

\paragraph{Minimax optimisation.}
Each outer iteration comprises $k_{\tau}=5$ projected gradient descent steps on~$\tau$ and $k_{\mathrm{adv}}=2$ adversary steps. Both players use vanilla (unaccelerated) projected gradient updates with constant step sizes
$\mathrm{lr}_{\tau} = \mathrm{lr}_{\mathrm{adv}} = 0.005$ for ATE and ATCE, and $\mathrm{lr}_{\tau} = \mathrm{lr}_{\mathrm{adv}} = 0.001$ for LiLUCAS (both objectives) and Portland. Under corner enumeration, the environment-mean adversary uses a step four times larger than the base adversary rate ($4\times0.005=0.02$ for ATE and ATCE, $4\times0.001=0.004$ for the Gaussian LiLUCAS runs), which we found necessary for the mean and mechanism players to converge on comparable timescales.

\paragraph{Corner enumeration for the entrywise box.}
For the entrywise-box ambiguity set used in every benchmark with nontrivial mechanism ambiguity: ATE, ATCE, and LiLUCAS, but not Portland, whose root-only shift leaves the mechanism set structurally empty; the shifted mechanism has at most two free coefficients, so the box has at most $2^2=4$ vertices, and we exploit this rather than running generic projected ascent on $\Delta W$. The mechanism adversary evaluates the loss at all corners of the box, selects the worst case, and refines it with three gradient-ascent steps; the $\tau$-update, in turn, computes the transport gradient at
every corner and averages them with uniform weights. This corner averaging is what ties the learned map to the box geometry. At a symmetric box ($\delta=0$), the opposing corners $+B$ and $-B$ contribute equal and opposite $\tau$-gradients that cancel under uniform weighting, so the identity is a stationary point of the $\tau$-update; at a directional box ($\delta\neq0$), the corner at $\delta+B$ lies farther from the origin and contributes a larger gradient, so the average no longer cancels and $\tau$ is driven away from the identity. The construction therefore makes the directional map non-trivial by design, independently of the loss, the algorithmic counterpart of the corner-sum argument of Section~\ref{subsec:mechanism-ambiguity}.

\paragraph{Convergence criterion.}
Because the problem is a saddle-point, a loss-descent stopping rule is inappropriate: the objective can increase toward the equilibrium when the adversary is active, so a criterion that stops when the loss ceases to decrease would terminate prematurely. We therefore monitor two quantities per outer iteration: the relative change in the objective,
\begin{equation}
  r_t
  =
  \frac{\lvert L_{t-1} - L_t\rvert}{\lvert L_t\rvert + \epsilon_{\mathrm{rel}}},
  \qquad \epsilon_{\mathrm{rel}} = 10^{-8},
\end{equation}
and the duality gap $g_t = L^{\mathrm{adv}}_t - L_t$ between the adversary's and the map's objective. After a warm-up floor of $30$ iterations, optimisation terminates once \emph{either} the objective has settled ($r_t < \mathrm{tol}$, with $\mathrm{tol}=10^{-5}$) \emph{or} the duality gap has effectively closed ($\lvert g_t\rvert < 10^{-6}$), \emph{and} in either case, the gap is below the confirmation threshold ($\lvert g_t\rvert < 10^{-4}$). The gap-confirmation conjunct is required so that convergence is never declared while the adversary is still being exploited: it tracks true saddle-point convergence rather than mere loss stationarity and guards against premature termination on high-magnitude objectives where the relative change is small but the gap is still decreasing. A hard cap of $5{,}000$ iterations is retained as a safety net. Every trained configuration in the reported results satisfies the gap confirmation condition at termination.

\section{Note on relation to DiRoCA}
\label{app:relation-diroca}
TraCA shares its optimisation core with DiRoCA \citep{felekis2026distributionallyrobustcausalabstractions} but solves a different problem. DiRoCA robustifies a causal abstraction between a low- and a high-level model \emph{within one system}, hedging against environmental noise and misspecification; TraCA transports a causal model \emph{across two same-level populations}: a source and a target. The shared machinery is the min-max formulation over $2$-Wasserstein ambiguity sets, its finite-dimensional reduction to Frobenius-budgeted perturbations $\lVert\Theta\rVert_F\le\varepsilon\sqrt N$ of the abduced exogenous residuals, and the alternating projected gradient descent-ascent solver. The problems differ in what the ambiguity set is placed on. DiRoCA centers a \emph{product} set on the joint low/high environment of a single system and interpolates between exact and uniform abstraction; TraCA places a \emph{single} set on a source model and asks which target models a common transport map can reach.

On that reformulation, TraCA builds its contribution: a hierarchy of source-target consistency notions (Section~\ref{subsec:hierarchy}), exact existence characterisations for a common constructive transport operator in the Markovian and semi-Markovian cases, together with the compatibility gap separating query- from model-level transport (Section~\ref{sec:characterizing}), and certified query \emph{intervals} $\ell\le Q_t\le L$ (Section~\ref{sec:provable-robustness}). Where DiRoCA bounds a scalar abstraction error, TraCA turns abstraction error into an interval for a causal query and does so exactly where classical transportability is silent: $\mathcal R_1$, when transportability does not hold, or $\mathcal R_2$, where no target data exists.

\section{Supplementary experimental results}
\label{app:supplementary-results}
This appendix collects results deferred from Section~\ref{sec:experiments}, organised by benchmark. All quantities are computed on the query family defined in Section~\ref{subsec:benchmarks}, under the protocol of Section~\ref{subsec:setup}. Each figure below is referenced from the corresponding paragraph of Section~\ref{subsec:results}, where the claim it supports is stated; here we give the full profile behind that claim.

\subsection{ATE}
\begin{figure}[h]
  \centering
  \includegraphics[width=.9\textwidth]{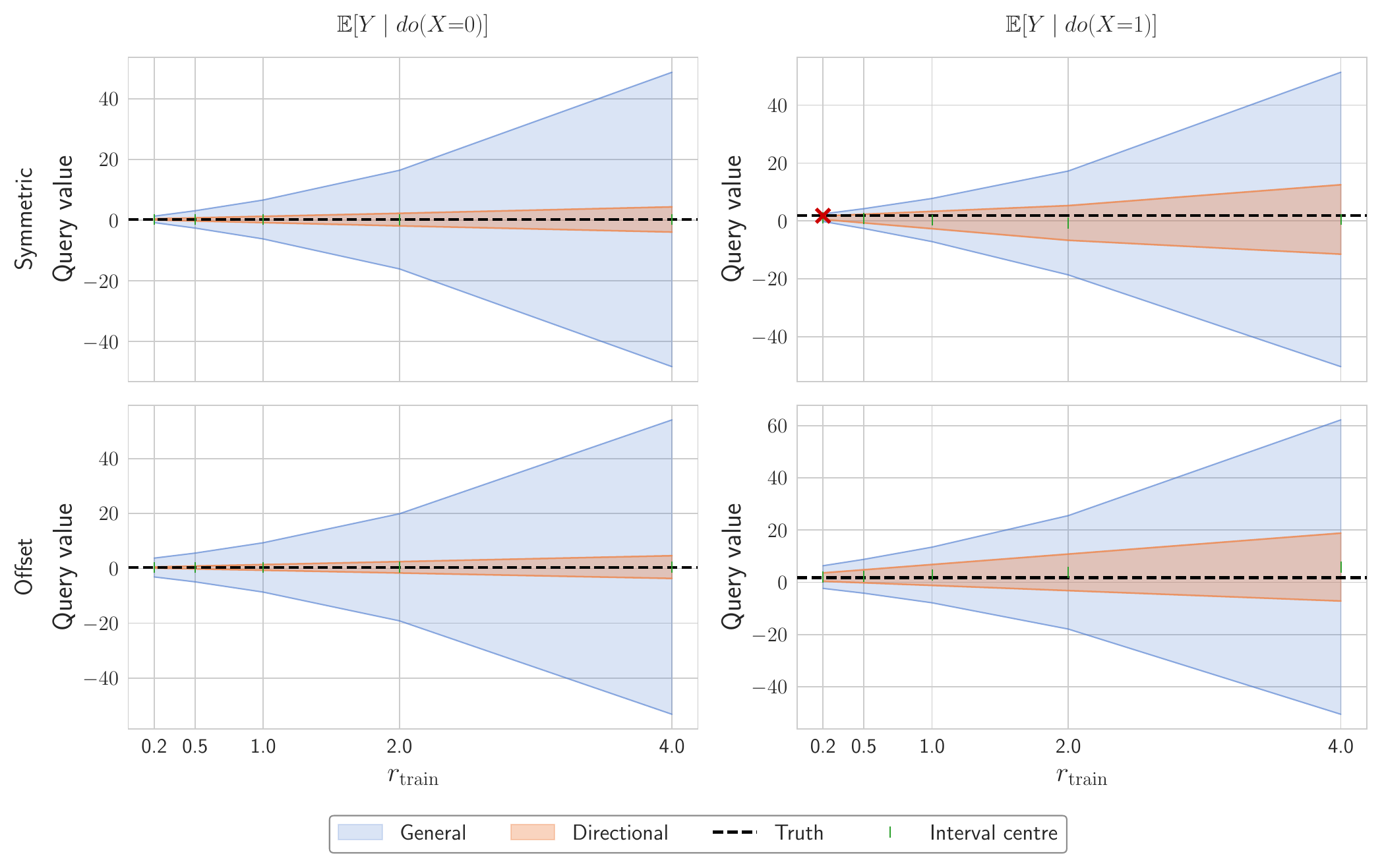}
  \caption{ATE: nested certified intervals along the diagonal $r_{\mathrm{train}} = \varepsilon{=}\eta$, for both variants and both queries. Blue = general, orange = directional, dashed = truth. The band widths are read more precisely from Figure~\ref{fig:ate-width}.}
  \label{fig:ate-shaded}
\end{figure}

Section~\ref{subsec:results} reports certificate widths at the observationally selected cell $r_{\mathrm{train}}{=}0.2$. Figure~\ref{fig:ate-shaded} provides the full picture along the diagonal: both certificates are nested at every radius, both widen with $r_{\mathrm{train}}$, and the general one does so faster. The truth lies within the directional band under the offset variant at every radius shown and outside it under the symmetric variant at the smallest radius, and the coverage gap of $0.08$ is discussed in the main text.

\subsection{ATCE}
\begin{figure}[h]
  \centering
  \includegraphics[width=.9\textwidth]{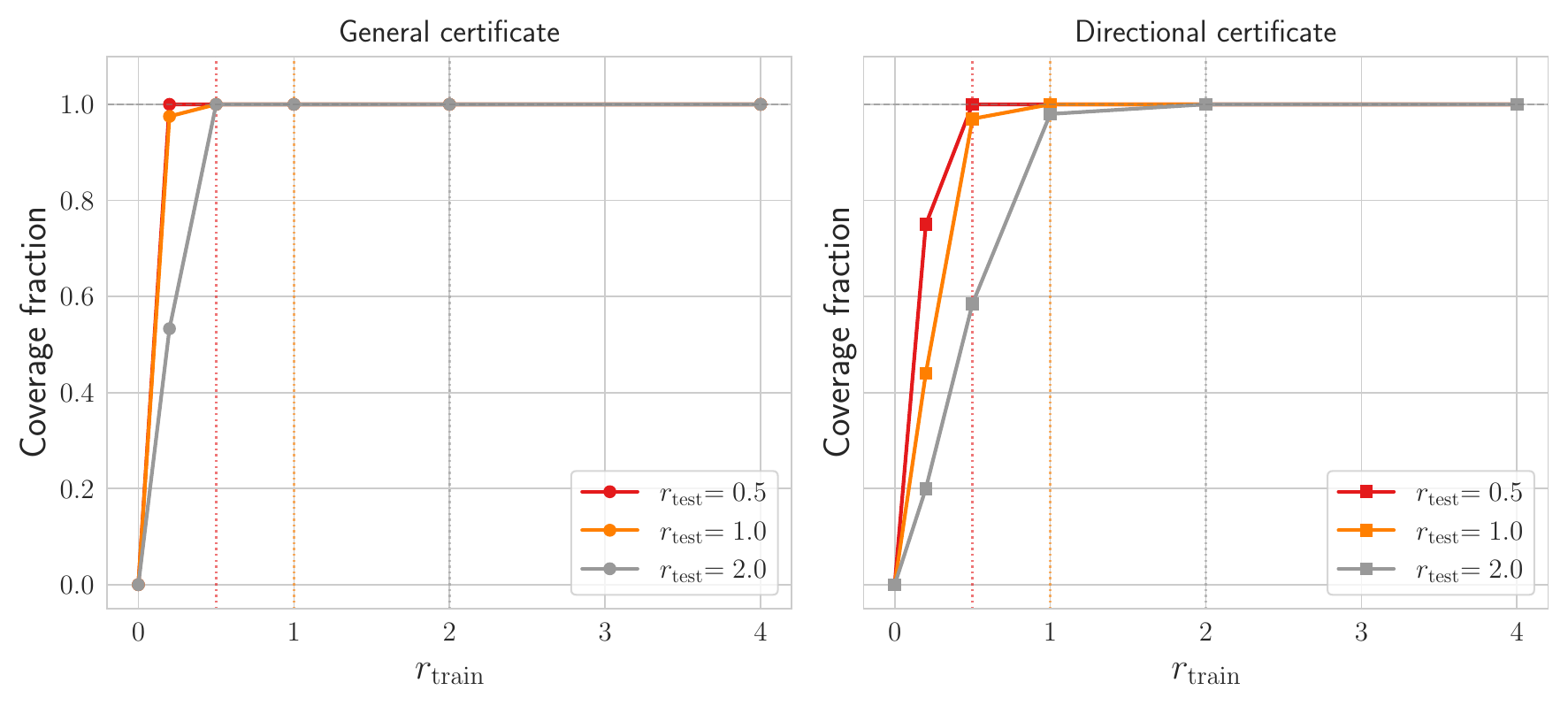}
  \caption{ATCE (sub-family): coverage against $r_{\mathrm{train}}$ at three fixed
  $r_{\mathrm{test}}$ values, for the general (left) and directional (right)
  certificates. Directional coverage rises to one at
  $r_{\mathrm{train}}{=}r_{\mathrm{test}}$; the general certificate already covers
  below that boundary.}
  \label{fig:atce-coverage}
\end{figure}

The main text describes the coverage transition at a single test magnitude.
Figure~\ref{fig:atce-coverage} shows it at three, and the pattern is the same in each: directional coverage reaches one exactly as $r_{\mathrm{train}}$ meets
$r_{\mathrm{test}}$, tracking the in-ball boundary, whereas the general certificate covers well below it. The wider bound is simply large enough to admit targets outside the ball for which it was built.

\subsection{LiLUCAS}
\begin{figure}[t]
\centering
\begin{subfigure}[b]{0.48\textwidth}
  \centering
  \includegraphics[width=\textwidth]{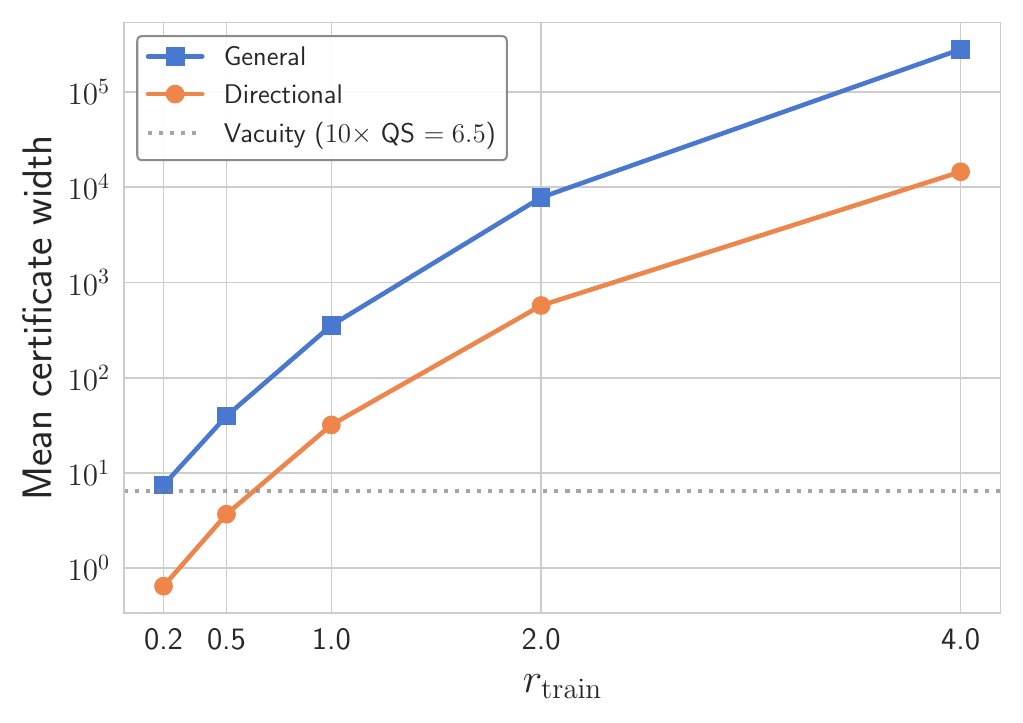}
  \subcaption{Empirical symmetric}
  \label{fig:lilucas-ewsym-width}
\end{subfigure}
\hfill
\begin{subfigure}[b]{0.48\textwidth}
  \centering
  \includegraphics[width=\textwidth]{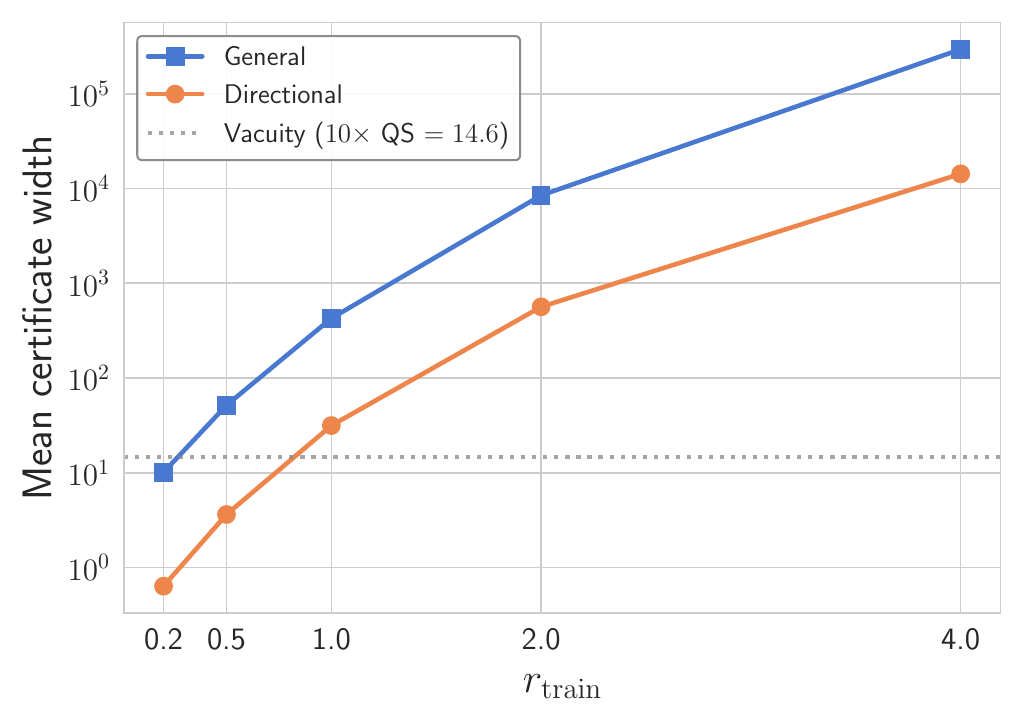}
  \subcaption{Gaussian symmetric}
  \label{fig:lilucas-gau-width}
\end{subfigure}
\caption{LiLUCAS: certificate width against $r_{\mathrm{train}}$ (log scale), with the vacuity threshold $10\times\text{QS}$. \textbf{(a)} Under the empirical objective the general certificate is vacuous across the entire grid; only the directional certificate is informative, and only at the two smallest radii. \textbf{(b)} Under the Gaussian objective the general certificate is vacuous from $r_{\mathrm{train}}{=}0.5$ onward, while the directional certificate stays roughly an order of magnitude narrower and remains informative longer. Note the thresholds differ ($6.5$ and $14.6$): the empirical query scale is smaller, so a narrower interval can still be vacuous.}
\label{fig:lilucas-width}
\end{figure}

Section~\ref{subsec:results} quotes widths at two radii for each objective; Figure~\ref{fig:lilucas-width} gives the profiles. Placing the two objectives side by side makes the query-scale effect visible: the empirical general certificate is narrower in absolute terms than the Gaussian one at every radius; yet, it is vacuous throughout because its vacuity threshold is less than half as large. Absolute width is not what determines whether a certificate says anything.

\subsection{Portland}
\begin{figure}[t]
\centering
\begin{subfigure}[b]{0.48\textwidth}
  \centering
  \includegraphics[width=\textwidth]{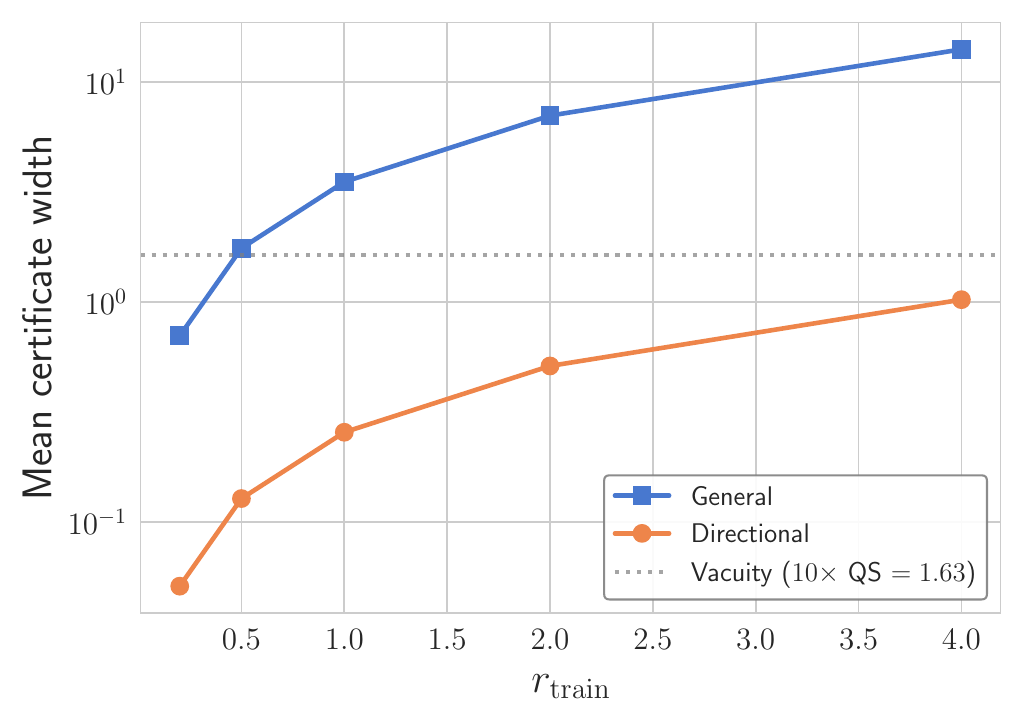}
  \subcaption{Certificate width}
  \label{fig:portland-width}
\end{subfigure}
\hfill
\begin{subfigure}[b]{0.48\textwidth}
  \centering
  \includegraphics[width=\textwidth]{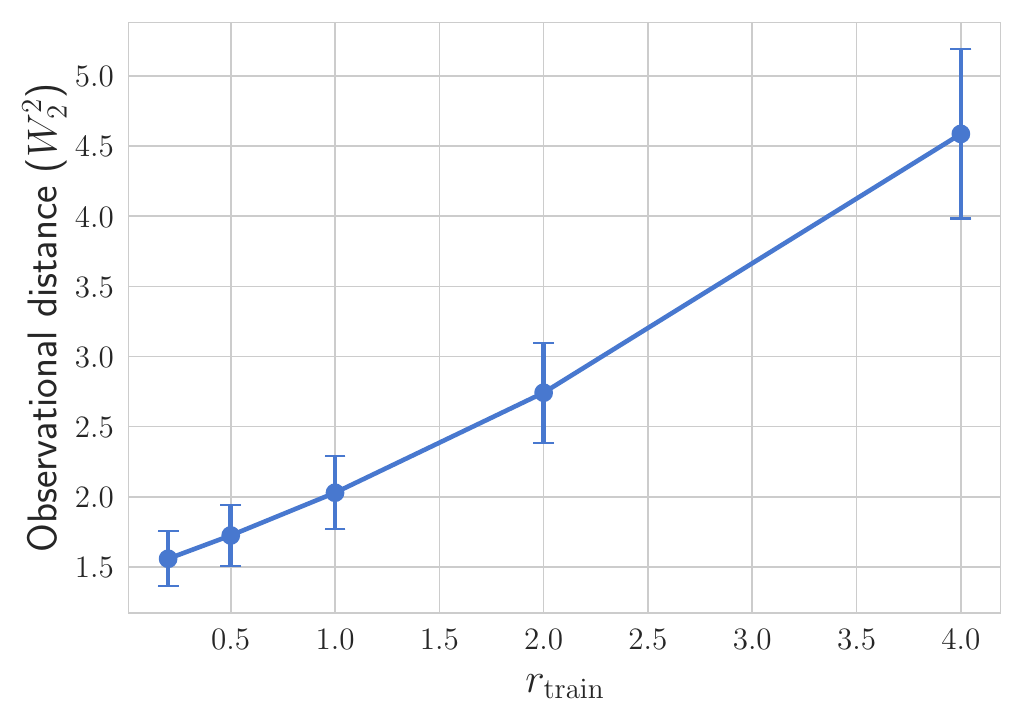}
  \subcaption{Observational transport distance}
  \label{fig:portland-obs}
\end{subfigure}
\caption{Portland, against $r_{\mathrm{train}}$. \textbf{(a)} Certificate width (log scale): the general certificate is vacuous from $r_{\mathrm{train}}{=}0.5$ onward, while the directional stays informative throughout. \textbf{(b)} Observational transport distance with fold-level error bars: it increases monotonically, so the observational selector returns the smallest $\varepsilon$, at which coverage is zero.}
\label{fig:portland-width-obs}
\end{figure}

The two panels of Figure~\ref{fig:portland-width-obs} show why radius selection needs care on this benchmark. Coverage of the interventional query requires a radius large enough to contain the realised shift, and the directional certificate remains informative well past that point. The observational distance, however, moves monotonically in $r_{\mathrm{train}}$, so it carries no interior optimum and cannot indicate where that point lies. The guarantee is unaffected; it holds at whatever radius is declared, but the radius must come from domain knowledge rather than from the observational fit, unlike ATE where the two coincide.

\end{document}